\documentclass[11pt]{article}
\newif\ifanonymous \anonymousfalse
\makeatletter

\RequirePackage{iftex}
\newif\ifmd@ot \ifXeTeX \md@ottrue\fi
\ifLuaTeX
  \ifnum\directlua{tex.sprint(luaotfload and luaotfload.add_fallback and 1 or 0)}=1 \md@ottrue\fi
\fi
\def\md@fb{0.55} \def\md@fbkey{} \ifdim\md@fb pt>0pt \edef\md@fbkey{FakeBold=\md@fb,}\fi

\RequirePackage{xcolor}
\RequirePackage[letterpaper,margin=1in,footskip=0.5in]{geometry} \RequirePackage{amsmath,amssymb} 

\ifmd@ot
  \RequirePackage{unicode-math}
  \defaultfontfeatures{Scale=MatchLowercase}
  \defaultfontfeatures[\rmfamily]{Ligatures=TeX,Scale=1}
  \expanded{\setmainfont{et-book-roman-line-figures.ttf}[\md@fbkey BoldFont=et-book-semi-bold-old-style-figures.ttf,ItalicFont=et-book-display-italic-old-style-figures.ttf]}
  \expanded{\setmonofont{JetBrainsMono-Regular.ttf}[\md@fbkey]}
  \expanded{\setmathfont{STIXTwoMath-Regular.otf}[\md@fbkey]}
  \newfontfamily\mdArxivBoldFile{et-book-semi-bold-old-style-figures.ttf}
  \newfontfamily\mdArxivItalicFile{et-book-display-italic-old-style-figures.ttf}
  
  \let\HeadingFont\rmfamily
  \DeclareTextCommand{\textvisiblespace}{TU}{{\ttfamily\char"2423\relax}}
  \DeclareTextCommand{\v}{TU}[1]{{\expandafter\let\expandafter\md@base\the\accent"02C7 \md@base #1}}
\else
  \PackageWarning{md2pdf}{Type 1 branch (Palatino); no OpenType engine here. Run lualatex with luaotfload for the md2pdf fonts}
  \RequirePackage[T1]{fontenc}
  \RequirePackage{lmodern}   \RequirePackage{mathpazo}  \let\HeadingFont\rmfamily
\fi
\RequirePackage{microtype}
\UseMicrotypeSet[protrusion]{basicmath} 

\definecolor{PaperIvory}{RGB}{255,253,245}
\pagecolor{PaperIvory} \RequirePackage{fancyhdr}
\RequirePackage[indent=1em]{parskip} 
\RequirePackage{setspace}
\def\md@body{11}
\newdimen\md@bodysize \md@bodysize=\md@body pt
\newcommand*\md@size[1]{\the\dimexpr #1\md@bodysize\relax}
\let\md@normalsize\normalsize
\renewcommand\normalsize{\md@normalsize\@setfontsize\normalsize{\md@size{1}}{\md@size{1.2}}}
\normalsize

\RequirePackage{titlesec}
\titleformat{\section}{\HeadingFont\fontsize{\md@size{1.5}}{\md@size{1.8}}\selectfont\bfseries}{\thesection}{1em}{}
\titleformat{\subsection}{\HeadingFont\fontsize{\md@size{1.333}}{\md@size{1.6}}\selectfont\bfseries}{\thesubsection}{1em}{}
\titleformat{\subsubsection}{\HeadingFont\fontsize{\md@size{1.167}}{\md@size{1.4}}\selectfont\bfseries}{\thesubsubsection}{1em}{}
\titleformat{\paragraph}{\HeadingFont\fontsize{\md@size{1}}{\md@size{1.2}}\selectfont\bfseries}{\theparagraph}{1em}{}
\titleformat{\subparagraph}{\HeadingFont\fontsize{\md@size{1}}{\md@size{1.2}}\selectfont\bfseries}{\thesubparagraph}{1em}{}
\RequirePackage{titling}
\pretitle{\begin{center}\HeadingFont\fontsize{\md@size{1.667}}{\md@size{2}}\selectfont\bfseries}
\posttitle{\par\end{center}}
\apptocmd\maketitle{\setstretch{1.2}}{}{} 

\RequirePackage{enumitem}
\setlist{nosep} 

\RequirePackage{fvextra}
\fvset{breaklines,breakanywhere,fontsize=\fontsize{\md@size{1}}{\md@size{1.2}}\selectfont,baselinestretch=0.8}
\DefineVerbatimEnvironment{verbatim}{Verbatim}{} 

\RequirePackage{caption}
\PassOptionsToPackage{hyphens}{url}
\RequirePackage{xurl}
\PassOptionsToPackage{unicode,colorlinks,linkcolor=blue,urlcolor=blue,citecolor=blue,filecolor=blue,breaklinks}{hyperref}

\makeatother
 \usepackage{amsmath,amssymb,amsthm}
\usepackage{booktabs}
\usepackage{array}
\usepackage{tikz}\usepackage{nicematrix}
\NiceMatrixOptions{custom-line = {letter = ;, command = dottedrule, tikz = {line width = 0.5pt, line cap = round, dash pattern = on 0pt off 2pt, color = black!40}, total-width = 0pt},
  xdots/radius = 0.25pt, xdots/inter = 2pt}
\usepackage{graphicx}
\usepackage{placeins}\graphicspath{{figures/}}
\usepackage{etoolbox}
\usepackage[authoryear,round]{natbib}
\usepackage{hyperref}
\usepackage{refcount}\makeatletter
\newcommand*{\appx@here}{\ifnum\c@subsection>\z@\thesubsection\else\thesection\fi}
\newcommand*{\locatelabels}{\AddToHookWithArguments{label}{\protected@write\@auxout{}{\string\newlabel{##1@appendix}{{\appx@here}{\thepage}{}{\@currentHref}{}}}}}
\newcommand*{\appx@where}[1]{\begingroup\edef\@tempa{\getrefnumber{#1@appendix}}\edef\@tempb{\appx@here}\ifx\@tempa\@tempb\else\ in Appendix~\ref{#1@appendix}\fi\endgroup}
\makeatother
\NewDocumentCommand{\appref}{m}{\ref{#1}\csname appx@where\endcsname{#1}}
\NewDocumentCommand{\appeqref}{m}{\eqref{#1}\csname appx@where\endcsname{#1}}

\newcommand{\tnFloorQwenThreeEightTwoSevenB}{\ensuremath{0.31}}    \newcommand{\tnStepShareMin}{\ensuremath{0.95}}                         \newcommand{\tnEOneOpeningZeroWidthSpacesPAtZeroMuseGlimmer}{\ensuremath{0.438}}    \newcommand{\tnEOneOpeningZeroWidthSpacesPAtFourMuseGlimmer}{\ensuremath{0.531}} \newcommand{\tnEOneOpeningZeroWidthSpacesPAtFiveMuseGlimmer}{\ensuremath{0.500}}                                       \newcommand{\tnEOneNewlinesRangeFGemmaFourThreeOneB}{\ensuremath{5.19}}                 \newcommand{\tnEOneOpeningNewlinesPAtZeroGemmaFourOneTwoB}{\ensuremath{0.915}}     \newcommand{\tnEOneOpeningNewlinesPAtFiveGemmaFourOneTwoB}{\ensuremath{0.593}}                \newcommand{\tnEOneNewlinesMinPGemmaFourOneTwoB}{\ensuremath{0.469}} \newcommand{\tnEOneNewlinesMaxPGemmaFourOneTwoB}{\ensuremath{0.967}}  \newcommand{\tnEOneNewlinesRangeFGemmaFourOneTwoB}{\ensuremath{3.50}}            \newcommand{\tnEOneBasePGemmaFourOneTwoB}{\ensuremath{0.915}}           \newcommand{\tnETwoAttributeShareMedianMuseGlimmer}{\ensuremath{8.6}}      \newcommand{\tnETwoAttributeShareMedianGemmaFourThreeOneB}{\ensuremath{37.7}}      \newcommand{\tnETwoAttributeShareMedianGemmaFourOneTwoB}{\ensuremath{26.3}}                                          \newcommand{\tnEFiveAttributeShareMedianSmartphoneGemmaFourThreeOneB}{\ensuremath{40.1}}                     \newcommand{\tnETwoExampleTopOtherPhiMuseGlimmer}{\ensuremath{-1.30}}                                                                                                                                                                                                                                                         \newcommand{\tnESixAttributeShareMedianLevelTwoGemmaFourOneTwoB}{\ensuremath{40.8}}          \newcommand{\tnGridCalibrationMaxAbsErrorTOneZeroZero}{\ensuremath{0.21}}  \newcommand{\tnGridCalibrationMaxAbsErrorTOneZeroZeroZero}{\ensuremath{0.071}}  \newcommand{\tnGridCalibrationMaxAbsErrorTOneZeroZeroZeroZero}{\ensuremath{0.025}}  \newcommand{\tnGridConvergenceMaxChange}{\ensuremath{0.063}} \newcommand{\tnGridConvergenceOutside}{\ensuremath{1}}  \newcommand{\tnGridShareMedianWord}{\ensuremath{19.1}} \newcommand{\tnGridShareMedianEmpty}{\ensuremath{24.6}}      \newcommand{\tnGridShareMedianProbability}{\ensuremath{16.4}} \newcommand{\tnGridShareMedianLogProbability}{\ensuremath{24.2}} \newcommand{\tnGridShareMedianSpan}{\ensuremath{40.8}} \newcommand{\tnGridShareMedianAdversarial}{\ensuremath{64.7}} \newcommand{\tnRetainedEditGameEditsFlight}{\ensuremath{10}}   \newcommand{\tnRetainedEditGameEditsHotel}{\ensuremath{11}}   \newcommand{\tnRetainedEditGameEditsSmartphone}{\ensuremath{11}}   \newcommand{\tnExampleBaseP}{\ensuremath{0.438}} \newcommand{\tnExampleFramedP}{\ensuremath{0.531}} \newcommand{\tnExampleDirectEffect}{\ensuremath{0.38}}      \newcommand{\tnBenchmarkEqualShare}{\ensuremath{8.2}}  \newcommand{\tnExampleAttributeShare}{\ensuremath{11.2}} \newcommand{\tnExampleAttributeShareLow}{\ensuremath{9.8}} \newcommand{\tnExampleAttributeShareHigh}{\ensuremath{12.4}}    \newcommand{\tnExamplePhrasePhi}{\ensuremath{0.07}}   \newcommand{\tnExampleContexts}{\ensuremath{1{,}024}} \newcommand{\tnExampleFramePhi}{\ensuremath{0.03}} \newcommand{\tnExampleContextsMean}{\ensuremath{-0.01}} \newcommand{\tnExampleContextsNegative}{\ensuremath{429}} \newcommand{\tnExampleContextsZero}{\ensuremath{248}} \newcommand{\tnExampleContextsPositive}{\ensuremath{347}}  \newcommand{\tnExamplePriceAlone}{\ensuremath{-3.25}} \newcommand{\tnExamplePriceWithEdits}{\ensuremath{-2.50}}                \newcommand{\tnExampleWordingGameEditPhiOpZeroEight}{\ensuremath{-0.08}}             \newcommand{\tnExampleAttributeRemovalMin}{\ensuremath{1.50}} \newcommand{\tnExampleAttributeRemovalMax}{\ensuremath{2.31}}   \newcommand{\tnExampleContrastTime}{\ensuremath{0.17}} \newcommand{\tnExampleContrastTimeSe}{\ensuremath{0.04}} \newcommand{\tnExamplePhiPriceA}{\ensuremath{0.19}} \newcommand{\tnExamplePhiPriceAFramed}{\ensuremath{0.07}} \newcommand{\tnExampleChangePriceA}{\ensuremath{-0.12}} \newcommand{\tnExampleChangePriceASe}{\ensuremath{0.03}}     \newcommand{\tnExamplePhiPriceB}{\ensuremath{-0.03}}                                                                                                  \newcommand{\tnRetainedPartShareMedianAnswerInstructionMuseGlimmer}{\ensuremath{40.1}}                           \newcommand{\tnRetainedWtpMinFlightMuseGlimmer}{\ensuremath{46.95}} \newcommand{\tnRetainedWtpMaxFlightMuseGlimmer}{\ensuremath{85.52}}                                                                    \newcommand{\tnProcedureDirectEvaluations}{\ensuremath{2}}      \newcommand{\tnProcedureEditsEvaluations}{\ensuremath{2{,}048}}  \newcommand{\tnProcedurePooledEvaluations}{\ensuremath{32}}  \newcommand{\tnProcedureSpansEvaluations}{\ensuremath{1{,}024}}  \newcommand{\tnProcedureWordsEvaluations}{\ensuremath{50{,}000}}    \newcommand{\tnDiagnosticsPairsQwenThreeEightTwoSevenB}{\ensuremath{32}} \newcommand{\tnDiagnosticsUnresolvedQwenThreeEightTwoSevenB}{\ensuremath{3}}   \newcommand{\tnDiagnosticsMinMarginMuseGlimmer}{\ensuremath{2.38}} \newcommand{\tnDiagnosticsPairsMuseGlimmer}{\ensuremath{32}} \newcommand{\tnDiagnosticsUnresolvedMuseGlimmer}{\ensuremath{2}}    \newcommand{\tnDiagnosticsPairsGemmaFourThreeOneB}{\ensuremath{32}} \newcommand{\tnDiagnosticsUnresolvedGemmaFourThreeOneB}{\ensuremath{0}}    \newcommand{\tnDiagnosticsPairsGemmaFourOneTwoB}{\ensuremath{32}} \newcommand{\tnDiagnosticsUnresolvedGemmaFourOneTwoB}{\ensuremath{5}}  
\newcommand*{\floatnotes}[1]{}\newcommand*{\figurenotes}[1]{}
\newcommand*{\floatNotesHere}{\ifdefined\floatNotesBody\par\smallskip
  \begin{minipage}{\linewidth}\footnotesize\setstretch{1}\raggedright\floatNotesBody\end{minipage}\fi}
\AtEndEnvironment{table}{\floatNotesHere}
\AtEndEnvironment{figure}{\floatNotesHere}
\makeatletter
\newcommand*{\logFloatMeasurement}{\typeout{FLOATMEASURE \@captype\space\csname the\@captype\endcsname\space ht=\the\ht\@currbox\space dp=\the\dp\@currbox\space textheight=\the\textheight}}
\AddToHook{cmd/@endfloatbox/after}{\logFloatMeasurement}
\makeatother
 \AtBeginEnvironment{tabular}{\setstretch{1}}

\newcommand{\openingInvisibleModel}{Muse Glimmer}
\newcommand{\openingLayoutModel}{Gemma 4 12B}

\title{Token Noise and the Illusion of Preference:\\ A Shapley Value Framework for\\ Auditing LLM-Simulated Choice Experiments}
\ifanonymous
\author{}
\preauthor{}\postauthor{}
\else
\author{Behnam Mohammadi\thanks{University of Texas at Dallas (UTD), \href{mailto:bxm180038@utdallas.edu}{\texttt{bxm180038@utdallas.edu}}.}}
\fi
\date{}
\predate{}\postdate{}

\begin{document}
\maketitle
\setstretch{1.444}
\ifdefined\NoSections\else
\ifdefined\NoAbstract\else
\begin{abstract}
Large language models are increasingly used to simulate human respondents in choice experiments, yet their answers depend on the exact tokens of a prompt rather than on the offered products alone. I show that invisible characters, blank lines, or minor wording changes shift or reverse simulated choice probabilities. This sensitivity, which I term \emph{token noise}, creates an illusion of preference and exposes studies to \emph{prompt hacking}. I introduce a Shapley value audit that divides a model's choice response among the pieces of its prompt and reads each share beside the effect of removing that piece. The \emph{edit game} divides the effect of several wording and formatting changes beside an actual price change. The \emph{word game} divides a prompt's response across its words or tokens and summarizes the credit earned by the product attributes as the attribute share. Across four open-weight models and three product categories, wording and formatting edits could rival a substantive price change, while product attributes consistently received a minority of the word-level credit, sometimes less than the answer instruction. \emph{Framing} cues could take credit from the very price they emphasized and frequently vanished or reversed across equivalent wordings of the question. A \emph{persona} (even a bare statement that the respondent is a consumer) also shifted choice probabilities and the attributes' credit. Declared with its ``players,'' reference, scale, and uncertainty, my audit separates what a simulated respondent prefers from how the question was asked, the distinction a study must draw before it reads a simulated choice as a preference.
\end{abstract}

 \clearpage
\fi
\section{Introduction}\label{sec:introduction}

Ask a language model which of two flights it prefers. Flight A costs \$200 and takes five hours; Flight B costs \$350 and takes two; the model must answer with a single letter, and I capture the probability that it answers ``A''. Now insert four zero-width spaces after the question (invisible characters used in software to mark where a line may break), and ask again. For one model (\openingInvisibleModel{}), the probability of choosing the cheaper flight rises from \tnEOneOpeningZeroWidthSpacesPAtZeroMuseGlimmer{} to \tnEOneOpeningZeroWidthSpacesPAtFourMuseGlimmer{}, so the answer the model is more likely to give changes, although no reader could tell the two prompts apart. Insert five blank lines instead, and for a second model (\openingLayoutModel{}), the probability of choosing the cheaper flight falls from \tnEOneOpeningNewlinesPAtZeroGemmaFourOneTwoB{} to \tnEOneOpeningNewlinesPAtFiveGemmaFourOneTwoB{} (Figure~\ref{fig:opening}). In both cases, the prices and travel times are untouched. Each check asks the same question in two versions that differ in one detail. I chose these two cases to show that a model's choice can move when nothing about the flights has moved.

\begingroup\long\def\floatNotesBody{
Flight task 7 offers \$200 and 5 hours against \$350 and 2 hours. Left: Muse Glimmer with zero-width spaces, characters with no visible glyph. Right: Gemma 4 12B with blank lines, a visible change of layout; the square marks five lines. Full sweeps: Figure~\appref{fig:e1}.
}%
\begin{figure}[htbp]
\centering
\includegraphics[width=\textwidth]{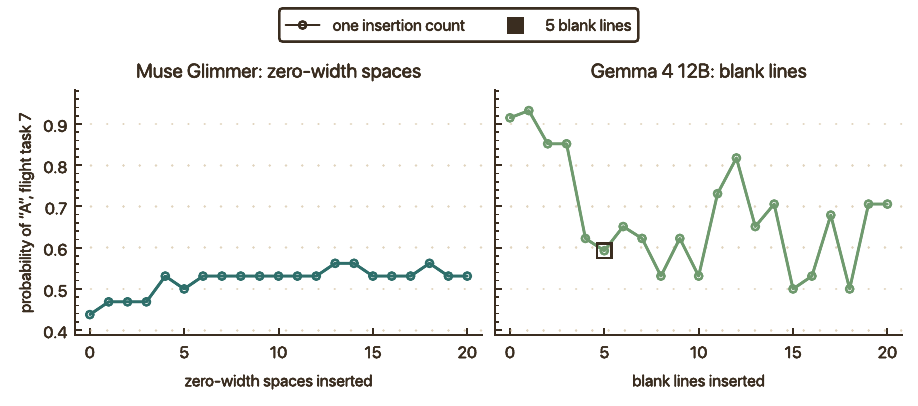}
\caption{Probability of ``A'' on flight task 7 as zero-width spaces (left) or blank lines (right) are inserted after the question.}\label{fig:opening}
\end{figure}
\endgroup{}
\figurenotes{opening}

This would simply remain a curiosity if language models were not used to simulate human respondents. In recent years, economists have used language models as simulated economic agents \citep{horton2023large}, and political scientists have used them as synthetic samples of respondents conditioned on demographic descriptions \citep{argyle2023out}. Marketing applications of the same idea include preference elicitation \citep{goli2024can}, perceptual analysis of brands and products \citep{li2024determining}, digital twins of individual respondents built from their answers to many questions \citep{toubia2025twin}, and the augmentation of observed choice data with generated choices \citep{wang2026large}. In each application, the model stands in for a person, and its answer is read as one that person might have given. These applications therefore have to establish whether simulated respondents choose as real customers do, and the evidence so far is mixed. \citet{brand2026usingllms} found willingness-to-pay estimates derived from language-model responses sometimes comparable with estimates from human studies, but often inaccurate and in some cases wrong-signed, and they positioned generated data as a supplement to human studies, not a substitute. But Figure~\ref{fig:opening} shows that, before asking whether simulated respondents choose as real customers do, a researcher must know what a simulated respondent is responding to. If a model's answer moves when invisible characters are inserted, the answer is not, or not only, a response to the flights, and checking it against real customers would be checking the wrong thing. This paper therefore asks how much of a simulated choice is a response to the products and how much is a response to the way the question was written, and it shows how to measure each.

Every choice experiment assumes something so ordinary that questionnaire designers rarely state it. The respondent reacts to the products, not to the typography of the question. Choice models make this assumption explicit by treating a choice as responsive to the displayed attributes and invariant to equivalent descriptions of the same options \citep{green1990conjoint,train2009discrete}. A language model has no such invariance built in. It is an autoregressive model, which means that, given the text so far, it assigns a probability to every possible next token (the unit of text it processes, which may be a word, part of a word, or a single character). The probability of answering ``A'' is therefore a fixed function of the exact text of the prompt, and that text includes every line break, every punctuation mark, and every invisible character. This is also why the natural safeguard (asking the same question many times and checking that the answer is stable) cannot detect this dependence. Repeating an unchanged prompt draws answers from a distribution that does not change. The distribution changes only when the text changes, as Figure~\ref{fig:opening} shows.

When the answer depends on tokens that do not change the choice being offered, I call that dependence \emph{token noise}. Token noise does not mean that instructions, labels, or layout are errors. It means only that a response that moves with them is not a response to the products alone. Token noise can create an \emph{illusion of preference} when a researcher reads the model's responses as revealing a stable preference over the offered alternatives. Because the researcher writes the prompt, the illusion is not harmless. The same question can be written in many ways, with different wording, labels, punctuation, and answer instructions. If the way the question is written changes the answer, the wording becomes part of the result that the question was meant to reveal, and choosing a wording for the result it produces, deliberately or not, is what I call \emph{prompt hacking}. Not every change in wording is noise, however. Adding ``only'' before a price, or telling the model whom it is simulating, changes the information supplied on purpose, and an audit must say which changes it treats as presentation (the way the same question is written) and which as interventions.

Checks such as those in Figure~\ref{fig:opening} are the right first step, but they leave two questions unanswered. The first arises when several wording choices are in play at once. Their effects depend on one another, since the effect of the word ``costs,'' for instance, can depend on whether the price follows it. Asking the question in its original and its fully edited versions then shows the combined effect of everything that differs between them, but not how much of that effect belongs to each choice. The second arises for a single question as written. A check can show that the answer moves when one detail changes, but it cannot say how much of the answer comes from the product attributes that the experiment varies and how much from the labels, instructions, and layout that carry them. In both cases, several pieces of the prompt produce the answer together, their effects depend on one another, and the researcher needs to divide the credit for the answer among them. Dividing a jointly produced total among contributors whose effects interact is a problem of allocation.

I solve this allocation problem with the Shapley value \citep{shapley1953value}, a rule from cooperative game theory for dividing a jointly produced total among the contributors (called ``players'') that produced it. The rule credits each player with the amount the total changes when that player joins the others, averaged over every order in which the players could have joined, which is the player's average marginal contribution. Readers may know the rule from advertising attribution, where it credits each exposure for a conversion \citep{berman2018beyond,singal2022shapley}, or from machine learning, where it credits each input feature for a prediction \citep{strumbelj2014explaining,lundberg2017unified}. I propose a new use of it, to decide what a simulated respondent is responding to. The players are pieces of the prompt. The total is the model's response, measured as the log-odds of answering ``A'' rather than ``B'', a scale on which contributions add.\footnote{For a prompt $x$, the log-odds of answering ``A'' is $f(x)=\log\bigl(p(\mathrm{A};x)/p(\mathrm{B};x)\bigr)$, where $p(\mathrm{A};x)$ and $p(\mathrm{B};x)$ are the probabilities of answering ``A'' and ``B''. It is positive when ``A'' is more likely and negative when ``B'' is.} A piece's Shapley value is the average change in the log-odds when that piece is added to the prompt, over all orders in which the pieces can be added.

I build the game in two versions, because the two unanswered questions call for different players. For the first question (how a change divides among several wording choices), I let the players be candidate edits to an intact prompt, such as replacing ``Flight'' with ``Option'' or a comma with a semicolon, and I include an actual change in a price as one more player, so that the wording effects can be read beside a substantive one. I call this the \emph{edit game}. It divides the change produced by applying all the edits among the edits. For the second question (how much of a prompt's answer comes from each of its words), I let the players be the words of the prompt itself, each shown as a placeholder until it is added. I call this the \emph{word game}. It divides the response of the complete prompt, relative to the prompt with every word replaced by a placeholder, among the words. The same game runs on the model's tokens instead of words. I use words because marketing researchers care about the product attributes and descriptions that a respondent reads, and those are written in words. Because the Shapley values add up to the total, the word game yields one summary number, the share of the credit in magnitude that goes to the attribute values (the prices and travel times). I call it the \emph{attribute share}.

Applied to \openingInvisibleModel{}'s answer to the flight question of Figure~\ref{fig:opening}, the word game gave an unexpected result. The two prices and the two travel times together received \tnExampleAttributeShare\% of the credit. The other forty-five pieces of the prompt (the question, the labels, the connecting words, the instruction, and the line breaks) received the rest. Taken one at a time, a price or a travel time carried more credit than a typical other piece, but there are only four of them against forty-five others. The single most influential piece of the prompt was not a price or a travel time at all. It was the letter ``B'' in the instruction ``Answer with a single letter, A or B.'' Across the sixteen versions of the flight question that vary the two prices and the two travel times, the median attribute share was \tnETwoAttributeShareMedianMuseGlimmer\% for \openingInvisibleModel{}, and the highest median among the four models tested in this paper was \tnETwoAttributeShareMedianGemmaFourThreeOneB\%, for Gemma 4 31B. The credit assigned to the instruction is not an error or noise, since the question needs an instruction. But a customer's choice between two flights is not supposed to rest on it, and in every model tested here, most of the credit went to parts of the question that choice models assume a customer looks past.

These credits are Shapley values, and a Shapley value averages a piece's contribution over many versions of the prompt, most of which the researcher never wrote. I therefore also report, for each piece, what it does to the prompt as written (the effect of removing it) and what it does on its own (the effect of adding it alone to the prompt of placeholders). The three numbers can disagree. Removing a single price or travel time from the complete flight question moves the answer several times as much as that value's Shapley value, because the Shapley value also averages over versions in which few other words are present and a lone price means little. I report the three side by side, so that a reader sees both what a piece earns on average and how much the prompt as written depends on it.

I then turn to a question raised by earlier studies, which report that language models show cognitive biases familiar from human decision making, such as reacting to how the same situation is framed \citep{mei2024turing,cheung2025amplified}. I use my audit to ask whether such a bias belongs to the model or is an artifact of the particular wording that disappears when the same question is worded differently. My test case is \emph{framing}. Writing ``costs only \$200'' instead of ``costs \$200'' is a classic cue meant to make a price look small, and in \openingInvisibleModel{}'s answer to the flight question it works. At the original wording, the cue raised the probability of choosing the cheaper flight from \tnExampleBaseP{} to \tnExampleFramedP{} and reversed the more likely answer. To see whether this effect belongs to the cue or to the wording it was added to, I added the cue to each of the \tnExampleContexts{} versions of the flight question that the ten presentation edits of the edit game produce in combination. Every version asks the same question in slightly different words. The cue lowered the probability of the cheaper flight in \tnExampleContextsNegative{} versions, left it unchanged in \tnExampleContextsZero{}, and raised it in \tnExampleContextsPositive{}. The original wording is one of those \tnExampleContextsPositive{}, yet the cue lowered the probability in more versions than it raised it. Under one wording, with ``costs'' replaced by ``is priced at,'' the cue had no measurable effect at all. For both Gemma models, the cue lowered the probability of the cheaper flight in nearly every version, the opposite of what it is meant to do. A framing effect found in a simulated customer can therefore be an artifact of the wording it was found under, and a researcher who had chosen another reasonable wording could have reported the opposite effect.

The second extension is a \emph{persona}, which describes whom the model is simulating and is placed before the question. Personas are how simulated respondents are given demographics and tastes, and they are meant to change the answer. The surprise is how little is needed. A single sentence stating only that the respondent is a consumer, with no demographics or tastes at all, lowered Gemma 4 12B's probability of choosing the cheaper flight from \tnEOneBasePGemmaFourOneTwoB{} to near one half, and a two-sentence description of a budget-conscious consumer raised it to almost one. Yet the persona itself received little of the credit. The one-sentence persona took about 1\% of the prompt's credit, and across the sixteen flight questions, the median credit that a description took was at most 8\% for every model. A persona thus changes how the model responds to the flights, and its own small share of the credit understates that change.

Finally, robustness checks bound the attribute share. For Qwen 3.8 27B on the sixteen flight questions, the median share was \tnGridShareMedianWord\% when every word was a separate player. It stayed between \tnGridShareMedianProbability\% and \tnGridShareMedianEmpty\% when I replaced the placeholder with an empty string or with random tokens, or when I measured the response as a probability or its logarithm instead of log-odds. The share rose when I grouped words into larger players, because a value grouped with its neighbors takes their credit and contributions of opposite sign cancel inside a pooled player. Even when the four values were the only separate players and every other word was pooled into one, the words that present the products kept more than a third of the credit. I therefore report every share with the players that produced it.

Taken together, these results show that a simulated choice can rest more on how the question is written than on the products it offers. My contribution is an audit that measures this dependence in simulated choice experiments. For a given model and prompt, it says how much of the simulated choice rests on the products and how much on the wording, with one consistent rule for both edits and words. I demonstrate it on three product categories (flights, hotels, and smartphones), with sixteen questions per category that vary two attributes, and on four open-weight models from three developers (Qwen 3.8 27B, Muse Glimmer, Gemma 4 31B, and Gemma 4 12B), whose public parameters let me read exact probabilities and reproduce every number. My audit applies to any model that reports the probabilities of the two answer labels. It describes the model's response. It does not explain the model's internal workings, and it does not tell a researcher whether the model's choices match those of real customers. That question still needs real customers, and my audit tells the researcher what the model was responding to when its choices are checked against theirs. The cost grows with the question. Testing one change needs the model to answer two prompts. An edit game with ten edits and a price change needs about two thousand, and a word game for a prompt of about fifty words needs tens of thousands.

Section~\ref{sec:literature} relates my audit to prior work on simulated respondents, prompt sensitivity, and model explanation. Section~\ref{sec:method} defines the two games and the attribute share and describes the models and the choice questions. Section~\ref{sec:results} reports the presentation sweeps, the edit game beside a real price change, the word profile of the flight question, the framing and persona extensions, and the robustness checks of the attribute share. Section~\ref{sec:discussion} discusses what my proposed audit adds to changing one detail at a time, says how a researcher can run it and at what cost, states the scope of my evidence and the questions left for future research, and concludes with what should accompany a simulated choice, for the researcher who runs the experiment and for the manager who reads it as market research.
 \section{Related literature}\label{sec:literature}

Studies that validate simulated respondents compare their answers with human answers on a chosen task. \citet{goli2024can} did so for choices between sooner and later rewards and found discrepancies that depend on the task. Synthetic survey answers shift with question wording and model version \citep{bisbee2024synthetic}, and language models react to changes in a question that leave human answers unchanged \citep{tjuatja2024llms}. To learn when synthetic data can be trusted, recent work asks where synthetic respondents fail \citep{leng2026prism}, how to evaluate them without a human answer for every question \citep{netzer2026synthetic}, and how a study's choices of model, prompt, and setting each change the simulated data \citep{cummins2026threat}. Nineteen preregistered studies of digital twins, each built from a person's answers to hundreds of questions, found the twins only modestly more accurate than the base model and weakly correlated with the people they stand for \citep{peng2026funhouse}. \citet{gao2025take} advised caution in using language models as surrogates for humans at all, because nearly every approach they tested failed to reproduce the human distribution of play in a money-request game. None of these studies says which part of the prompt produced the answer, which is the measurement my audit adds.

A second line of work treats the prompt itself as the instrument. \citet{brucks2025prompt} showed that the architecture of a prompt (how a request is laid out) induces artifacts in the answers. \citet{gui2023challenge} explained that a prompt that changes a treatment, such as a price, can also change what the model infers about variables it never states. In natural language processing, equivalent formats \citep{sclar2024quantifying}, paraphrased instructions \citep{mizrahi2024state}, and the order of examples in a prompt \citep{lu2022fantastically} all change a model's measured performance on benchmark tasks. The dependence reaches decisions as well. In classic economic games, chatbots change their play under different framings of the same situation \citep{mei2024turing}, most language models flip a moral decision with the wording of the question \citep{cheung2025amplified}, and shopping agents built on language models shift their product choices with the position of a product, a sponsored tag, and small changes to its description \citep{allouah2026agent}. Each of these findings comes from changing one detail of a prompt at a time, and \citet{qin2026interactions} explained such sensitivity through interactions among the parts of a prompt. My audit goes from measuring how much the answer changes to dividing that change among the edits or words that produced it. It asks the researcher to state in advance which of them only present the products (presentation) and which are meant to change the answer (interventions, such as a different price or a framing cue), and it reports how much credit each one receives.

I divide the credit with the Shapley value \citep{shapley1953value}, a rule that is established in advertising attribution \citep{berman2018beyond,singal2022shapley} and in explaining the predictions of machine-learning models \citep{strumbelj2014explaining,lundberg2017unified}. I propose a new use of the rule, in which it divides a simulated respondent's answer among the pieces of the prompt.\footnote{My proposed framework, which divides the credit for a language model's answer among the pieces of its prompt by their Shapley values, first appeared in my working paper, submitted to arXiv on March 29, 2024 (\href{https://arxiv.org/abs/2404.01332v1}{arXiv:2404.01332v1}). Similar token-level Shapley attribution methods for language models proposed after that date are not reviewed here.} After my working paper of March 2024, \citet{ye2025fair} applied the Shapley value to language models for a different purpose. They divided the value of a model's summary among the documents that the summary drew on, so that the creators of those documents can be compensated, and they contrasted this economic goal with that of my working paper. Their question is what each document is worth, and mine is what the model responded to.

The machine-learning literature also shows what a Shapley value depends on, and my audit follows it on two points. First, a Shapley value is measured against a \emph{reference}, which is the input that the model sees when every piece is absent (in my word game, the prompt with a placeholder for every word), and the reference has to be stated \citep{merrick2020explanation}. Second, the same model and the same prompt can receive different Shapley values, depending on how the game is defined \citep{sundararajan2020many}. In my games, that definition consists of the players, the reference, and the scale on which the answer is measured. I state all three with every division of credit that I report, and the robustness checks of Section~\ref{sec:grouping} repeat one model's word games under other choices of each.

My word game is also related to a simpler measure of a word's importance. \citet{hackmann2024word} masked the words of a prompt's instructions one at a time and recorded how the model's outputs changed. Masking one word gives the effect of removing it from the complete prompt, which is not its Shapley value, since a Shapley value averages the word's contribution over many versions of the prompt in which other words are absent as well. My audit reports the effect of removing each word beside that word's Shapley value.

Finally, my attribute share reports how much of the credit for a simulated answer goes to the attribute values that an experiment varies and how much to the words that present them. Conjoint analysis motivates the line between the two \citep{green1990conjoint}, since it describes a product by the levels of its attributes and explains a respondent's choices by those levels. Whether simulated choices match those of real customers is the question of the validation studies above, and my audit complements them. It tells such a study what a simulated answer was responding to, so that when simulated and real choices agree, the study can also ask whether they agree because of the products or because of the way the question was written.
 \section{Framework and design}\label{sec:method}

\subsection{The choice question and its score}\label{sec:setup}

A choice experiment shows a respondent two options, describes each option by its value on every attribute under study, and records which one the respondent picks. The researcher writes the question once and changes only the attribute values between versions, so that a change in the choice can be read as a response to the values. I call the question with an empty slot for each option's value of each attribute a \emph{template}, and the template with its slots filled a \emph{task}. Nothing in my proposed audit depends on how many attributes a template has. The demonstrations in this paper describe each option by two attributes, the smallest number that allows a trade-off. The question of Figure~\ref{fig:opening} is the example I use throughout.

\begin{quote}\small\ttfamily
Which of the following flights from Chicago to Atlanta do you prefer?\\
Flight A: costs \$200, with a total travel time of 5 hours.\\
Flight B: costs \$350, with a total travel time of 2 hours.\\
Answer with a single letter, A or B.
\end{quote}

This template describes each flight by two attributes (price and travel time) and so has four slots. The sixteen flight tasks fill them with every combination of \$200 or \$350 and 2 or 5 hours (Table~\appref{app:prompts:tasks:flight}). Four are ties, ten have one flight better on an attribute and no worse on the other, and two are trade-offs, among them task 7 above. The hotel template compares two hotels in Boston by price per night and distance from the city center, and the smartphone template two phones by price and storage, each with sixteen tasks built the same way (Tables~\ref{app:prompts:tasks:hotel} and~\appref{app:prompts:tasks:smartphone}).

I call everything in a prompt other than the attribute values its \emph{presentation} (the question, the option labels, the connecting words, the answer instruction, and the line breaks), because it presents the values to the respondent. A language model's answer can change when only the presentation changes, as the sweeps of Section~\ref{sec:presentation} show. That dependence is what Section~\ref{sec:introduction} calls token noise, and my audit measures it. The presentation itself is not an error, since no question can be asked without one.

I measure the model's response to a prompt by one number, how strongly the model favors one option over the other. The instruction asks for a single letter, so the model's reply is the first token it generates. Because the models I study have public weights, I do not need to sample that reply. I read the probabilities that the model assigns to the tokens ``A'' and ``B'' at that position and rescale the two to sum to one. I write the rescaled probabilities as $p(\mathrm{A};x)$ and $p(\mathrm{B};x)$ for a prompt $x$, and I take their log-odds:
\begin{equation}\label{eq:payoff}
 f(x)=\log\frac{p(\mathrm{A};x)}{p(\mathrm{B};x)}.
\end{equation}
Here $p(\mathrm{A};x)$ is the probability of choosing ``A'' that a researcher would record. Its log-odds, $f(x)$, is the quantity my games divide, and I call it the \emph{score} of the prompt. The score is positive when ``A'' is the more likely answer and negative when ``B'' is. Rescaling gives the probability that the model would assign to ``A'' if the instruction's two letters were its only possible replies, and Table~\appref{tab:repro:validation} checks this by running every model under that restriction. I divide log-odds rather than probabilities because contributions add on that scale, and because the probability scale compresses answers near zero and one, where a large change of score moves the probability very little. I report probabilities when I describe a choice and log-odds when I divide one.

My proposed audit can also be applied to a closed model such as Claude or GPT. Closed models often do not reveal the full log probabilities of the next token (some return none, and others return only the few most likely tokens, which may not include both ``A'' and ``B''). A researcher would then estimate $p(\mathrm{A};x)$ and $p(\mathrm{B};x)$ by sampling the reply many times under a restriction to the two letters, at the cost of more calls and a sampling error of its own. Requiring a structured answer, as is common with these models, changes the prompt itself, because the required structure, for example a JavaScript Object Notation (JSON) schema, is given to the model as additional tokens, so the requirement must be declared. Everything else in my audit would remain the same. I chose open-weight models because they give exact probabilities and because each of them is a fixed checkpoint (a published file of the model's weights). This gives the paper a lasting advantage, since every result can be reproduced easily from these files years from now, whereas a closed model can change behind the scenes or become obsolete within months.

\subsection{Players and the allocation rule}\label{sec:allocation}

The \emph{players} of my games are edits, words, and line breaks. A game consists of a set $N$ of $n$ players and, for every subset $S$ of $N$ (a \emph{coalition}), a prompt $x_S$ in which exactly the players of $S$ are present. The prompt of the empty coalition, $x_\varnothing$, is the \emph{reference}, and the prompt of the full coalition, $x_N$, is the complete prompt. The value of a coalition is the score of its prompt minus the score of the reference, $v(S)=f(x_S)-f(x_\varnothing)$, with $f$ the score of Equation~\eqref{eq:payoff}, so $v(\varnothing)=0$ and $v(N)$ is the total to be divided. The Shapley value of player $i$ is its average marginal contribution \citep{shapley1953value}:
\begin{equation}\label{eq:shapley-permutation}
 \phi_i(v)=\frac{1}{n!}\sum_{\omega\in\Omega}
 \left[v\bigl(S_i(\omega)\cup\{i\}\bigr)-v\bigl(S_i(\omega)\bigr)\right].
\end{equation}
The sum runs over the set $\Omega$ of the $n!$ orders $\omega$ in which the players can be added, $S_i(\omega)$ is the set of players added before $i$ in order $\omega$, and the bracket is $i$'s \emph{marginal contribution} in that order, the change of score when $i$ joins the players already present. A value is positive when adding the player moves the answer toward ``A'' on average and negative when it moves the answer toward ``B'' on average, and its size is the average movement in log-odds. \citet{shapley1953value} proved that this is the only rule under which the values add up to the total, two players that add the same amount to every coalition that contains neither of them receive the same value, a player that never changes the score receives zero, and adding two games adds their values. Shapley's theorem gives my audit a firm mathematical foundation. Once the players, the reference, and the scale of the score are stated, the division is settled. These three choices define the question that a game answers, just as the attributes and their levels define the question that a conjoint study answers, and I state them with every result.

Beside each Shapley value, I report the player's marginal contribution when it joins first and when it joins last. I report these two because each is a check that a researcher can run directly, by changing one detail of a prompt. Joining first gives the player's \emph{alone} effect, $v(\{i\})$, the change in the score of the reference when the player is the only one added to it. It says what the player does without the others. Joining last gives its \emph{left-out} effect, $v(N)-v(N\setminus\{i\})$, the score that the complete prompt loses when the player is removed. It says how much the prompt as written depends on the player. If the players did not interact, every order would give a player the same marginal contribution, and the alone effect, the left-out effect, and the Shapley value would be equal. When players interact, as when the credit of ``\$200'' depends on whether ``costs'' precedes it, the three numbers differ, and reporting all three shows how much a player's credit depends on which other players are present.

\subsection{The edit game}\label{sec:formB}

The \emph{edit game} answers the first of the Introduction's two questions, how the combined effect of several wording choices divides among them. Its players are edits to an intact prompt, and the prompt $x_S$ of coalition $S$ is the original prompt with exactly the edits in $S$ applied. The reference is the unedited prompt, so the value $v(S)$ is the change in score that the edits in $S$ produce together. Every prompt in this game is a complete question that a researcher could have written, unlike the prompts of the word game, which show absent words as placeholders. What the game needs is the edits, written in advance as defensible equivalents of the original wording.

I call an edit that changes the presentation of a question and leaves its attribute values as they are a \emph{presentation edit}. The edit games use ten presentation edits for the flight template and eleven each for the hotel and smartphone templates. Four of the flight edits change words (``the following'' becomes ``these'' in the question, ``Flight'' becomes ``Option'' and ``costs'' becomes ``is priced at'' in both option lines, and ``Answer with a single letter'' becomes ``Reply with one letter'' in the instruction). The other six change only the formatting (a space after the question mark, a blank line after the question, two spaces after the colon of each option label, a semicolon in place of the comma after each price, one line instead of two for the options, and parentheses around the instruction's two letters). Table~\appref{app:prompts:operations:flight} lists all ten with their positions. A credit in log-odds is hard to judge without something to set it against, so every edit game has one more player, a change that the experiment is designed to measure. This player, the \emph{price flip}, sets the price of the first option to its other level (for Flight A in task 7, from \$200 to \$350). It is an \emph{intervention} rather than a presentation edit, because it changes the choice offered. Because the edits act at fixed positions that do not overlap, every coalition renders exactly one prompt, whatever the order in which its edits are applied (Appendix~\ref{part:formal} states the condition). A game with $n$ players has $2^n$ coalitions, since each player is either present or absent. A flight game, with its ten edits and the price flip, thus has $2^{11}=2{,}048$ coalitions, and a hotel or smartphone game, with one edit more, has twice as many ($2^{12}=4{,}096$).

\subsection{The word game}\label{sec:formA}

The \emph{word game} answers the second question, how much of one prompt's answer comes from each of its pieces. Its players are the pieces of the prompt itself, which I call its \emph{components}. A component could be a single token, and the game runs on tokens in the same way. The components in this paper are words and line breaks instead, for two reasons. First, words are what a researcher writes and can change, so the credit of a word is the credit of something that the researcher controls. Second, every model receives the same words, whereas each model splits the text into tokens in its own way, so a division among words can be set side by side for different models. The prompt $x_S$ of coalition $S$ shows the text of every component in $S$ in its place and a placeholder in place of every other component. The reference $x_\varnothing$ is then the prompt with every component replaced, and $x_N$ is the original prompt. Adding a component to a coalition puts its text back in its place and leaves every other component as it was, so every prompt of the game keeps the components of the original in their original order. As in every game, the value of a coalition is $v(S)=f(x_S)-f(x_\varnothing)$, where $f$ is the score of Equation~\eqref{eq:payoff}, so the word game divides the difference between the scores of the original prompt and the reference among the components.

Unlike the edit game, the word game needs no alternative wordings. It needs only a rule that partitions a prompt into components and a rule for what replaces a component when it is absent. I partition a prompt as follows. Each attribute value with its unit is one component, so ``\$200'' and ``5 hours'' are single players, since a value and its unit mean one thing together. Punctuation after a value is a component of its own, so that the player for the value holds only the value. Every other word is one component with any punctuation attached to it, and each line break is one component. The spaces between words are not players, because a space cannot be absent without joining the words on either side of it, and the effects of spacing are measured directly by the sweeps of Section~\ref{sec:presentation} and by the spacing edits of the edit game. The rule gives 49 components for the flight prompt, 45 for the hotel prompt, and 40 for the smartphone prompt.

An absent word is shown as an underscore (\texttt{\_}), so that the gap is visible. An absent line break or split punctuation mark is removed instead, because an underscore in place of the comma after ``\$200'' would fuse with the price. The reference built this way is a defined baseline, not a neutral respondent \citep{merrick2020explanation}, because a model can respond strongly to a line of underscores and can infer missing information from the words that remain. A prompt with many words replaced is also unfamiliar text, so the model's response to it is an extrapolation \citep{hooker2021unrestricted}. A word's value thus measures the prompt's dependence on that word under the declared replacement, and a different replacement defines a different question \citep{sundararajan2020many}. The robustness checks of Section~\ref{sec:grouping} show that the central finding about the attribute values depends neither on the replacement nor on the scale of the score. They repeated one model's word games under other replacements and on other scales, and under every one of them, the attribute values still received a minority of the credit (Tables~\ref{tab:grid-references} and~\appref{tab:grid-payoffs}).

\subsection{The attribute share}\label{sec:share}

I call the components that hold the attribute values the \emph{attribute players} and write $\mathcal A$ for their set, which has one component for each slot of the template (in the flight prompt, the two prices and the two travel times). Write $\phi_i$ for the Shapley value of component $i$ in the word game. In every game, the Shapley values of all the players add up to the total being divided, $v(N)$, and in the word game that total is the score of the complete prompt minus the score of the reference. Splitting the players into the attribute players and all the others gives
\begin{equation}\label{eq:decomposition}
 f(x_N)=f(x_\varnothing)+\sum_{i\in\mathcal A}\phi_i+\sum_{i\in N\setminus\mathcal A}\phi_i .
\end{equation}
In words, the score of the complete prompt equals the score of the reference, plus the credit of the attribute values, plus the credit of every other component, and those other components make up the presentation. Choice models motivate this split, because they expect a choice to respond to the attribute values and not to change when the same options are presented differently \citep{green1990conjoint,train2009discrete}. The split does not make the credit of the attribute values the respondent's utility and the credit of the presentation an error, however. The presentation includes the answer instruction and the option labels, without which there would be no question to answer. Equation~\eqref{eq:decomposition} describes how the word game divides the score between the attribute values and their presentation, and I read it only as that description.

The two sums of Equation~\eqref{eq:decomposition} do not show how much credit each group receives, because credit toward ``A'' and credit toward ``B'' cancel within a sum. Two prices that move the score equally far in opposite directions, for example, would add up to zero, as if the prices had received no credit. To summarize the division in one number, I count credit without regard to its direction. The \emph{mass} of a group of components is the sum of the absolute values of their Shapley values. I write $M_{\mathcal A}$ for the mass of the attribute players and $M_{\mathrm{other}}$ for the mass of all other components, and I call the attribute players' fraction of the total mass, $q$, the \emph{attribute share}:
\begin{equation}\label{eq:attribute-share}
 M_{\mathcal A}=\sum_{i\in\mathcal A}|\phi_i|,\qquad
 M_{\mathrm{other}}=\sum_{i\in N\setminus\mathcal A}|\phi_i|,\qquad
 q=\frac{M_{\mathcal A}}{M_{\mathcal A}+M_{\mathrm{other}}}.
\end{equation}
That is, the attribute share is the share of all credit, in magnitude, that falls on the attribute values. When every Shapley value is zero, the share is undefined.

The same masses describe any part of the prompt. To make the 49 values of the flight prompt readable, Section~\ref{sec:word-results} shows them grouped by the part of the prompt to which each component belongs (the question, the route, the option labels, the attribute phrases, the answer instruction, and the line breaks), together with the mass of each part. Grouping in this way only adds up values that the word game has already produced, and it changes none of them.

The attribute share also depends on how many players share the credit. If every component of the flight prompt received credit of the same size, the four attribute players would hold four of the 49 equal parts, a share of \tnBenchmarkEqualShare\%. That figure is arithmetic, not a null hypothesis or a threshold, but it shows that a share should be read together with the players that produced it.

When a prompt opens with a persona (a description of the respondent placed before the question), I add the persona's clauses to the word game as players, one for each clause, and report two shares instead of one. The first is the attribute share of the task, whose denominator counts the mass of the question's components but not the mass of the persona, so that it stays comparable with the share of the same task without a persona. The second is the \emph{persona share}, the persona's mass divided by the mass of the whole prompt, persona included, and it says how much of all the credit the description of the respondent takes.

\subsection{Estimation, models, and studies}\label{sec:estimation}

I scored the prompt of every coalition of an edit game, so its Shapley values are exact. A word game is too large for that, since the 49 components of the flight prompt form $2^{49}$ coalitions, and I estimated its Shapley values from a random sample of orders instead \citep{castro2009polynomial}. I drew 1,000 orders of the players at random and, in each order, added the players to the reference one at a time, scoring the prompt after each addition, which gives every player one marginal contribution per order. A player's estimated Shapley value is the average of its 1,000 contributions, and their spread gives the estimate's Monte Carlo standard error. For a mass or an attribute share, I drew 1,000 resamples of the 1,000 orders with replacement, recomputed the quantity in each, and took the range that holds the middle 95\% of the recomputed values as its 95\% interval. These standard errors and intervals reflect only the sampling of orders, since every prompt was fixed and scored once. They say how far an estimate may lie from the exact Shapley value of the same game, and nothing about other tasks, other models, or human respondents. Appendix~\ref{part:measurement} gives the estimators and checks them against a longer run of orders (Figure~\appref{fig:grid-convergence}) and against a game small enough to solve exactly (Table~\appref{tab:grid-calibration}).

I ran every study on four instruction-tuned models (models trained to answer a user's message) from three developers. They are Qwen 3.8 27B, Meta's Muse Glimmer, and Google's Gemma 4 31B and Gemma 4 12B. These were the strongest open-weight models I could run locally, and they demonstrate my proposed audit rather than rank models. Each question was one user message in the model's own chat format, with no example answers, and reasoning was switched off, or in Muse Glimmer set to its lowest strength behind a fixed assistant prefix addressing the user, so that the first token generated was the answer (Table~\appref{tab:repro:models}).

Table~\ref{tab:roadmap} lists the five studies of Section~\ref{sec:results}, with what each varies, its reference, and its players. Two of them have not been described above. The presentation sweeps changed one detail of the flight question at a time and divided nothing, and the framing study added a price cue (``only'' before the cheaper price or ``as much as'' before the pricier one) as one more player in the edit game and in the word game.

\begingroup\long\def\floatNotesBody{
A player is one unit that a game credits: an edit, a word, or a clause. The reference is the prompt that receives no credit. Framing runs two games on the same prompts; every game runs on all four models.
}%
\begin{table}[tbp]
\centering
\caption{The five studies: what each varies and holds fixed, its reference, its players, and where its full results sit online.}\label{tab:roadmap}
\footnotesize\setlength{\tabcolsep}{3pt}
\begin{NiceTabular}{@{}>{\raggedright\arraybackslash}p{1.75cm};>{\raggedright\arraybackslash}p{3.1cm};>{\raggedright\arraybackslash}p{1.6cm};>{\raggedright\arraybackslash}p{2.1cm};>{\raggedright\arraybackslash}p{2.1cm};>{\raggedright\arraybackslash}p{1.75cm};>{\raggedright\arraybackslash}p{2.6cm}@{}}
\toprule
Study & What varies & Held fixed & Reference & Players & Models and tasks & Online \\
\midrule
Presentation sensitivity & characters inserted after the question: 0 to 200 extra line breaks, 0 to 50 trailing spaces, and 0 to 20 zero-width spaces; the long counts are stress tests & every word & the unchanged prompt & none; one change at a time & 4 models; flight task 7 at every count, all 16 flight tasks at 1, 5, 20, and 100 line breaks & Figure~\appref{fig:e1} \\
\dottedrule
Edit attribution & 10 or 11 retained wording and formatting edits and the price flip, in every combination & the rest of the intact prompt & the unedited prompt & 11 or 12; exact over 2,048 or 4,096 combinations & 4 models; the 16 tasks of each of three templates & Figures~\ref{fig:edit-games-hotel} and~\ref{fig:edit-games-smartphone} and Table~\appref{tab:edit-games} \\
\dottedrule
Word attribution & which words are present & word order and spacing & every word masked by \texttt{\protect\frenchspacing \_}, line breaks and split punctuation removed & 49, 45, or 40 components; 1,000 sampled orders & 4 models; all 48 tasks & Figures~\ref{fig:shapley-e2-flight-qwen3-8-27b} to~\appref{fig:shapley-e5-smartphone-gemma4-12b} \\
\dottedrule
Framing & the phrase ``only'' before the cheaper price and ``as much as'' before the pricier price: in the edit game, with and without each combination of the ten retained edits; in the word game, which words of the framed prompt are present & the attribute values & edit game: the same prompt without the phrase; word game: every word masked by \texttt{\protect\frenchspacing \_} & edit game: the phrase and 10 edits, 1,024 paired contexts; word game: the 50 components, the phrase among them, 1,000 sampled orders & 4 models; the eight flight tasks with unequal prices & Tables~\ref{tab:framing-summary} and~\ref{tab:framed-words} and Figure~\appref{fig:e4-framing} \\
\dottedrule
Personas & a system message describing the respondent: none, or one of three descriptions of 1, 2, or 5 clauses & the flight task & every word and clause masked & the 49 components and the clauses; 1,000 sampled orders & 4 models; all 16 flight tasks & Table~\ref{tab:persona-shares} and Figure~\appref{fig:e6} \\
\bottomrule
\end{NiceTabular}
\end{table}
\endgroup{}
\floatnotes{roadmap}
 \section{Results}\label{sec:results}

In this section, I report the five studies listed in Table~\ref{tab:roadmap}, in that order. I begin each study with flight task 7 (the question of Figure~\ref{fig:opening}) and then turn to the other tasks. I use one model, Muse Glimmer, for the examples of the edit game, the word game, and framing, so that a reader can follow one model's answer to one question through all three studies. The persona example uses Gemma 4 12B. I end the section with robustness checks of the attribute share, which vary the choices behind it (the placeholder for an absent word, the scale of the score, the division of the prompt into players, and what counts as an attribute) for Qwen 3.8 27B on all sixteen flight tasks.

\subsection{Presentation dependence}\label{sec:presentation}

Without changing a word, I inserted zero-width spaces, trailing spaces, or blank lines after the question of flight task 7 and scored the prompt at every count for each of the four models (Figure~\appref{fig:e1}). I read each probability directly from the model instead of estimating it from sampled answers, so none of the movement below is sampling variation.

Invisible characters move the answer, and not in proportion to their number. Four zero-width spaces raised Muse Glimmer's probability of choosing Flight A (the cheaper flight) from \tnEOneOpeningZeroWidthSpacesPAtZeroMuseGlimmer{} to \tnEOneOpeningZeroWidthSpacesPAtFourMuseGlimmer{} and reversed the more likely answer (Figure~\ref{fig:opening}), and a fifth brought it to an even \tnEOneOpeningZeroWidthSpacesPAtFiveMuseGlimmer. One more character that no reader can see can push the answer either way.

Blank lines can move the probability far without changing the recorded choice. Five blank lines lowered Gemma 4 12B's probability of the cheaper flight from \tnEOneOpeningNewlinesPAtZeroGemmaFourOneTwoB{} to \tnEOneOpeningNewlinesPAtFiveGemmaFourOneTwoB, a drop that a researcher who records only the more likely answer never sees. Over the whole sweep, the probability ranged from \tnEOneNewlinesMinPGemmaFourOneTwoB{} to \tnEOneNewlinesMaxPGemmaFourOneTwoB{} (a range of \tnEOneNewlinesRangeFGemmaFourOneTwoB{} log-odds), and at some counts ``B'' became the more likely answer.

A probability that looks fixed can hide the largest movement. Under line breaks, Gemma 4 31B gave the cheaper flight a probability of almost zero at every count (never above one in a thousand), so ``B'' was always the more likely answer. Yet its score moved across \tnEOneNewlinesRangeFGemmaFourThreeOneB{} log-odds, the widest range of the four models, because the probability scale compresses answers near zero and one (Section~\ref{sec:setup}). A study that read only the probabilities would call this model stable.

None of this is peculiar to task 7. For Muse Glimmer and both Gemma models, at least one of 1, 5, 20, or 100 extra line breaks changed the score on every one of the sixteen flight tasks (Figure~\appref{fig:e1}, last column). A sweep shows that the answer moves, but not which parts of the prompt the model is responding to, and that is what the two games divide.

\subsection{The edit game beside a price change}\label{sec:edit-results}

Wording that leaves the choice untouched can earn as much credit as a change in price. In the edit game, the ten presentation edits of Section~\ref{sec:formB} compete for credit with a \$150 change in Flight A's price (Figure~\ref{fig:edit-games}). For Qwen 3.8 27B and Muse Glimmer, the ten edits together received more credit in magnitude than the price change on half of the sixteen flight tasks. For the two Gemma models, the price change outweighed all ten edits together on three quarters of the tasks. The hotel and smartphone templates repeated the pattern (Figures~\ref{fig:edit-games-hotel} and~\ref{fig:edit-games-smartphone} and Table~\appref{tab:edit-games}). The credit measures how far each change moves the answer, not what a word is worth in dollars.

\begingroup\long\def\floatNotesBody{
Points are tasks; bars are medians. Values are exact and allocate the change from the unedited prompt. The flip raises the price of ``A'' on tasks 1 to 8 and lowers it on tasks 9 to 16. The axis is linear within $\pm0.1$, logarithmic beyond.
}%
\begin{figure}[tbp]
\centering
\includegraphics[width=\textwidth]{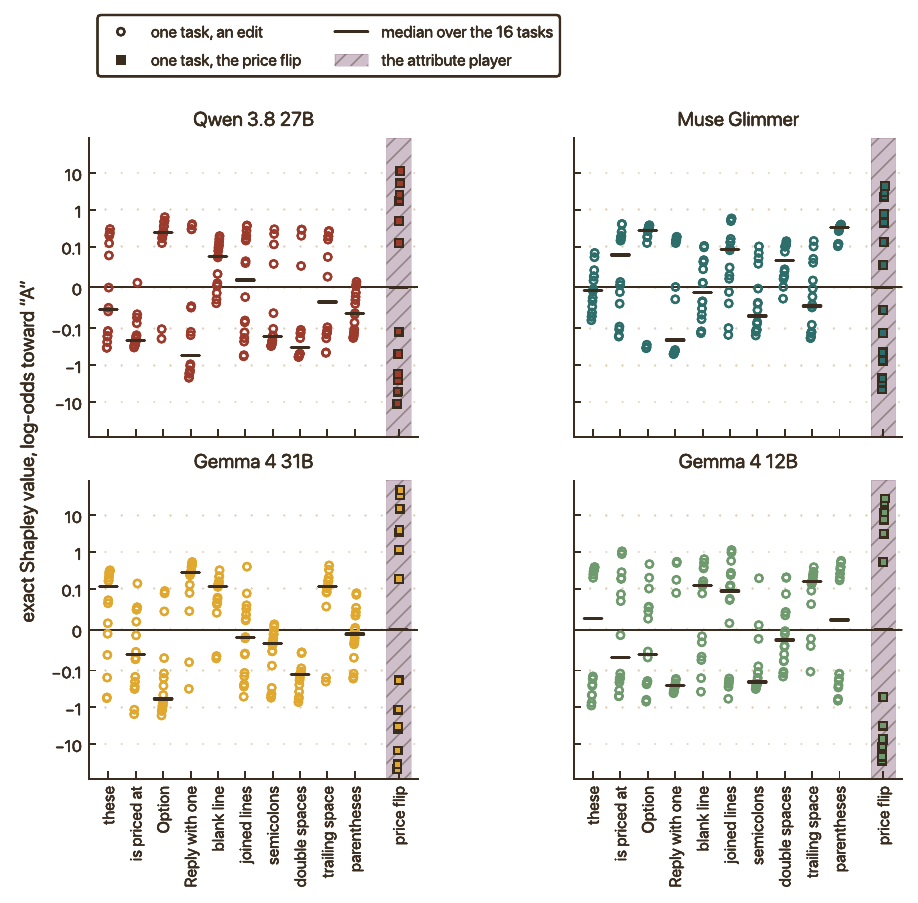}
\caption{Exact Shapley values of the ten retained edits and the price flip over the 16 flight tasks, by model.}\label{fig:edit-games}
\end{figure}
\endgroup{}
\figurenotes{edit_games_flight}

The wording also changes what the price does. Raising the price of ``A'' by \$150 moved Muse Glimmer's score on task 7 by \tnExamplePriceAlone{} log-odds at the original wording and by \tnExamplePriceWithEdits{} once all ten edits were in place (Table~\ref{tab:matched}). The effect of the one change the experiment is designed to measure thus differs by about a quarter between two wordings of the same question.

An edit's effect can even reverse with the wording around it. Joining the two option lines raised Muse Glimmer's score on task 7 when it was the only edit made, and it lowered the score by the same amount when it was the last of the ten edits to be added (Table~\ref{tab:matched}). Its Shapley value in the game of the ten edits alone (\tnExampleWordingGameEditPhiOpZeroEight{} log-odds) averages those two contexts and every one between. The effect of an edit at the original wording is not its effect in an edited prompt, and the average is neither. Across models, templates, and tasks, an edit's Shapley value agreed in sign with its alone and left-out effects on most edits but not on all (Table~\appref{tab:matched-summary}).

\begingroup\long\def\floatNotesBody{
Values are log-odds toward ``A''. Panel A's Shapley values come from the ten-edit game at the original price; Figure~\ref{fig:edit-games} adds the price flip as a player, so the price row holds only its alone and left-out effects. Alone: the edit on the unedited prompt, or the word restored to the all-masked prompt. Left out: the score the complete prompt loses without the component. se: Monte Carlo standard error.
}%
\begin{table}[tbp]
\centering
\caption{Shapley values beside the alone and left-out effects on Muse Glimmer's flight task 7: the retained edits and the price flip (A) and the words (B).}\label{tab:matched}
\footnotesize\setlength{\tabcolsep}{4pt}
\begin{NiceTabular}{@{}l;r;r;r;r@{}}
\toprule
Component & Shapley value & se & Alone & Left out \\
\midrule
\multicolumn{5}{@{}l}{\textit{A. Edits at the original price, exact over $1{,}024$ combinations; price flip: alone and left-out effects only}} \\
``the following'' to ``these'' & $-0.09$ &  & $0.00$ & $-0.25$ \\
\dottedrule
``costs'' to ``is priced at'' & $0.09$ &  & $0.38$ & $-0.12$ \\
\dottedrule
``Flight'' to ``Option'' & $0.34$ &  & $0.50$ & $0.25$ \\
\dottedrule
``Answer with a single'' to ``Reply with one'' & $-0.15$ &  & $-0.25$ & $-0.12$ \\
\dottedrule
\textit{blank line after the question} & $0.03$ &  & $0.12$ & $0.00$ \\
\dottedrule
\textit{join the option lines} & $-0.08$ &  & $0.38$ & $-0.38$ \\
\dottedrule
\textit{comma to semicolon after the prices} & $0.11$ &  & $0.12$ & $0.00$ \\
\dottedrule
\textit{two spaces after the label colons} & $0.06$ &  & $0.25$ & $0.00$ \\
\dottedrule
\textit{trailing space after the question mark} & $-0.01$ &  & $0.25$ & $0.00$ \\
\dottedrule
\textit{options in parentheses} & $0.20$ &  & $0.25$ & $0.25$ \\
\dottedrule
price of ``A'' flipped &  &  & $-3.25$ & $-2.50$ \\
\addlinespace
\multicolumn{5}{@{}l}{\textit{B. Words: the attribute values and the four largest other words, $1{,}000$ sampled orders}} \\
``\$200'' & $0.19$ & $0.02$ & $0.25$ & $1.88$ \\
\dottedrule
``5 hours'' & $-0.20$ & $0.02$ & $0.25$ & $-1.94$ \\
\dottedrule
``\$350'' & $-0.03$ & $0.02$ & $-0.25$ & $1.50$ \\
\dottedrule
``2 hours'' & $-0.37$ & $0.03$ & $0.06$ & $-2.31$ \\
\dottedrule
``B.'' & $-1.30$ & $0.05$ & $-4.94$ & $-2.00$ \\
\dottedrule
``B:'' & $-0.95$ & $0.05$ & $-3.88$ & $-1.75$ \\
\dottedrule
``or'' & $-0.72$ & $0.04$ & $-0.12$ & $-3.06$ \\
\dottedrule
``A'' & $0.44$ & $0.04$ & $0.69$ & $-1.13$ \\
\bottomrule
\end{NiceTabular}
\end{table}
\endgroup{}
\floatnotes{matched}

\subsection{The word profile of one prompt}\label{sec:word-results}

The prices and travel times received little of the credit for Muse Glimmer's answer to task 7. The four attribute values together held \tnExampleAttributeShare\% of all credit in magnitude (95\% interval from \tnExampleAttributeShareLow{} to \tnExampleAttributeShareHigh\%), and the other 45 components of the prompt held the rest (Figure~\ref{fig:word-profile}). The single most influential component was the ``B.'' that ends the instruction, with a Shapley value of \tnETwoExampleTopOtherPhiMuseGlimmer{} log-odds (toward ``B''). The instruction as a whole held more credit than any other part of this prompt, and a median \tnRetainedPartShareMedianAnswerInstructionMuseGlimmer\% of all credit across Muse Glimmer's sixteen flight tasks. This model's answer rests more on how it was told to answer than on the flights.

\begingroup\long\def\floatNotesBody{
Filled squares are the attribute values, circles context words, and triangles function words, punctuation, and line breaks; bars span the 95\% Monte Carlo interval. Shaded blocks are the prompt's parts. Values allocate the change from the all-masked reference to the complete prompt.
}%
\begin{figure}[tbp]
\centering
\includegraphics[width=\textwidth]{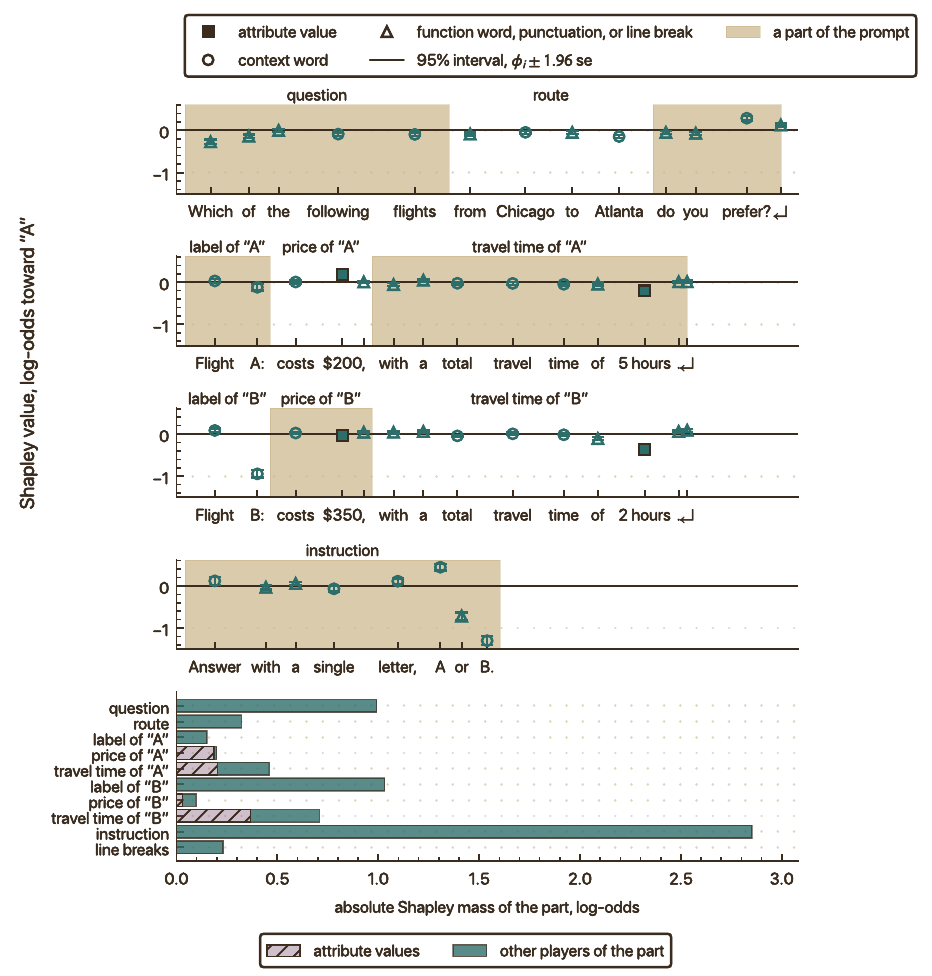}
\caption{Muse Glimmer's word profile on flight task 7: each word's Shapley value, grouped by prompt part, and each part's absolute mass.}\label{fig:word-profile}
\end{figure}
\endgroup{}
\figurenotes{word_profile}

The low share does not mean that the models cannot read the prices. On the ten flight tasks in which one flight is better on an attribute and no worse on the other, every model chose the better flight on all ten, by a margin of at least \tnDiagnosticsMinMarginMuseGlimmer{} log-odds, so that it gave the better flight a probability above 0.9 (Table~\appref{tab:dominance}). When one flight is plainly better, the models choose it decisively. What the share adds is that most of the credit for an answer, relative to a prompt of underscores, goes to words other than the prices and travel times.

The share stayed a minority in every word game I ran (four models, three templates, and sixteen tasks each), and it depended more on the model than on the product. Muse Glimmer's median share was below a tenth on flights, hotels, and smartphones alike, and the highest of the twelve medians (four models on three templates) was \tnEFiveAttributeShareMedianSmartphoneGemmaFourThreeOneB\%, for Gemma 4 31B on smartphones (Figures~\ref{fig:shapley-e2-flight-qwen3-8-27b} to~\appref{fig:shapley-e5-smartphone-gemma4-12b}).

A Shapley value can also understate how much the prompt as written depends on a word. Removing any one attribute value from Muse Glimmer's complete prompt moved the score by \tnExampleAttributeRemovalMin{} to \tnExampleAttributeRemovalMax{} log-odds, several times the value's Shapley value (Table~\ref{tab:matched}, panel B). The Shapley value also averages over prompts in which few words are present, and there a lone price or travel time means little. The two can even disagree in sign. The \$350 of ``B'' received a Shapley value of \tnExamplePhiPriceB{} log-odds (slightly toward ``B''), yet in the complete prompt it pushed the score toward ``A'' by more than a full log-odds unit. Table~\appref{tab:matched-summary} shows such disagreements across models and templates.

\subsection{Framing}\label{sec:framing}

Framing is the first of the two manipulations made on purpose. I inserted ``only'' before the \$200 of ``A'' in Muse Glimmer's task 7 (a cue meant to make the price look small). At the original wording, the cue raised the probability of the cheaper flight from \tnExampleBaseP{} to \tnExampleFramedP{} and reversed the more likely answer. A researcher who tested the cue on this one wording would report a framing effect.

For Muse Glimmer, that effect belongs to the wording, not to the cue. I added the cue to each of the \tnExampleContexts{} wordings of the question that the ten presentation edits produce in combination, and in each I measured what inserting ``only'' did (Figure~\ref{fig:framing}, right). The cue lowered the probability of the cheaper flight in \tnExampleContextsNegative{} wordings, left it unchanged in \tnExampleContextsZero, and raised it in \tnExampleContextsPositive. The original wording is one of the \tnExampleContextsPositive{} in which the cue raised the probability, yet the cue lowered it in more wordings than it raised it. Averaged over the wordings, the cue's effect was close to zero, both as a plain mean (\tnExampleContextsMean{} log-odds) and as its Shapley value (\tnExampleFramePhi). With ``costs'' replaced by ``is priced at'', the prompt received the same score with the cue and without it. A researcher who had written ``is priced at \$200'' would have found no framing effect, and one who had chosen any of those \tnExampleContextsNegative{} other wordings would have found the opposite effect.

\begingroup\long\def\floatNotesBody{
Left: the attribute values, the phrase, and each word whose change exceeds twice its contrast standard error, with 95\% intervals. Right: the phrase's effect in each of the \tnExampleContexts{} edit contexts; marked: the unedited context, the mean, and the frame's Shapley value.
}%
\begin{figure}[tbp]
\centering
\includegraphics[width=\textwidth]{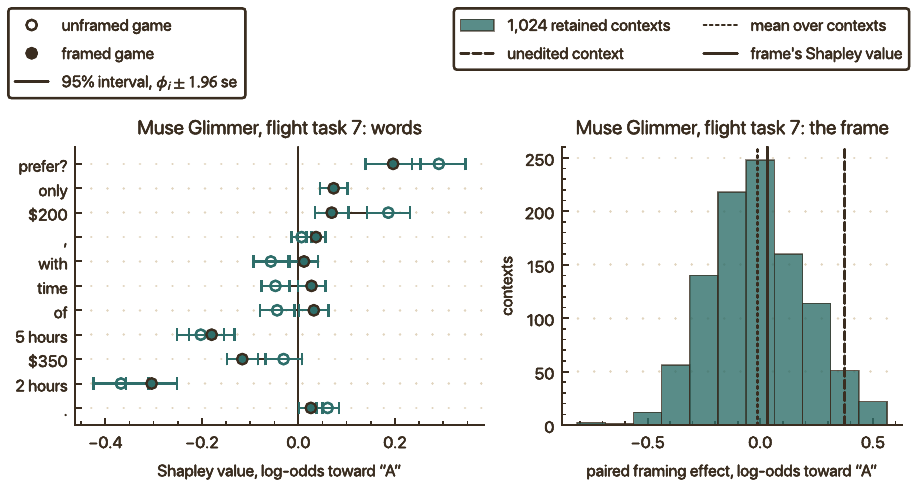}
\caption{The phrase ``only'' before the price of ``A'' on Muse Glimmer's flight task 7: the word game before and after (left) and the paired framing effect over retained contexts (right).}\label{fig:framing}
\end{figure}
\endgroup{}
\figurenotes{framing}

The cue also takes credit away from the very price it was meant to emphasize. In the word game of the framed prompt, the price of ``A'' fell from \tnExamplePhiPriceA{} to \tnExamplePhiPriceAFramed{} log-odds of credit (a change of \tnExampleChangePriceA, standard error \tnExampleChangePriceASe), and ``only'' itself received \tnExamplePhrasePhi{} (Figure~\ref{fig:framing}, left). The cue does not add its own credit on top of an unchanged answer. It moves credit across the prompt. The framed word game has one more player than the unframed one, and so one more underscore in its reference, which is why its profile sits beside the unframed one rather than dividing the cue's effect.

Beyond this example, the cue often backfired. For Muse Glimmer, ``only'' lowered the probability of the cheaper flight at the original wording on seven of the eight flight tasks with unequal prices, and task 7 was the one exception. For both Gemma models, it lowered the probability of the cheaper flight on task 7 itself in nearly every one of the wordings above (Table~\appref{tab:framing-summary}). Figure~\appref{fig:e4-framing} repeats the analysis with a second cue (``as much as'' before the pricier price), and Table~\appref{tab:framed-words} gives the word game of every framed prompt.

\subsection{Personas}\label{sec:personas}

The second manipulation is a persona, which describes the respondent in a system message placed before the question so that the model answers as that person. Even a description that says nothing about the respondent changes the answer. Without a persona, Gemma 4 12B strongly preferred the cheaper flight of task 7. The one-sentence persona ``You are a consumer'', which gives the model no tastes and no demographics, pulled that probability down to near one half while taking about 1\% of the prompt's credit (Table~\ref{tab:personas}). A sentence that tells the model nothing about the person turns a strong preference into a toss-up.

\begingroup\long\def\floatNotesBody{
Attribute share: value mass divided by task-word mass; persona share: clause mass divided by whole-prompt mass. Brackets: 95\% joint-row bootstrap intervals. Without a persona, its share is structurally zero. Near-one probabilities show their distance from one.
}%
\begin{table}[tbp]
\centering
\caption{The persona example: Gemma 4 12B on flight task 7 without a persona and under three descriptions of the respondent.}\label{tab:personas}
\footnotesize\setlength{\tabcolsep}{4pt}
\begin{NiceTabular}{@{}>{\raggedright\arraybackslash}p{6.6cm};r;r;r@{}}
\toprule
System message & \begin{tabular}[t]{@{}r@{}}Probability\\of ``A''\end{tabular} & \begin{tabular}[t]{@{}r@{}}Attribute share\\of the task (\%)\end{tabular} & \begin{tabular}[t]{@{}r@{}}Persona share\\of the prompt (\%)\end{tabular} \\
\midrule
none & $0.915$ & \begin{tabular}[t]{@{}c@{}}$21.2$\\$[19.2, 22.8]$\end{tabular} & \begin{tabular}[t]{@{}c@{}}$0.0$\\$[0.0, 0.0]$\end{tabular} \\
\dottedrule
\texttt{\protect\frenchspacing You are a consumer.} & $0.562$ & \begin{tabular}[t]{@{}c@{}}$33.5$\\$[31.2, 35.5]$\end{tabular} & \begin{tabular}[t]{@{}c@{}}$1.0$\\$[0.2, 2.0]$\end{tabular} \\
\dottedrule
\texttt{\protect\frenchspacing You are a budget-\allowbreak{}conscious consumer. You compare prices before you buy.} & $1 - 1.9\times10^{-7}$ & \begin{tabular}[t]{@{}c@{}}$45.8$\\$[42.6, 48.2]$\end{tabular} & \begin{tabular}[t]{@{}c@{}}$12.6$\\$[10.9, 14.0]$\end{tabular} \\
\dottedrule
\texttt{\protect\frenchspacing You are a budget-\allowbreak{}conscious consumer. You compare prices before you buy. You are 35 years old. You work full time as an accountant. You travel a few times a year.} & $1 - 8.0\times10^{-8}$ & \begin{tabular}[t]{@{}c@{}}$42.5$\\$[39.8, 44.7]$\end{tabular} & \begin{tabular}[t]{@{}c@{}}$8.1$\\$[6.7, 9.8]$\end{tabular} \\
\bottomrule
\end{NiceTabular}
\end{table}
\endgroup{}
\floatnotes{personas}

The two descriptions that call the respondent budget-conscious pushed the probability of the cheaper flight to almost one, while taking about a tenth of the credit for themselves. They also doubled the share of the task's credit that goes to the prices and travel times. Across the sixteen flight tasks, every model's median attribute share was higher under each description than without one, while the description itself took at most 8\% of the credit (Table~\ref{tab:persona-shares} and Figure~\appref{fig:e6}). For Gemma 4 12B, the median rose from \tnETwoAttributeShareMedianGemmaFourOneTwoB\% without a persona to as much as \tnESixAttributeShareMedianLevelTwoGemmaFourOneTwoB\% with one. A persona is thus not a layer added on top of a fixed response to the flights. It changes how the model responds to the flights, and its own credit understates that change.

\subsection{Robustness checks of the attribute share}\label{sec:grouping}

I built the word game on three choices (the placeholder for an absent word, the scale of the score, and the division of the prompt into players), and the share adds a fourth (what counts as an attribute). I held Qwen 3.8 27B and the sixteen flight prompts fixed and changed one choice at a time.

The placeholder and the scale barely move the share. With an underscore for each absent word and the score in log-odds, as in Section~\ref{sec:word-results}, the median share was \tnGridShareMedianWord\%. Replacing the underscore with an empty string or with random tokens from the model's vocabulary (five separate draws), or replacing the log-odds with the probability of ``A'' or its logarithm, gave eight alternatives, and under all of them the median stayed between \tnGridShareMedianProbability{} and \tnGridShareMedianEmpty\% (Tables~\ref{tab:grid-references} and~\appref{tab:grid-payoffs}). The minority that the prices and travel times hold does not rest on the underscore or on the log-odds scale.

Grouping words into larger players raises the share. Grouping each value with the words around it (such as ``costs \$200'') into ten spans raised the median share to \tnGridShareMedianSpan\%, because a span's credit includes that of the surrounding words. Keeping the four values as the only separate players and pooling every other word into one raised it to \tnGridShareMedianAdversarial\%, because pooling lets opposing contributions cancel inside one player and changes how the credit for interactions is split. Even that single player kept more than a third of the credit (Table~\ref{tab:grouping}), and pooling can also lower a share (Appendix~\ref{part:formal} gives a case). Across the three groupings, the words that present the products thus kept between a third and about four fifths of the credit (Tables~\ref{tab:grid-resolutions} and~\appref{tab:grid-adversarial} give every grouping).

\begingroup\long\def\floatNotesBody{
Medians and bracketed interquartile ranges across tasks. The word-level values are Monte Carlo estimates, and the ten-span and pooled games are exact. Attribute spans include surrounding words. One model establishes no invariance across models. Span membership: Table~\appref{tab:span-membership}.
}%
\begin{table}[tbp]
\centering
\caption{The attribute share under three groupings of the same 16 flight prompts into players, Qwen 3.8 27B.}\label{tab:grouping}
\footnotesize\setlength{\tabcolsep}{4pt}
\begin{NiceTabular}{@{}l;r;r@{}}
\toprule
Grouping & Players & Attribute share (\%) \\
\midrule
every word a player & $49$ & \begin{tabular}[t]{@{}c@{}}$19.1$\\$[14.6, 24.5]$\end{tabular} \\
\dottedrule
ten spans, each value with its surrounding words & $10$ & \begin{tabular}[t]{@{}c@{}}$40.8$\\$[35.8, 44.1]$\end{tabular} \\
\dottedrule
the four values and one pooled remainder & $5$ & \begin{tabular}[t]{@{}c@{}}$64.7$\\$[53.9, 71.9]$\end{tabular} \\
\bottomrule
\end{NiceTabular}
\end{table}
\endgroup{}
\floatnotes{grouping}

What counts as an attribute matters as well. When I also counted as attributes the content words that every task shares (among them ``flights'', ``costs'', the option labels, and the ``A'' and ``B'' of the instruction), the median share rose to a majority (Table~\appref{tab:grid-classifications}). Those words are the same in every task, so the rise measures the credit they hold, while the prices and travel times kept the same \tnGridShareMedianWord\%.
 \section{Discussion and Conclusion}\label{sec:discussion}

\subsection{What my proposed audit adds to changing one detail at a time}

A researcher who simulates a choice experiment can always change one detail of the prompt and ask the model again, and Figure~\ref{fig:opening} needs nothing more to show that the answer moves when only the text moves. Such a check cannot divide a change among several details, however, because the effect of one detail depends on the others. Take two changes to the flight question of Figure~\ref{fig:opening}, a \$150 increase in the price of Flight A and the ten presentation edits of Section~\ref{sec:edit-results} taken together. A check of the price increase at the original wording found that it moved the log-odds that Muse Glimmer chooses Flight A by \tnExamplePriceAlone{}. If effects simply added up, the price increase would move the log-odds by the same amount after the ten edits have been applied. It moved them by \tnExamplePriceWithEdits{} instead. Making both changes therefore does not move the log-odds by the price's effect at the original wording plus the edits' effect at the original price, and these two checks cannot say how much of the combined effect belongs to the price and how much to the wording. The Shapley value answers that question. It gives each change one number, its effect averaged over every order in which the changes can be made, and these numbers do add up to the combined effect. This is what allows Section~\ref{sec:edit-results} to say that the ten presentation edits together can receive more credit than the price change. My games need no new kind of data for this, only more versions of the same question. They also need no probability distribution over the ways in which a question could be written, which sets them apart from the sensitivity analysis of simulation models, where the inputs are treated as random and the variance of the output is divided among them \citep{owen2014sobol,song2016shapley}.

The Shapley value is an average, and that is also its limit. A word or an edit can receive a Shapley value near zero in two different cases. In the first case, it changes the answer very little whichever other words or edits are present. In the second case, it pushes the answer toward ``A'' or toward ``B'' depending on which of them are present, and the pushes offset each other in the average. The edit that joins the two option lines is an example of the second case (Section~\ref{sec:edit-results}). A Shapley value does not show which of the two cases produced it. For that reason, I report two more numbers with every Shapley value, the effect of the word or edit when no other is present (its alone effect) and what the complete prompt loses without it (its left-out effect).

\subsection{Running my proposed audit and its cost}

I propose that a researcher who simulates a choice experiment audit it in three steps and stop at the step that answers the researcher's question. The first step changes one detail of the prompt at a time, such as a line break or a synonym, and asks whether the answer survives changes that should not matter. The second step is the edit game, which says how the effect of several such changes divides among them. When the design has a change that the experiment is meant to measure, such as a change in price, the researcher should make that change one of the players. The game then measures the price change and the wording edits in the same unit, so the researcher can see which of them receives more credit (as in Section~\ref{sec:edit-results}). The third step is the word game, which says how much of the answer to one prompt depends on the attribute values. Before the model answers any prompt, the researcher should write down a plan. The plan lists the details to be changed in the first step and the players of the edit game, and it says of each of them whether it only rewords the question or is meant to move the answer. For the word game, it states how the prompt is divided into pieces and what stands in for a missing piece.

A researcher who decides after seeing the answers can, knowingly or not, keep the wording that gives the hoped-for result, which is the prompt hacking concern of the Introduction. False findings are already common in experiments on real customers. In the randomized tests that firms run on their websites, about one in five of the effects that reach significance at the 5\% level is estimated to be null \citep{berman2022false}. A simulated experiment adds a source of its own, since its wording can be changed until an effect appears.

Table~\ref{tab:procedure} gives the cost of each step for one choice question, counted as an upper bound on the number of prompts that the model must answer. Changing one detail takes \tnProcedureDirectEvaluations{} prompts, an edit game with ten edits and a price change takes \tnProcedureEditsEvaluations{}, and a word game for the 49 pieces of the flight question takes \tnProcedureWordsEvaluations{} at 1,000 sampled orders. A researcher can lower the cost in two ways. The first is to use larger players, such as ten spans of words (\tnProcedureSpansEvaluations{} prompts) or the four attribute values and one player for everything else (\tnProcedurePooledEvaluations{} prompts). Larger players do more than lower the cost, however. They change the game that is played, and the new game divides the credit differently (Section~\ref{sec:grouping} and Appendix~\ref{part:constructions}). The second is to audit a sample of the questions in a study and to report how the sample was chosen, because an audit of one question describes that question only.

\begingroup\long\def\floatNotesBody{
Evaluations count logical configurations per task before deduplication; a prompt shared across configurations is scored once.
}%
\begin{table}[tbp]
\centering
\caption{Analyses for one choice task, from a single change to the full word profile, with their logical evaluations and their use.}\label{tab:procedure}
\footnotesize\setlength{\tabcolsep}{4pt}
\begin{NiceTabular}{@{}>{\raggedright\arraybackslash}p{4.8cm};r;r;>{\raggedright\arraybackslash}p{5.7cm}@{}}
\toprule
Analysis & Players & Evaluations per task & When useful \\
\midrule
One edit or one insertion, scored with and without it & $1$ & $2$ & one observed change of the response; nothing is allocated \\
\dottedrule
The six formatting edits & $6$ & $64$ & formatting sensitivity of one wording; willingness to pay across formattings \\
\dottedrule
The formatting edits and the price flip & $7$ & $128$ & a price change beside the formatting edits \\
\dottedrule
The ten retained edits and the price flip & $11$ & $2{,}048$ & allocation across specified edits and their combinations, the price flip as yardstick \\
\dottedrule
The four attribute values and one pooled remainder & $5$ & $32$ & the coarsest word game: values against the rest, with nothing inside the remainder examined \\
\dottedrule
Ten spans & $10$ & $1{,}024$ & parts of the prompt; the words around a value count with it \\
\dottedrule
Every word, 1,000 sampled orders & $49$ & $50{,}000$ & the full word profile; estimated, with Monte Carlo error \\
\bottomrule
\end{NiceTabular}
\end{table}
\endgroup{}
\floatnotes{procedure}

\subsection{Scope and future research}

I ran my audit on short questions that describe each option by two attributes and are answered with a single letter, which is what allowed me to change one price or one word at a time and to tie the movement of the answer to that change. Simulated studies in marketing often go further, with longer interviews, questionnaires that adapt to earlier answers, and more product categories than the three I used. My two games are defined for any prompt and apply to such studies as they stand, but the limits of my hardware kept me from running them there. A model that answers in a sentence instead of a letter would also need a different measure of its answer. My audit also describes what a model responds to and not why, so it cannot say whether a model's architecture, its training data, or its tokenizer is behind a response.

Two questions are left for future research. The first is validation. A study that checks simulated choices against real ones could add my audit, to learn whether the simulated choices responded to the products or to the way the question was written. The second is progress in the models. A newer model may depend less on the presentation of a question, and my audit, run on that model, would show how much less.

\subsection{Conclusion}\label{sec:conclusion}

This paper opened with a flight question whose answer moved when invisible characters were inserted after it. My contribution is an audit that says what such an answer was responding to. It treats the pieces of a prompt as players (edits in one version and words in the other), divides the simulated choice among them with the Shapley value, reads the division beside the effects of adding each piece alone and of removing it, computed from the same answers of the model, and summarizes it in the attribute share (the credit that falls on the products). In four models and three product categories, that share was a minority. A framing effect reversed under many equivalent wordings of the flight question, and the share (a minority under every reference and scale I tried) moved with the choice of players. None of these results identifies a wording under which a model's answer stops depending on how the question is written, so clear writing remains good practice and not a repair. The practical message (for a researcher who runs a simulated choice experiment and for a manager who reads one as market research) is to report a simulated choice with what produced it. That means the effects of single wording changes, which show whether the choice survives equivalent wordings, and, where the question is how the choice divides among the pieces of the prompt, the Shapley values with their players, reference, scale, and uncertainty. A reader can then tell a preference from its presentation.
 \fi

\bibliographystyle{plainnat}

\clearpage
\setstretch{1.2}
\setcounter{section}{0}\setcounter{figure}{0}\setcounter{table}{0}\setcounter{equation}{0}
\renewcommand{\thesection}{\Alph{section}}
\counterwithin{figure}{section}\renewcommand{\thefigure}{\thesection\arabic{figure}}
\counterwithin{table}{section}\renewcommand{\thetable}{\thesection\arabic{table}}
\counterwithin{equation}{section}
\locatelabels
\makeatletter\renewcommand{\table}[1][]{\@float{table}[htp]}\renewcommand{\figure}[1][]{\@float{figure}[htp]}\makeatother
\AddToHook{cmd/section/before}{\FloatBarrier}\AddToHook{cmd/subsection/before}{\FloatBarrier}
\renewcommand{\theHsection}{\Alph{section}}
\renewcommand{\refname}{Appendix references}
\phantomsection\addcontentsline{toc}{part}{Online Appendix}
\begin{center}\HeadingFont\fontsize{20pt}{24pt}\selectfont\bfseries Online Appendix\par\end{center}
\suppressfloats[t]
\vspace{1em}
\section{Prompts and designs}\label{part:prompts}

I give the exact inputs and the rules that turn them into attribution games. Read the prompt tables together with the task levels and edit spans: wording, whitespace, and component boundaries all enter the evaluated input. Table~\appref{tab:study-names} maps descriptive study names to the codes used in the data. The tables are generated from the design specification that Appendix~\ref{part:reproducibility} describes, which also marks the excluded operations.

\begingroup\long\def\floatNotesBody{
The design file, \texttt{code/config/design.yaml}, and every file under \texttt{data/} carry the codes; the main text carries the names.
}%
\begin{table}[tbp]
\centering
\caption{The study codes of the design and data files, and the names the main text uses for them.}\label{tab:study-names}
\footnotesize\setlength{\tabcolsep}{4pt}
\begin{NiceTabular}{@{}l;>{\raggedright\arraybackslash}p{5.2cm};>{\raggedright\arraybackslash}p{7.2cm}@{}}
\toprule
Code & Name in the main text & Title in the design file \\
\midrule
E1 & presentation sensitivity: the insertion sweeps & invisible perturbations \\
\dottedrule
E2 & word attribution on the flight tasks & designed flight tasks, word-level Formulation A \\
\dottedrule
E3 & edit attribution beside the price flip & Formulation B with hand-authored operations and the WTP capstone \\
\dottedrule
E4 & framing & framing \\
\dottedrule
E5 & word attribution on the hotel and smartphone tasks & hotel and smartphone templates \\
\dottedrule
E6 & personas & four-level persona gradient \\
\dottedrule
grid & robustness checks & the alternative references, payoffs, groupings, classifications, and edit set on Qwen 3.8 27B \\
\bottomrule
\end{NiceTabular}
\end{table}
\endgroup{}
\floatnotes{study_names}

\subsection{Tasks, components, and classification}

Tables~\ref{app:prompts:templates:flight}, \ref{app:prompts:templates:hotel}, and~\appref{app:prompts:templates:smartphone} show the example flight, hotel, and smartphone prompts with each component's classification. Tier~1 contains the displayed attribute values, including their units. Tier~2 contains substantive context, labels, and instruction words. Tier~3 contains the remaining connecting words, punctuation, and line breaks. These are descriptive categories; Tier~2 and Tier~3 contributions can help communicate the task.

The primary attribute share uses Tier~1. A broader classification includes Tier~2 as well. I keep this change of category membership separate from merging players. Each price or second-attribute value is one player. Punctuation split from a value and each line break are separate players; other words retain attached punctuation. Ordinary spaces are fixed separators. The flight, hotel, and smartphone templates have 49, 45, and 40 players, respectively. These readable components need not match internal tokenizer tokens.

An absent word becomes \texttt{\_}; an absent line break or split punctuation becomes the empty string. Restoring components preserves their order, not their internal tokenizer positions. The prompt-part displays collect the existing word values into question, route or purpose, option-label, attribute-phrase, instruction, and line-break categories. They do not solve a new span game.

\begingroup\long\def\floatNotesBody{
1: attribute value; 2: context word; 3: function word, punctuation, or line break. \ensuremath{\hookleftarrow} marks a line break; spaces are fixed. Absent words become \texttt{\protect\frenchspacing \_}; line breaks and split punctuation become empty strings. Values retain units.
}%
\begin{table}[htbp]
\centering
\caption{Example prompt of the flight template, flight task 7, a trade-off task.}\label{app:prompts:templates:flight}
\small
\begin{NiceTabular}{@{}r;>{\raggedright\arraybackslash}p{0.9\linewidth}@{}}
\toprule
Line & Prompt \\
\midrule
1 & \texttt{\protect\frenchspacing \begin{tabular}[t]{@{}c@{}}Which\\\textrm{\scriptsize 3}\end{tabular} \begin{tabular}[t]{@{}c@{}}of\\\textrm{\scriptsize 3}\end{tabular} \begin{tabular}[t]{@{}c@{}}the\\\textrm{\scriptsize 3}\end{tabular} \begin{tabular}[t]{@{}c@{}}following\\\textrm{\scriptsize 2}\end{tabular} \begin{tabular}[t]{@{}c@{}}flights\\\textrm{\scriptsize 2}\end{tabular} \begin{tabular}[t]{@{}c@{}}from\\\textrm{\scriptsize 3}\end{tabular} \begin{tabular}[t]{@{}c@{}}Chicago\\\textrm{\scriptsize 2}\end{tabular} \begin{tabular}[t]{@{}c@{}}to\\\textrm{\scriptsize 3}\end{tabular} \begin{tabular}[t]{@{}c@{}}Atlanta\\\textrm{\scriptsize 2}\end{tabular} \begin{tabular}[t]{@{}c@{}}do\\\textrm{\scriptsize 3}\end{tabular} \begin{tabular}[t]{@{}c@{}}you\\\textrm{\scriptsize 3}\end{tabular} \begin{tabular}[t]{@{}c@{}}prefer?\\\textrm{\scriptsize 2}\end{tabular}\begin{tabular}[t]{@{}c@{}}\ensuremath{\hookleftarrow}\\\textrm{\scriptsize 3}\end{tabular}} \\
\dottedrule
2 & \texttt{\protect\frenchspacing \begin{tabular}[t]{@{}c@{}}Flight\\\textrm{\scriptsize 2}\end{tabular} \begin{tabular}[t]{@{}c@{}}A:\\\textrm{\scriptsize 2}\end{tabular} \begin{tabular}[t]{@{}c@{}}costs\\\textrm{\scriptsize 2}\end{tabular} \begin{tabular}[t]{@{}c@{}}\$200\\\textrm{\scriptsize 1}\end{tabular}\begin{tabular}[t]{@{}c@{}},\\\textrm{\scriptsize 3}\end{tabular} \begin{tabular}[t]{@{}c@{}}with\\\textrm{\scriptsize 3}\end{tabular} \begin{tabular}[t]{@{}c@{}}a\\\textrm{\scriptsize 3}\end{tabular} \begin{tabular}[t]{@{}c@{}}total\\\textrm{\scriptsize 2}\end{tabular} \begin{tabular}[t]{@{}c@{}}travel\\\textrm{\scriptsize 2}\end{tabular} \begin{tabular}[t]{@{}c@{}}time\\\textrm{\scriptsize 2}\end{tabular} \begin{tabular}[t]{@{}c@{}}of\\\textrm{\scriptsize 3}\end{tabular} \begin{tabular}[t]{@{}c@{}}5 hours\\\textrm{\scriptsize 1}\end{tabular}\begin{tabular}[t]{@{}c@{}}.\\\textrm{\scriptsize 3}\end{tabular}\begin{tabular}[t]{@{}c@{}}\ensuremath{\hookleftarrow}\\\textrm{\scriptsize 3}\end{tabular}} \\
\dottedrule
3 & \texttt{\protect\frenchspacing \begin{tabular}[t]{@{}c@{}}Flight\\\textrm{\scriptsize 2}\end{tabular} \begin{tabular}[t]{@{}c@{}}B:\\\textrm{\scriptsize 2}\end{tabular} \begin{tabular}[t]{@{}c@{}}costs\\\textrm{\scriptsize 2}\end{tabular} \begin{tabular}[t]{@{}c@{}}\$350\\\textrm{\scriptsize 1}\end{tabular}\begin{tabular}[t]{@{}c@{}},\\\textrm{\scriptsize 3}\end{tabular} \begin{tabular}[t]{@{}c@{}}with\\\textrm{\scriptsize 3}\end{tabular} \begin{tabular}[t]{@{}c@{}}a\\\textrm{\scriptsize 3}\end{tabular} \begin{tabular}[t]{@{}c@{}}total\\\textrm{\scriptsize 2}\end{tabular} \begin{tabular}[t]{@{}c@{}}travel\\\textrm{\scriptsize 2}\end{tabular} \begin{tabular}[t]{@{}c@{}}time\\\textrm{\scriptsize 2}\end{tabular} \begin{tabular}[t]{@{}c@{}}of\\\textrm{\scriptsize 3}\end{tabular} \begin{tabular}[t]{@{}c@{}}2 hours\\\textrm{\scriptsize 1}\end{tabular}\begin{tabular}[t]{@{}c@{}}.\\\textrm{\scriptsize 3}\end{tabular}\begin{tabular}[t]{@{}c@{}}\ensuremath{\hookleftarrow}\\\textrm{\scriptsize 3}\end{tabular}} \\
\dottedrule
4 & \texttt{\protect\frenchspacing \begin{tabular}[t]{@{}c@{}}Answer\\\textrm{\scriptsize 2}\end{tabular} \begin{tabular}[t]{@{}c@{}}with\\\textrm{\scriptsize 3}\end{tabular} \begin{tabular}[t]{@{}c@{}}a\\\textrm{\scriptsize 3}\end{tabular} \begin{tabular}[t]{@{}c@{}}single\\\textrm{\scriptsize 2}\end{tabular} \begin{tabular}[t]{@{}c@{}}letter,\\\textrm{\scriptsize 2}\end{tabular} \begin{tabular}[t]{@{}c@{}}A\\\textrm{\scriptsize 2}\end{tabular} \begin{tabular}[t]{@{}c@{}}or\\\textrm{\scriptsize 3}\end{tabular} \begin{tabular}[t]{@{}c@{}}B.\\\textrm{\scriptsize 2}\end{tabular}} \\
\bottomrule
\end{NiceTabular}
\end{table}
\endgroup{}
\floatnotes{prompts_templates_flight}

\begingroup\long\def\floatNotesBody{
1: attribute value; 2: context word; 3: function word, punctuation, or line break. \ensuremath{\hookleftarrow} marks a line break; spaces are fixed. Absent words become \texttt{\protect\frenchspacing \_}; line breaks and split punctuation become empty strings. Values retain units.
}%
\begin{table}[htbp]
\centering
\caption{Example prompt of the hotel template, hotel task 7, a trade-off task.}\label{app:prompts:templates:hotel}
\small
\begin{NiceTabular}{@{}r;>{\raggedright\arraybackslash}p{0.9\linewidth}@{}}
\toprule
Line & Prompt \\
\midrule
1 & \texttt{\protect\frenchspacing \begin{tabular}[t]{@{}c@{}}Which\\\textrm{\scriptsize 3}\end{tabular} \begin{tabular}[t]{@{}c@{}}of\\\textrm{\scriptsize 3}\end{tabular} \begin{tabular}[t]{@{}c@{}}the\\\textrm{\scriptsize 3}\end{tabular} \begin{tabular}[t]{@{}c@{}}following\\\textrm{\scriptsize 2}\end{tabular} \begin{tabular}[t]{@{}c@{}}hotels\\\textrm{\scriptsize 2}\end{tabular} \begin{tabular}[t]{@{}c@{}}in\\\textrm{\scriptsize 3}\end{tabular} \begin{tabular}[t]{@{}c@{}}Boston\\\textrm{\scriptsize 2}\end{tabular} \begin{tabular}[t]{@{}c@{}}do\\\textrm{\scriptsize 3}\end{tabular} \begin{tabular}[t]{@{}c@{}}you\\\textrm{\scriptsize 3}\end{tabular} \begin{tabular}[t]{@{}c@{}}prefer?\\\textrm{\scriptsize 2}\end{tabular}\begin{tabular}[t]{@{}c@{}}\ensuremath{\hookleftarrow}\\\textrm{\scriptsize 3}\end{tabular}} \\
\dottedrule
2 & \texttt{\protect\frenchspacing \begin{tabular}[t]{@{}c@{}}Hotel\\\textrm{\scriptsize 2}\end{tabular} \begin{tabular}[t]{@{}c@{}}A:\\\textrm{\scriptsize 2}\end{tabular} \begin{tabular}[t]{@{}c@{}}costs\\\textrm{\scriptsize 2}\end{tabular} \begin{tabular}[t]{@{}c@{}}\$120\\\textrm{\scriptsize 1}\end{tabular} \begin{tabular}[t]{@{}c@{}}per\\\textrm{\scriptsize 3}\end{tabular} \begin{tabular}[t]{@{}c@{}}night,\\\textrm{\scriptsize 2}\end{tabular} \begin{tabular}[t]{@{}c@{}}located\\\textrm{\scriptsize 2}\end{tabular} \begin{tabular}[t]{@{}c@{}}8 miles\\\textrm{\scriptsize 1}\end{tabular} \begin{tabular}[t]{@{}c@{}}from\\\textrm{\scriptsize 3}\end{tabular} \begin{tabular}[t]{@{}c@{}}the\\\textrm{\scriptsize 3}\end{tabular} \begin{tabular}[t]{@{}c@{}}city\\\textrm{\scriptsize 2}\end{tabular} \begin{tabular}[t]{@{}c@{}}center.\\\textrm{\scriptsize 2}\end{tabular}\begin{tabular}[t]{@{}c@{}}\ensuremath{\hookleftarrow}\\\textrm{\scriptsize 3}\end{tabular}} \\
\dottedrule
3 & \texttt{\protect\frenchspacing \begin{tabular}[t]{@{}c@{}}Hotel\\\textrm{\scriptsize 2}\end{tabular} \begin{tabular}[t]{@{}c@{}}B:\\\textrm{\scriptsize 2}\end{tabular} \begin{tabular}[t]{@{}c@{}}costs\\\textrm{\scriptsize 2}\end{tabular} \begin{tabular}[t]{@{}c@{}}\$210\\\textrm{\scriptsize 1}\end{tabular} \begin{tabular}[t]{@{}c@{}}per\\\textrm{\scriptsize 3}\end{tabular} \begin{tabular}[t]{@{}c@{}}night,\\\textrm{\scriptsize 2}\end{tabular} \begin{tabular}[t]{@{}c@{}}located\\\textrm{\scriptsize 2}\end{tabular} \begin{tabular}[t]{@{}c@{}}2 miles\\\textrm{\scriptsize 1}\end{tabular} \begin{tabular}[t]{@{}c@{}}from\\\textrm{\scriptsize 3}\end{tabular} \begin{tabular}[t]{@{}c@{}}the\\\textrm{\scriptsize 3}\end{tabular} \begin{tabular}[t]{@{}c@{}}city\\\textrm{\scriptsize 2}\end{tabular} \begin{tabular}[t]{@{}c@{}}center.\\\textrm{\scriptsize 2}\end{tabular}\begin{tabular}[t]{@{}c@{}}\ensuremath{\hookleftarrow}\\\textrm{\scriptsize 3}\end{tabular}} \\
\dottedrule
4 & \texttt{\protect\frenchspacing \begin{tabular}[t]{@{}c@{}}Answer\\\textrm{\scriptsize 2}\end{tabular} \begin{tabular}[t]{@{}c@{}}with\\\textrm{\scriptsize 3}\end{tabular} \begin{tabular}[t]{@{}c@{}}a\\\textrm{\scriptsize 3}\end{tabular} \begin{tabular}[t]{@{}c@{}}single\\\textrm{\scriptsize 2}\end{tabular} \begin{tabular}[t]{@{}c@{}}letter,\\\textrm{\scriptsize 2}\end{tabular} \begin{tabular}[t]{@{}c@{}}A\\\textrm{\scriptsize 2}\end{tabular} \begin{tabular}[t]{@{}c@{}}or\\\textrm{\scriptsize 3}\end{tabular} \begin{tabular}[t]{@{}c@{}}B.\\\textrm{\scriptsize 2}\end{tabular}} \\
\bottomrule
\end{NiceTabular}
\end{table}
\endgroup{}
\floatnotes{prompts_templates_hotel}

\begingroup\long\def\floatNotesBody{
1: attribute value; 2: context word; 3: function word, punctuation, or line break. \ensuremath{\hookleftarrow} marks a line break; spaces are fixed. Absent words become \texttt{\protect\frenchspacing \_}; line breaks and split punctuation become empty strings. Values retain units.
}%
\begin{table}[htbp]
\centering
\caption{Example prompt of the smartphone template, smartphone task 7, a trade-off task.}\label{app:prompts:templates:smartphone}
\small
\begin{NiceTabular}{@{}r;>{\raggedright\arraybackslash}p{0.9\linewidth}@{}}
\toprule
Line & Prompt \\
\midrule
1 & \texttt{\protect\frenchspacing \begin{tabular}[t]{@{}c@{}}Which\\\textrm{\scriptsize 3}\end{tabular} \begin{tabular}[t]{@{}c@{}}of\\\textrm{\scriptsize 3}\end{tabular} \begin{tabular}[t]{@{}c@{}}the\\\textrm{\scriptsize 3}\end{tabular} \begin{tabular}[t]{@{}c@{}}following\\\textrm{\scriptsize 2}\end{tabular} \begin{tabular}[t]{@{}c@{}}smartphones\\\textrm{\scriptsize 2}\end{tabular} \begin{tabular}[t]{@{}c@{}}do\\\textrm{\scriptsize 3}\end{tabular} \begin{tabular}[t]{@{}c@{}}you\\\textrm{\scriptsize 3}\end{tabular} \begin{tabular}[t]{@{}c@{}}prefer\\\textrm{\scriptsize 2}\end{tabular} \begin{tabular}[t]{@{}c@{}}for\\\textrm{\scriptsize 3}\end{tabular} \begin{tabular}[t]{@{}c@{}}everyday\\\textrm{\scriptsize 2}\end{tabular} \begin{tabular}[t]{@{}c@{}}use?\\\textrm{\scriptsize 2}\end{tabular}\begin{tabular}[t]{@{}c@{}}\ensuremath{\hookleftarrow}\\\textrm{\scriptsize 3}\end{tabular}} \\
\dottedrule
2 & \texttt{\protect\frenchspacing \begin{tabular}[t]{@{}c@{}}Phone\\\textrm{\scriptsize 2}\end{tabular} \begin{tabular}[t]{@{}c@{}}A:\\\textrm{\scriptsize 2}\end{tabular} \begin{tabular}[t]{@{}c@{}}costs\\\textrm{\scriptsize 2}\end{tabular} \begin{tabular}[t]{@{}c@{}}\$499\\\textrm{\scriptsize 1}\end{tabular}\begin{tabular}[t]{@{}c@{}},\\\textrm{\scriptsize 3}\end{tabular} \begin{tabular}[t]{@{}c@{}}with\\\textrm{\scriptsize 3}\end{tabular} \begin{tabular}[t]{@{}c@{}}128 GB\\\textrm{\scriptsize 1}\end{tabular} \begin{tabular}[t]{@{}c@{}}of\\\textrm{\scriptsize 3}\end{tabular} \begin{tabular}[t]{@{}c@{}}storage.\\\textrm{\scriptsize 2}\end{tabular}\begin{tabular}[t]{@{}c@{}}\ensuremath{\hookleftarrow}\\\textrm{\scriptsize 3}\end{tabular}} \\
\dottedrule
3 & \texttt{\protect\frenchspacing \begin{tabular}[t]{@{}c@{}}Phone\\\textrm{\scriptsize 2}\end{tabular} \begin{tabular}[t]{@{}c@{}}B:\\\textrm{\scriptsize 2}\end{tabular} \begin{tabular}[t]{@{}c@{}}costs\\\textrm{\scriptsize 2}\end{tabular} \begin{tabular}[t]{@{}c@{}}\$799\\\textrm{\scriptsize 1}\end{tabular}\begin{tabular}[t]{@{}c@{}},\\\textrm{\scriptsize 3}\end{tabular} \begin{tabular}[t]{@{}c@{}}with\\\textrm{\scriptsize 3}\end{tabular} \begin{tabular}[t]{@{}c@{}}512 GB\\\textrm{\scriptsize 1}\end{tabular} \begin{tabular}[t]{@{}c@{}}of\\\textrm{\scriptsize 3}\end{tabular} \begin{tabular}[t]{@{}c@{}}storage.\\\textrm{\scriptsize 2}\end{tabular}\begin{tabular}[t]{@{}c@{}}\ensuremath{\hookleftarrow}\\\textrm{\scriptsize 3}\end{tabular}} \\
\dottedrule
4 & \texttt{\protect\frenchspacing \begin{tabular}[t]{@{}c@{}}Answer\\\textrm{\scriptsize 2}\end{tabular} \begin{tabular}[t]{@{}c@{}}with\\\textrm{\scriptsize 3}\end{tabular} \begin{tabular}[t]{@{}c@{}}a\\\textrm{\scriptsize 3}\end{tabular} \begin{tabular}[t]{@{}c@{}}single\\\textrm{\scriptsize 2}\end{tabular} \begin{tabular}[t]{@{}c@{}}letter,\\\textrm{\scriptsize 2}\end{tabular} \begin{tabular}[t]{@{}c@{}}A\\\textrm{\scriptsize 2}\end{tabular} \begin{tabular}[t]{@{}c@{}}or\\\textrm{\scriptsize 3}\end{tabular} \begin{tabular}[t]{@{}c@{}}B.\\\textrm{\scriptsize 2}\end{tabular}} \\
\bottomrule
\end{NiceTabular}
\end{table}
\endgroup{}
\floatnotes{prompts_templates_smartphone}

Tables~\ref{app:prompts:tasks:flight}, \ref{app:prompts:tasks:hotel}, and~\appref{app:prompts:tasks:smartphone} specify every attribute level. Each template crosses two levels in four slots, giving sixteen tasks with ties, tasks in which one option dominates, and trade-offs. Lower price, travel time, and hotel distance are preferred; higher storage is preferred. Level index zero denotes the preferred level. With level indices $a,b$ for ``A'' and $c,d$ for ``B'', the task number is $1+8a+4b+2c+d$.

\begingroup\long\def\floatNotesBody{
Preferred levels: \$200 and 2 hours. Dominance (``A'' or ``B'') marks an option at least as good on both attributes and better on one. In tasks 4 and 13, the dominant option is better on both.
}%
\begin{table}[htbp]
\centering
\caption{The 16 tasks of the flight template, the full $2^{4}$ factorial over its attribute slots.}\label{app:prompts:tasks:flight}
\footnotesize
\begin{NiceTabular}{@{}l;l;l;l;l;l@{}}
\toprule
 & \multicolumn{2}{c}{Option A} & \multicolumn{2}{c}{Option B} & \\
\cmidrule(lr){2-3}\cmidrule(lr){4-5}
Task & Price & Travel time & Price & Travel time & Category \\
\midrule
1 & \$200 & 2 hours & \$200 & 2 hours & tie \\
\dottedrule
2 & \$200 & 2 hours & \$200 & 5 hours & dominance (``A'') \\
\dottedrule
3 & \$200 & 2 hours & \$350 & 2 hours & dominance (``A'') \\
\dottedrule
4 & \$200 & 2 hours & \$350 & 5 hours & dominance (``A'') \\
\dottedrule
5 & \$200 & 5 hours & \$200 & 2 hours & dominance (``B'') \\
\dottedrule
6 & \$200 & 5 hours & \$200 & 5 hours & tie \\
\dottedrule
7 & \$200 & 5 hours & \$350 & 2 hours & trade-off \\
\dottedrule
8 & \$200 & 5 hours & \$350 & 5 hours & dominance (``A'') \\
\dottedrule
9 & \$350 & 2 hours & \$200 & 2 hours & dominance (``B'') \\
\dottedrule
10 & \$350 & 2 hours & \$200 & 5 hours & trade-off \\
\dottedrule
11 & \$350 & 2 hours & \$350 & 2 hours & tie \\
\dottedrule
12 & \$350 & 2 hours & \$350 & 5 hours & dominance (``A'') \\
\dottedrule
13 & \$350 & 5 hours & \$200 & 2 hours & dominance (``B'') \\
\dottedrule
14 & \$350 & 5 hours & \$200 & 5 hours & dominance (``B'') \\
\dottedrule
15 & \$350 & 5 hours & \$350 & 2 hours & dominance (``B'') \\
\dottedrule
16 & \$350 & 5 hours & \$350 & 5 hours & tie \\
\bottomrule
\end{NiceTabular}
\end{table}
\endgroup{}
\floatnotes{prompts_tasks_flight}

\begingroup\long\def\floatNotesBody{
Preferred levels: \$120 and 2 miles. Dominance (``A'' or ``B'') marks an option at least as good on both attributes and better on one. In tasks 4 and 13, the dominant option is better on both.
}%
\begin{table}[htbp]
\centering
\caption{The 16 tasks of the hotel template, the full $2^{4}$ factorial over its attribute slots.}\label{app:prompts:tasks:hotel}
\footnotesize
\begin{NiceTabular}{@{}l;l;l;l;l;l@{}}
\toprule
 & \multicolumn{2}{c}{Option A} & \multicolumn{2}{c}{Option B} & \\
\cmidrule(lr){2-3}\cmidrule(lr){4-5}
Task & Price & Distance & Price & Distance & Category \\
\midrule
1 & \$120 & 2 miles & \$120 & 2 miles & tie \\
\dottedrule
2 & \$120 & 2 miles & \$120 & 8 miles & dominance (``A'') \\
\dottedrule
3 & \$120 & 2 miles & \$210 & 2 miles & dominance (``A'') \\
\dottedrule
4 & \$120 & 2 miles & \$210 & 8 miles & dominance (``A'') \\
\dottedrule
5 & \$120 & 8 miles & \$120 & 2 miles & dominance (``B'') \\
\dottedrule
6 & \$120 & 8 miles & \$120 & 8 miles & tie \\
\dottedrule
7 & \$120 & 8 miles & \$210 & 2 miles & trade-off \\
\dottedrule
8 & \$120 & 8 miles & \$210 & 8 miles & dominance (``A'') \\
\dottedrule
9 & \$210 & 2 miles & \$120 & 2 miles & dominance (``B'') \\
\dottedrule
10 & \$210 & 2 miles & \$120 & 8 miles & trade-off \\
\dottedrule
11 & \$210 & 2 miles & \$210 & 2 miles & tie \\
\dottedrule
12 & \$210 & 2 miles & \$210 & 8 miles & dominance (``A'') \\
\dottedrule
13 & \$210 & 8 miles & \$120 & 2 miles & dominance (``B'') \\
\dottedrule
14 & \$210 & 8 miles & \$120 & 8 miles & dominance (``B'') \\
\dottedrule
15 & \$210 & 8 miles & \$210 & 2 miles & dominance (``B'') \\
\dottedrule
16 & \$210 & 8 miles & \$210 & 8 miles & tie \\
\bottomrule
\end{NiceTabular}
\end{table}
\endgroup{}
\floatnotes{prompts_tasks_hotel}

\begingroup\long\def\floatNotesBody{
Preferred levels: \$499 and 512 GB. Dominance (``A'' or ``B'') marks an option at least as good on both attributes and better on one. In tasks 4 and 13, the dominant option is better on both.
}%
\begin{table}[htbp]
\centering
\caption{The 16 tasks of the smartphone template, the full $2^{4}$ factorial over its attribute slots.}\label{app:prompts:tasks:smartphone}
\footnotesize
\begin{NiceTabular}{@{}l;l;l;l;l;l@{}}
\toprule
 & \multicolumn{2}{c}{Option A} & \multicolumn{2}{c}{Option B} & \\
\cmidrule(lr){2-3}\cmidrule(lr){4-5}
Task & Price & Storage & Price & Storage & Category \\
\midrule
1 & \$499 & 512 GB & \$499 & 512 GB & tie \\
\dottedrule
2 & \$499 & 512 GB & \$499 & 128 GB & dominance (``A'') \\
\dottedrule
3 & \$499 & 512 GB & \$799 & 512 GB & dominance (``A'') \\
\dottedrule
4 & \$499 & 512 GB & \$799 & 128 GB & dominance (``A'') \\
\dottedrule
5 & \$499 & 128 GB & \$499 & 512 GB & dominance (``B'') \\
\dottedrule
6 & \$499 & 128 GB & \$499 & 128 GB & tie \\
\dottedrule
7 & \$499 & 128 GB & \$799 & 512 GB & trade-off \\
\dottedrule
8 & \$499 & 128 GB & \$799 & 128 GB & dominance (``A'') \\
\dottedrule
9 & \$799 & 512 GB & \$499 & 512 GB & dominance (``B'') \\
\dottedrule
10 & \$799 & 512 GB & \$499 & 128 GB & trade-off \\
\dottedrule
11 & \$799 & 512 GB & \$799 & 512 GB & tie \\
\dottedrule
12 & \$799 & 512 GB & \$799 & 128 GB & dominance (``A'') \\
\dottedrule
13 & \$799 & 128 GB & \$499 & 512 GB & dominance (``B'') \\
\dottedrule
14 & \$799 & 128 GB & \$499 & 128 GB & dominance (``B'') \\
\dottedrule
15 & \$799 & 128 GB & \$799 & 512 GB & dominance (``B'') \\
\dottedrule
16 & \$799 & 128 GB & \$799 & 128 GB & tie \\
\bottomrule
\end{NiceTabular}
\end{table}
\endgroup{}
\floatnotes{prompts_tasks_smartphone}

\subsection{Retained edits and rendering}

Tables~\ref{app:prompts:operations:flight}, \ref{app:prompts:operations:hotel}, and~\appref{app:prompts:operations:smartphone} list both retained and excluded edits. I exclude ``do you prefer'' to ``would you choose'' in every category and ``from Chicago to Atlanta'' to ``between Chicago and Atlanta'' for flights because these are disputed changes of meaning, not defensible equivalents of the intended choice problem. The hotel and smartphone operations with the same latter identifier change different text and remain. The ``Wording'' column marks changes to words or values, including the substantive price flip; it does not certify equivalent meaning.

\begin{samepage}
Replacing ``Flight'' by ``Option'' in the option lines gives this complete edited prompt, with the transportation context retained in the question:
\begin{quote}\small\ttfamily
Which of the following flights from Chicago to Atlanta do you prefer?\\
Option A: costs \$200, with a total travel time of 5 hours.\\
Option B: costs \$350, with a total travel time of 2 hours.\\
Answer with a single letter, A or B.
\end{quote}
\end{samepage}

Excluded edits are held absent in every retained coalition. I recompute the smaller game's values; deleting entries from the original attribution vector would retain the wrong context weights. The flight game contains \tnRetainedEditGameEditsFlight{} presentation edits plus the price flip; hotels and smartphones contain \tnRetainedEditGameEditsHotel{} and \tnRetainedEditGameEditsSmartphone{} presentation edits, respectively. The price flip changes the price of ``A'' to its other design level and is a substantive intervention.

Edit offsets count Unicode code points from zero in the original rendered text. The half-open span $[g,h)$ contains positions $g$ through $h-1$. Replacements are applied simultaneously at these locations. Insertions replace a neighboring character, avoiding competing insertions at one empty span. Visible escape marks in the tables identify whitespace; they are not literal input text. Characters outside the basic Latin character set are identified by code point, including the zero-width space U+200B.

\begingroup\long\def\floatNotesBody{
$[g,h)$ indexes original-message characters from zero. Wording marks changes to words or values; formatting changes spacing or punctuation. Retained marks edits kept in reduced games. Price rows are alternatives. \textvisiblespace{} and \ensuremath{\hookleftarrow} mark spaces and line breaks.
}%
\begin{table}[htbp]
\centering
\caption{The 13 operations of the flight template, one row per edit, with the design file's note on an operation in italics beneath its edits.}\label{app:prompts:operations:flight}
\scriptsize\setlength{\tabcolsep}{4pt}
\begin{NiceTabular}{@{}>{\raggedright\arraybackslash}p{2.4cm};l;l;l;>{\raggedright\arraybackslash}p{2.6cm};>{\raggedright\arraybackslash}p{2.6cm};l@{}}
\toprule
Edit & Kind & Wording & Retained & Source & Target & Span \\
\midrule
would you choose & lexical & yes & no & \texttt{\protect\frenchspacing do you prefer} & \texttt{\protect\frenchspacing would you choose} & $[55, 68)$ \\
\dottedrule
 & \multicolumn{6}{>{\raggedright\arraybackslash}p{\dimexpr\textwidth-2.4cm-2\tabcolsep\relax}@{}}{\textit{held off in the reduced games: a disputed meaning change, not a defensible meaning-preserving edit}} \\
\dottedrule
these & lexical & yes & yes & \texttt{\protect\frenchspacing the following} & \texttt{\protect\frenchspacing these} & $[9, 22)$ \\
\dottedrule
between Chicago and Atlanta & syntactic & yes & no & \texttt{\protect\frenchspacing from Chicago to Atlanta} & \texttt{\protect\frenchspacing between Chicago and Atlanta} & $[31, 54)$ \\
\dottedrule
 & \multicolumn{6}{>{\raggedright\arraybackslash}p{\dimexpr\textwidth-2.4cm-2\tabcolsep\relax}@{}}{\textit{held off in the reduced games: a disputed meaning change of the route}} \\
\dottedrule
is priced at & lexical & yes & yes & \texttt{\protect\frenchspacing costs} & \texttt{\protect\frenchspacing is priced at} & $[80, 85)$ \\
\dottedrule
 &  &  &  & \texttt{\protect\frenchspacing costs} & \texttt{\protect\frenchspacing is priced at} & $[139, 144)$ \\
\dottedrule
Option & lexical & yes & yes & \texttt{\protect\frenchspacing Flight} & \texttt{\protect\frenchspacing Option} & $[70, 76)$ \\
\dottedrule
 &  &  &  & \texttt{\protect\frenchspacing Flight} & \texttt{\protect\frenchspacing Option} & $[129, 135)$ \\
\dottedrule
Reply with one & lexical & yes & yes & \texttt{\protect\frenchspacing Answer with a single} & \texttt{\protect\frenchspacing Reply with one} & $[188, 208)$ \\
\dottedrule
blank line after the question & formatting & no & yes & \texttt{\protect\frenchspacing \ensuremath{\hookleftarrow}} & \texttt{\protect\frenchspacing \ensuremath{\hookleftarrow}\ensuremath{\hookleftarrow}} & $[69, 70)$ \\
\dottedrule
 & \multicolumn{6}{>{\raggedright\arraybackslash}p{\dimexpr\textwidth-2.4cm-2\tabcolsep\relax}@{}}{\textit{blank line after the question}} \\
\dottedrule
join the option lines & formatting & no & yes & \texttt{\protect\frenchspacing \ensuremath{\hookleftarrow}} & \texttt{\protect\frenchspacing \textvisiblespace{}} & $[128, 129)$ \\
\dottedrule
 & \multicolumn{6}{>{\raggedright\arraybackslash}p{\dimexpr\textwidth-2.4cm-2\tabcolsep\relax}@{}}{\textit{join the option lines}} \\
\dottedrule
comma to semicolon after the prices & formatting & no & yes & \texttt{\protect\frenchspacing ,} & \texttt{\protect\frenchspacing ;} & $[90, 91)$ \\
\dottedrule
 &  &  &  & \texttt{\protect\frenchspacing ,} & \texttt{\protect\frenchspacing ;} & $[149, 150)$ \\
\dottedrule
 & \multicolumn{6}{>{\raggedright\arraybackslash}p{\dimexpr\textwidth-2.4cm-2\tabcolsep\relax}@{}}{\textit{comma to semicolon after the prices}} \\
\dottedrule
two spaces after the label colons & formatting & no & yes & \texttt{\protect\frenchspacing \textvisiblespace{}} & \texttt{\protect\frenchspacing \textvisiblespace{}\textvisiblespace{}} & $[79, 80)$ \\
\dottedrule
 &  &  &  & \texttt{\protect\frenchspacing \textvisiblespace{}} & \texttt{\protect\frenchspacing \textvisiblespace{}\textvisiblespace{}} & $[138, 139)$ \\
\dottedrule
 & \multicolumn{6}{>{\raggedright\arraybackslash}p{\dimexpr\textwidth-2.4cm-2\tabcolsep\relax}@{}}{\textit{two spaces after the label colons}} \\
\dottedrule
trailing space after the question mark & formatting & no & yes & \texttt{\protect\frenchspacing ?} & \texttt{\protect\frenchspacing ?\textvisiblespace{}} & $[68, 69)$ \\
\dottedrule
 & \multicolumn{6}{>{\raggedright\arraybackslash}p{\dimexpr\textwidth-2.4cm-2\tabcolsep\relax}@{}}{\textit{trailing space after the question mark}} \\
\dottedrule
options in parentheses & formatting & no & yes & \texttt{\protect\frenchspacing , A or B} & \texttt{\protect\frenchspacing \textvisiblespace{}(A or B)} & $[215, 223)$ \\
\dottedrule
 & \multicolumn{6}{>{\raggedright\arraybackslash}p{\dimexpr\textwidth-2.4cm-2\tabcolsep\relax}@{}}{\textit{options in parentheses}} \\
\dottedrule
price flip & attribute & yes & yes & \texttt{\protect\frenchspacing \$200} & \texttt{\protect\frenchspacing \$350} & $[86, 90)$ \\
\dottedrule
 &  &  &  & \texttt{\protect\frenchspacing \$350} & \texttt{\protect\frenchspacing \$200} & $[86, 90)$ \\
\dottedrule
 & \multicolumn{6}{>{\raggedright\arraybackslash}p{\dimexpr\textwidth-2.4cm-2\tabcolsep\relax}@{}}{\textit{price flip: the price of ``A'' set to its other level}} \\
\bottomrule
\end{NiceTabular}
\end{table}
\endgroup{}
\floatnotes{prompts_operations_flight}

\begingroup\long\def\floatNotesBody{
$[g,h)$ indexes original-message characters from zero. Wording marks changes to words or values; formatting changes spacing or punctuation. Retained marks edits kept in reduced games. Price rows are alternatives. \textvisiblespace{} and \ensuremath{\hookleftarrow} mark spaces and line breaks.
}%
\begin{table}[htbp]
\centering
\caption{The 13 operations of the hotel template, one row per edit, with the design file's note on an operation in italics beneath its edits.}\label{app:prompts:operations:hotel}
\scriptsize\setlength{\tabcolsep}{4pt}
\begin{NiceTabular}{@{}>{\raggedright\arraybackslash}p{2.4cm};l;l;l;>{\raggedright\arraybackslash}p{2.6cm};>{\raggedright\arraybackslash}p{2.6cm};l@{}}
\toprule
Edit & Kind & Wording & Retained & Source & Target & Span \\
\midrule
would you choose & lexical & yes & no & \texttt{\protect\frenchspacing do you prefer} & \texttt{\protect\frenchspacing would you choose} & $[40, 53)$ \\
\dottedrule
 & \multicolumn{6}{>{\raggedright\arraybackslash}p{\dimexpr\textwidth-2.4cm-2\tabcolsep\relax}@{}}{\textit{held off in the reduced games: a disputed meaning change, not a defensible meaning-preserving edit}} \\
\dottedrule
these & lexical & yes & yes & \texttt{\protect\frenchspacing the following} & \texttt{\protect\frenchspacing these} & $[9, 22)$ \\
\dottedrule
Boston hotels & syntactic & yes & yes & \texttt{\protect\frenchspacing hotels in Boston} & \texttt{\protect\frenchspacing Boston hotels} & $[23, 39)$ \\
\dottedrule
is priced at & lexical & yes & yes & \texttt{\protect\frenchspacing costs} & \texttt{\protect\frenchspacing is priced at} & $[64, 69)$ \\
\dottedrule
 &  &  &  & \texttt{\protect\frenchspacing costs} & \texttt{\protect\frenchspacing is priced at} & $[133, 138)$ \\
\dottedrule
Option & lexical & yes & yes & \texttt{\protect\frenchspacing Hotel} & \texttt{\protect\frenchspacing Option} & $[55, 60)$ \\
\dottedrule
 &  &  &  & \texttt{\protect\frenchspacing Hotel} & \texttt{\protect\frenchspacing Option} & $[124, 129)$ \\
\dottedrule
Reply with one & lexical & yes & yes & \texttt{\protect\frenchspacing Answer with a single} & \texttt{\protect\frenchspacing Reply with one} & $[193, 213)$ \\
\dottedrule
blank line after the question & formatting & no & yes & \texttt{\protect\frenchspacing \ensuremath{\hookleftarrow}} & \texttt{\protect\frenchspacing \ensuremath{\hookleftarrow}\ensuremath{\hookleftarrow}} & $[54, 55)$ \\
\dottedrule
 & \multicolumn{6}{>{\raggedright\arraybackslash}p{\dimexpr\textwidth-2.4cm-2\tabcolsep\relax}@{}}{\textit{blank line after the question}} \\
\dottedrule
join the option lines & formatting & no & yes & \texttt{\protect\frenchspacing \ensuremath{\hookleftarrow}} & \texttt{\protect\frenchspacing \textvisiblespace{}} & $[123, 124)$ \\
\dottedrule
 & \multicolumn{6}{>{\raggedright\arraybackslash}p{\dimexpr\textwidth-2.4cm-2\tabcolsep\relax}@{}}{\textit{join the option lines}} \\
\dottedrule
comma to semicolon after the prices & formatting & no & yes & \texttt{\protect\frenchspacing ,} & \texttt{\protect\frenchspacing ;} & $[84, 85)$ \\
\dottedrule
 &  &  &  & \texttt{\protect\frenchspacing ,} & \texttt{\protect\frenchspacing ;} & $[153, 154)$ \\
\dottedrule
 & \multicolumn{6}{>{\raggedright\arraybackslash}p{\dimexpr\textwidth-2.4cm-2\tabcolsep\relax}@{}}{\textit{comma to semicolon after the prices}} \\
\dottedrule
two spaces after the label colons & formatting & no & yes & \texttt{\protect\frenchspacing \textvisiblespace{}} & \texttt{\protect\frenchspacing \textvisiblespace{}\textvisiblespace{}} & $[63, 64)$ \\
\dottedrule
 &  &  &  & \texttt{\protect\frenchspacing \textvisiblespace{}} & \texttt{\protect\frenchspacing \textvisiblespace{}\textvisiblespace{}} & $[132, 133)$ \\
\dottedrule
 & \multicolumn{6}{>{\raggedright\arraybackslash}p{\dimexpr\textwidth-2.4cm-2\tabcolsep\relax}@{}}{\textit{two spaces after the label colons}} \\
\dottedrule
trailing space after the question mark & formatting & no & yes & \texttt{\protect\frenchspacing ?} & \texttt{\protect\frenchspacing ?\textvisiblespace{}} & $[53, 54)$ \\
\dottedrule
 & \multicolumn{6}{>{\raggedright\arraybackslash}p{\dimexpr\textwidth-2.4cm-2\tabcolsep\relax}@{}}{\textit{trailing space after the question mark}} \\
\dottedrule
options in parentheses & formatting & no & yes & \texttt{\protect\frenchspacing , A or B} & \texttt{\protect\frenchspacing \textvisiblespace{}(A or B)} & $[220, 228)$ \\
\dottedrule
 & \multicolumn{6}{>{\raggedright\arraybackslash}p{\dimexpr\textwidth-2.4cm-2\tabcolsep\relax}@{}}{\textit{options in parentheses}} \\
\dottedrule
price flip & attribute & yes & yes & \texttt{\protect\frenchspacing \$120} & \texttt{\protect\frenchspacing \$210} & $[70, 74)$ \\
\dottedrule
 &  &  &  & \texttt{\protect\frenchspacing \$210} & \texttt{\protect\frenchspacing \$120} & $[70, 74)$ \\
\dottedrule
 & \multicolumn{6}{>{\raggedright\arraybackslash}p{\dimexpr\textwidth-2.4cm-2\tabcolsep\relax}@{}}{\textit{price flip: the price of ``A'' set to its other level}} \\
\bottomrule
\end{NiceTabular}
\end{table}
\endgroup{}
\floatnotes{prompts_operations_hotel}

\begingroup\long\def\floatNotesBody{
$[g,h)$ indexes original-message characters from zero. Wording marks changes to words or values; formatting changes spacing or punctuation. Retained marks edits kept in reduced games. Price rows are alternatives. \textvisiblespace{} and \ensuremath{\hookleftarrow} mark spaces and line breaks.
}%
\begin{table}[htbp]
\centering
\caption{The 13 operations of the smartphone template, one row per edit, with the design file's note on an operation in italics beneath its edits.}\label{app:prompts:operations:smartphone}
\scriptsize\setlength{\tabcolsep}{4pt}
\begin{NiceTabular}{@{}>{\raggedright\arraybackslash}p{2.4cm};l;l;l;>{\raggedright\arraybackslash}p{2.6cm};>{\raggedright\arraybackslash}p{2.6cm};l@{}}
\toprule
Edit & Kind & Wording & Retained & Source & Target & Span \\
\midrule
would you choose & lexical & yes & no & \texttt{\protect\frenchspacing do you prefer} & \texttt{\protect\frenchspacing would you choose} & $[35, 48)$ \\
\dottedrule
 & \multicolumn{6}{>{\raggedright\arraybackslash}p{\dimexpr\textwidth-2.4cm-2\tabcolsep\relax}@{}}{\textit{held off in the reduced games: a disputed meaning change, not a defensible meaning-preserving edit}} \\
\dottedrule
these & lexical & yes & yes & \texttt{\protect\frenchspacing the following} & \texttt{\protect\frenchspacing these} & $[9, 22)$ \\
\dottedrule
purpose phrase moved to the front & syntactic & yes & yes & \texttt{\protect\frenchspacing Which} & \texttt{\protect\frenchspacing For everyday use, which} & $[0, 5)$ \\
\dottedrule
 &  &  &  & \texttt{\protect\frenchspacing \textvisiblespace{}for everyday use} & \texttt{\protect\frenchspacing \textrm{(empty)}} & $[48, 65)$ \\
\dottedrule
 & \multicolumn{6}{>{\raggedright\arraybackslash}p{\dimexpr\textwidth-2.4cm-2\tabcolsep\relax}@{}}{\textit{purpose phrase moved to the front}} \\
\dottedrule
is priced at & lexical & yes & yes & \texttt{\protect\frenchspacing costs} & \texttt{\protect\frenchspacing is priced at} & $[76, 81)$ \\
\dottedrule
 &  &  &  & \texttt{\protect\frenchspacing costs} & \texttt{\protect\frenchspacing is priced at} & $[121, 126)$ \\
\dottedrule
Option & lexical & yes & yes & \texttt{\protect\frenchspacing Phone} & \texttt{\protect\frenchspacing Option} & $[67, 72)$ \\
\dottedrule
 &  &  &  & \texttt{\protect\frenchspacing Phone} & \texttt{\protect\frenchspacing Option} & $[112, 117)$ \\
\dottedrule
Reply with one & lexical & yes & yes & \texttt{\protect\frenchspacing Answer with a single} & \texttt{\protect\frenchspacing Reply with one} & $[157, 177)$ \\
\dottedrule
blank line after the question & formatting & no & yes & \texttt{\protect\frenchspacing \ensuremath{\hookleftarrow}} & \texttt{\protect\frenchspacing \ensuremath{\hookleftarrow}\ensuremath{\hookleftarrow}} & $[66, 67)$ \\
\dottedrule
 & \multicolumn{6}{>{\raggedright\arraybackslash}p{\dimexpr\textwidth-2.4cm-2\tabcolsep\relax}@{}}{\textit{blank line after the question}} \\
\dottedrule
join the option lines & formatting & no & yes & \texttt{\protect\frenchspacing \ensuremath{\hookleftarrow}} & \texttt{\protect\frenchspacing \textvisiblespace{}} & $[111, 112)$ \\
\dottedrule
 & \multicolumn{6}{>{\raggedright\arraybackslash}p{\dimexpr\textwidth-2.4cm-2\tabcolsep\relax}@{}}{\textit{join the option lines}} \\
\dottedrule
comma to semicolon after the prices & formatting & no & yes & \texttt{\protect\frenchspacing ,} & \texttt{\protect\frenchspacing ;} & $[86, 87)$ \\
\dottedrule
 &  &  &  & \texttt{\protect\frenchspacing ,} & \texttt{\protect\frenchspacing ;} & $[131, 132)$ \\
\dottedrule
 & \multicolumn{6}{>{\raggedright\arraybackslash}p{\dimexpr\textwidth-2.4cm-2\tabcolsep\relax}@{}}{\textit{comma to semicolon after the prices}} \\
\dottedrule
two spaces after the label colons & formatting & no & yes & \texttt{\protect\frenchspacing \textvisiblespace{}} & \texttt{\protect\frenchspacing \textvisiblespace{}\textvisiblespace{}} & $[75, 76)$ \\
\dottedrule
 &  &  &  & \texttt{\protect\frenchspacing \textvisiblespace{}} & \texttt{\protect\frenchspacing \textvisiblespace{}\textvisiblespace{}} & $[120, 121)$ \\
\dottedrule
 & \multicolumn{6}{>{\raggedright\arraybackslash}p{\dimexpr\textwidth-2.4cm-2\tabcolsep\relax}@{}}{\textit{two spaces after the label colons}} \\
\dottedrule
trailing space after the question mark & formatting & no & yes & \texttt{\protect\frenchspacing ?} & \texttt{\protect\frenchspacing ?\textvisiblespace{}} & $[65, 66)$ \\
\dottedrule
 & \multicolumn{6}{>{\raggedright\arraybackslash}p{\dimexpr\textwidth-2.4cm-2\tabcolsep\relax}@{}}{\textit{trailing space after the question mark}} \\
\dottedrule
options in parentheses & formatting & no & yes & \texttt{\protect\frenchspacing , A or B} & \texttt{\protect\frenchspacing \textvisiblespace{}(A or B)} & $[184, 192)$ \\
\dottedrule
 & \multicolumn{6}{>{\raggedright\arraybackslash}p{\dimexpr\textwidth-2.4cm-2\tabcolsep\relax}@{}}{\textit{options in parentheses}} \\
\dottedrule
price flip & attribute & yes & yes & \texttt{\protect\frenchspacing \$499} & \texttt{\protect\frenchspacing \$799} & $[82, 86)$ \\
\dottedrule
 &  &  &  & \texttt{\protect\frenchspacing \$799} & \texttt{\protect\frenchspacing \$499} & $[82, 86)$ \\
\dottedrule
 & \multicolumn{6}{>{\raggedright\arraybackslash}p{\dimexpr\textwidth-2.4cm-2\tabcolsep\relax}@{}}{\textit{price flip: the price of ``A'' set to its other level}} \\
\bottomrule
\end{NiceTabular}
\end{table}
\endgroup{}
\floatnotes{prompts_operations_smartphone}

\subsection{Personas, presentation sweeps, and framing}

Table~\appref{app:prompts:personas:texts} gives four conditions: no persona and three system-message descriptions. Their one, two, or five clause players precede the task components in the allocation. Table~\appref{app:prompts:personas:operations} lists the persona rewordings. These join the six task-formatting edits and price flip where applicable. Changing the respondent's substantive description differs from rephrasing that description.

\begingroup\long\def\floatNotesBody{
Each clause is a persona-clause player; removing it replaces each word with \texttt{\protect\frenchspacing \_}. $I$ counts task and clause players. $K$ counts listed persona edits plus all formatting and price operations. The system message applies to every flight task.
}%
\begin{table}[htbp]
\centering
\caption{The personas, one per description.}\label{app:prompts:personas:texts}
\footnotesize\setlength{\tabcolsep}{4pt}
\begin{NiceTabular}{@{}>{\raggedright\arraybackslash}p{2.6cm};>{\raggedright\arraybackslash}p{5.2cm};r;r;>{\raggedright\arraybackslash}p{4.0cm};r@{}}
\toprule
Description & Persona & Clauses & $I$ & Operations & $K$ \\
\midrule
the generic description & \texttt{\protect\frenchspacing You are a consumer.} & 1 & 50 & none & 7 \\
\dottedrule
the price-conscious description & \texttt{\protect\frenchspacing You are a budget-\allowbreak{}conscious consumer. You compare prices before you buy.} & 2 & 51 & cost-conscious and before buying & 9 \\
\dottedrule
the detailed description & \texttt{\protect\frenchspacing You are a budget-\allowbreak{}conscious consumer. You compare prices before you buy. You are 35 years old. You work full time as an accountant. You travel a few times a year.} & 5 & 54 & cost-conscious, before buying, 35 years of age, are employed, and per year & 12 \\
\bottomrule
\end{NiceTabular}
\end{table}
\endgroup{}
\floatnotes{prompts_personas_texts}

\begingroup\long\def\floatNotesBody{
Offsets count characters of the detailed description from zero. An edit applies where its source occurs at the same offsets. Wording marks changes to words or values; formatting changes spacing or punctuation.
}%
\begin{table}[htbp]
\centering
\caption{The five persona operations for personas, one row per edit.}\label{app:prompts:personas:operations}
\footnotesize\setlength{\tabcolsep}{4pt}
\begin{NiceTabular}{@{}>{\raggedright\arraybackslash}p{2.0cm};l;l;>{\raggedright\arraybackslash}p{2.6cm};>{\raggedright\arraybackslash}p{2.2cm};>{\raggedright\arraybackslash}p{2.2cm};l@{}}
\toprule
Edit & Kind & Wording & Descriptions & Source & Target & Span \\
\midrule
cost-conscious & lexical & yes & the price-conscious description and the detailed description & \texttt{\protect\frenchspacing budget-\allowbreak{}conscious} & \texttt{\protect\frenchspacing cost-\allowbreak{}conscious} & $[10, 26)$ \\
\dottedrule
before buying & syntactic & yes & the price-conscious description and the detailed description & \texttt{\protect\frenchspacing before you buy} & \texttt{\protect\frenchspacing before buying} & $[56, 70)$ \\
\dottedrule
35 years of age & lexical & yes & the detailed description & \texttt{\protect\frenchspacing 35 years old} & \texttt{\protect\frenchspacing 35 years of age} & $[80, 92)$ \\
\dottedrule
are employed & lexical & yes & the detailed description & \texttt{\protect\frenchspacing work} & \texttt{\protect\frenchspacing are employed} & $[98, 102)$ \\
\dottedrule
per year & lexical & yes & the detailed description & \texttt{\protect\frenchspacing a year} & \texttt{\protect\frenchspacing per year} & $[154, 160)$ \\
\bottomrule
\end{NiceTabular}
\end{table}
\endgroup{}
\floatnotes{prompts_personas_operations}

Table~\appref{app:prompts:manipulations:e1} specifies the newline, trailing-space, and zero-width-space sweeps, including the additional newline counts across all flight tasks. Zero-width characters can leave the rendered question unchanged; blank lines visibly change layout. Table~\appref{app:prompts:manipulations:e4} specifies the framing insertions on unequal-price tasks. The price flip is omitted from framing games to preserve the price configuration defining the cheaper option. Flipping a price can create a tie; it need not make that option pricier. Chat-template handling is documented with Table~\appref{tab:repro:models}.

\begingroup\long\def\floatNotesBody{
Insertions adjoin the question's line break on line 1, characters $[69, 70)$. Position means before or after that break; Unicode identifies the character. Count 0 is unchanged. Ranges include every integer count from $m$ to $n$.
}%
\begin{table}[htbp]
\centering
\caption{The presentation sweeps on the flight template.}\label{app:prompts:manipulations:e1}
\footnotesize
\begin{NiceTabular}{@{}l;l;l;l;l@{}}
\toprule
Sweep & Code point & Position & Counts & Tasks \\
\midrule
Line breaks & U+000A & after & 0 to 200 & 7 \\
\dottedrule
Trailing spaces & U+0020 & before & 0 to 50 & 7 \\
\dottedrule
Zero-width spaces & U+200B & before & 0 to 20 & 7 \\
\dottedrule
Line breaks & U+000A & after & 0, 1, 5, 20, and 100 & 1 to 16 \\
\bottomrule
\end{NiceTabular}
\end{table}
\endgroup{}
\floatnotes{prompts_manipulations_e1}

\begingroup\long\def\floatNotesBody{
Edits insert “only” before the cheaper price or “as much as” before the pricier price, replacing spaces after “costs” at $[85,86)$ (``A'') or $[144,145)$ (``B''). Cells identify options and edited words. Each phrase is one framing-phrase player: 50 players total.
}%
\begin{table}[htbp]
\centering
\caption{The framing edits on the eight flight tasks with unequal prices.}\label{app:prompts:manipulations:e4}
\footnotesize
\begin{NiceTabular}{@{}l;l;l;l;l@{}}
\toprule
 & \multicolumn{2}{c}{Price} & \multicolumn{2}{c}{Framing edit} \\
\cmidrule(lr){2-3}\cmidrule(lr){4-5}
Task & ``A'' & ``B'' & ``only'' & ``as much as'' \\
\midrule
3 & \$200 & \$350 & ``A'', \texttt{\protect\frenchspacing costs only \$200,} & ``B'', \texttt{\protect\frenchspacing costs as much as \$350,} \\
\dottedrule
4 & \$200 & \$350 & ``A'', \texttt{\protect\frenchspacing costs only \$200,} & ``B'', \texttt{\protect\frenchspacing costs as much as \$350,} \\
\dottedrule
7 & \$200 & \$350 & ``A'', \texttt{\protect\frenchspacing costs only \$200,} & ``B'', \texttt{\protect\frenchspacing costs as much as \$350,} \\
\dottedrule
8 & \$200 & \$350 & ``A'', \texttt{\protect\frenchspacing costs only \$200,} & ``B'', \texttt{\protect\frenchspacing costs as much as \$350,} \\
\dottedrule
9 & \$350 & \$200 & ``B'', \texttt{\protect\frenchspacing costs only \$200,} & ``A'', \texttt{\protect\frenchspacing costs as much as \$350,} \\
\dottedrule
10 & \$350 & \$200 & ``B'', \texttt{\protect\frenchspacing costs only \$200,} & ``A'', \texttt{\protect\frenchspacing costs as much as \$350,} \\
\dottedrule
13 & \$350 & \$200 & ``B'', \texttt{\protect\frenchspacing costs only \$200,} & ``A'', \texttt{\protect\frenchspacing costs as much as \$350,} \\
\dottedrule
14 & \$350 & \$200 & ``B'', \texttt{\protect\frenchspacing costs only \$200,} & ``A'', \texttt{\protect\frenchspacing costs as much as \$350,} \\
\bottomrule
\end{NiceTabular}
\end{table}
\endgroup{}
\floatnotes{prompts_manipulations_e4}
 \clearpage
\section{Formal details}\label{part:formal}

I state the games and their allocation properties, then use hypothetical examples to delimit their interpretation. Each example specifies its own response function; none reports model measurements. The common rule compares marginal contributions within a chosen game. It does not make different intervention questions interchangeable. Table~\appref{tab:notation} collects the notation for the games, allocation, and measurement.

\subsection{Games and allocation}

Let $f(x)$ be the log-odds of ``A'' against ``B'' at prompt $x$ under the two-label response rule. A game has a set $N$ of $n=|N|$ players and, for each coalition $S\subseteq N$, a prompt $x_S$ in which exactly the players of $S$ are present. In the edit game, $x_S$ is the intact prompt with the compatible edits in $S$ applied, so the reference $x_\varnothing$ is the intact prompt. In the word game, $x_S$ restores the components in $S$ and replaces the rest, so the reference $x_\varnothing$ has every component replaced. Both games have the value
\begin{equation}\label{eq:online-games}
 v(S)=f(x_S)-f(x_\varnothing),
\end{equation}
so $v(\varnothing)=0$. The coalition form of the Shapley value is
\begin{equation}\label{eq:online-shapley}
 \phi_i(v)=\sum_{S\subseteq N\setminus\{i\}}
 \frac{|S|!(n-|S|-1)!}{n!}\,[v(S\cup\{i\})-v(S)].
\end{equation}
Exactly $|S|!(n-|S|-1)!$ of the $n!$ inclusion orders put precisely $S$ before $i$. Thus this formula equals the uniform-order average \citep{app:shapley1953value}.

For games $v,w$ on $N$ and real constants $a,b$, the four properties have the following formal statements:
\begin{align}
 \sum_{i\in N}\phi_i(v)&=v(N), &&\text{efficiency};\label{eq:online-efficiency}\\
 v(S\cup\{i\})=v(S\cup\{j\})\ \forall S\subseteq N\setminus\{i,j\}
 &\ \Longrightarrow\ \phi_i(v)=\phi_j(v), &&\text{symmetry};\label{eq:online-symmetry}\\
 v(S\cup\{i\})=v(S)\ \forall S\subseteq N\setminus\{i\}
 &\ \Longrightarrow\ \phi_i(v)=0, &&\text{null player};\label{eq:online-null}\\
 \phi_i(av+bw)&=a\phi_i(v)+b\phi_i(w), &&\text{linearity}.\label{eq:online-linearity}
\end{align}
Efficiency follows by telescoping along each complete order. Exchanging symmetric players pairs equal contributions; null contributions vanish; averaging preserves linear combinations. These properties characterize the allocation rule on finite games, without endorsing a reference or classification.

\begin{table}[t]
\centering
\caption{Notation for the games, allocation, and measurement.}\label{tab:notation}
\footnotesize\setlength{\tabcolsep}{4pt}
\begin{NiceTabular}{@{}>{\raggedright\arraybackslash}p{2.4cm};p{7.0cm};>{\raggedright\arraybackslash}p{4.3cm}@{}}
\toprule
Symbol & Meaning & Defined in \\
\midrule
$x,x_S$ & A prompt, and the prompt of coalition $S$ & Equation~\appeqref{eq:online-games} \\
\dottedrule
$x_\varnothing,x_N$ & The reference (the prompt of the empty coalition) and the complete prompt & Equation~\appeqref{eq:online-games} \\
\dottedrule
$p(\mathrm{A};x)$, $p(\mathrm{B};x)$ & Conditional probabilities of answering ``A'' and ``B'' at prompt $x$, rescaled to sum to one & Equation~\appeqref{eq:online-extraction} \\
\dottedrule
$f(x)$ & Log-odds of ``A'' against ``B'' at prompt $x$ & Equation~\appeqref{eq:online-extraction} \\
\dottedrule
$p_{\mathrm{raw}}(\mathrm{A};x)$, $p_{\mathrm{raw}}(\mathrm{B};x)$, $c(x)$ & Raw next-token probabilities of the two answer labels at prompt $x$, and their sum, the unrestricted mass on both labels & Equation~\appeqref{eq:online-extraction} \\
\dottedrule
$N,n$ & Players of the game under discussion and their number, $n=|N|$ & Equation~\appeqref{eq:online-shapley} \\
\dottedrule
$S$ & Coalition, a subset of the game's players & Equation~\appeqref{eq:online-games} \\
\dottedrule
$v(S)$ & Value of coalition $S$, the score of its prompt minus the score of the reference & Equation~\appeqref{eq:online-games} \\
\dottedrule
$v,\phi_i(v)$ & A game and player $i$'s Shapley allocation & Equation~\appeqref{eq:online-shapley} \\
\dottedrule
$\omega\in\Omega,S_i(\omega)$ & An inclusion order in the set of all orders, and the players preceding $i$ in it & Main text, \emph{Framework and design} \\
\dottedrule
$\mathcal A$ & Attribute-value players & Main text, \emph{Framework and design} \\
\dottedrule
$M_{\mathcal A},M_{\mathrm{other}}$ & Absolute final attribution masses for attribute and remaining players & Main text, \emph{Framework and design} \\
\dottedrule
$q$ & Attribute mass divided by total absolute attribution mass & Main text, \emph{Framework and design} \\
\dottedrule
$R$ & Attribute-to-other mass ratio, when defined & Appendix~\ref{part:counts-pooling} \\
\dottedrule
$T,\Delta_i^{(t)}$ & Number of sampled orders, and player $i$'s contribution in the $t$-th sampled order $\omega_t$ & Equation~\appeqref{eq:online-monte-carlo} \\
\dottedrule
$\widehat\phi_i,\widehat{\mathrm{se}}$ & Estimated allocation and its Monte Carlo standard error & Equation~\appeqref{eq:online-monte-carlo} \\
\dottedrule
$f_{\mathrm{res}}$ & Empirical measurement floor from engine checks & Appendix~\ref{part:engine-precision} \\
\bottomrule
\end{NiceTabular}
\end{table}

\subsection{Compatible edits}

Represent an edit as $(g,h,u)$, replacing original characters in the nonempty half-open span $[g,h)$ with text $u$. For disjoint spans sorted by location, simultaneous application to a prompt $x$ of length $L$ is
\begin{equation}\label{eq:online-apply}
 x[0{:}g_1]u_1x[h_1{:}g_2]u_2\cdots u_mx[h_m{:}L].
\end{equation}
Each operation may contain several such edits. Sequential application gives the same string if later coordinates are shifted by the net length changes to their left. Every untouched segment and replacement then appears once in its original positional order. Disjointness is sufficient, including adjacent spans. Overlapping substitutions need an explicit joint rule, such as a composite player. Two insertions at the same empty span are not automatically compatible. Calculation orders never resolve conflicting text edits.

\subsection{Relations between games}

A player bijection preserving every coalition payoff guarantees identical attributions. This condition is sufficient, not necessary. Masked-prompt scores alone do not determine responses to edited prompts outside that evaluated set.

Align two word games' players on a common index set, adding null players if needed. Let the two games be $v$ and $v'$, with complete prompts $x_N,x'_N$ and references $x_\varnothing,x'_\varnothing$. Subtracting their efficiency identities gives
\begin{equation}\label{eq:online-aligned}
 \sum_i[\phi_i(v')-\phi_i(v)]
 =f(x'_N)-f(x_N)-[f(x'_\varnothing)-f(x_\varnothing)].
\end{equation}
A common reference payoff removes the bracketed adjustment. Both games must be evaluated; this accounting identity neither predicts the edited response nor equates word-value changes with edit-player allocations.

\paragraph*{Hypothetical nonagreement.}
Let $a\in\{0,1\}$ denote an attribute and $t=0,1,2$ masked, original, and equivalent text. For a constant $M$ and $f(a,t)=a+M\mathbf1[t=2]$, where the indicator $\mathbf1[\cdot]$ equals one when its condition holds, the word game from $(0,0)$ to $(1,1)$ assigns $(1,0)$, giving attribute share one. The intact wording edit has value $M$, unrestricted by that word game. Conversely, set $f(a,0)=-C$ and $f(a,1)=f(a,2)=a$, with $C\ge0$. Word values are $(1/2,C+1/2)$, and the attribute share $1/[2(C+1)]$ approaches zero. The intact wording effect remains zero.

\paragraph*{Hypothetical complementarity and redundancy.}
For indicators $y_1,y_2$ of ``costs'' and ``\$200,'' $f(y_1,y_2)=2y_1y_2$ assigns one to each. A context word receives credit through interaction. If the coalition scores are instead $(0,2,2,2)$ in the order neither, first alone, second alone, both, each complete-prompt removal effect is zero while each allocation is one. Removal measures necessity in the full context; allocation divides the endpoint contrast.

\paragraph*{Hypothetical cancellation and additivity.}
For two edit indicators, $f(y_1,y_2)=-1+2y_1+2y_2-4y_1y_2$ gives each edit marginal effects $+2$ and $-2$, hence zero attribution. Taking absolute values before averaging would instead give two. The share uses absolute final averages and is undefined when all are zero. Formatting can also act additively: $f(a,t)=a+0.2t$ has zero interaction contrast and nonzero formatting credit.

\subsection{Counts and pooling}\label{part:counts-pooling}

For the attribute set $\mathcal A$, define absolute masses $M_{\mathcal A}=\sum_{i\in\mathcal A}|\phi_i|$ and $M_{\mathrm{other}}=\sum_{i\in N\setminus\mathcal A}|\phi_i|$. The attribute share is $q=M_{\mathcal A}/(M_{\mathcal A}+M_{\mathrm{other}})$ when total mass is positive. The ratio $R=M_{\mathcal A}/M_{\mathrm{other}}$, when defined, gives $q=R/(1+R)$. These definitions take magnitudes after signed averaging.

\paragraph*{Hypothetical counts.}
Four attribute players and 45 others, each with unit magnitude, give share $4/49$. Giving each attribute magnitude ten gives $40/85$, despite a tenfold larger magnitude per attribute. Similarly, a price magnitude ten and ten edit magnitudes two give price-to-edits ratio $1/2$, although each edit is smaller. These are arithmetic illustrations, not benchmarks for consumer validity.

\paragraph*{Hypothetical pooling counterexample.}
With attribute player $a$ and other players $b,c$, let
\begin{equation}\label{eq:online-pooling}
 v(S)=\mathbf1[a\in S]-\tfrac13\mathbf1[b\in S]
       -\mathbf1[\{a,b,c\}\subseteq S].
\end{equation}
Values $(2/3,-2/3,-1/3)$ give attribute-to-other mass ratio $2/3$. Pooling $b,c$ gives $(1/2,-5/6)$ and ratio $3/5$; the attribute share falls from $2/5$ to $3/8$. Under the sufficient restriction $v(S)=v_{\mathcal A}(S\cap\mathcal A)+v_{\mathrm{other}}(S\cap(N\setminus\mathcal A))$ for attribute and other blocks, pooling $N\setminus\mathcal A$ preserves attribute values and replaces its mass by $|\sum_{i\in N\setminus\mathcal A}\phi_i|\le\sum_{i\in N\setminus\mathcal A}|\phi_i|$. Under that restriction the share weakly increases; without one, no direction follows. Displaying sums of existing word magnitudes is a different operation.
 \clearpage
\section{Measurement and numerical uncertainty}\label{part:measurement}

A language model assigns a probability to each possible next token given every token of its input, including whitespace and characters with no visible glyph. I read those probabilities directly and make no claim about how the model computes them.

I distinguish extracting the declared response, numerical differences in model scoring, and approximation of the allocation. Table~\appref{tab:repro:validation} addresses the first two; Figure~\ref{fig:grid-convergence} and Tables~\ref{tab:grid-calibration} and~\appref{tab:efficiency-residuals} address the estimator. Their checks have different targets and cannot substitute for one another. Log-odds also distinguish response movements near the ends of the probability scale, where a sizable score change can produce a small probability change.

\subsection{The two-label response rule}

At the answer position, I request the raw log-probabilities of the two labels, $\log p_{\mathrm{raw}}(\mathrm{A};x)$ and $\log p_{\mathrm{raw}}(\mathrm{B};x)$, by their token identifiers. ``Raw'' means before temperature scaling or sampling processors. The conditional response and unrestricted label mass are
\begin{equation}\label{eq:online-extraction}
\begin{gathered}
 f(x)=\log p_{\mathrm{raw}}(\mathrm{A};x)-\log p_{\mathrm{raw}}(\mathrm{B};x),\qquad p(\mathrm{A};x)=\frac{1}{1+\exp(-f(x))}=1-p(\mathrm{B};x),\\
 c(x)=p_{\mathrm{raw}}(\mathrm{A};x)+p_{\mathrm{raw}}(\mathrm{B};x).
\end{gathered}
\end{equation}
The field called coverage stores $c(x)$. A small value describes little unrestricted mass on the allowed labels; it does not invalidate the conditional definition of $p(\mathrm{A};x)$. I read both labels directly, without discarding generated free-form answers. An unrestricted leading token could begin a longer answer, which this experiment does not observe.

A hard output mask allowing only ``A'' and ``B'' at the same context renormalizes their raw probabilities and leaves $f$ unchanged in exact arithmetic. Table~\appref{tab:repro:validation} compares this constrained output with raw-label scoring on the base prompts. The largest Qwen discrepancy is \tnFloorQwenThreeEightTwoSevenB{} log-odds; the other models' discrepancies are much smaller. This supports the conditional interpretation while exposing an engine qualification. It establishes no equivalence to a different answer prefix, structured response, or free-form protocol. An absent label in an interface's truncated token list is not a zero probability.

\subsection{Engine precision}\label{part:engine-precision}

The recorded checks vary sampling temperature, compare constrained scoring, append each label to check its extracted log-probability, change batch composition, and compare warm and cold caches. Temperature invariance checks that extraction precedes sampling transformations. The operational floor $f_{\mathrm{res}}$ is the maximum difference across the other checks: log-odds differences for masks, batches, and caches, but per-label log-probability differences for the appended-label check. It is an empirical diagnostic for these settings, not a universal error bound.

For Muse and both Gemma models, scoring is configured so that a prompt's score does not depend on the other prompts in its batch. That configuration is unavailable for Qwen's architecture, and its recorded floor is materially larger. Between-prompt score granularity is separate from repeatability: the table also reports the largest power-of-two step containing at least \tnStepShareMin{} of the flight word-game scores. The log-probability normalizer cancels in $f$, leaving differences of finite-precision logits. A repeatable score can therefore still move on a coarse grid.

\begingroup\long\def\floatNotesBody{
Test rows give maximum absolute changes in the indicated quantity. The floor is the maximum appended-label, mask, batch, or cache discrepancy, an empirical diagnostic. Coverage is unrestricted mass on ``A'' and ``B''. The step contains at least 0.95 of flight word-game scores; it describes between-prompt granularity.
}%
\begin{table}[htbp]
\centering
\caption{Validation results and measurement floors on the 48 base prompts.}\label{tab:repro:validation}
\footnotesize
\begin{NiceTabular}{@{}l;r;r;r;r@{}}
\toprule
 & Qwen 3.8 27B & Muse Glimmer & Gemma 4 31B & Gemma 4 12B \\
\midrule
Temperature, label log-probability & $0$ & $0$ & $0$ & $0$ \\
\dottedrule
Two-label mask, $f$ & $0.31$ & $1.9\times10^{-7}$ & $9.5\times10^{-7}$ & $7.2\times10^{-7}$ \\
\dottedrule
Appended label, label log-probability & $0.25$ & $0$ & $0$ & $0$ \\
\dottedrule
Batch composition, $f$ & $0.25$ & $0$ & $0$ & $0$ \\
\dottedrule
Warm cache, $f$ & $0$ & $0$ & $0$ & $0$ \\
\addlinespace
Coverage $c$, minimum & $0.990$ & $0.774$ & $0.947$ & $0.054$ \\
\dottedrule
Coverage $c$, median & $1 - 9.2\times10^{-4}$ & $0.937$ & $1 - 6.0\times10^{-7}$ & $1 - 3.5\times10^{-5}$ \\
\dottedrule
Coverage $c$, maximum & $1 - 6.9\times10^{-5}$ & $0.971$ & $1 - 1.2\times10^{-7}$ & $1 - 2.0\times10^{-7}$ \\
\addlinespace
Floor $f_{\mathrm{res}}$ & $0.31$ & $1.9\times10^{-7}$ & $9.5\times10^{-7}$ & $7.2\times10^{-7}$ \\
\dottedrule
Log-odds step & $1/16$ & $1/16$ & $1/128$ & $1/32$ \\
\bottomrule
\end{NiceTabular}
\end{table}
\endgroup{}
\floatnotes{repro_validation}

\subsection{Order sampling and contrasts}

For the $t$-th sampled order $\omega_t$, let $\Delta_i^{(t)}$ be player $i$'s marginal contribution in it. I sample $T$ orders $\omega_1,\dots,\omega_T$ uniformly and estimate
\begin{equation}\label{eq:online-monte-carlo}
 \widehat\phi_i=\frac1T\sum_{t=1}^T\Delta_i^{(t)},\qquad
 \widehat{\mathrm{se}}(\widehat\phi_i)
 =\sqrt{\frac{\sum_t(\Delta_i^{(t)}-\widehat\phi_i)^2}{T(T-1)}}.
\end{equation}
This is permutation Monte Carlo averaging \citep{app:castro2009polynomial}. The main word games use 1,000 orders. Component intervals use $\widehat\phi_i\pm1.96\widehat{\mathrm{se}}(\widehat\phi_i)$. Exact enumeration of smaller games removes order-sampling error alone.

For a within-game contrast, I compute each row's difference before its standard error. Equivalently, its variance includes the covariance between components. For independent base and framed streams, the variance of a signed difference is the sum of their estimated variances. Equal row indices do not create pairing. Alignment follows text and semantic role, including the inserted framing component explicitly.

In Muse's unframed running example, the Shapley value of the travel time of ``A'' minus that of ``B'' is \tnExampleContrastTime{} log-odds, with standard error \tnExampleContrastTimeSe{} computed from their row-wise differences.

For masses and shares, I draw 1,000 bootstrap samples of complete rows with replacement. Each resample first averages the signed contributions, then takes absolute values and the desired ratio. Percentile endpoints at 2.5\% and 97.5\% give the reported interval. For independent games, resampling occurs separately within each. This preserves within-row dependence. Absolute values are nondifferentiable at zero; cancellation and a small denominator can make the approximation unreliable. An all-zero denominator is undefined.

These intervals condition on stored payoffs. Reused scores can carry shared engine error that resampling orders does not remove. A single-reading floor is not the standard error of an averaged attribution. Selected contrasts do not establish simultaneous certainty about all component ranks. The interquartile ranges in Table~\appref{tab:persona-shares} describe heterogeneity across the sixteen fixed tasks, not order-sampling or human-population uncertainty.

\subsection{Convergence and an exact check}

Figure~\appref{fig:grid-convergence} extends Qwen's flight task 7 word-game stream from 1,000 to 10,000 orders. The largest component change is \tnGridConvergenceMaxChange{} log-odds; \tnGridConvergenceOutside{} enlarged-run estimate lies outside its earlier interval. The longer calculation contains the earlier orders. It is neither independent nor the exact word-game solution.

\begingroup\long\def\floatNotesBody{
Left: running player estimates, with thick squares for attributes and a measurement-floor band. Right: standard errors on log-log axes. The reference is $\kappa/\sqrt{T}$; $\kappa$ is the median of standard error times $\sqrt{T}$ across players at the last checkpoint.
}%
\begin{figure}[tbp]
\centering
\includegraphics[width=\textwidth]{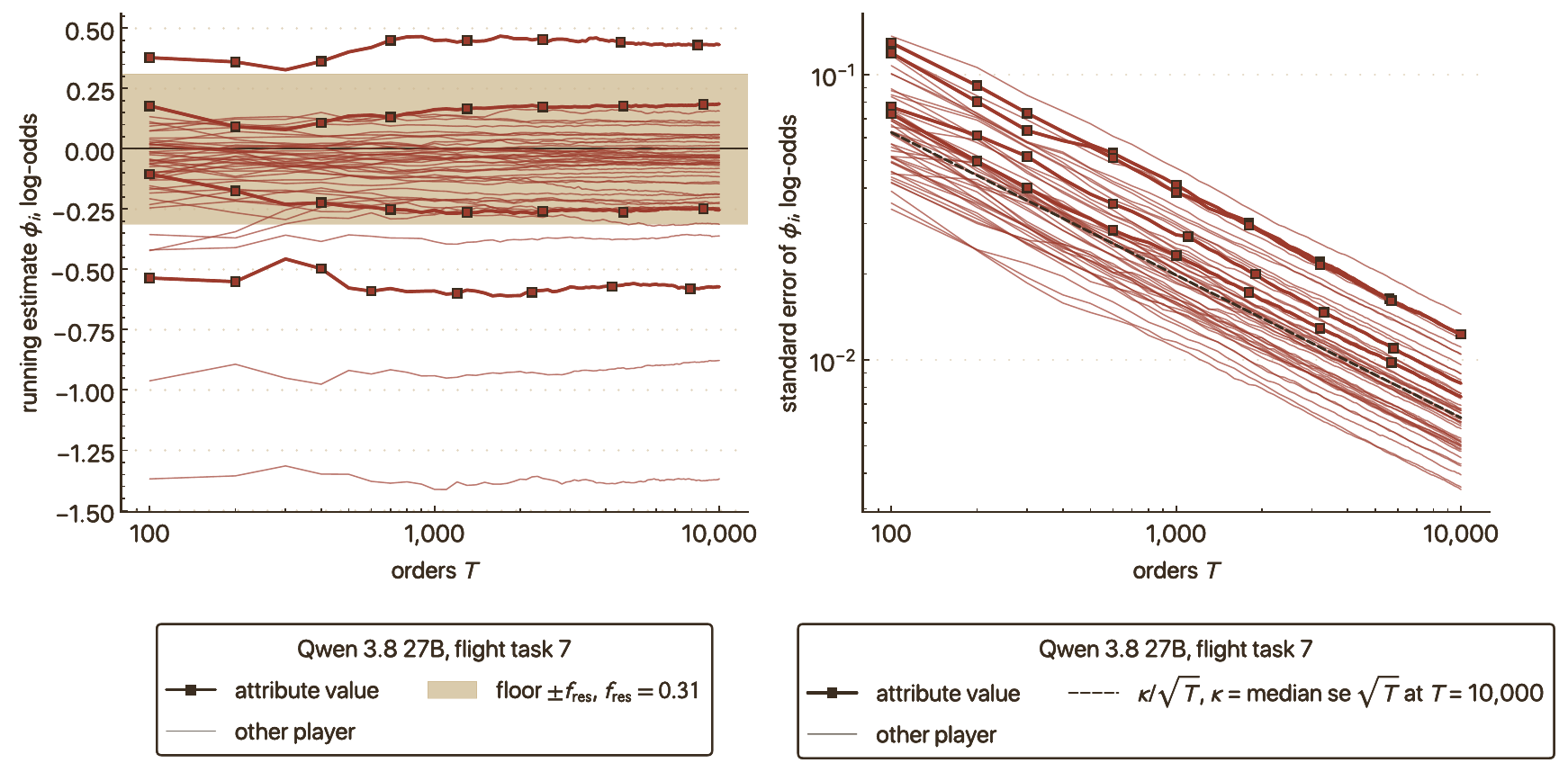}
\caption{Running Shapley estimates and their standard errors on the convergence game, flight task 7 on Qwen 3.8 27B, against the number of sampled orders $T$.}\label{fig:grid-convergence}
\end{figure}
\endgroup{}
\figurenotes{grid_convergence}

Table~\appref{tab:grid-calibration} compares sampled attribution with the exact ten-span game. Maximum errors at 100, 1,000, and 10,000 orders are \tnGridCalibrationMaxAbsErrorTOneZeroZero, \tnGridCalibrationMaxAbsErrorTOneZeroZeroZero, and \tnGridCalibrationMaxAbsErrorTOneZeroZeroZeroZero{} log-odds. All exact component values fall within the displayed intervals in these checks. This selected estimator-calibration example does not establish nominal interval coverage in general. The familiar $T^{-1/2}$ approximation rate concerns averaging fixed payoffs.

\begingroup\long\def\floatNotesBody{
The ten-span game is exact. Largest error is the maximum absolute component error in log-odds; the next column multiplies it by $\sqrt{T}$. Coverage counts exact values within 1.96 standard errors of estimates; this does not establish general interval coverage.
}%
\begin{table}[htbp]
\centering
\caption{Monte Carlo estimates against exact Shapley values in the calibration game, the span-level game of flight task 7 on Qwen 3.8 27B.}\label{tab:grid-calibration}
\footnotesize
\begin{NiceTabular}{@{}r;r;r;r@{}}
\toprule
$T$ & Largest error & Error $\times \sqrt{T}$ & Coverage \\
\midrule
$100$ & $0.21$ & $2.1$ & $10$ of $10$ \\
\dottedrule
$1{,}000$ & $0.071$ & $2.2$ & $10$ of $10$ \\
\dottedrule
$10{,}000$ & $0.025$ & $2.5$ & $10$ of $10$ \\
\bottomrule
\end{NiceTabular}
\end{table}
\endgroup{}
\floatnotes{grid_calibration}

Table~\appref{tab:efficiency-residuals} checks the signed sum. Each complete order telescopes to the endpoint contrast, so even one order can satisfy efficiency while estimating individual contributions poorly. This check establishes arithmetic consistency, not convergence.

\begingroup\long\def\floatNotesBody{
Cells give the maximum absolute gap between summed attribution and the endpoint contrast, in log-odds. Parentheses count games per model; n/a means none. Edit games use the retained set. Aggregation rejects residuals above $1.0\times10^{-6}$. Efficiency alone does not establish precise component estimates.
}%
\begin{table}[htbp]
\centering
\caption{Largest efficiency residual by study and model.}\label{tab:efficiency-residuals}
\footnotesize\setlength{\tabcolsep}{3pt}
\begin{NiceTabular}{@{}l;r;r;r;r@{}}
\toprule
Games & Qwen 3.8 27B & Muse Glimmer & Gemma 4 31B & Gemma 4 12B \\
\midrule
Word games, flights ($16$) & $1.8\times10^{-15}$ & $8.9\times10^{-16}$ & $2.1\times10^{-14}$ & $7.1\times10^{-15}$ \\
\dottedrule
Reduced edit games ($48$) & $1.8\times10^{-15}$ & $8.9\times10^{-16}$ & $7.1\times10^{-15}$ & $3.6\times10^{-15}$ \\
\dottedrule
Framed word games ($16$) & $1.8\times10^{-15}$ & $8.9\times10^{-16}$ & $1.8\times10^{-14}$ & $7.1\times10^{-15}$ \\
\dottedrule
Reduced framing edit games ($16$) & $4.4\times10^{-16}$ & $2.2\times10^{-16}$ & $8.9\times10^{-16}$ & $8.9\times10^{-16}$ \\
\dottedrule
Word games, hotels, smartphones ($32$) & $3.6\times10^{-15}$ & $8.9\times10^{-16}$ & $2.1\times10^{-14}$ & $7.1\times10^{-15}$ \\
\dottedrule
Persona word games ($48$) & $3.6\times10^{-15}$ & $8.9\times10^{-16}$ & $3.2\times10^{-14}$ & $1.1\times10^{-14}$ \\
\dottedrule
Persona edit games ($48$) & $3.6\times10^{-15}$ & $8.9\times10^{-16}$ & $1.4\times10^{-14}$ & $3.6\times10^{-15}$ \\
\dottedrule
Grid, references ($96$) & $1.8\times10^{-15}$ & n/a & n/a & n/a \\
\dottedrule
Grid, spans ($16$) & $1.8\times10^{-15}$ & n/a & n/a & n/a \\
\dottedrule
Grid, pooled grouping ($16$) & $1.3\times10^{-15}$ & n/a & n/a & n/a \\
\bottomrule
\end{NiceTabular}
\end{table}
\endgroup{}
\floatnotes{efficiency_residuals}
 \clearpage
\section{Full retained results}\label{part:results}

I collect the full displays behind the selected demonstrations. Read signed effects with their response orientation, absolute shares with their denominator, and intervals with their source of uncertainty. These are results for fixed tasks and protocols, not estimates of prevalence across marketing applications.

\subsection{Presentation and edit attribution}\label{part:presentation}\label{part:edits}

Figure~\appref{fig:e1} gives all insertion sweeps and, in its last column, 1, 5, 20, and 100 extra line breaks on every flight task. The larger Gemma's newline score range is \tnEOneNewlinesRangeFGemmaFourThreeOneB{} although its probabilities remain close to zero. Qwen's smaller movements require its larger engine floor in Table~\appref{tab:repro:validation}. Probability movement and crossing the equal-probability boundary are different summaries.

\begingroup\long\def\floatNotesBody{
Rows are models. First three panels: the probability $p(\mathrm{A};x)$ of ``A'' on flight task 7, with model-specific logit limits. Fourth: $f(n)-f(0)$ across all 16 flight tasks by category, at 1, 5, 20, 100 line breaks. Tan bands: plus or minus the measurement floor.
}%
\begin{figure}[tbp]
\centering
\includegraphics[width=\textwidth]{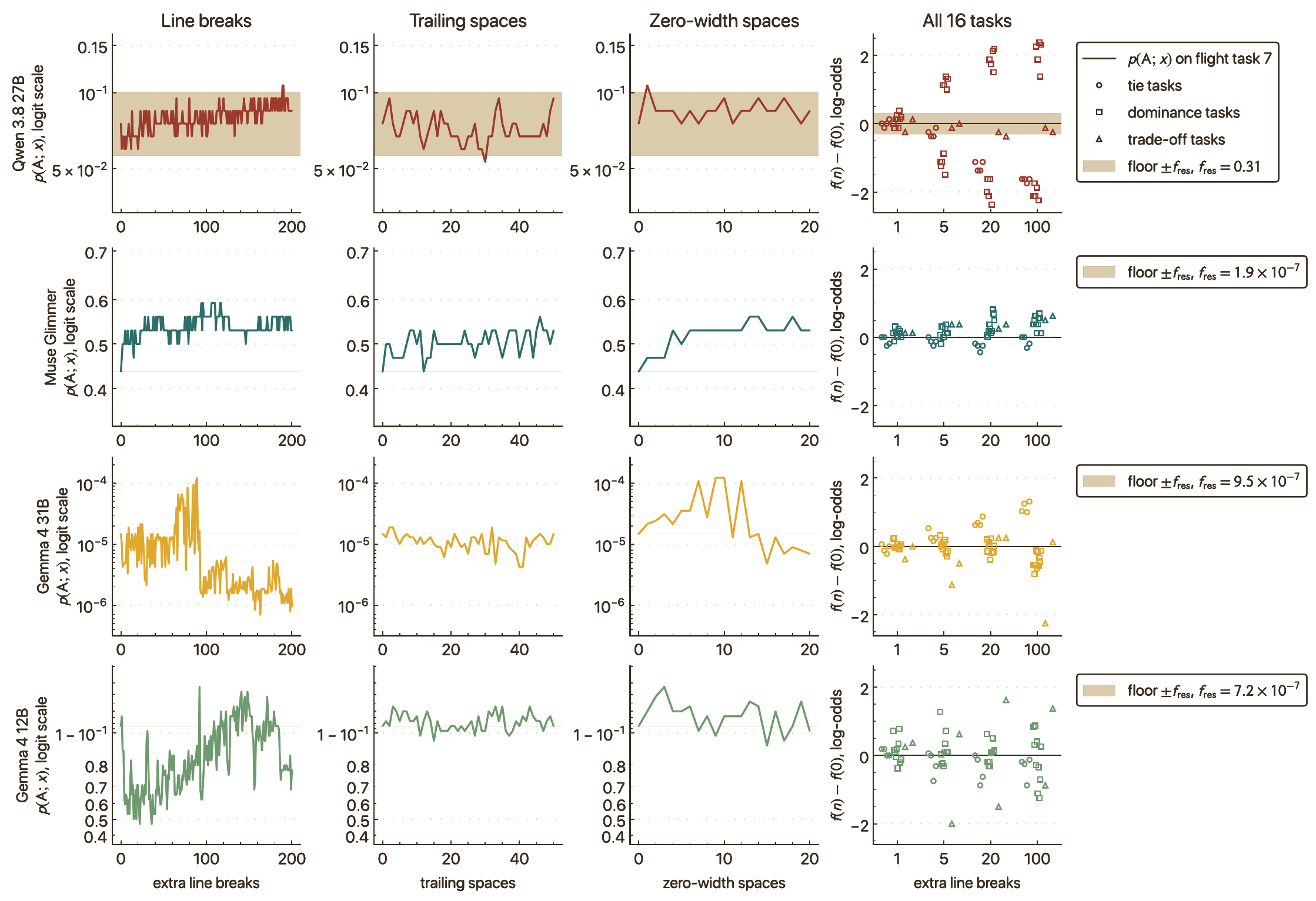}
\caption{The insertion sweeps: the probability $p(\mathrm{A};x)$ of ``A'' on flight task 7 under each inserted character, and the change in log-odds on every flight task, by model.}\label{fig:e1}
\end{figure}
\endgroup{}
\figurenotes{e1_sweeps}

Table~\appref{tab:edit-games} summarizes every retained edit by template and model. Figures~\ref{fig:edit-games-hotel} and~\appref{fig:edit-games-smartphone} retain the hotel and smartphone task distributions. The price player flips the price of ``A'' to its other level, so its direction depends on the starting task. The combined price-to-edits ratio compares one absolute allocation with a sum across edits; it does not describe a typical individual rewrite.

\begingroup\long\def\floatNotesBody{
Cells give median absolute Shapley values across 16 tasks, with bracketed interquartile ranges, in log-odds. Games are enumerated exactly. The final row summarizes the price magnitude divided by total edit magnitude. Italic labels name the formatting edits; the italic last row is a ratio, not a player.
}%
\begin{table}[htbp]
\centering
\caption{The reduced edit games by edit, template, and model, one panel per template.}\label{tab:edit-games}
\footnotesize\setlength{\tabcolsep}{4pt}
\begin{NiceTabular}{@{}>{\raggedright\arraybackslash}p{4.2cm};c;c;c;c@{}}
\toprule
Edit & \begin{tabular}[t]{@{}c@{}}Qwen 3.8\\27B\end{tabular} & \begin{tabular}[t]{@{}c@{}}Muse\\Glimmer\end{tabular} & \begin{tabular}[t]{@{}c@{}}Gemma 4\\31B\end{tabular} & \begin{tabular}[t]{@{}c@{}}Gemma 4\\12B\end{tabular} \\
\midrule
\multicolumn{5}{@{}l}{\textit{Flight}} \\
``the following'' to ``these'' & \begin{tabular}[t]{@{}c@{}}$0.16$\\$[0.11, 0.24]$\end{tabular} & \begin{tabular}[t]{@{}c@{}}$0.04$\\$[0.02, 0.07]$\end{tabular} & \begin{tabular}[t]{@{}c@{}}$0.19$\\$[0.07, 0.30]$\end{tabular} & \begin{tabular}[t]{@{}c@{}}$0.32$\\$[0.24, 0.41]$\end{tabular} \\
\dottedrule
``costs'' to ``is priced at'' & \begin{tabular}[t]{@{}c@{}}$0.21$\\$[0.14, 0.28]$\end{tabular} & \begin{tabular}[t]{@{}c@{}}$0.15$\\$[0.08, 0.19]$\end{tabular} & \begin{tabular}[t]{@{}c@{}}$0.07$\\$[0.05, 0.22]$\end{tabular} & \begin{tabular}[t]{@{}c@{}}$0.19$\\$[0.15, 0.38]$\end{tabular} \\
\dottedrule
``Flight'' to ``Option'' & \begin{tabular}[t]{@{}c@{}}$0.24$\\$[0.19, 0.38]$\end{tabular} & \begin{tabular}[t]{@{}c@{}}$0.31$\\$[0.27, 0.34]$\end{tabular} & \begin{tabular}[t]{@{}c@{}}$0.59$\\$[0.11, 0.99]$\end{tabular} & \begin{tabular}[t]{@{}c@{}}$0.20$\\$[0.06, 0.22]$\end{tabular} \\
\dottedrule
``Answer with a single'' to ``Reply with one'' & \begin{tabular}[t]{@{}c@{}}$0.68$\\$[0.26, 1.51]$\end{tabular} & \begin{tabular}[t]{@{}c@{}}$0.29$\\$[0.15, 0.44]$\end{tabular} & \begin{tabular}[t]{@{}c@{}}$0.28$\\$[0.21, 0.33]$\end{tabular} & \begin{tabular}[t]{@{}c@{}}$0.27$\\$[0.21, 0.36]$\end{tabular} \\
\dottedrule
\textit{blank line after the question} & \begin{tabular}[t]{@{}c@{}}$0.08$\\$[0.03, 0.10]$\end{tabular} & \begin{tabular}[t]{@{}c@{}}$0.06$\\$[0.04, 0.10]$\end{tabular} & \begin{tabular}[t]{@{}c@{}}$0.11$\\$[0.07, 0.14]$\end{tabular} & \begin{tabular}[t]{@{}c@{}}$0.15$\\$[0.08, 0.36]$\end{tabular} \\
\dottedrule
\textit{join the option lines} & \begin{tabular}[t]{@{}c@{}}$0.17$\\$[0.08, 0.29]$\end{tabular} & \begin{tabular}[t]{@{}c@{}}$0.11$\\$[0.05, 0.24]$\end{tabular} & \begin{tabular}[t]{@{}c@{}}$0.06$\\$[0.03, 0.17]$\end{tabular} & \begin{tabular}[t]{@{}c@{}}$0.21$\\$[0.12, 0.35]$\end{tabular} \\
\dottedrule
\textit{comma to semicolon after the prices} & \begin{tabular}[t]{@{}c@{}}$0.22$\\$[0.15, 0.28]$\end{tabular} & \begin{tabular}[t]{@{}c@{}}$0.09$\\$[0.06, 0.11]$\end{tabular} & \begin{tabular}[t]{@{}c@{}}$0.03$\\$[0.01, 0.14]$\end{tabular} & \begin{tabular}[t]{@{}c@{}}$0.21$\\$[0.11, 0.24]$\end{tabular} \\
\dottedrule
\textit{two spaces after the label colons} & \begin{tabular}[t]{@{}c@{}}$0.39$\\$[0.17, 0.53]$\end{tabular} & \begin{tabular}[t]{@{}c@{}}$0.07$\\$[0.03, 0.10]$\end{tabular} & \begin{tabular}[t]{@{}c@{}}$0.13$\\$[0.11, 0.28]$\end{tabular} & \begin{tabular}[t]{@{}c@{}}$0.07$\\$[0.04, 0.11]$\end{tabular} \\
\dottedrule
\textit{trailing space after the question mark} & \begin{tabular}[t]{@{}c@{}}$0.17$\\$[0.12, 0.26]$\end{tabular} & \begin{tabular}[t]{@{}c@{}}$0.10$\\$[0.05, 0.14]$\end{tabular} & \begin{tabular}[t]{@{}c@{}}$0.14$\\$[0.11, 0.17]$\end{tabular} & \begin{tabular}[t]{@{}c@{}}$0.16$\\$[0.10, 0.22]$\end{tabular} \\
\dottedrule
\textit{options in parentheses} & \begin{tabular}[t]{@{}c@{}}$0.06$\\$[0.02, 0.13]$\end{tabular} & \begin{tabular}[t]{@{}c@{}}$0.33$\\$[0.23, 0.37]$\end{tabular} & \begin{tabular}[t]{@{}c@{}}$0.04$\\$[0.03, 0.09]$\end{tabular} & \begin{tabular}[t]{@{}c@{}}$0.27$\\$[0.18, 0.42]$\end{tabular} \\
\dottedrule
price of ``A'' flipped & \begin{tabular}[t]{@{}c@{}}$3.88$\\$[1.45, 6.73]$\end{tabular} & \begin{tabular}[t]{@{}c@{}}$1.51$\\$[0.37, 3.35]$\end{tabular} & \begin{tabular}[t]{@{}c@{}}$9.42$\\$[2.83, 38.51]$\end{tabular} & \begin{tabular}[t]{@{}c@{}}$9.74$\\$[6.24, 23.45]$\end{tabular} \\
\dottedrule
\textit{price over the summed edits} & \begin{tabular}[t]{@{}c@{}}$1.4$\\$[0.41, 3.8]$\end{tabular} & \begin{tabular}[t]{@{}c@{}}$1.1$\\$[0.26, 2.2]$\end{tabular} & \begin{tabular}[t]{@{}c@{}}$3.4$\\$[1.3, 14]$\end{tabular} & \begin{tabular}[t]{@{}c@{}}$4.2$\\$[2.4, 9.8]$\end{tabular} \\
\bottomrule
\end{NiceTabular}
\end{table}
\begin{table}[htbp]
\ContinuedFloat
\centering
\caption{The reduced edit games by edit, template, and model, continued.}\label{tab:edit-games:hotel}
\footnotesize\setlength{\tabcolsep}{4pt}
\begin{NiceTabular}{@{}>{\raggedright\arraybackslash}p{4.2cm};c;c;c;c@{}}
\toprule
Edit & \begin{tabular}[t]{@{}c@{}}Qwen 3.8\\27B\end{tabular} & \begin{tabular}[t]{@{}c@{}}Muse\\Glimmer\end{tabular} & \begin{tabular}[t]{@{}c@{}}Gemma 4\\31B\end{tabular} & \begin{tabular}[t]{@{}c@{}}Gemma 4\\12B\end{tabular} \\
\midrule
\multicolumn{5}{@{}l}{\textit{Hotel}} \\
``the following'' to ``these'' & \begin{tabular}[t]{@{}c@{}}$0.15$\\$[0.04, 0.19]$\end{tabular} & \begin{tabular}[t]{@{}c@{}}$0.06$\\$[0.04, 0.08]$\end{tabular} & \begin{tabular}[t]{@{}c@{}}$0.19$\\$[0.10, 0.35]$\end{tabular} & \begin{tabular}[t]{@{}c@{}}$0.63$\\$[0.44, 1.19]$\end{tabular} \\
\dottedrule
``hotels in Boston'' to ``Boston hotels'' & \begin{tabular}[t]{@{}c@{}}$0.11$\\$[0.07, 0.17]$\end{tabular} & \begin{tabular}[t]{@{}c@{}}$0.10$\\$[0.03, 0.18]$\end{tabular} & \begin{tabular}[t]{@{}c@{}}$0.11$\\$[0.06, 0.19]$\end{tabular} & \begin{tabular}[t]{@{}c@{}}$0.44$\\$[0.26, 0.51]$\end{tabular} \\
\dottedrule
``costs'' to ``is priced at'' & \begin{tabular}[t]{@{}c@{}}$0.14$\\$[0.10, 0.18]$\end{tabular} & \begin{tabular}[t]{@{}c@{}}$0.08$\\$[0.05, 0.14]$\end{tabular} & \begin{tabular}[t]{@{}c@{}}$0.34$\\$[0.16, 0.41]$\end{tabular} & \begin{tabular}[t]{@{}c@{}}$0.38$\\$[0.24, 0.52]$\end{tabular} \\
\dottedrule
``Hotel'' to ``Option'' & \begin{tabular}[t]{@{}c@{}}$0.11$\\$[0.07, 0.17]$\end{tabular} & \begin{tabular}[t]{@{}c@{}}$0.36$\\$[0.24, 0.40]$\end{tabular} & \begin{tabular}[t]{@{}c@{}}$0.39$\\$[0.24, 0.87]$\end{tabular} & \begin{tabular}[t]{@{}c@{}}$0.28$\\$[0.10, 0.50]$\end{tabular} \\
\dottedrule
``Answer with a single'' to ``Reply with one'' & \begin{tabular}[t]{@{}c@{}}$0.57$\\$[0.23, 1.28]$\end{tabular} & \begin{tabular}[t]{@{}c@{}}$0.28$\\$[0.06, 0.52]$\end{tabular} & \begin{tabular}[t]{@{}c@{}}$0.38$\\$[0.20, 0.53]$\end{tabular} & \begin{tabular}[t]{@{}c@{}}$0.20$\\$[0.11, 0.37]$\end{tabular} \\
\dottedrule
\textit{blank line after the question} & \begin{tabular}[t]{@{}c@{}}$0.13$\\$[0.11, 0.16]$\end{tabular} & \begin{tabular}[t]{@{}c@{}}$0.03$\\$[0.01, 0.08]$\end{tabular} & \begin{tabular}[t]{@{}c@{}}$0.13$\\$[0.06, 0.17]$\end{tabular} & \begin{tabular}[t]{@{}c@{}}$0.14$\\$[0.10, 0.26]$\end{tabular} \\
\dottedrule
\textit{join the option lines} & \begin{tabular}[t]{@{}c@{}}$0.11$\\$[0.07, 0.16]$\end{tabular} & \begin{tabular}[t]{@{}c@{}}$0.21$\\$[0.11, 0.29]$\end{tabular} & \begin{tabular}[t]{@{}c@{}}$0.19$\\$[0.11, 0.22]$\end{tabular} & \begin{tabular}[t]{@{}c@{}}$0.37$\\$[0.18, 0.49]$\end{tabular} \\
\dottedrule
\textit{comma to semicolon after the prices} & \begin{tabular}[t]{@{}c@{}}$0.11$\\$[0.03, 0.15]$\end{tabular} & \begin{tabular}[t]{@{}c@{}}$0.03$\\$[0.02, 0.06]$\end{tabular} & \begin{tabular}[t]{@{}c@{}}$0.05$\\$[0.02, 0.14]$\end{tabular} & \begin{tabular}[t]{@{}c@{}}$0.07$\\$[0.03, 0.10]$\end{tabular} \\
\dottedrule
\textit{two spaces after the label colons} & \begin{tabular}[t]{@{}c@{}}$0.15$\\$[0.08, 0.28]$\end{tabular} & \begin{tabular}[t]{@{}c@{}}$0.06$\\$[0.05, 0.09]$\end{tabular} & \begin{tabular}[t]{@{}c@{}}$0.10$\\$[0.04, 0.14]$\end{tabular} & \begin{tabular}[t]{@{}c@{}}$0.20$\\$[0.11, 0.33]$\end{tabular} \\
\dottedrule
\textit{trailing space after the question mark} & \begin{tabular}[t]{@{}c@{}}$0.27$\\$[0.04, 0.34]$\end{tabular} & \begin{tabular}[t]{@{}c@{}}$0.06$\\$[0.04, 0.08]$\end{tabular} & \begin{tabular}[t]{@{}c@{}}$0.09$\\$[0.03, 0.15]$\end{tabular} & \begin{tabular}[t]{@{}c@{}}$0.13$\\$[0.04, 0.23]$\end{tabular} \\
\dottedrule
\textit{options in parentheses} & \begin{tabular}[t]{@{}c@{}}$0.09$\\$[0.07, 0.11]$\end{tabular} & \begin{tabular}[t]{@{}c@{}}$0.29$\\$[0.24, 0.31]$\end{tabular} & \begin{tabular}[t]{@{}c@{}}$0.12$\\$[0.09, 0.25]$\end{tabular} & \begin{tabular}[t]{@{}c@{}}$0.51$\\$[0.30, 0.63]$\end{tabular} \\
\dottedrule
price of ``A'' flipped & \begin{tabular}[t]{@{}c@{}}$3.80$\\$[1.92, 6.79]$\end{tabular} & \begin{tabular}[t]{@{}c@{}}$1.70$\\$[0.55, 3.27]$\end{tabular} & \begin{tabular}[t]{@{}c@{}}$10.99$\\$[3.99, 27.76]$\end{tabular} & \begin{tabular}[t]{@{}c@{}}$9.65$\\$[3.62, 19.92]$\end{tabular} \\
\dottedrule
\textit{price over the summed edits} & \begin{tabular}[t]{@{}c@{}}$1.6$\\$[0.73, 4.9]$\end{tabular} & \begin{tabular}[t]{@{}c@{}}$0.96$\\$[0.35, 2.1]$\end{tabular} & \begin{tabular}[t]{@{}c@{}}$2.1$\\$[1.3, 10]$\end{tabular} & \begin{tabular}[t]{@{}c@{}}$1.8$\\$[0.99, 5.0]$\end{tabular} \\
\bottomrule
\end{NiceTabular}
\end{table}
\begin{table}[htbp]
\ContinuedFloat
\centering
\caption{The reduced edit games by edit, template, and model, continued.}\label{tab:edit-games:smartphone}
\footnotesize\setlength{\tabcolsep}{4pt}
\begin{NiceTabular}{@{}>{\raggedright\arraybackslash}p{4.2cm};c;c;c;c@{}}
\toprule
Edit & \begin{tabular}[t]{@{}c@{}}Qwen 3.8\\27B\end{tabular} & \begin{tabular}[t]{@{}c@{}}Muse\\Glimmer\end{tabular} & \begin{tabular}[t]{@{}c@{}}Gemma 4\\31B\end{tabular} & \begin{tabular}[t]{@{}c@{}}Gemma 4\\12B\end{tabular} \\
\midrule
\multicolumn{5}{@{}l}{\textit{Smartphone}} \\
``the following'' to ``these'' & \begin{tabular}[t]{@{}c@{}}$0.06$\\$[0.04, 0.11]$\end{tabular} & \begin{tabular}[t]{@{}c@{}}$0.06$\\$[0.04, 0.10]$\end{tabular} & \begin{tabular}[t]{@{}c@{}}$0.49$\\$[0.41, 0.69]$\end{tabular} & \begin{tabular}[t]{@{}c@{}}$0.52$\\$[0.27, 0.79]$\end{tabular} \\
\dottedrule
\textit{purpose phrase moved to the front} & \begin{tabular}[t]{@{}c@{}}$0.64$\\$[0.07, 1.09]$\end{tabular} & \begin{tabular}[t]{@{}c@{}}$0.15$\\$[0.05, 0.28]$\end{tabular} & \begin{tabular}[t]{@{}c@{}}$0.41$\\$[0.29, 0.95]$\end{tabular} & \begin{tabular}[t]{@{}c@{}}$0.74$\\$[0.63, 1.02]$\end{tabular} \\
\dottedrule
``costs'' to ``is priced at'' & \begin{tabular}[t]{@{}c@{}}$0.19$\\$[0.13, 0.25]$\end{tabular} & \begin{tabular}[t]{@{}c@{}}$0.15$\\$[0.09, 0.24]$\end{tabular} & \begin{tabular}[t]{@{}c@{}}$0.33$\\$[0.11, 0.98]$\end{tabular} & \begin{tabular}[t]{@{}c@{}}$0.40$\\$[0.19, 0.80]$\end{tabular} \\
\dottedrule
``Phone'' to ``Option'' & \begin{tabular}[t]{@{}c@{}}$0.18$\\$[0.09, 0.26]$\end{tabular} & \begin{tabular}[t]{@{}c@{}}$0.26$\\$[0.08, 0.44]$\end{tabular} & \begin{tabular}[t]{@{}c@{}}$0.71$\\$[0.39, 1.16]$\end{tabular} & \begin{tabular}[t]{@{}c@{}}$0.66$\\$[0.47, 0.86]$\end{tabular} \\
\dottedrule
``Answer with a single'' to ``Reply with one'' & \begin{tabular}[t]{@{}c@{}}$0.59$\\$[0.17, 1.03]$\end{tabular} & \begin{tabular}[t]{@{}c@{}}$0.12$\\$[0.05, 0.37]$\end{tabular} & \begin{tabular}[t]{@{}c@{}}$0.23$\\$[0.17, 0.38]$\end{tabular} & \begin{tabular}[t]{@{}c@{}}$0.20$\\$[0.10, 0.48]$\end{tabular} \\
\dottedrule
\textit{blank line after the question} & \begin{tabular}[t]{@{}c@{}}$0.05$\\$[0.03, 0.11]$\end{tabular} & \begin{tabular}[t]{@{}c@{}}$0.06$\\$[0.02, 0.09]$\end{tabular} & \begin{tabular}[t]{@{}c@{}}$0.19$\\$[0.13, 0.26]$\end{tabular} & \begin{tabular}[t]{@{}c@{}}$0.07$\\$[0.05, 0.16]$\end{tabular} \\
\dottedrule
\textit{join the option lines} & \begin{tabular}[t]{@{}c@{}}$0.07$\\$[0.01, 0.15]$\end{tabular} & \begin{tabular}[t]{@{}c@{}}$0.06$\\$[0.04, 0.16]$\end{tabular} & \begin{tabular}[t]{@{}c@{}}$0.09$\\$[0.03, 0.26]$\end{tabular} & \begin{tabular}[t]{@{}c@{}}$0.24$\\$[0.09, 0.34]$\end{tabular} \\
\dottedrule
\textit{comma to semicolon after the prices} & \begin{tabular}[t]{@{}c@{}}$0.10$\\$[0.06, 0.13]$\end{tabular} & \begin{tabular}[t]{@{}c@{}}$0.04$\\$[0.02, 0.07]$\end{tabular} & \begin{tabular}[t]{@{}c@{}}$0.15$\\$[0.05, 0.32]$\end{tabular} & \begin{tabular}[t]{@{}c@{}}$0.17$\\$[0.07, 0.36]$\end{tabular} \\
\dottedrule
\textit{two spaces after the label colons} & \begin{tabular}[t]{@{}c@{}}$0.25$\\$[0.07, 0.31]$\end{tabular} & \begin{tabular}[t]{@{}c@{}}$0.17$\\$[0.09, 0.27]$\end{tabular} & \begin{tabular}[t]{@{}c@{}}$0.20$\\$[0.16, 0.33]$\end{tabular} & \begin{tabular}[t]{@{}c@{}}$0.25$\\$[0.14, 0.42]$\end{tabular} \\
\dottedrule
\textit{trailing space after the question mark} & \begin{tabular}[t]{@{}c@{}}$0.17$\\$[0.14, 0.24]$\end{tabular} & \begin{tabular}[t]{@{}c@{}}$0.05$\\$[0.01, 0.08]$\end{tabular} & \begin{tabular}[t]{@{}c@{}}$0.08$\\$[0.06, 0.13]$\end{tabular} & \begin{tabular}[t]{@{}c@{}}$0.10$\\$[0.03, 0.18]$\end{tabular} \\
\dottedrule
\textit{options in parentheses} & \begin{tabular}[t]{@{}c@{}}$0.09$\\$[0.03, 0.14]$\end{tabular} & \begin{tabular}[t]{@{}c@{}}$0.35$\\$[0.23, 0.40]$\end{tabular} & \begin{tabular}[t]{@{}c@{}}$0.16$\\$[0.04, 0.43]$\end{tabular} & \begin{tabular}[t]{@{}c@{}}$0.36$\\$[0.12, 0.45]$\end{tabular} \\
\dottedrule
price of ``A'' flipped & \begin{tabular}[t]{@{}c@{}}$4.92$\\$[2.27, 9.40]$\end{tabular} & \begin{tabular}[t]{@{}c@{}}$2.33$\\$[0.62, 3.48]$\end{tabular} & \begin{tabular}[t]{@{}c@{}}$17.11$\\$[6.26, 37.36]$\end{tabular} & \begin{tabular}[t]{@{}c@{}}$11.96$\\$[5.46, 21.51]$\end{tabular} \\
\dottedrule
\textit{price over the summed edits} & \begin{tabular}[t]{@{}c@{}}$1.7$\\$[0.71, 5.8]$\end{tabular} & \begin{tabular}[t]{@{}c@{}}$1.4$\\$[0.47, 1.9]$\end{tabular} & \begin{tabular}[t]{@{}c@{}}$2.6$\\$[1.5, 8.4]$\end{tabular} & \begin{tabular}[t]{@{}c@{}}$3.1$\\$[1.2, 6.8]$\end{tabular} \\
\bottomrule
\end{NiceTabular}
\end{table}
\endgroup{}
\floatnotes{edit_games}

\begingroup\long\def\floatNotesBody{
Points are tasks; bars are medians. Values are exact and allocate the change from the unedited prompt. The flip raises the price of ``A'' on tasks 1 to 8 and lowers it on tasks 9 to 16. The axis is linear within $\pm0.1$, logarithmic beyond.
}%
\begin{figure}[tbp]
\centering
\includegraphics[width=\textwidth]{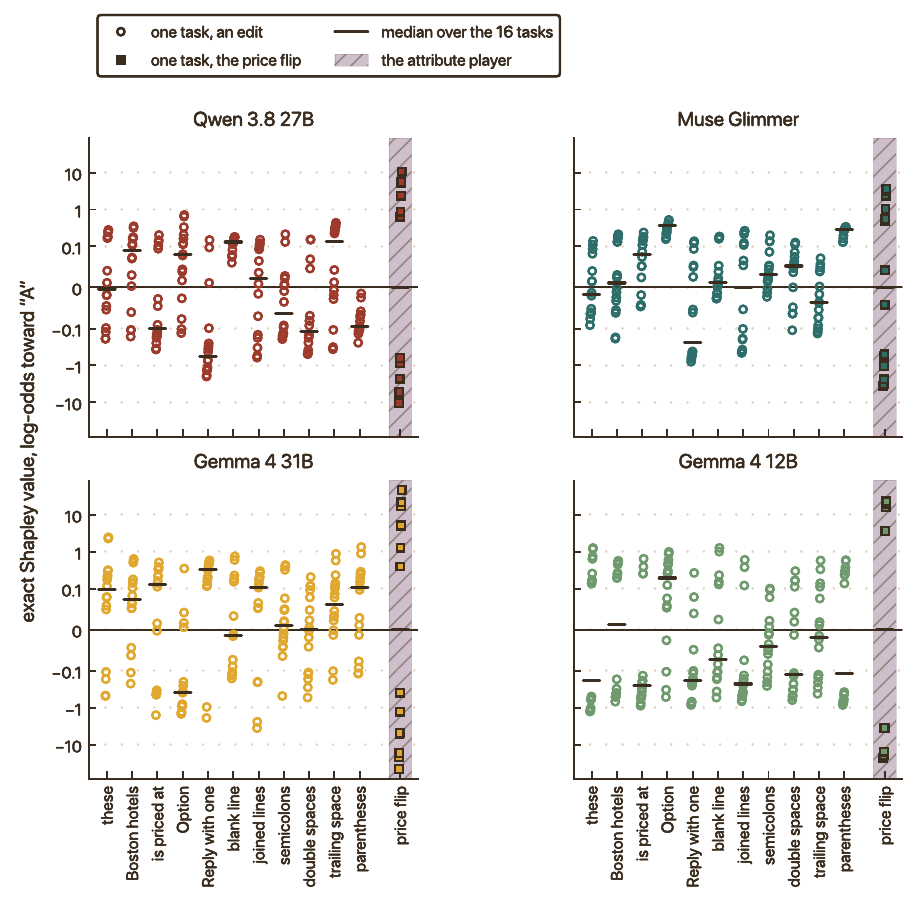}
\caption{Exact Shapley values of the 11 retained edits and the price flip over the 16 hotel tasks, by model.}\label{fig:edit-games-hotel}
\end{figure}
\endgroup{}
\figurenotes{edit_games_hotel}

\begingroup\long\def\floatNotesBody{
Points are tasks; bars are medians. Values are exact and allocate the change from the unedited prompt. The flip raises the price of ``A'' on tasks 1 to 8 and lowers it on tasks 9 to 16. The axis is linear within $\pm0.1$, logarithmic beyond.
}%
\begin{figure}[tbp]
\centering
\includegraphics[width=\textwidth]{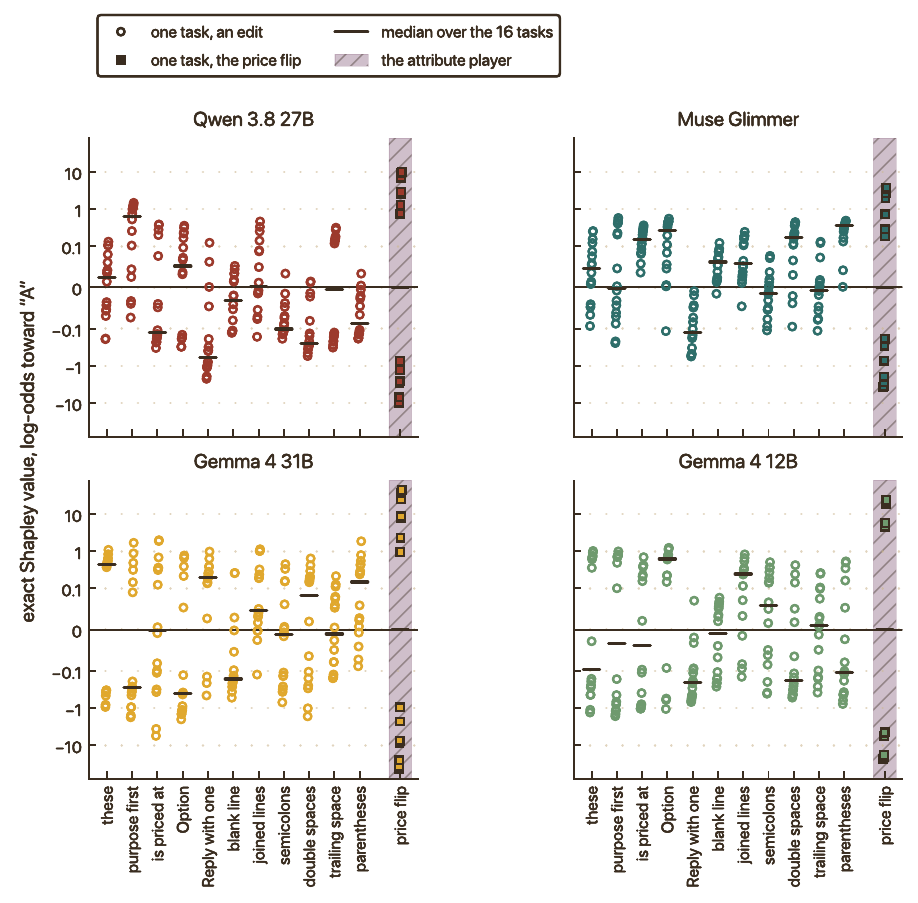}
\caption{Exact Shapley values of the 11 retained edits and the price flip over the 16 smartphone tasks, by model.}\label{fig:edit-games-smartphone}
\end{figure}
\endgroup{}
\figurenotes{edit_games_smartphone}

\subsection{Shapley values beside the alone and left-out effects}

Table~\appref{tab:matched-summary} summarizes, for every model and product template, how far the Shapley values agree with the alone and left-out effects within a task. Each statistic sets the signed Shapley values against one of the two effects; the summaries describe heterogeneity across the fixed tasks. Table~\appref{tab:matched-words} supplies every player of Muse Glimmer's flight task 7, including the words omitted from the main display.

\begingroup\long\def\floatNotesBody{
Medians [interquartile ranges] of per-task statistics; Tasks counts defined statistics. Floor-small magnitudes become zero. Sign agreement excludes pairs containing zero; zero/available and unavailable counts sum across tasks. Signed Spearman includes all available pairs, zeros tied; constants are undefined. Edits hold original price, excluding its row. Alone: the edit on the unedited prompt, or the word restored to the all-masked prompt. Left out: the score the complete prompt loses without the component.
}%
\begin{table}[htbp]
\centering
\caption{Within-task agreement between Shapley values and the alone and left-out effects.}\label{tab:matched-summary}
\footnotesize\setlength{\tabcolsep}{4pt}
\begin{NiceTabular}{@{}l;l;r;c;r;c;r;r@{}}
\toprule
Game & Contrast & \begin{tabular}[t]{@{}c@{}}Sign agreement\\(\%)\end{tabular} & Tasks & \begin{tabular}[t]{@{}c@{}}Signed rank\\correlation\end{tabular} & Tasks & \begin{tabular}[t]{@{}c@{}}Zero / available\\pairs\end{tabular} & \begin{tabular}[t]{@{}c@{}}Unavailable\\pairs\end{tabular} \\
\midrule
\multicolumn{8}{@{}l}{\textit{Qwen 3.8 27B, flight}} \\
Edits & Alone & \begin{tabular}[t]{@{}c@{}}$100.0$\\$[100.0, 100.0]$\end{tabular} & 15 & \begin{tabular}[t]{@{}c@{}}$0.76$\\$[0.59, 0.84]$\end{tabular} & 16 & 119/160 & 0 \\
\dottedrule
Edits & Left out & \begin{tabular}[t]{@{}c@{}}$100.0$\\$[100.0, 100.0]$\end{tabular} & 15 & \begin{tabular}[t]{@{}c@{}}$0.79$\\$[0.67, 0.85]$\end{tabular} & 16 & 116/160 & 0 \\
\dottedrule
Words & Alone & \begin{tabular}[t]{@{}c@{}}$50.0$\\$[48.2, 100.0]$\end{tabular} & 16 & \begin{tabular}[t]{@{}c@{}}$0.25$\\$[0.13, 0.39]$\end{tabular} & 16 & 655/784 & 0 \\
\dottedrule
Words & Left out & \begin{tabular}[t]{@{}c@{}}$75.0$\\$[72.9, 100.0]$\end{tabular} & 16 & \begin{tabular}[t]{@{}c@{}}$0.46$\\$[0.37, 0.50]$\end{tabular} & 16 & 643/784 & 0 \\
\addlinespace
\multicolumn{8}{@{}l}{\textit{Qwen 3.8 27B, hotel}} \\
Edits & Alone & \begin{tabular}[t]{@{}c@{}}$100.0$\\$[100.0, 100.0]$\end{tabular} & 14 & \begin{tabular}[t]{@{}c@{}}$0.62$\\$[0.57, 0.74]$\end{tabular} & 14 & 145/176 & 0 \\
\dottedrule
Edits & Left out & \begin{tabular}[t]{@{}c@{}}$100.0$\\$[100.0, 100.0]$\end{tabular} & 13 & \begin{tabular}[t]{@{}c@{}}$0.74$\\$[0.58, 0.77]$\end{tabular} & 13 & 150/176 & 0 \\
\dottedrule
Words & Alone & \begin{tabular}[t]{@{}c@{}}$80.0$\\$[59.3, 88.4]$\end{tabular} & 16 & \begin{tabular}[t]{@{}c@{}}$0.32$\\$[0.05, 0.38]$\end{tabular} & 16 & 587/720 & 0 \\
\dottedrule
Words & Left out & \begin{tabular}[t]{@{}c@{}}$88.9$\\$[74.1, 100.0]$\end{tabular} & 16 & \begin{tabular}[t]{@{}c@{}}$0.55$\\$[0.41, 0.58]$\end{tabular} & 16 & 569/720 & 0 \\
\addlinespace
\multicolumn{8}{@{}l}{\textit{Qwen 3.8 27B, smartphone}} \\
Edits & Alone & \begin{tabular}[t]{@{}c@{}}$100.0$\\$[100.0, 100.0]$\end{tabular} & 14 & \begin{tabular}[t]{@{}c@{}}$0.72$\\$[0.55, 0.77]$\end{tabular} & 16 & 143/176 & 0 \\
\dottedrule
Edits & Left out & \begin{tabular}[t]{@{}c@{}}$100.0$\\$[100.0, 100.0]$\end{tabular} & 14 & \begin{tabular}[t]{@{}c@{}}$0.61$\\$[0.52, 0.81]$\end{tabular} & 16 & 144/176 & 0 \\
\dottedrule
Words & Alone & \begin{tabular}[t]{@{}c@{}}$87.5$\\$[65.6, 100.0]$\end{tabular} & 16 & \begin{tabular}[t]{@{}c@{}}$0.28$\\$[0.21, 0.35]$\end{tabular} & 16 & 544/640 & 0 \\
\dottedrule
Words & Left out & \begin{tabular}[t]{@{}c@{}}$77.5$\\$[66.7, 100.0]$\end{tabular} & 16 & \begin{tabular}[t]{@{}c@{}}$0.46$\\$[0.37, 0.65]$\end{tabular} & 16 & 508/640 & 0 \\
\bottomrule
\end{NiceTabular}
\end{table}
\begin{table}[htbp]
\ContinuedFloat
\centering
\caption{Shapley values beside the alone and left-out effects, continued.}\label{tab:matched-summary:1}
\footnotesize\setlength{\tabcolsep}{4pt}
\begin{NiceTabular}{@{}l;l;r;c;r;c;r;r@{}}
\toprule
Game & Contrast & \begin{tabular}[t]{@{}c@{}}Sign agreement\\(\%)\end{tabular} & Tasks & \begin{tabular}[t]{@{}c@{}}Signed rank\\correlation\end{tabular} & Tasks & \begin{tabular}[t]{@{}c@{}}Zero / available\\pairs\end{tabular} & \begin{tabular}[t]{@{}c@{}}Unavailable\\pairs\end{tabular} \\
\midrule
\multicolumn{8}{@{}l}{\textit{Muse Glimmer, flight}} \\
Edits & Alone & \begin{tabular}[t]{@{}c@{}}$86.6$\\$[82.5, 89.2]$\end{tabular} & 16 & \begin{tabular}[t]{@{}c@{}}$0.81$\\$[0.74, 0.88]$\end{tabular} & 16 & 31/160 & 0 \\
\dottedrule
Edits & Left out & \begin{tabular}[t]{@{}c@{}}$76.4$\\$[69.2, 88.9]$\end{tabular} & 16 & \begin{tabular}[t]{@{}c@{}}$0.73$\\$[0.54, 0.88]$\end{tabular} & 16 & 21/160 & 0 \\
\dottedrule
Words & Alone & \begin{tabular}[t]{@{}c@{}}$58.5$\\$[36.2, 62.2]$\end{tabular} & 16 & \begin{tabular}[t]{@{}c@{}}$0.20$\\$[0.14, 0.23]$\end{tabular} & 16 & 32/784 & 0 \\
\dottedrule
Words & Left out & \begin{tabular}[t]{@{}c@{}}$70.1$\\$[66.7, 71.7]$\end{tabular} & 16 & \begin{tabular}[t]{@{}c@{}}$0.48$\\$[0.42, 0.50]$\end{tabular} & 16 & 41/784 & 0 \\
\addlinespace
\multicolumn{8}{@{}l}{\textit{Muse Glimmer, hotel}} \\
Edits & Alone & \begin{tabular}[t]{@{}c@{}}$87.5$\\$[80.0, 89.4]$\end{tabular} & 16 & \begin{tabular}[t]{@{}c@{}}$0.80$\\$[0.74, 0.84]$\end{tabular} & 16 & 21/176 & 0 \\
\dottedrule
Edits & Left out & \begin{tabular}[t]{@{}c@{}}$80.9$\\$[63.4, 89.4]$\end{tabular} & 16 & \begin{tabular}[t]{@{}c@{}}$0.73$\\$[0.58, 0.87]$\end{tabular} & 16 & 24/176 & 0 \\
\dottedrule
Words & Alone & \begin{tabular}[t]{@{}c@{}}$65.9$\\$[49.4, 71.0]$\end{tabular} & 16 & \begin{tabular}[t]{@{}c@{}}$0.41$\\$[0.36, 0.44]$\end{tabular} & 16 & 8/720 & 0 \\
\dottedrule
Words & Left out & \begin{tabular}[t]{@{}c@{}}$66.3$\\$[63.7, 72.3]$\end{tabular} & 16 & \begin{tabular}[t]{@{}c@{}}$0.47$\\$[0.35, 0.59]$\end{tabular} & 16 & 40/720 & 0 \\
\addlinespace
\multicolumn{8}{@{}l}{\textit{Muse Glimmer, smartphone}} \\
Edits & Alone & \begin{tabular}[t]{@{}c@{}}$78.9$\\$[74.4, 83.6]$\end{tabular} & 16 & \begin{tabular}[t]{@{}c@{}}$0.78$\\$[0.64, 0.81]$\end{tabular} & 16 & 39/176 & 0 \\
\dottedrule
Edits & Left out & \begin{tabular}[t]{@{}c@{}}$77.8$\\$[68.4, 83.6]$\end{tabular} & 16 & \begin{tabular}[t]{@{}c@{}}$0.82$\\$[0.70, 0.89]$\end{tabular} & 16 & 21/176 & 0 \\
\dottedrule
Words & Alone & \begin{tabular}[t]{@{}c@{}}$67.5$\\$[61.2, 73.1]$\end{tabular} & 16 & \begin{tabular}[t]{@{}c@{}}$0.38$\\$[0.24, 0.45]$\end{tabular} & 16 & 8/640 & 0 \\
\dottedrule
Words & Left out & \begin{tabular}[t]{@{}c@{}}$70.4$\\$[67.5, 76.9]$\end{tabular} & 16 & \begin{tabular}[t]{@{}c@{}}$0.58$\\$[0.41, 0.69]$\end{tabular} & 16 & 45/640 & 0 \\
\bottomrule
\end{NiceTabular}
\end{table}
\begin{table}[htbp]
\ContinuedFloat
\centering
\caption{Shapley values beside the alone and left-out effects, continued.}\label{tab:matched-summary:2}
\footnotesize\setlength{\tabcolsep}{4pt}
\begin{NiceTabular}{@{}l;l;r;c;r;c;r;r@{}}
\toprule
Game & Contrast & \begin{tabular}[t]{@{}c@{}}Sign agreement\\(\%)\end{tabular} & Tasks & \begin{tabular}[t]{@{}c@{}}Signed rank\\correlation\end{tabular} & Tasks & \begin{tabular}[t]{@{}c@{}}Zero / available\\pairs\end{tabular} & \begin{tabular}[t]{@{}c@{}}Unavailable\\pairs\end{tabular} \\
\midrule
\multicolumn{8}{@{}l}{\textit{Gemma 4 31B, flight}} \\
Edits & Alone & \begin{tabular}[t]{@{}c@{}}$70.0$\\$[60.0, 80.0]$\end{tabular} & 16 & \begin{tabular}[t]{@{}c@{}}$0.59$\\$[0.45, 0.70]$\end{tabular} & 16 & 6/160 & 0 \\
\dottedrule
Edits & Left out & \begin{tabular}[t]{@{}c@{}}$80.0$\\$[70.0, 90.0]$\end{tabular} & 16 & \begin{tabular}[t]{@{}c@{}}$0.73$\\$[0.69, 0.84]$\end{tabular} & 16 & 3/160 & 0 \\
\dottedrule
Words & Alone & \begin{tabular}[t]{@{}c@{}}$53.1$\\$[42.7, 67.7]$\end{tabular} & 16 & \begin{tabular}[t]{@{}c@{}}$0.23$\\$[0.14, 0.27]$\end{tabular} & 16 & 16/784 & 0 \\
\dottedrule
Words & Left out & \begin{tabular}[t]{@{}c@{}}$66.0$\\$[59.0, 74.0]$\end{tabular} & 16 & \begin{tabular}[t]{@{}c@{}}$0.52$\\$[0.24, 0.72]$\end{tabular} & 16 & 3/784 & 0 \\
\addlinespace
\multicolumn{8}{@{}l}{\textit{Gemma 4 31B, hotel}} \\
Edits & Alone & \begin{tabular}[t]{@{}c@{}}$72.7$\\$[68.4, 80.5]$\end{tabular} & 16 & \begin{tabular}[t]{@{}c@{}}$0.63$\\$[0.51, 0.76]$\end{tabular} & 16 & 4/176 & 0 \\
\dottedrule
Edits & Left out & \begin{tabular}[t]{@{}c@{}}$81.8$\\$[72.7, 90.2]$\end{tabular} & 16 & \begin{tabular}[t]{@{}c@{}}$0.87$\\$[0.80, 0.88]$\end{tabular} & 16 & 2/176 & 0 \\
\dottedrule
Words & Alone & \begin{tabular}[t]{@{}c@{}}$54.5$\\$[36.4, 68.2]$\end{tabular} & 16 & \begin{tabular}[t]{@{}c@{}}$0.10$\\$[0.08, 0.13]$\end{tabular} & 16 & 16/720 & 0 \\
\dottedrule
Words & Left out & \begin{tabular}[t]{@{}c@{}}$68.9$\\$[63.9, 71.1]$\end{tabular} & 16 & \begin{tabular}[t]{@{}c@{}}$0.58$\\$[0.34, 0.65]$\end{tabular} & 16 & 5/720 & 0 \\
\addlinespace
\multicolumn{8}{@{}l}{\textit{Gemma 4 31B, smartphone}} \\
Edits & Alone & \begin{tabular}[t]{@{}c@{}}$72.7$\\$[52.3, 74.5]$\end{tabular} & 16 & \begin{tabular}[t]{@{}c@{}}$0.68$\\$[0.43, 0.81]$\end{tabular} & 16 & 2/176 & 0 \\
\dottedrule
Edits & Left out & \begin{tabular}[t]{@{}c@{}}$85.9$\\$[81.4, 90.9]$\end{tabular} & 16 & \begin{tabular}[t]{@{}c@{}}$0.85$\\$[0.80, 0.90]$\end{tabular} & 16 & 2/176 & 0 \\
\dottedrule
Words & Alone & \begin{tabular}[t]{@{}c@{}}$48.7$\\$[40.4, 59.0]$\end{tabular} & 16 & \begin{tabular}[t]{@{}c@{}}$0.17$\\$[0.09, 0.25]$\end{tabular} & 16 & 16/640 & 0 \\
\dottedrule
Words & Left out & \begin{tabular}[t]{@{}c@{}}$67.5$\\$[65.0, 78.1]$\end{tabular} & 16 & \begin{tabular}[t]{@{}c@{}}$0.56$\\$[0.41, 0.65]$\end{tabular} & 16 & 2/640 & 0 \\
\bottomrule
\end{NiceTabular}
\end{table}
\begin{table}[htbp]
\ContinuedFloat
\centering
\caption{Shapley values beside the alone and left-out effects, continued.}\label{tab:matched-summary:3}
\footnotesize\setlength{\tabcolsep}{4pt}
\begin{NiceTabular}{@{}l;l;r;c;r;c;r;r@{}}
\toprule
Game & Contrast & \begin{tabular}[t]{@{}c@{}}Sign agreement\\(\%)\end{tabular} & Tasks & \begin{tabular}[t]{@{}c@{}}Signed rank\\correlation\end{tabular} & Tasks & \begin{tabular}[t]{@{}c@{}}Zero / available\\pairs\end{tabular} & \begin{tabular}[t]{@{}c@{}}Unavailable\\pairs\end{tabular} \\
\midrule
\multicolumn{8}{@{}l}{\textit{Gemma 4 12B, flight}} \\
Edits & Alone & \begin{tabular}[t]{@{}c@{}}$80.0$\\$[69.2, 88.9]$\end{tabular} & 16 & \begin{tabular}[t]{@{}c@{}}$0.76$\\$[0.65, 0.81]$\end{tabular} & 16 & 7/160 & 0 \\
\dottedrule
Edits & Left out & \begin{tabular}[t]{@{}c@{}}$89.4$\\$[80.0, 90.0]$\end{tabular} & 16 & \begin{tabular}[t]{@{}c@{}}$0.69$\\$[0.58, 0.85]$\end{tabular} & 16 & 4/160 & 0 \\
\dottedrule
Words & Alone & \begin{tabular}[t]{@{}c@{}}$74.5$\\$[31.1, 79.6]$\end{tabular} & 16 & \begin{tabular}[t]{@{}c@{}}$0.38$\\$[0.23, 0.43]$\end{tabular} & 16 & 0/784 & 0 \\
\dottedrule
Words & Left out & \begin{tabular}[t]{@{}c@{}}$74.2$\\$[69.4, 76.3]$\end{tabular} & 16 & \begin{tabular}[t]{@{}c@{}}$0.62$\\$[0.59, 0.65]$\end{tabular} & 16 & 12/784 & 0 \\
\addlinespace
\multicolumn{8}{@{}l}{\textit{Gemma 4 12B, hotel}} \\
Edits & Alone & \begin{tabular}[t]{@{}c@{}}$78.9$\\$[63.6, 81.8]$\end{tabular} & 16 & \begin{tabular}[t]{@{}c@{}}$0.73$\\$[0.60, 0.79]$\end{tabular} & 16 & 9/176 & 0 \\
\dottedrule
Edits & Left out & \begin{tabular}[t]{@{}c@{}}$81.8$\\$[68.4, 90.2]$\end{tabular} & 16 & \begin{tabular}[t]{@{}c@{}}$0.73$\\$[0.60, 0.85]$\end{tabular} & 16 & 7/176 & 0 \\
\dottedrule
Words & Alone & \begin{tabular}[t]{@{}c@{}}$69.3$\\$[18.2, 75.6]$\end{tabular} & 16 & \begin{tabular}[t]{@{}c@{}}$0.49$\\$[0.27, 0.54]$\end{tabular} & 16 & 16/720 & 0 \\
\dottedrule
Words & Left out & \begin{tabular}[t]{@{}c@{}}$77.5$\\$[71.8, 80.0]$\end{tabular} & 16 & \begin{tabular}[t]{@{}c@{}}$0.56$\\$[0.51, 0.60]$\end{tabular} & 16 & 9/720 & 0 \\
\addlinespace
\multicolumn{8}{@{}l}{\textit{Gemma 4 12B, smartphone}} \\
Edits & Alone & \begin{tabular}[t]{@{}c@{}}$81.8$\\$[70.5, 89.2]$\end{tabular} & 16 & \begin{tabular}[t]{@{}c@{}}$0.69$\\$[0.61, 0.79]$\end{tabular} & 16 & 4/176 & 0 \\
\dottedrule
Edits & Left out & \begin{tabular}[t]{@{}c@{}}$80.0$\\$[75.8, 90.9]$\end{tabular} & 16 & \begin{tabular}[t]{@{}c@{}}$0.81$\\$[0.76, 0.84]$\end{tabular} & 16 & 7/176 & 0 \\
\dottedrule
Words & Alone & \begin{tabular}[t]{@{}c@{}}$65.0$\\$[41.2, 80.0]$\end{tabular} & 16 & \begin{tabular}[t]{@{}c@{}}$0.51$\\$[0.35, 0.54]$\end{tabular} & 16 & 0/640 & 0 \\
\dottedrule
Words & Left out & \begin{tabular}[t]{@{}c@{}}$67.5$\\$[63.9, 73.1]$\end{tabular} & 16 & \begin{tabular}[t]{@{}c@{}}$0.49$\\$[0.41, 0.64]$\end{tabular} & 16 & 10/640 & 0 \\
\bottomrule
\end{NiceTabular}
\end{table}
\endgroup{}
\floatnotes{matched_summary}

\begingroup\long\def\floatNotesBody{
Positions count all players from the prompt's start. Values are log-odds toward ``A''; se is Monte Carlo standard error. Alone restores one player to the all-masked reference; Left out is the score the complete prompt loses when it is removed. Unavailable effects print n/a.
}%
\begin{table}[htbp]
\centering
\caption{Every word's Shapley value and its alone and left-out effects: Muse Glimmer, flight task 7.}\label{tab:matched-words}
\footnotesize\setlength{\tabcolsep}{4pt}
\begin{NiceTabular}{@{}r;l;r;r;r;r@{}}
\toprule
Position & Component & Shapley value & se & Alone & Left out \\
\midrule
\multicolumn{6}{@{}l}{\textit{}} \\
1 & ``Which'' & $-0.27$ & $0.03$ & $-0.31$ & $0.00$ \\
\dottedrule
2 & ``of'' & $-0.13$ & $0.02$ & $0.38$ & $-0.12$ \\
\dottedrule
3 & ``the'' & $-6.7\times10^{-4}$ & $0.02$ & $0.25$ & $0.00$ \\
\dottedrule
4 & ``following'' & $-0.09$ & $0.02$ & $0.75$ & $-0.25$ \\
\dottedrule
5 & ``flights'' & $-0.09$ & $0.02$ & $0.31$ & $-0.12$ \\
\dottedrule
6 & ``from'' & $-0.09$ & $0.02$ & $0.12$ & $-0.12$ \\
\dottedrule
7 & ``Chicago'' & $-0.04$ & $0.02$ & $0.00$ & $-0.50$ \\
\dottedrule
8 & ``to'' & $-0.05$ & $0.02$ & $-0.06$ & $-0.25$ \\
\dottedrule
9 & ``Atlanta'' & $-0.14$ & $0.02$ & $0.81$ & $-0.50$ \\
\dottedrule
10 & ``do'' & $-0.05$ & $0.02$ & $0.00$ & $0.00$ \\
\dottedrule
11 & ``you'' & $-0.07$ & $0.02$ & $0.81$ & $-0.50$ \\
\dottedrule
12 & ``prefer?'' & $0.29$ & $0.03$ & $-0.06$ & $0.38$ \\
\dottedrule
13 & a line break & $0.13$ & $0.02$ & $-0.19$ & $0.00$ \\
\dottedrule
14 & ``Flight'' & $0.03$ & $0.02$ & $0.75$ & $-0.75$ \\
\dottedrule
15 & ``A:'' & $-0.12$ & $0.04$ & $0.25$ & $-0.50$ \\
\dottedrule
16 & ``costs'' & $6.4\times10^{-3}$ & $0.02$ & $0.31$ & $0.00$ \\
\dottedrule
17 & ``\$200'' & $0.19$ & $0.02$ & $0.25$ & $1.88$ \\
\dottedrule
18 & ``,'' & $6.5\times10^{-3}$ & $0.01$ & $0.25$ & $-0.38$ \\
\dottedrule
19 & ``with'' & $-0.06$ & $0.02$ & $1.19$ & $-0.25$ \\
\dottedrule
20 & ``a'' & $0.05$ & $0.01$ & $0.50$ & $-0.12$ \\
\dottedrule
21 & ``total'' & $-0.02$ & $0.02$ & $0.31$ & $0.12$ \\
\dottedrule
22 & ``travel'' & $-0.03$ & $0.02$ & $1.25$ & $-0.75$ \\
\dottedrule
23 & ``time'' & $-0.05$ & $0.01$ & $0.81$ & $-0.62$ \\
\dottedrule
24 & ``of'' & $-0.04$ & $0.02$ & $0.75$ & $-1.38$ \\
\dottedrule
25 & ``5 hours'' & $-0.20$ & $0.02$ & $0.25$ & $-1.94$ \\
\bottomrule
\end{NiceTabular}
\end{table}
\begin{table}[htbp]
\ContinuedFloat
\centering
\caption{Every word beside its alone and left-out effects, continued.}\label{tab:matched-words:1}
\footnotesize\setlength{\tabcolsep}{4pt}
\begin{NiceTabular}{@{}r;l;r;r;r;r@{}}
\toprule
Position & Component & Shapley value & se & Alone & Left out \\
\midrule
\multicolumn{6}{@{}l}{\textit{}} \\
26 & ``.'' & $0.01$ & $0.01$ & $-0.06$ & $-0.25$ \\
\dottedrule
27 & a line break & $0.01$ & $0.02$ & $0.44$ & $0.25$ \\
\dottedrule
28 & ``Flight'' & $0.09$ & $0.02$ & $0.94$ & $-0.25$ \\
\dottedrule
29 & ``B:'' & $-0.95$ & $0.05$ & $-3.88$ & $-1.75$ \\
\dottedrule
30 & ``costs'' & $0.03$ & $0.01$ & $0.31$ & $1.12$ \\
\dottedrule
31 & ``\$350'' & $-0.03$ & $0.02$ & $-0.25$ & $1.50$ \\
\dottedrule
32 & ``,'' & $0.04$ & $9.3\times10^{-3}$ & $0.38$ & $0.00$ \\
\dottedrule
33 & ``with'' & $0.04$ & $0.02$ & $1.44$ & $0.50$ \\
\dottedrule
34 & ``a'' & $0.06$ & $0.01$ & $0.06$ & $0.50$ \\
\dottedrule
35 & ``total'' & $-0.04$ & $0.01$ & $0.19$ & $0.50$ \\
\dottedrule
36 & ``travel'' & $5.8\times10^{-3}$ & $0.02$ & $1.44$ & $0.63$ \\
\dottedrule
37 & ``time'' & $-0.02$ & $0.01$ & $0.62$ & $0.63$ \\
\dottedrule
38 & ``of'' & $-0.11$ & $0.02$ & $1.12$ & $-0.88$ \\
\dottedrule
39 & ``2 hours'' & $-0.37$ & $0.03$ & $0.06$ & $-2.31$ \\
\dottedrule
40 & ``.'' & $0.06$ & $0.01$ & $-0.31$ & $0.00$ \\
\dottedrule
41 & a line break & $0.09$ & $0.02$ & $-0.12$ & $0.12$ \\
\dottedrule
42 & ``Answer'' & $0.12$ & $0.05$ & $-0.25$ & $0.13$ \\
\dottedrule
43 & ``with'' & $-0.03$ & $0.03$ & $0.81$ & $0.50$ \\
\dottedrule
44 & ``a'' & $0.06$ & $0.02$ & $0.38$ & $0.25$ \\
\dottedrule
45 & ``single'' & $-0.07$ & $0.03$ & $1.37$ & $0.00$ \\
\dottedrule
46 & ``letter,'' & $0.11$ & $0.03$ & $0.25$ & $-0.75$ \\
\dottedrule
47 & ``A'' & $0.44$ & $0.04$ & $0.69$ & $-1.13$ \\
\dottedrule
48 & ``or'' & $-0.72$ & $0.04$ & $-0.12$ & $-3.06$ \\
\dottedrule
49 & ``B.'' & $-1.30$ & $0.05$ & $-4.94$ & $-2.00$ \\
\bottomrule
\end{NiceTabular}
\end{table}
\endgroup{}
\floatnotes{matched_words}

\subsection{Framing}\label{part:framing-personas}

Table~\appref{tab:framing-summary} distinguishes the direct unedited effect, frame allocation, and uniform mean over contexts. Effects whose magnitude is at or below the model's measurement floor count as zero. The floors differ by model; Qwen 3.8 27B's coarse floor of about \tnFloorQwenThreeEightTwoSevenB{} log-odds necessarily turns many paired effects into zeros (Table~\appref{tab:repro:validation}). Positive table scores favor ``A''; Figure~\appref{fig:e4-framing} orients paired effects toward the cheaper option. With $m$ surrounding edits, the players $N\setminus\{F\}$ other than the frame $F$, and paired effect $d(S)$, the frame receives
\begin{equation}\label{eq:online-frame}
 \phi_F=\sum_{S\subseteq N\setminus\{F\}}\frac{d(S)}{(m+1)\binom{m}{|S|}}.
\end{equation}
This weighting differs from an equal average over subsets. Muse's flight task 7 “only” allocation is \tnExampleFramePhi, while its direct effect is \tnExampleDirectEffect. A zero paired effect after rewording need not restore the original unframed response. The current cheaper-price protocol also differs from the earlier public framing example; it is not an exact replication.

\begingroup\long\def\floatNotesBody{
Scores favor ``A''. Unedited is the direct effect; Frame is exact frame attribution; Mean averages paired effects uniformly over 1,024 contexts. Final columns count negative, zero, and positive effects. Cheaper identifies the lower-price option; the number below each model is its engine floor.
}%
\begin{table}[htbp]
\centering
\caption{The reduced framing games by framing, model, and task, one panel per framing.}\label{tab:framing-summary}
\footnotesize\setlength{\tabcolsep}{4pt}
\begin{NiceTabular}{@{}l;l;l;r;r;r;r;r;r@{}}
\toprule
\begin{tabular}[t]{@{}l@{}}Model\\$f_{\mathrm{res}}$\end{tabular} & Task & Cheaper & Unedited & Frame & Mean & Below 0 & At 0 & Above 0 \\
\midrule
\multicolumn{9}{@{}l}{\textit{``only'' before the cheaper price}} \\
Qwen 3.8 27B & 3 & ``A'' & $0.00$ & $-0.11$ & $-0.12$ & $258$ & $708$ & $58$ \\
\dottedrule
$0.31$ & 4 & ``A'' & $0.00$ & $0.08$ & $0.04$ & $90$ & $786$ & $148$ \\
\dottedrule
 & 7 & ``A'' & $-0.38$ & $-0.20$ & $-0.13$ & $334$ & $557$ & $133$ \\
\dottedrule
 & 8 & ``A'' & $0.00$ & $-0.15$ & $-0.18$ & $333$ & $652$ & $39$ \\
\dottedrule
 & 9 & ``B'' & $0.00$ & $0.19$ & $0.17$ & $1$ & $817$ & $206$ \\
\dottedrule
 & 10 & ``B'' & $0.62$ & $0.79$ & $0.70$ & $0$ & $66$ & $958$ \\
\dottedrule
 & 13 & ``B'' & $0.00$ & $0.27$ & $0.30$ & $0$ & $540$ & $484$ \\
\dottedrule
 & 14 & ``B'' & $0.00$ & $0.15$ & $0.17$ & $0$ & $844$ & $180$ \\
\addlinespace
Muse Glimmer & 3 & ``A'' & $-0.31$ & $-0.26$ & $-0.30$ & $909$ & $51$ & $64$ \\
\dottedrule
$1.9\times10^{-7}$ & 4 & ``A'' & $-0.06$ & $-0.12$ & $-0.16$ & $761$ & $137$ & $126$ \\
\dottedrule
 & 7 & ``A'' & $0.38$ & $0.03$ & $-0.01$ & $429$ & $248$ & $347$ \\
\dottedrule
 & 8 & ``A'' & $-0.38$ & $-0.37$ & $-0.38$ & $946$ & $41$ & $37$ \\
\dottedrule
 & 9 & ``B'' & $0.75$ & $0.21$ & $0.13$ & $213$ & $199$ & $612$ \\
\dottedrule
 & 10 & ``B'' & $0.12$ & $0.02$ & $0.01$ & $296$ & $336$ & $392$ \\
\dottedrule
 & 13 & ``B'' & $0.50$ & $0.20$ & $0.22$ & $83$ & $151$ & $790$ \\
\dottedrule
 & 14 & ``B'' & $0.94$ & $0.31$ & $0.23$ & $139$ & $127$ & $758$ \\
\bottomrule
\end{NiceTabular}
\end{table}
\begin{table}[htbp]
\ContinuedFloat
\centering
\caption{The reduced framing games by framing, model, and task, continued.}\label{tab:framing-summary:1}
\footnotesize\setlength{\tabcolsep}{4pt}
\begin{NiceTabular}{@{}l;l;l;r;r;r;r;r;r@{}}
\toprule
\begin{tabular}[t]{@{}l@{}}Model\\$f_{\mathrm{res}}$\end{tabular} & Task & Cheaper & Unedited & Frame & Mean & Below 0 & At 0 & Above 0 \\
\midrule
\multicolumn{9}{@{}l}{\textit{``only'' before the cheaper price}} \\
Gemma 4 31B & 3 & ``A'' & $-0.02$ & $0.02$ & $-0.01$ & $543$ & $14$ & $467$ \\
\dottedrule
$9.5\times10^{-7}$ & 4 & ``A'' & $0.24$ & $0.07$ & $0.02$ & $461$ & $11$ & $552$ \\
\dottedrule
 & 7 & ``A'' & $-3.44$ & $-2.96$ & $-3.02$ & $1021$ & $0$ & $3$ \\
\dottedrule
 & 8 & ``A'' & $0.19$ & $0.05$ & $6.2\times10^{-3}$ & $485$ & $15$ & $524$ \\
\dottedrule
 & 9 & ``B'' & $0.25$ & $1.5\times10^{-3}$ & $0.02$ & $441$ & $14$ & $569$ \\
\dottedrule
 & 10 & ``B'' & $3.37$ & $3.33$ & $3.87$ & $15$ & $4$ & $1005$ \\
\dottedrule
 & 13 & ``B'' & $0.02$ & $0.06$ & $0.07$ & $300$ & $17$ & $707$ \\
\dottedrule
 & 14 & ``B'' & $-0.03$ & $-0.03$ & $-0.02$ & $564$ & $17$ & $443$ \\
\addlinespace
Gemma 4 12B & 3 & ``A'' & $-0.06$ & $-0.23$ & $-0.27$ & $840$ & $5$ & $179$ \\
\dottedrule
$7.2\times10^{-7}$ & 4 & ``A'' & $0.91$ & $0.41$ & $0.35$ & $141$ & $21$ & $862$ \\
\dottedrule
 & 7 & ``A'' & $-1.00$ & $-2.74$ & $-3.04$ & $994$ & $7$ & $23$ \\
\dottedrule
 & 8 & ``A'' & $-0.25$ & $-0.37$ & $-0.37$ & $866$ & $11$ & $147$ \\
\dottedrule
 & 9 & ``B'' & $-0.53$ & $-0.12$ & $-0.04$ & $512$ & $15$ & $497$ \\
\dottedrule
 & 10 & ``B'' & $3.38$ & $1.87$ & $1.93$ & $6$ & $9$ & $1009$ \\
\dottedrule
 & 13 & ``B'' & $-0.38$ & $-0.73$ & $-0.75$ & $939$ & $6$ & $79$ \\
\dottedrule
 & 14 & ``B'' & $-0.48$ & $-0.20$ & $-0.14$ & $610$ & $10$ & $404$ \\
\bottomrule
\end{NiceTabular}
\end{table}
\begin{table}[htbp]
\ContinuedFloat
\centering
\caption{The reduced framing games by framing, model, and task, continued.}\label{tab:framing-summary:2}
\footnotesize\setlength{\tabcolsep}{4pt}
\begin{NiceTabular}{@{}l;l;l;r;r;r;r;r;r@{}}
\toprule
\begin{tabular}[t]{@{}l@{}}Model\\$f_{\mathrm{res}}$\end{tabular} & Task & Cheaper & Unedited & Frame & Mean & Below 0 & At 0 & Above 0 \\
\midrule
\multicolumn{9}{@{}l}{\textit{``as much as'' before the pricier price}} \\
Qwen 3.8 27B & 3 & ``A'' & $0.62$ & $0.29$ & $0.24$ & $86$ & $509$ & $429$ \\
\dottedrule
$0.31$ & 4 & ``A'' & $1.00$ & $0.36$ & $0.27$ & $197$ & $360$ & $467$ \\
\dottedrule
 & 7 & ``A'' & $0.75$ & $1.03$ & $1.18$ & $0$ & $36$ & $988$ \\
\dottedrule
 & 8 & ``A'' & $0.00$ & $-0.04$ & $-0.11$ & $358$ & $427$ & $239$ \\
\dottedrule
 & 9 & ``B'' & $1.00$ & $0.50$ & $0.46$ & $1$ & $355$ & $668$ \\
\dottedrule
 & 10 & ``B'' & $-1.12$ & $-1.30$ & $-1.37$ & $967$ & $52$ & $5$ \\
\dottedrule
 & 13 & ``B'' & $1.00$ & $0.43$ & $0.33$ & $6$ & $485$ & $533$ \\
\dottedrule
 & 14 & ``B'' & $0.75$ & $0.38$ & $0.38$ & $3$ & $452$ & $569$ \\
\addlinespace
Muse Glimmer & 3 & ``A'' & $-0.69$ & $-0.23$ & $-0.29$ & $796$ & $58$ & $170$ \\
\dottedrule
$1.9\times10^{-7}$ & 4 & ``A'' & $0.19$ & $0.19$ & $0.11$ & $298$ & $117$ & $609$ \\
\dottedrule
 & 7 & ``A'' & $-0.63$ & $-0.87$ & $-0.89$ & $1024$ & $0$ & $0$ \\
\dottedrule
 & 8 & ``A'' & $-0.37$ & $-0.15$ & $-0.17$ & $669$ & $96$ & $259$ \\
\dottedrule
 & 9 & ``B'' & $0.00$ & $-0.07$ & $-0.09$ & $562$ & $174$ & $288$ \\
\dottedrule
 & 10 & ``B'' & $-0.13$ & $-7.2\times10^{-4}$ & $0.04$ & $261$ & $344$ & $419$ \\
\dottedrule
 & 13 & ``B'' & $0.37$ & $0.11$ & $0.13$ & $176$ & $184$ & $664$ \\
\dottedrule
 & 14 & ``B'' & $0.00$ & $-0.10$ & $-0.12$ & $613$ & $166$ & $245$ \\
\bottomrule
\end{NiceTabular}
\end{table}
\begin{table}[htbp]
\ContinuedFloat
\centering
\caption{The reduced framing games by framing, model, and task, continued.}\label{tab:framing-summary:3}
\footnotesize\setlength{\tabcolsep}{4pt}
\begin{NiceTabular}{@{}l;l;l;r;r;r;r;r;r@{}}
\toprule
\begin{tabular}[t]{@{}l@{}}Model\\$f_{\mathrm{res}}$\end{tabular} & Task & Cheaper & Unedited & Frame & Mean & Below 0 & At 0 & Above 0 \\
\midrule
\multicolumn{9}{@{}l}{\textit{``as much as'' before the pricier price}} \\
Gemma 4 31B & 3 & ``A'' & $-1.04$ & $-0.49$ & $-0.45$ & $979$ & $2$ & $43$ \\
\dottedrule
$9.5\times10^{-7}$ & 4 & ``A'' & $-0.11$ & $-0.30$ & $-0.33$ & $939$ & $5$ & $80$ \\
\dottedrule
 & 7 & ``A'' & $-2.88$ & $-3.49$ & $-3.47$ & $991$ & $2$ & $31$ \\
\dottedrule
 & 8 & ``A'' & $-1.03$ & $-0.53$ & $-0.47$ & $981$ & $1$ & $42$ \\
\dottedrule
 & 9 & ``B'' & $0.38$ & $-0.10$ & $-0.14$ & $662$ & $8$ & $354$ \\
\dottedrule
 & 10 & ``B'' & $12.62$ & $3.99$ & $2.62$ & $243$ & $6$ & $775$ \\
\dottedrule
 & 13 & ``B'' & $-0.26$ & $-0.38$ & $-0.37$ & $919$ & $5$ & $100$ \\
\dottedrule
 & 14 & ``B'' & $0.18$ & $-0.17$ & $-0.19$ & $736$ & $5$ & $283$ \\
\addlinespace
Gemma 4 12B & 3 & ``A'' & $0.56$ & $0.09$ & $1.5\times10^{-3}$ & $516$ & $17$ & $491$ \\
\dottedrule
$7.2\times10^{-7}$ & 4 & ``A'' & $1.37$ & $0.75$ & $0.65$ & $22$ & $6$ & $996$ \\
\dottedrule
 & 7 & ``A'' & $-1.75$ & $-3.20$ & $-3.40$ & $1001$ & $2$ & $21$ \\
\dottedrule
 & 8 & ``A'' & $0.31$ & $-0.02$ & $-0.06$ & $590$ & $17$ & $417$ \\
\dottedrule
 & 9 & ``B'' & $-1.34$ & $-0.47$ & $-0.42$ & $814$ & $1$ & $209$ \\
\dottedrule
 & 10 & ``B'' & $-2.62$ & $-1.90$ & $-1.83$ & $783$ & $25$ & $216$ \\
\dottedrule
 & 13 & ``B'' & $-2.38$ & $-1.63$ & $-1.77$ & $1023$ & $0$ & $1$ \\
\dottedrule
 & 14 & ``B'' & $-0.89$ & $-0.24$ & $-0.29$ & $679$ & $6$ & $339$ \\
\bottomrule
\end{NiceTabular}
\end{table}
\endgroup{}
\floatnotes{framing_summary}

\begingroup\long\def\floatNotesBody{
Boxes show medians and interquartile ranges over retained edit contexts, signed toward the cheaper option; whiskers reach extremes. Diamonds mark the unedited context. Axes are linear within $\pm0.1$, logarithmic beyond. Tan bands show measurement floors.
}%
\begin{figure}[tbp]
\centering
\includegraphics[width=\textwidth]{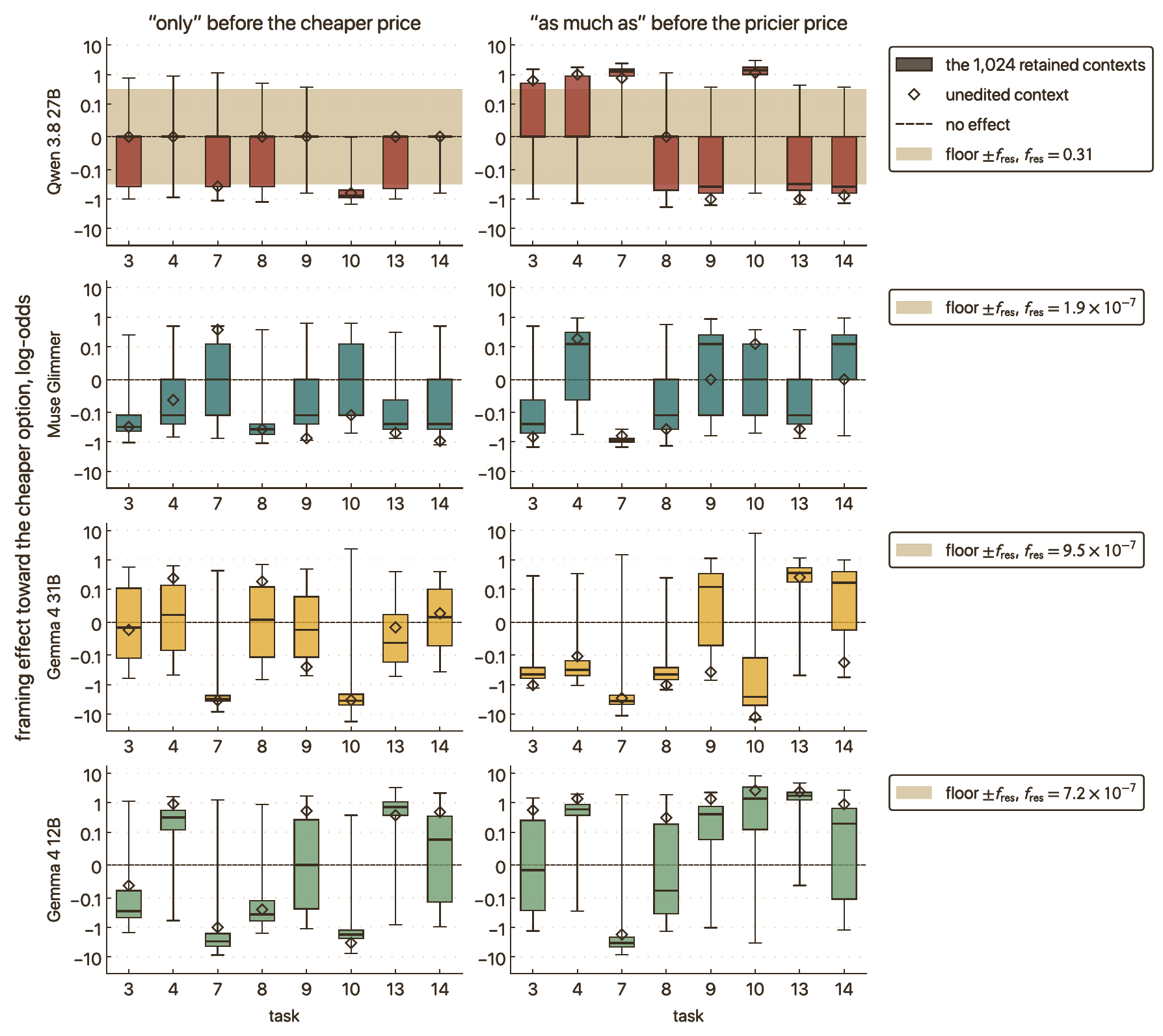}
\caption{Paired framing effects toward the cheaper option over the retained contexts, by model, framing, and task.}\label{fig:e4-framing}
\end{figure}
\endgroup{}
\figurenotes{e4_framing}

The framed word games use a replacement reference, unlike the framing edit game, whose reference is the intact prompt without the phrase. Table~\appref{tab:framed-words} gives each phrase's allocation and standard error together with the attribute share and its interval.

\begingroup\long\def\floatNotesBody{
Phrase values are log-odds toward ``A''; se is Monte Carlo standard error. Attribute share includes the framing phrase in total absolute mass. Brackets are 95\% joint-row bootstrap intervals. These word games use the all-masked reference.
}%
\begin{table}[htbp]
\centering
\caption{Phrase allocations and attribute shares in every framed word game.}\label{tab:framed-words}
\footnotesize\setlength{\tabcolsep}{4pt}
\begin{NiceTabular}{@{}l;r;r;r@{}}
\toprule
Task & Phrase value & se & Attribute share (\%) \\
\midrule
\multicolumn{4}{@{}l}{\textit{Qwen 3.8 27B, ``only'' before the cheaper price}} \\
flight task 3 & $0.07$ & $0.02$ & \begin{tabular}[t]{@{}c@{}}$24.8$\\$[22.8, 26.2]$\end{tabular} \\
\dottedrule
flight task 4 & $0.12$ & $0.02$ & \begin{tabular}[t]{@{}c@{}}$21.5$\\$[19.8, 23.0]$\end{tabular} \\
\dottedrule
flight task 7 & $-0.04$ & $0.02$ & \begin{tabular}[t]{@{}c@{}}$15.6$\\$[14.2, 16.8]$\end{tabular} \\
\dottedrule
flight task 8 & $0.10$ & $0.02$ & \begin{tabular}[t]{@{}c@{}}$21.4$\\$[19.6, 23.2]$\end{tabular} \\
\dottedrule
flight task 9 & $-0.22$ & $0.02$ & \begin{tabular}[t]{@{}c@{}}$15.8$\\$[14.5, 17.0]$\end{tabular} \\
\dottedrule
flight task 10 & $-0.12$ & $0.02$ & \begin{tabular}[t]{@{}c@{}}$16.9$\\$[15.3, 18.2]$\end{tabular} \\
\dottedrule
flight task 13 & $-0.15$ & $0.02$ & \begin{tabular}[t]{@{}c@{}}$15.5$\\$[14.3, 16.5]$\end{tabular} \\
\dottedrule
flight task 14 & $-0.24$ & $0.02$ & \begin{tabular}[t]{@{}c@{}}$16.4$\\$[15.0, 17.7]$\end{tabular} \\
\bottomrule
\end{NiceTabular}
\end{table}
\begin{table}[htbp]
\ContinuedFloat
\centering
\caption{Framed word games, continued.}\label{tab:framed-words:1}
\footnotesize\setlength{\tabcolsep}{4pt}
\begin{NiceTabular}{@{}l;r;r;r@{}}
\toprule
Task & Phrase value & se & Attribute share (\%) \\
\midrule
\multicolumn{4}{@{}l}{\textit{Qwen 3.8 27B, ``as much as'' before the pricier price}} \\
flight task 3 & $0.16$ & $0.02$ & \begin{tabular}[t]{@{}c@{}}$25.2$\\$[23.0, 26.7]$\end{tabular} \\
\dottedrule
flight task 4 & $0.19$ & $0.02$ & \begin{tabular}[t]{@{}c@{}}$23.6$\\$[21.7, 25.3]$\end{tabular} \\
\dottedrule
flight task 7 & $0.10$ & $0.02$ & \begin{tabular}[t]{@{}c@{}}$18.3$\\$[16.8, 19.7]$\end{tabular} \\
\dottedrule
flight task 8 & $0.20$ & $0.02$ & \begin{tabular}[t]{@{}c@{}}$26.0$\\$[23.9, 27.6]$\end{tabular} \\
\dottedrule
flight task 9 & $-0.08$ & $0.02$ & \begin{tabular}[t]{@{}c@{}}$17.1$\\$[15.7, 18.7]$\end{tabular} \\
\dottedrule
flight task 10 & $-0.10$ & $0.02$ & \begin{tabular}[t]{@{}c@{}}$18.2$\\$[16.6, 19.5]$\end{tabular} \\
\dottedrule
flight task 13 & $2.2\times10^{-3}$ & $0.02$ & \begin{tabular}[t]{@{}c@{}}$14.6$\\$[13.6, 15.6]$\end{tabular} \\
\dottedrule
flight task 14 & $-0.10$ & $0.02$ & \begin{tabular}[t]{@{}c@{}}$18.2$\\$[16.9, 19.5]$\end{tabular} \\
\bottomrule
\end{NiceTabular}
\end{table}
\begin{table}[htbp]
\ContinuedFloat
\centering
\caption{Framed word games, continued.}\label{tab:framed-words:2}
\footnotesize\setlength{\tabcolsep}{4pt}
\begin{NiceTabular}{@{}l;r;r;r@{}}
\toprule
Task & Phrase value & se & Attribute share (\%) \\
\midrule
\multicolumn{4}{@{}l}{\textit{Muse Glimmer, ``only'' before the cheaper price}} \\
flight task 3 & $0.03$ & $0.01$ & \begin{tabular}[t]{@{}c@{}}$5.7$\\$[4.8, 6.8]$\end{tabular} \\
\dottedrule
flight task 4 & $0.02$ & $0.01$ & \begin{tabular}[t]{@{}c@{}}$5.8$\\$[4.7, 6.9]$\end{tabular} \\
\dottedrule
flight task 7 & $0.07$ & $0.01$ & \begin{tabular}[t]{@{}c@{}}$9.7$\\$[8.4, 10.7]$\end{tabular} \\
\dottedrule
flight task 8 & $0.03$ & $0.01$ & \begin{tabular}[t]{@{}c@{}}$6.8$\\$[5.8, 7.8]$\end{tabular} \\
\dottedrule
flight task 9 & $-0.07$ & $0.02$ & \begin{tabular}[t]{@{}c@{}}$8.4$\\$[7.4, 9.3]$\end{tabular} \\
\dottedrule
flight task 10 & $0.01$ & $0.01$ & \begin{tabular}[t]{@{}c@{}}$6.9$\\$[5.8, 8.0]$\end{tabular} \\
\dottedrule
flight task 13 & $4.1\times10^{-3}$ & $0.01$ & \begin{tabular}[t]{@{}c@{}}$9.6$\\$[8.4, 10.6]$\end{tabular} \\
\dottedrule
flight task 14 & $-0.11$ & $0.02$ & \begin{tabular}[t]{@{}c@{}}$8.9$\\$[7.9, 9.8]$\end{tabular} \\
\bottomrule
\end{NiceTabular}
\end{table}
\begin{table}[htbp]
\ContinuedFloat
\centering
\caption{Framed word games, continued.}\label{tab:framed-words:3}
\footnotesize\setlength{\tabcolsep}{4pt}
\begin{NiceTabular}{@{}l;r;r;r@{}}
\toprule
Task & Phrase value & se & Attribute share (\%) \\
\midrule
\multicolumn{4}{@{}l}{\textit{Muse Glimmer, ``as much as'' before the pricier price}} \\
flight task 3 & $-5.7\times10^{-3}$ & $0.01$ & \begin{tabular}[t]{@{}c@{}}$6.7$\\$[5.8, 8.0]$\end{tabular} \\
\dottedrule
flight task 4 & $-0.01$ & $0.01$ & \begin{tabular}[t]{@{}c@{}}$7.0$\\$[5.9, 8.1]$\end{tabular} \\
\dottedrule
flight task 7 & $-0.02$ & $0.02$ & \begin{tabular}[t]{@{}c@{}}$10.9$\\$[9.5, 12.2]$\end{tabular} \\
\dottedrule
flight task 8 & $-0.02$ & $0.02$ & \begin{tabular}[t]{@{}c@{}}$6.7$\\$[5.7, 7.9]$\end{tabular} \\
\dottedrule
flight task 9 & $-0.14$ & $0.02$ & \begin{tabular}[t]{@{}c@{}}$12.3$\\$[11.1, 13.4]$\end{tabular} \\
\dottedrule
flight task 10 & $-0.09$ & $0.02$ & \begin{tabular}[t]{@{}c@{}}$9.6$\\$[8.4, 10.6]$\end{tabular} \\
\dottedrule
flight task 13 & $-0.07$ & $0.02$ & \begin{tabular}[t]{@{}c@{}}$12.4$\\$[11.3, 13.4]$\end{tabular} \\
\dottedrule
flight task 14 & $-0.11$ & $0.02$ & \begin{tabular}[t]{@{}c@{}}$11.8$\\$[10.6, 12.8]$\end{tabular} \\
\bottomrule
\end{NiceTabular}
\end{table}
\begin{table}[htbp]
\ContinuedFloat
\centering
\caption{Framed word games, continued.}\label{tab:framed-words:4}
\footnotesize\setlength{\tabcolsep}{4pt}
\begin{NiceTabular}{@{}l;r;r;r@{}}
\toprule
Task & Phrase value & se & Attribute share (\%) \\
\midrule
\multicolumn{4}{@{}l}{\textit{Gemma 4 31B, ``only'' before the cheaper price}} \\
flight task 3 & $0.16$ & $0.10$ & \begin{tabular}[t]{@{}c@{}}$31.0$\\$[28.7, 33.1]$\end{tabular} \\
\dottedrule
flight task 4 & $-0.07$ & $0.08$ & \begin{tabular}[t]{@{}c@{}}$22.7$\\$[20.7, 24.5]$\end{tabular} \\
\dottedrule
flight task 7 & $0.09$ & $0.09$ & \begin{tabular}[t]{@{}c@{}}$35.8$\\$[33.2, 38.3]$\end{tabular} \\
\dottedrule
flight task 8 & $-0.02$ & $0.10$ & \begin{tabular}[t]{@{}c@{}}$32.9$\\$[30.3, 34.9]$\end{tabular} \\
\dottedrule
flight task 9 & $-0.14$ & $0.10$ & \begin{tabular}[t]{@{}c@{}}$28.5$\\$[26.6, 30.1]$\end{tabular} \\
\dottedrule
flight task 10 & $-0.80$ & $0.11$ & \begin{tabular}[t]{@{}c@{}}$36.1$\\$[33.5, 38.2]$\end{tabular} \\
\dottedrule
flight task 13 & $0.17$ & $0.08$ & \begin{tabular}[t]{@{}c@{}}$25.1$\\$[23.2, 26.4]$\end{tabular} \\
\dottedrule
flight task 14 & $-6.0\times10^{-3}$ & $0.10$ & \begin{tabular}[t]{@{}c@{}}$28.8$\\$[26.8, 30.2]$\end{tabular} \\
\bottomrule
\end{NiceTabular}
\end{table}
\begin{table}[htbp]
\ContinuedFloat
\centering
\caption{Framed word games, continued.}\label{tab:framed-words:5}
\footnotesize\setlength{\tabcolsep}{4pt}
\begin{NiceTabular}{@{}l;r;r;r@{}}
\toprule
Task & Phrase value & se & Attribute share (\%) \\
\midrule
\multicolumn{4}{@{}l}{\textit{Gemma 4 31B, ``as much as'' before the pricier price}} \\
flight task 3 & $0.11$ & $0.08$ & \begin{tabular}[t]{@{}c@{}}$36.1$\\$[33.2, 38.5]$\end{tabular} \\
\dottedrule
flight task 4 & $0.24$ & $0.08$ & \begin{tabular}[t]{@{}c@{}}$23.4$\\$[21.4, 25.1]$\end{tabular} \\
\dottedrule
flight task 7 & $0.32$ & $0.08$ & \begin{tabular}[t]{@{}c@{}}$41.7$\\$[38.3, 43.8]$\end{tabular} \\
\dottedrule
flight task 8 & $0.20$ & $0.08$ & \begin{tabular}[t]{@{}c@{}}$34.7$\\$[32.1, 36.9]$\end{tabular} \\
\dottedrule
flight task 9 & $0.39$ & $0.09$ & \begin{tabular}[t]{@{}c@{}}$33.9$\\$[31.7, 35.6]$\end{tabular} \\
\dottedrule
flight task 10 & $1.15$ & $0.14$ & \begin{tabular}[t]{@{}c@{}}$39.8$\\$[37.1, 41.9]$\end{tabular} \\
\dottedrule
flight task 13 & $0.60$ & $0.08$ & \begin{tabular}[t]{@{}c@{}}$25.9$\\$[24.2, 27.5]$\end{tabular} \\
\dottedrule
flight task 14 & $0.23$ & $0.10$ & \begin{tabular}[t]{@{}c@{}}$32.2$\\$[30.2, 33.9]$\end{tabular} \\
\bottomrule
\end{NiceTabular}
\end{table}
\begin{table}[htbp]
\ContinuedFloat
\centering
\caption{Framed word games, continued.}\label{tab:framed-words:6}
\footnotesize\setlength{\tabcolsep}{4pt}
\begin{NiceTabular}{@{}l;r;r;r@{}}
\toprule
Task & Phrase value & se & Attribute share (\%) \\
\midrule
\multicolumn{4}{@{}l}{\textit{Gemma 4 12B, ``only'' before the cheaper price}} \\
flight task 3 & $0.29$ & $0.04$ & \begin{tabular}[t]{@{}c@{}}$24.1$\\$[22.6, 25.8]$\end{tabular} \\
\dottedrule
flight task 4 & $0.37$ & $0.04$ & \begin{tabular}[t]{@{}c@{}}$19.6$\\$[18.2, 21.2]$\end{tabular} \\
\dottedrule
flight task 7 & $0.31$ & $0.04$ & \begin{tabular}[t]{@{}c@{}}$18.4$\\$[16.4, 20.0]$\end{tabular} \\
\dottedrule
flight task 8 & $0.31$ & $0.04$ & \begin{tabular}[t]{@{}c@{}}$25.0$\\$[23.2, 26.8]$\end{tabular} \\
\dottedrule
flight task 9 & $0.04$ & $0.04$ & \begin{tabular}[t]{@{}c@{}}$25.2$\\$[23.5, 27.0]$\end{tabular} \\
\dottedrule
flight task 10 & $0.16$ & $0.04$ & \begin{tabular}[t]{@{}c@{}}$23.5$\\$[21.0, 25.2]$\end{tabular} \\
\dottedrule
flight task 13 & $0.07$ & $0.04$ & \begin{tabular}[t]{@{}c@{}}$23.3$\\$[21.7, 24.8]$\end{tabular} \\
\dottedrule
flight task 14 & $0.08$ & $0.04$ & \begin{tabular}[t]{@{}c@{}}$26.0$\\$[24.0, 27.7]$\end{tabular} \\
\bottomrule
\end{NiceTabular}
\end{table}
\begin{table}[htbp]
\ContinuedFloat
\centering
\caption{Framed word games, continued.}\label{tab:framed-words:7}
\footnotesize\setlength{\tabcolsep}{4pt}
\begin{NiceTabular}{@{}l;r;r;r@{}}
\toprule
Task & Phrase value & se & Attribute share (\%) \\
\midrule
\multicolumn{4}{@{}l}{\textit{Gemma 4 12B, ``as much as'' before the pricier price}} \\
flight task 3 & $0.44$ & $0.04$ & \begin{tabular}[t]{@{}c@{}}$25.7$\\$[23.8, 27.5]$\end{tabular} \\
\dottedrule
flight task 4 & $0.45$ & $0.04$ & \begin{tabular}[t]{@{}c@{}}$18.7$\\$[17.3, 20.1]$\end{tabular} \\
\dottedrule
flight task 7 & $0.47$ & $0.04$ & \begin{tabular}[t]{@{}c@{}}$23.4$\\$[21.4, 25.3]$\end{tabular} \\
\dottedrule
flight task 8 & $0.47$ & $0.04$ & \begin{tabular}[t]{@{}c@{}}$26.5$\\$[24.6, 28.5]$\end{tabular} \\
\dottedrule
flight task 9 & $0.21$ & $0.04$ & \begin{tabular}[t]{@{}c@{}}$28.0$\\$[26.0, 29.6]$\end{tabular} \\
\dottedrule
flight task 10 & $0.22$ & $0.05$ & \begin{tabular}[t]{@{}c@{}}$28.5$\\$[26.2, 30.6]$\end{tabular} \\
\dottedrule
flight task 13 & $0.22$ & $0.04$ & \begin{tabular}[t]{@{}c@{}}$22.7$\\$[21.1, 24.1]$\end{tabular} \\
\dottedrule
flight task 14 & $0.18$ & $0.04$ & \begin{tabular}[t]{@{}c@{}}$28.5$\\$[26.6, 30.2]$\end{tabular} \\
\bottomrule
\end{NiceTabular}
\end{table}
\endgroup{}
\floatnotes{framed_words}

\subsection{Word profiles and choice diagnostics}\label{part:words}

The flight profiles are Figures~\ref{fig:shapley-e2-flight-qwen3-8-27b}, \ref{fig:shapley-e2-flight-muse-glimmer}, \ref{fig:shapley-e2-flight-gemma4-31b}, and~\appref{fig:shapley-e2-flight-gemma4-12b}. Hotel profiles are Figures~\ref{fig:shapley-e5-hotel-qwen3-8-27b}, \ref{fig:shapley-e5-hotel-muse-glimmer}, \ref{fig:shapley-e5-hotel-gemma4-31b}, and~\appref{fig:shapley-e5-hotel-gemma4-12b}; smartphone profiles are Figures~\ref{fig:shapley-e5-smartphone-qwen3-8-27b}, \ref{fig:shapley-e5-smartphone-muse-glimmer}, \ref{fig:shapley-e5-smartphone-gemma4-31b}, and~\appref{fig:shapley-e5-smartphone-gemma4-12b}. Read each strip against its task's attributes, using the common replacement rule.

Table~\appref{tab:agreement} compares aligned word profiles across models. Its cosine uses the absolute vectors $u_i=|\phi_i|$, namely $u^\top u'/(\|u\|_2\|u'\|_2)$; Spearman correlation ranks signed values. A zero norm makes cosine undefined. Neither statistic identifies an architectural cause.

Figure~\appref{fig:e2-rationality} separates attribute-value allocations from complete-prompt responses. Favoring the displayed dominant option is a direct choice check; every model does so on the flight dominance tasks. Ordering an attribute player's value across its two levels instead concerns its reference-relative allocation, not monotonicity of the complete response. The unresolved counts are \tnDiagnosticsUnresolvedQwenThreeEightTwoSevenB{} of \tnDiagnosticsPairsQwenThreeEightTwoSevenB{} pairs for Qwen 3.8 27B, \tnDiagnosticsUnresolvedMuseGlimmer{} of \tnDiagnosticsPairsMuseGlimmer{} for Muse Glimmer, \tnDiagnosticsUnresolvedGemmaFourThreeOneB{} of \tnDiagnosticsPairsGemmaFourThreeOneB{} for Gemma 4 31B, and \tnDiagnosticsUnresolvedGemmaFourOneTwoB{} of \tnDiagnosticsPairsGemmaFourOneTwoB{} for Gemma 4 12B. Their point ordering does not establish a difference. Tie-task mean scores describe asymmetry with ``A'' also first. They do not isolate label from position or establish irrationality.

\begingroup\long\def\floatNotesBody{
Players follow prompt order. Shading and squares: attributes; circles: context; triangles: function words, punctuation, or line breaks. Bars: 95\% Monte Carlo intervals in log-odds. Tan bands: plus or minus the measurement floor.
}%
\begin{figure}[tbp]
\centering
\includegraphics[width=0.97\textwidth]{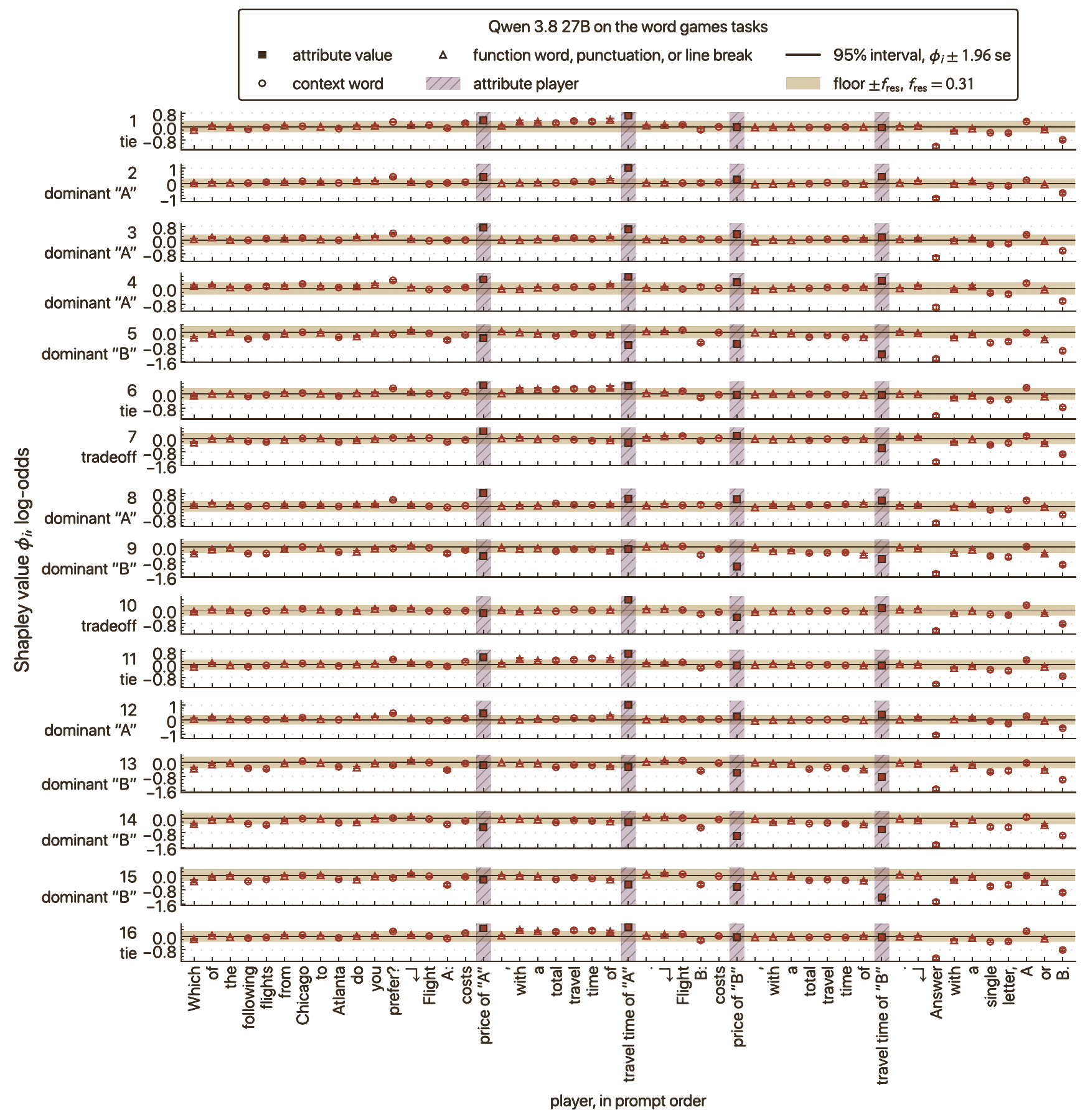}
\caption{Word-level Shapley values on every flight task for Qwen 3.8 27B (word games), in the task order of Table~\appref{app:prompts:tasks:flight}.}\label{fig:shapley-e2-flight-qwen3-8-27b}
\end{figure}
\endgroup{}
\figurenotes{shapley_e2_flight_qwen3-8-27b}

\begingroup%
\begin{figure}[tbp]
\centering
\includegraphics[width=0.97\textwidth]{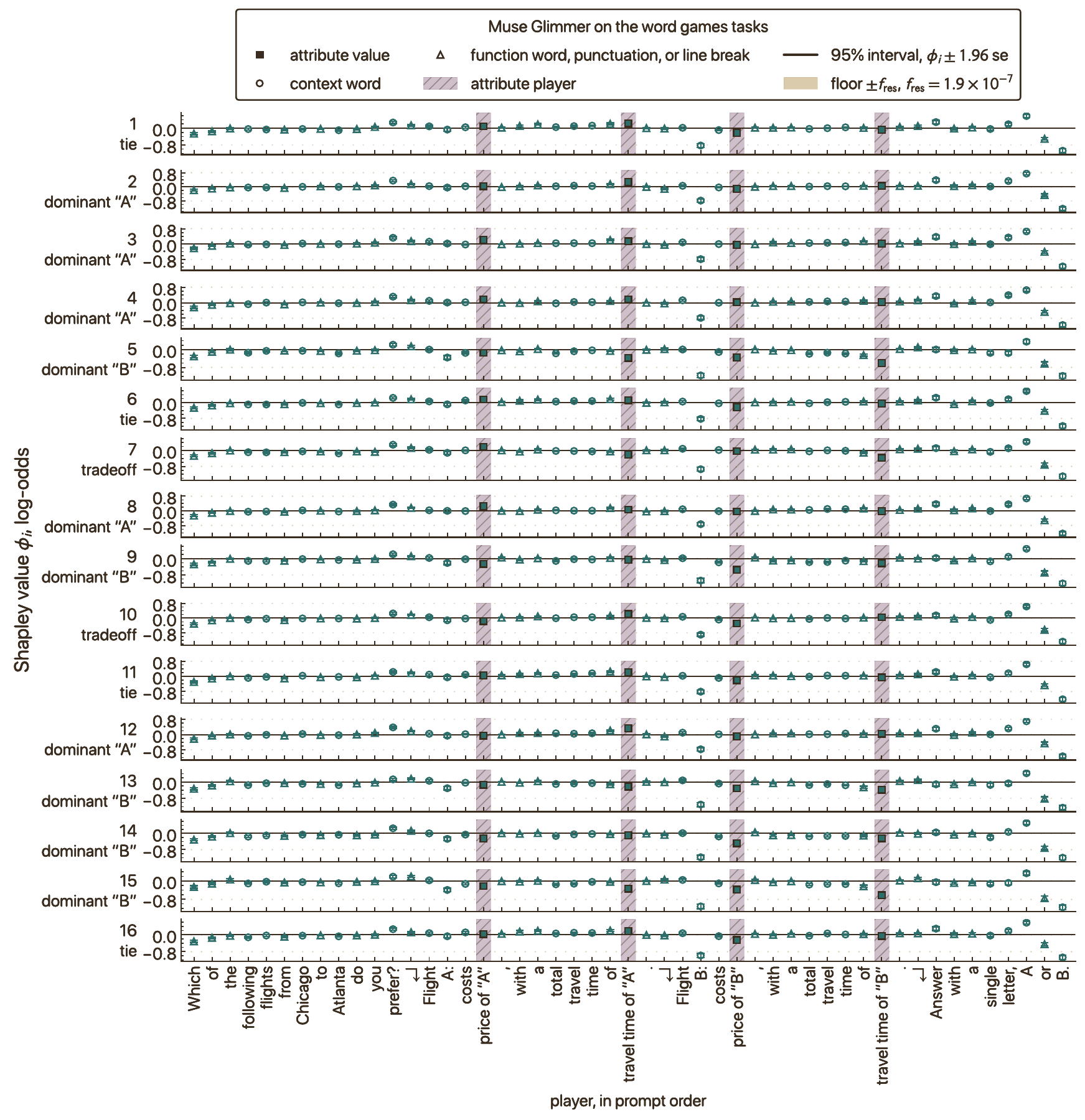}
\caption{Word-level Shapley values on every flight task for Muse Glimmer (word games), drawn as in Figure~\appref{fig:shapley-e2-flight-qwen3-8-27b}.}\label{fig:shapley-e2-flight-muse-glimmer}
\end{figure}
\endgroup{}

\begingroup%
\begin{figure}[tbp]
\centering
\includegraphics[width=0.97\textwidth]{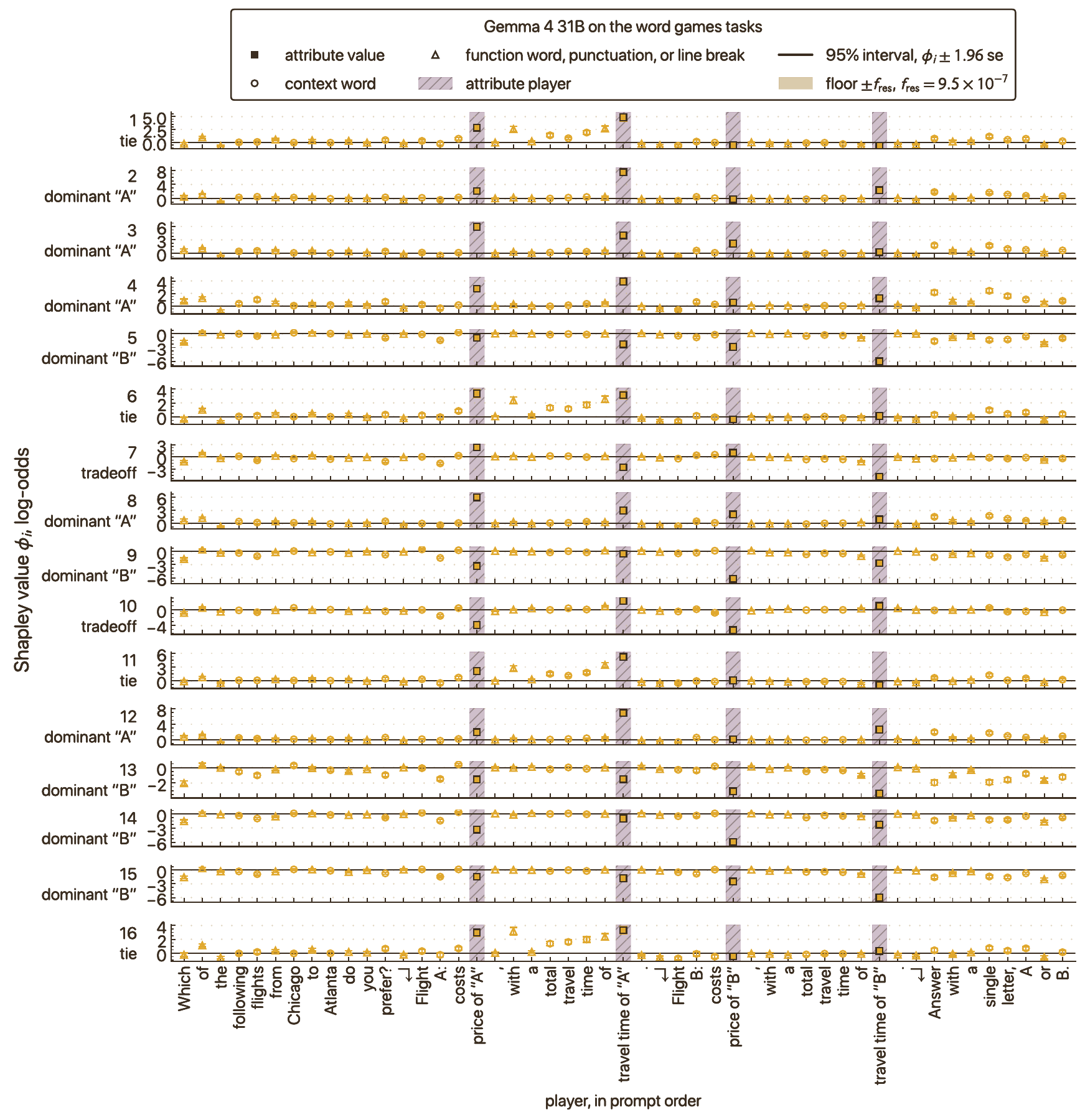}
\caption{Word-level Shapley values on every flight task for Gemma 4 31B (word games), drawn as in Figure~\appref{fig:shapley-e2-flight-qwen3-8-27b}.}\label{fig:shapley-e2-flight-gemma4-31b}
\end{figure}
\endgroup{}

\begingroup%
\begin{figure}[tbp]
\centering
\includegraphics[width=0.97\textwidth]{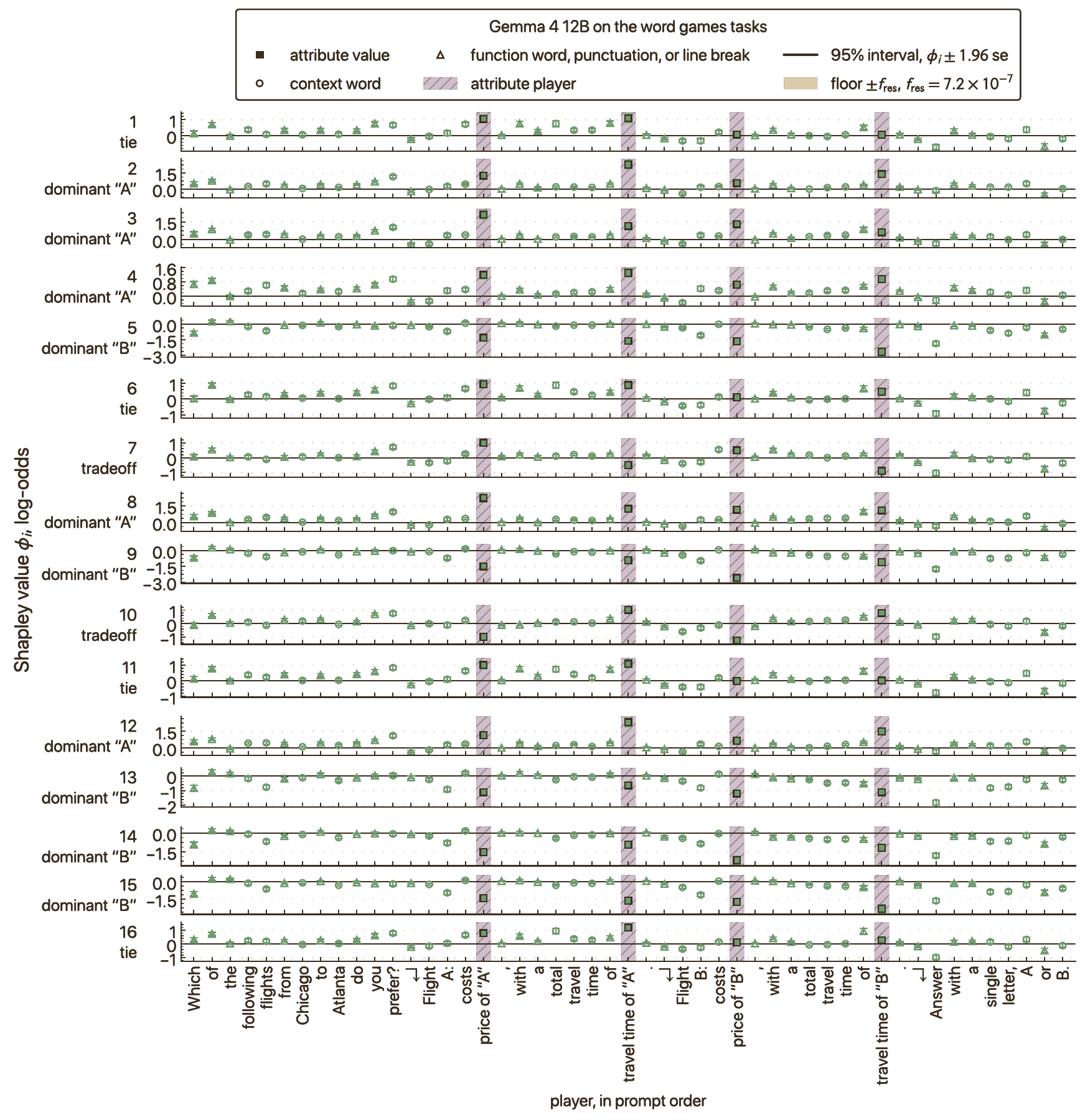}
\caption{Word-level Shapley values on every flight task for Gemma 4 12B (word games), drawn as in Figure~\appref{fig:shapley-e2-flight-qwen3-8-27b}.}\label{fig:shapley-e2-flight-gemma4-12b}
\end{figure}
\endgroup{}

\begingroup\long\def\floatNotesBody{
Players follow prompt order. Shading and squares: attributes; circles: context; triangles: function words, punctuation, or line breaks. Bars: 95\% Monte Carlo intervals in log-odds. Tan bands: plus or minus the measurement floor.
}%
\begin{figure}[tbp]
\centering
\includegraphics[width=0.97\textwidth]{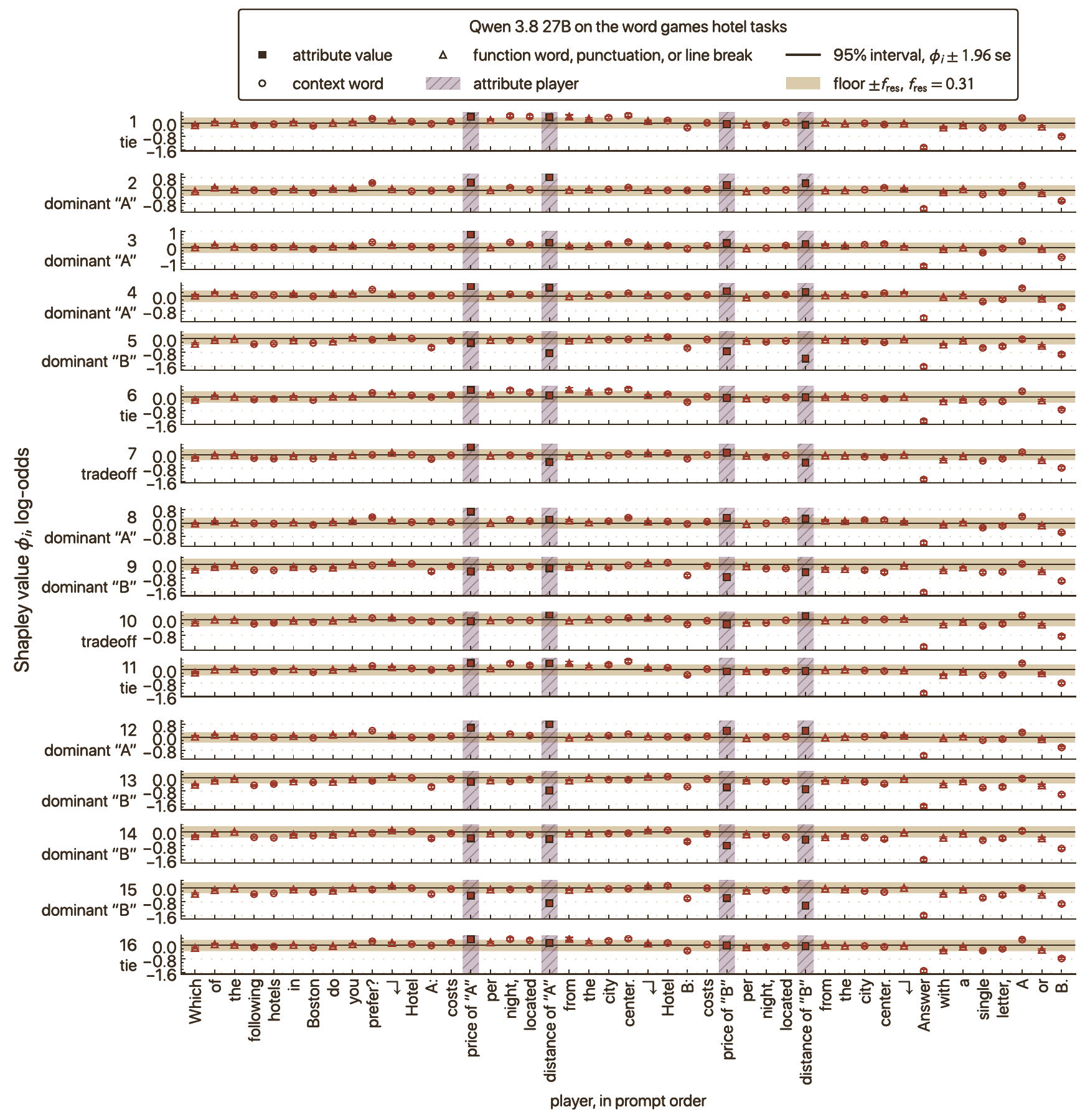}
\caption{Word-level Shapley values on every hotel task for Qwen 3.8 27B (word games), in the task order of Table~\appref{app:prompts:tasks:hotel}.}\label{fig:shapley-e5-hotel-qwen3-8-27b}
\end{figure}
\endgroup{}
\figurenotes{shapley_e5_hotel_qwen3-8-27b}

\begingroup%
\begin{figure}[tbp]
\centering
\includegraphics[width=0.97\textwidth]{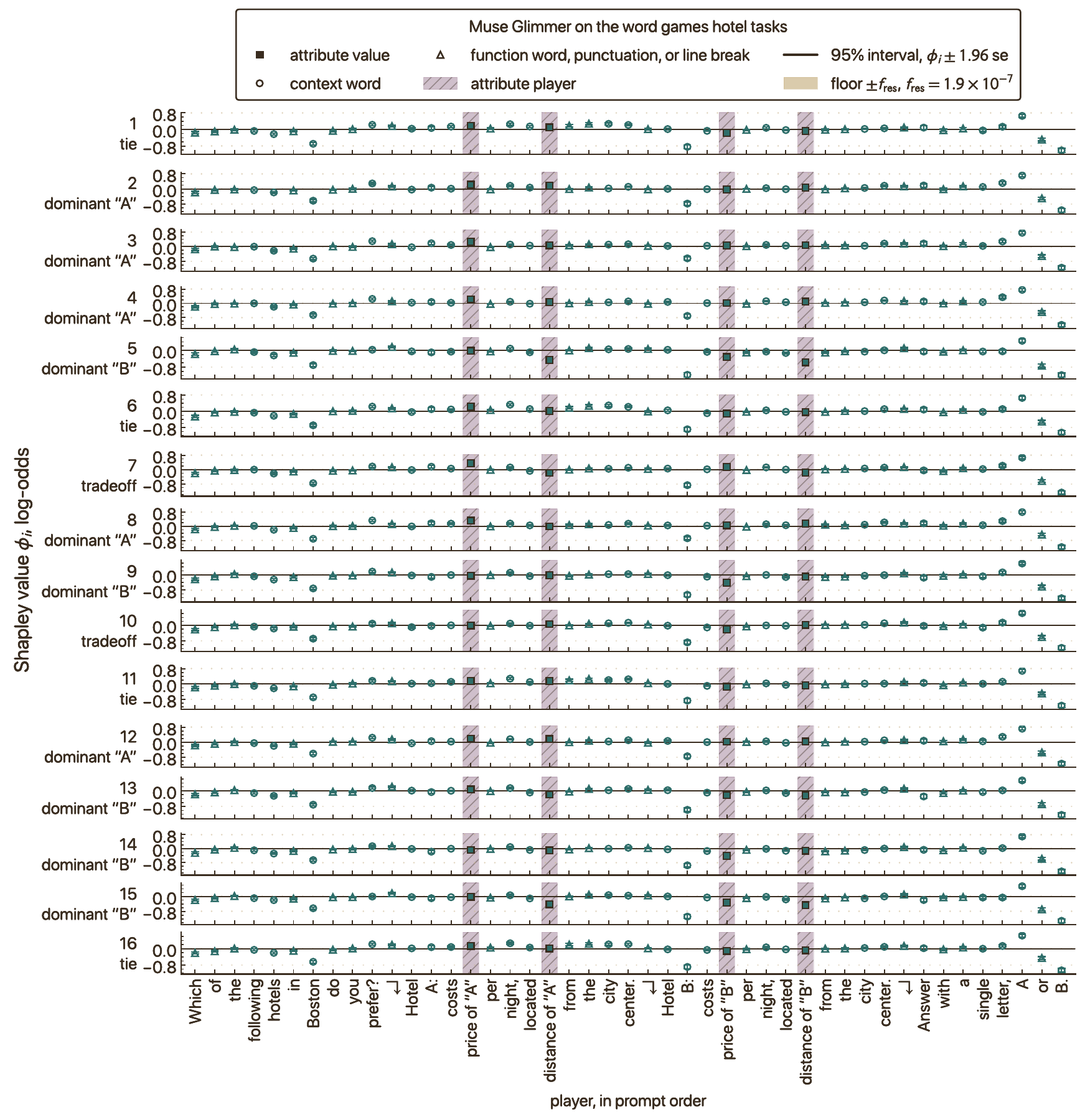}
\caption{Word-level Shapley values on every hotel task for Muse Glimmer (word games), drawn as in Figure~\appref{fig:shapley-e5-hotel-qwen3-8-27b}.}\label{fig:shapley-e5-hotel-muse-glimmer}
\end{figure}
\endgroup{}

\begingroup%
\begin{figure}[tbp]
\centering
\includegraphics[width=0.97\textwidth]{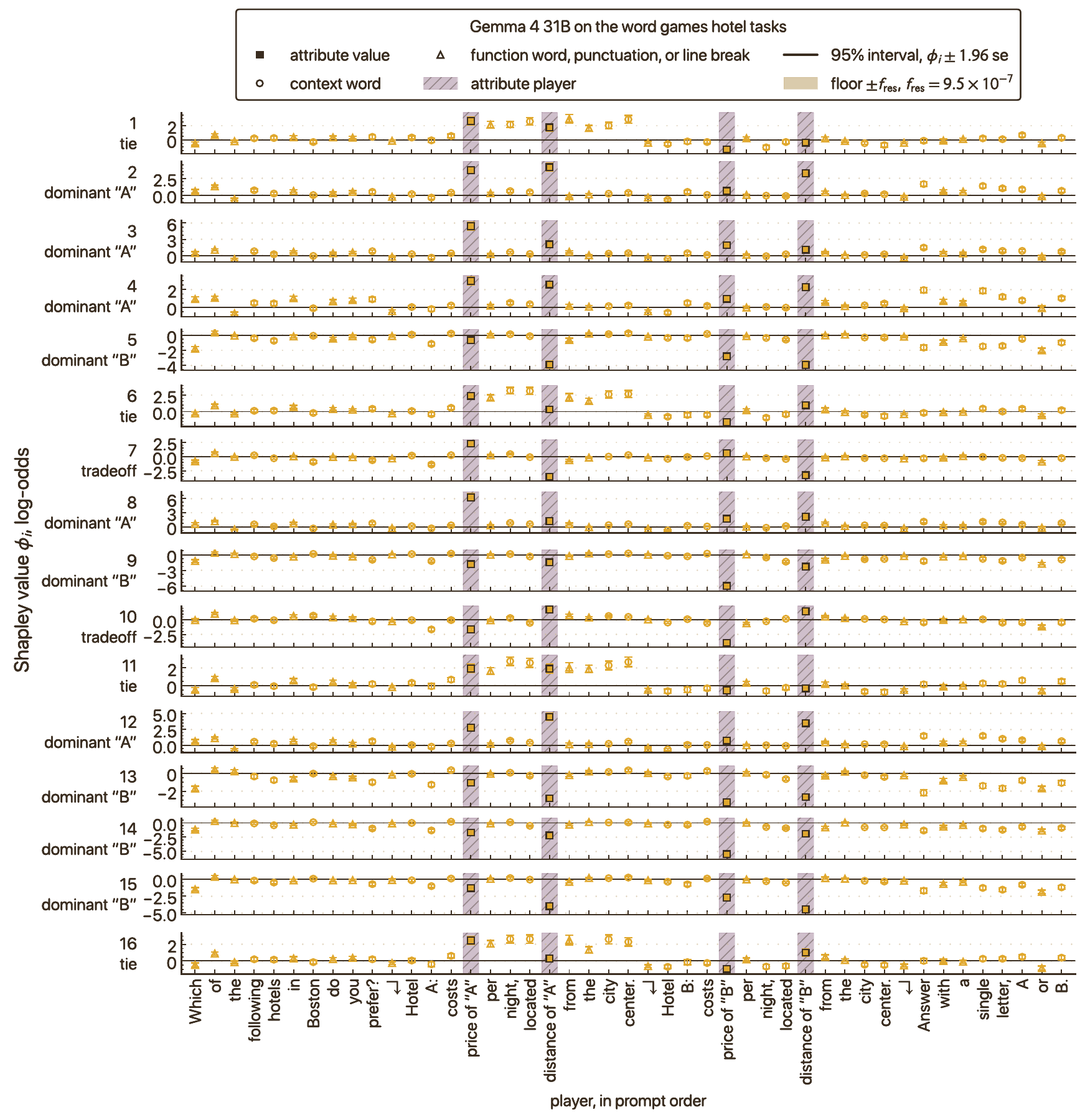}
\caption{Word-level Shapley values on every hotel task for Gemma 4 31B (word games), drawn as in Figure~\appref{fig:shapley-e5-hotel-qwen3-8-27b}.}\label{fig:shapley-e5-hotel-gemma4-31b}
\end{figure}
\endgroup{}

\begingroup%
\begin{figure}[tbp]
\centering
\includegraphics[width=0.97\textwidth]{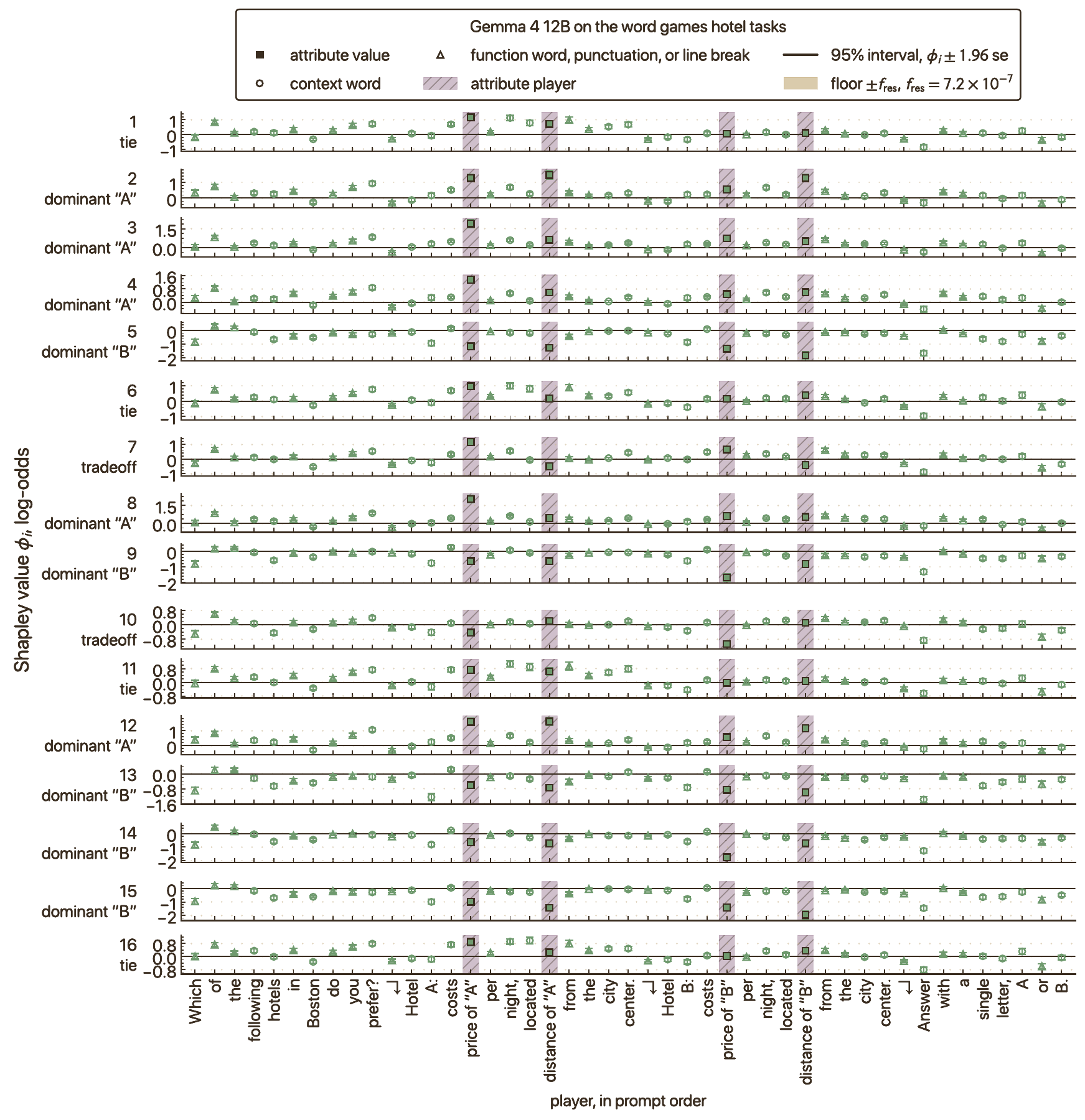}
\caption{Word-level Shapley values on every hotel task for Gemma 4 12B (word games), drawn as in Figure~\appref{fig:shapley-e5-hotel-qwen3-8-27b}.}\label{fig:shapley-e5-hotel-gemma4-12b}
\end{figure}
\endgroup{}

\begingroup%
\begin{figure}[tbp]
\centering
\includegraphics[width=0.97\textwidth]{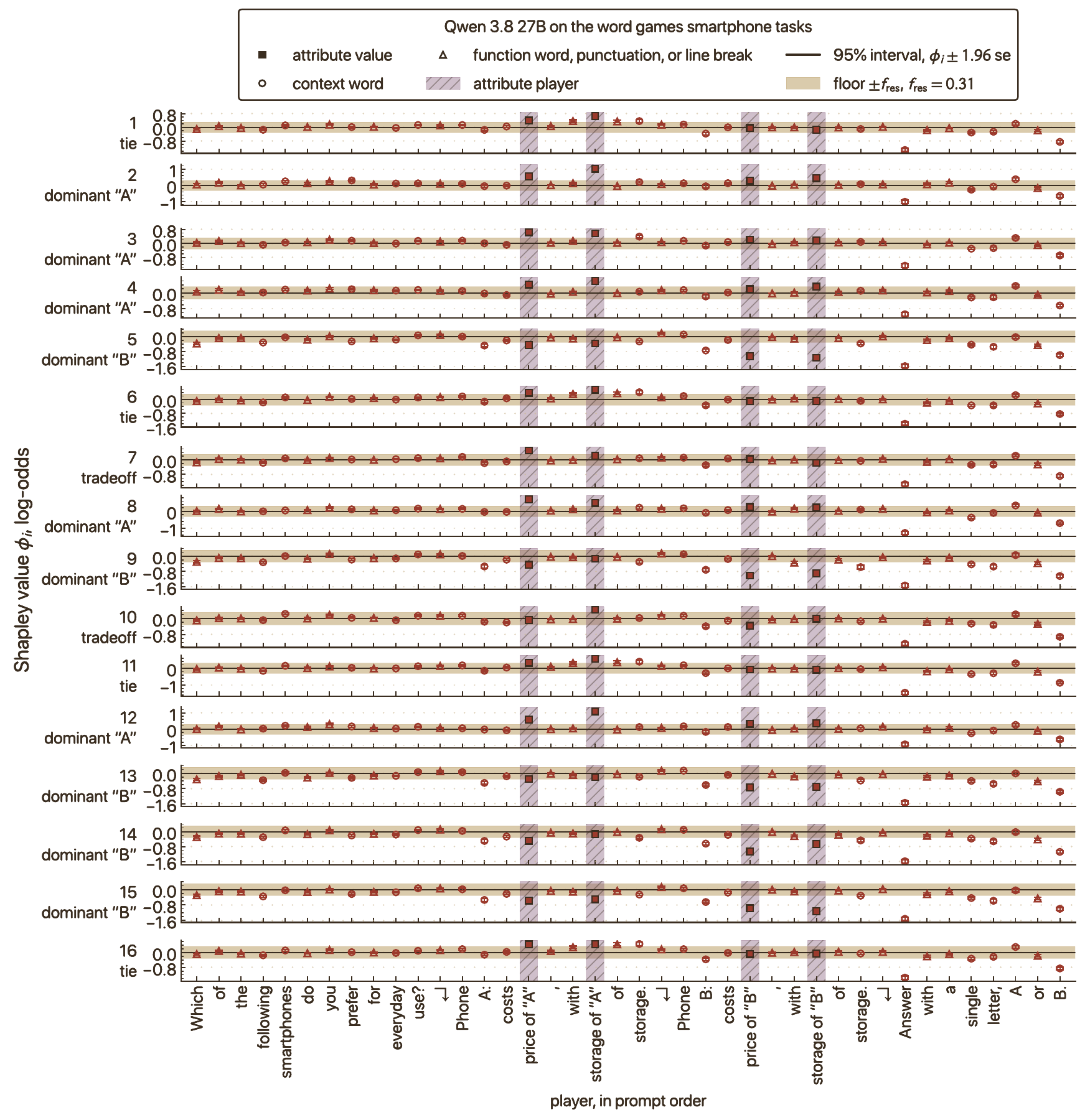}
\caption{Word-level Shapley values on every smartphone task for Qwen 3.8 27B (word games), drawn as in Figure~\appref{fig:shapley-e5-hotel-qwen3-8-27b}.}\label{fig:shapley-e5-smartphone-qwen3-8-27b}
\end{figure}
\endgroup{}

\begingroup%
\begin{figure}[tbp]
\centering
\includegraphics[width=0.97\textwidth]{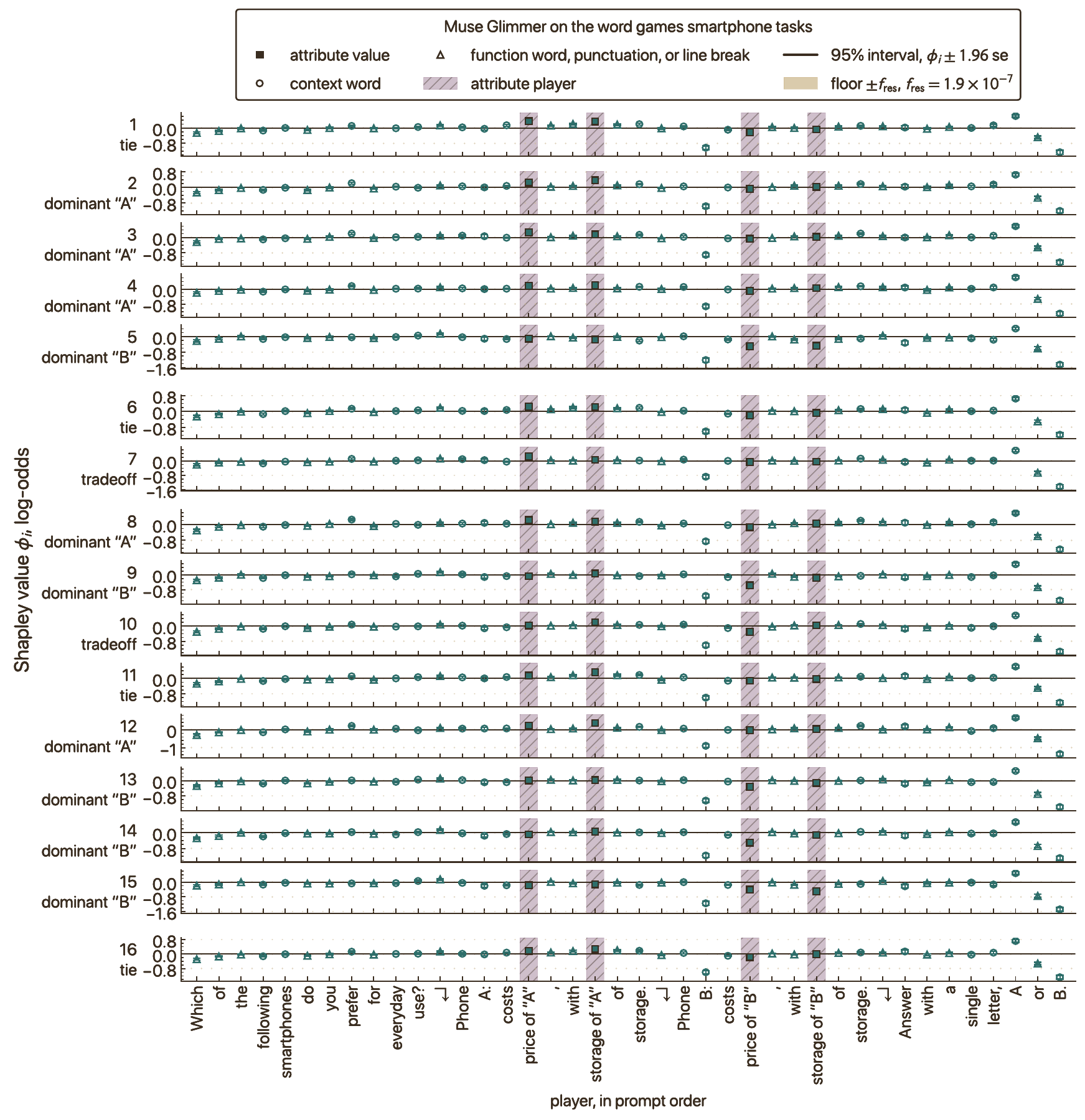}
\caption{Word-level Shapley values on every smartphone task for Muse Glimmer (word games), drawn as in Figure~\appref{fig:shapley-e5-hotel-qwen3-8-27b}.}\label{fig:shapley-e5-smartphone-muse-glimmer}
\end{figure}
\endgroup{}

\begingroup%
\begin{figure}[tbp]
\centering
\includegraphics[width=0.97\textwidth]{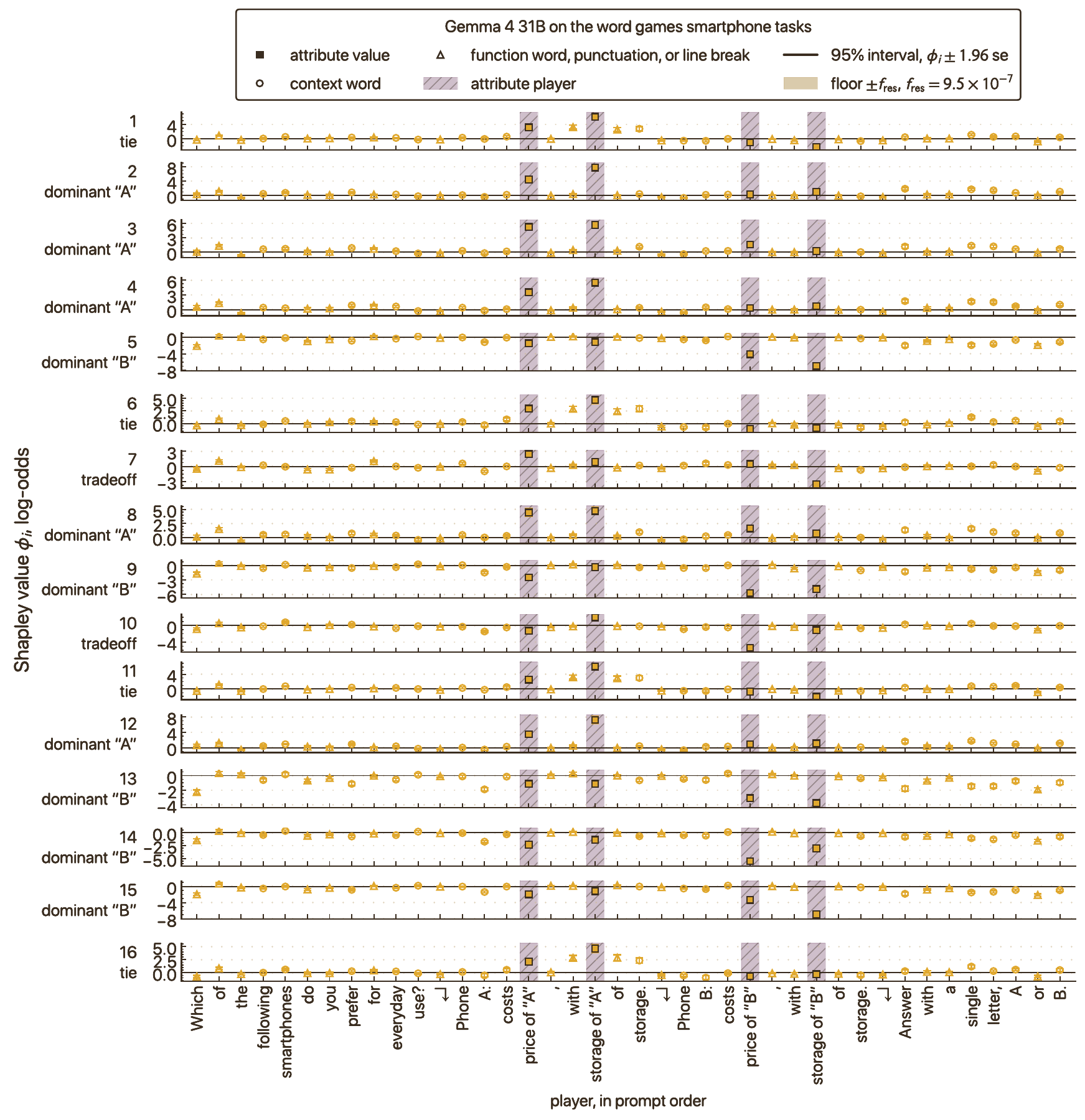}
\caption{Word-level Shapley values on every smartphone task for Gemma 4 31B (word games), drawn as in Figure~\appref{fig:shapley-e5-hotel-qwen3-8-27b}.}\label{fig:shapley-e5-smartphone-gemma4-31b}
\end{figure}
\endgroup{}

\begingroup%
\begin{figure}[tbp]
\centering
\includegraphics[width=0.97\textwidth]{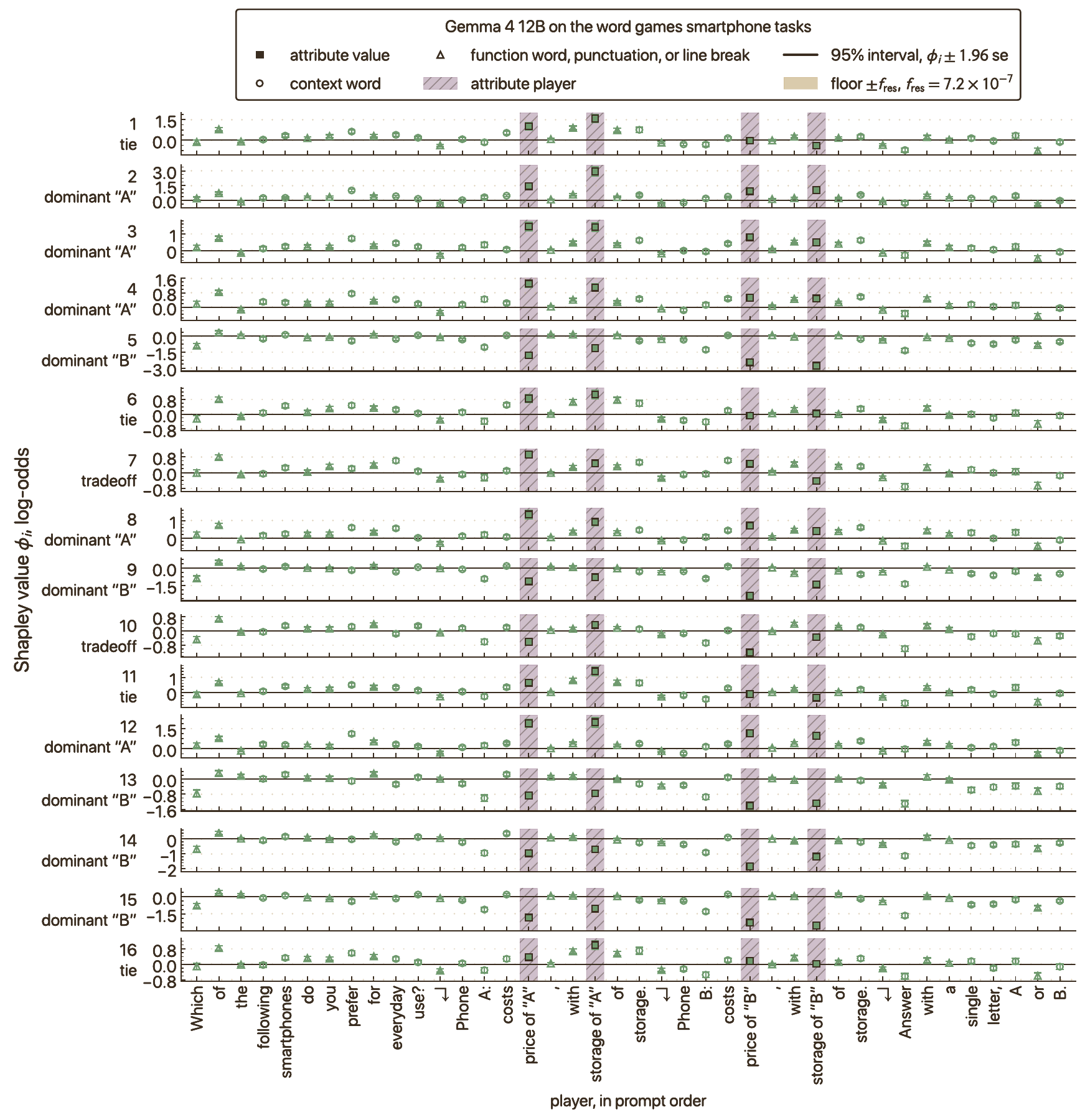}
\caption{Word-level Shapley values on every smartphone task for Gemma 4 12B (word games), drawn as in Figure~\appref{fig:shapley-e5-hotel-qwen3-8-27b}.}\label{fig:shapley-e5-smartphone-gemma4-12b}
\end{figure}
\endgroup{}

\begingroup\long\def\floatNotesBody{
Cosine is cosine similarity of absolute attribution vectors; Spearman correlates ranks of signed values within tasks. Similarity cells give task medians and bracketed interquartile ranges. $f_{\mathrm{res}}$ is each model's measurement floor in log-odds, from Table~\appref{tab:repro:validation}.
}%
\begin{table}[tbp]
\centering
\caption{Cross-model agreement of the word-level Shapley values on the flight tasks.}\label{tab:agreement}
\footnotesize\setlength{\tabcolsep}{4pt}
\begin{NiceTabular}{@{}l;l;r;r;r;c;c@{}}
\toprule
Model A & Model B & $f_{\mathrm{res}}$, A & $f_{\mathrm{res}}$, B & Tasks & Cosine & Spearman \\
\midrule
Qwen 3.8 27B & Muse Glimmer & $0.31$ & $1.9\times10^{-7}$ & $16$ & \begin{tabular}[t]{@{}c@{}}$0.64$\\$[0.63, 0.66]$\end{tabular} & \begin{tabular}[t]{@{}c@{}}$0.55$\\$[0.29, 0.67]$\end{tabular} \\
\dottedrule
Qwen 3.8 27B & Gemma 4 31B & $0.31$ & $9.5\times10^{-7}$ & $16$ & \begin{tabular}[t]{@{}c@{}}$0.78$\\$[0.64, 0.82]$\end{tabular} & \begin{tabular}[t]{@{}c@{}}$0.38$\\$[0.22, 0.66]$\end{tabular} \\
\dottedrule
Qwen 3.8 27B & Gemma 4 12B & $0.31$ & $7.2\times10^{-7}$ & $16$ & \begin{tabular}[t]{@{}c@{}}$0.72$\\$[0.70, 0.88]$\end{tabular} & \begin{tabular}[t]{@{}c@{}}$0.57$\\$[0.54, 0.67]$\end{tabular} \\
\dottedrule
Muse Glimmer & Gemma 4 31B & $1.9\times10^{-7}$ & $9.5\times10^{-7}$ & $16$ & \begin{tabular}[t]{@{}c@{}}$0.40$\\$[0.35, 0.52]$\end{tabular} & \begin{tabular}[t]{@{}c@{}}$0.37$\\$[0.09, 0.42]$\end{tabular} \\
\dottedrule
Muse Glimmer & Gemma 4 12B & $1.9\times10^{-7}$ & $7.2\times10^{-7}$ & $16$ & \begin{tabular}[t]{@{}c@{}}$0.49$\\$[0.43, 0.57]$\end{tabular} & \begin{tabular}[t]{@{}c@{}}$0.22$\\$[0.10, 0.49]$\end{tabular} \\
\dottedrule
Gemma 4 31B & Gemma 4 12B & $9.5\times10^{-7}$ & $7.2\times10^{-7}$ & $16$ & \begin{tabular}[t]{@{}c@{}}$0.82$\\$[0.75, 0.90]$\end{tabular} & \begin{tabular}[t]{@{}c@{}}$0.56$\\$[0.50, 0.80]$\end{tabular} \\
\bottomrule
\end{NiceTabular}
\end{table}
\endgroup{}
\floatnotes{agreement}

\begingroup\long\def\floatNotesBody{
Bars: 95\% Monte Carlo intervals. Pairs differ only in that slot. Filled pairs' absolute contrasts exceed $1.96\sqrt{\mathrm{se}_1^2+\mathrm{se}_2^2}$; hollow pairs are unresolved. Games use separate seeds. Dashes oppose the better-level direction. Last panel: tie scores and mean. Tan bands: measurement floors.
}%
\begin{figure}[tbp]
\centering
\includegraphics[width=\textwidth]{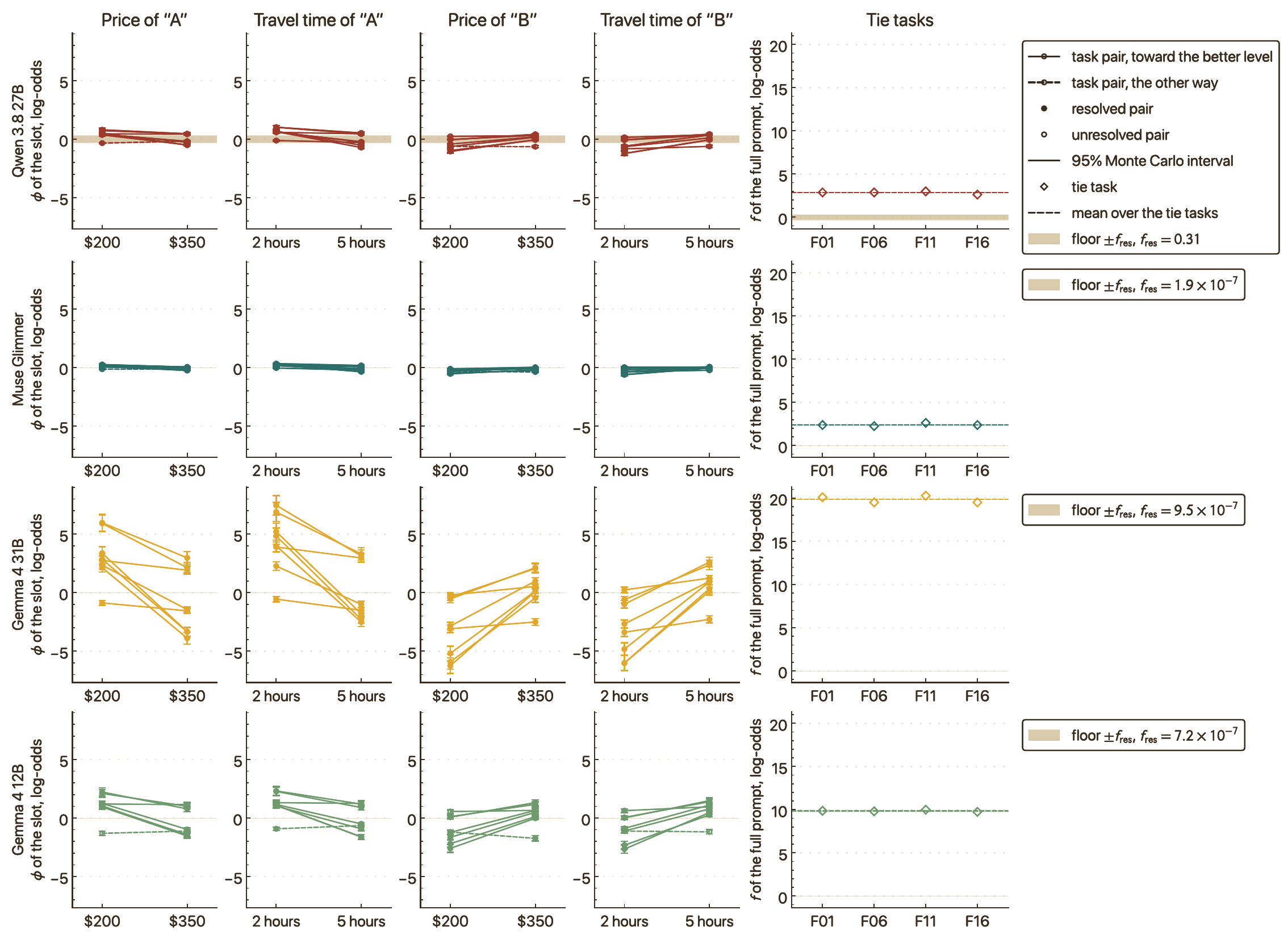}
\caption{Attribute-allocation ordering and complete-prompt tie scores on the flight tasks.}\label{fig:e2-rationality}
\end{figure}
\endgroup{}
\figurenotes{e2_rationality}
\begingroup\long\def\floatNotesBody{
Favored means the dominant option has positive log-odds against the other option. The margin is the smallest positive log-odds among favored tasks. These are direct complete-prompt scores; no Monte Carlo interval applies.
}%
\begin{table}[tbp]
\centering
\caption{Direct choice checks on the flight dominance tasks.}\label{tab:dominance}
\footnotesize\setlength{\tabcolsep}{4pt}
\begin{NiceTabular}{@{}l;r;r@{}}
\toprule
Model & Favored / tasks & Smallest margin (log-odds) \\
\midrule
Qwen 3.8 27B & $10/10$ & $4.88$ \\
\dottedrule
Muse Glimmer & $10/10$ & $2.38$ \\
\dottedrule
Gemma 4 31B & $10/10$ & $24.10$ \\
\dottedrule
Gemma 4 12B & $10/10$ & $14.91$ \\
\bottomrule
\end{NiceTabular}
\end{table}
\endgroup{}
\floatnotes{dominance}

\subsection{Personas}

Table~\ref{tab:persona-shares} and Figure~\appref{fig:e6} distinguish the attribute share within task components from the persona share of the whole prompt. No persona plus three descriptions gives four conditions. Their interquartile ranges describe heterogeneity across the sixteen tasks. Meaningful respondent descriptions can change both response and allocation without demonstrating a repair.

For the generic description, read Figures~\ref{fig:shapley-e6-flight-persona-1-qwen3-8-27b}, \ref{fig:shapley-e6-flight-persona-1-muse-glimmer}, \ref{fig:shapley-e6-flight-persona-1-gemma4-31b}, and~\appref{fig:shapley-e6-flight-persona-1-gemma4-12b}. The price-conscious description's profiles are Figures~\ref{fig:shapley-e6-flight-persona-2-qwen3-8-27b}, \ref{fig:shapley-e6-flight-persona-2-muse-glimmer}, \ref{fig:shapley-e6-flight-persona-2-gemma4-31b}, and~\appref{fig:shapley-e6-flight-persona-2-gemma4-12b}. The detailed description's profiles appear in Figures~\ref{fig:shapley-e6-flight-persona-3-qwen3-8-27b}, \ref{fig:shapley-e6-flight-persona-3-muse-glimmer}, \ref{fig:shapley-e6-flight-persona-3-gemma4-31b}, and~\appref{fig:shapley-e6-flight-persona-3-gemma4-12b}. Persona clauses precede task components in these displays.

\begingroup\long\def\floatNotesBody{
Cells show medians across 16 flight tasks and bracketed interquartile ranges. Attribute share divides attribute mass by task mass; persona share divides clause mass by whole-prompt mass. All masses sum absolute final values.
}%
\begin{table}[htbp]
\centering
\caption{Persona results by model and condition, one panel per model.}\label{tab:persona-shares}
\footnotesize\setlength{\tabcolsep}{4pt}
\begin{NiceTabular}{@{}l;c;c@{}}
\toprule
Condition & Attribute share of the task (\%) & Persona share of the prompt (\%) \\
\midrule
\multicolumn{3}{@{}l}{\textit{Qwen 3.8 27B, floor $f_{\mathrm{res}}$ $0.31$}} \\
no persona & \begin{tabular}[t]{@{}c@{}}$19.1$\\$[14.6, 24.5]$\end{tabular} & $0.0$ \\
\dottedrule
the generic description & \begin{tabular}[t]{@{}c@{}}$23.7$\\$[19.3, 29.4]$\end{tabular} & \begin{tabular}[t]{@{}c@{}}$2.2$\\$[0.6, 3.0]$\end{tabular} \\
\dottedrule
the price-conscious description & \begin{tabular}[t]{@{}c@{}}$29.7$\\$[22.7, 35.3]$\end{tabular} & \begin{tabular}[t]{@{}c@{}}$5.6$\\$[2.3, 6.5]$\end{tabular} \\
\dottedrule
the detailed description & \begin{tabular}[t]{@{}c@{}}$30.7$\\$[24.2, 37.9]$\end{tabular} & \begin{tabular}[t]{@{}c@{}}$7.9$\\$[6.3, 8.3]$\end{tabular} \\
\addlinespace
\multicolumn{3}{@{}l}{\textit{Muse Glimmer, floor $f_{\mathrm{res}}$ $1.9\times10^{-7}$}} \\
no persona & \begin{tabular}[t]{@{}c@{}}$8.6$\\$[7.2, 12.5]$\end{tabular} & $0.0$ \\
\dottedrule
the generic description & \begin{tabular}[t]{@{}c@{}}$9.3$\\$[8.0, 13.4]$\end{tabular} & \begin{tabular}[t]{@{}c@{}}$0.5$\\$[0.3, 0.6]$\end{tabular} \\
\dottedrule
the price-conscious description & \begin{tabular}[t]{@{}c@{}}$11.1$\\$[7.7, 15.5]$\end{tabular} & \begin{tabular}[t]{@{}c@{}}$1.8$\\$[1.4, 2.2]$\end{tabular} \\
\dottedrule
the detailed description & \begin{tabular}[t]{@{}c@{}}$11.9$\\$[9.5, 14.6]$\end{tabular} & \begin{tabular}[t]{@{}c@{}}$4.2$\\$[3.9, 4.5]$\end{tabular} \\
\bottomrule
\end{NiceTabular}
\end{table}
\begin{table}[htbp]
\ContinuedFloat
\centering
\caption{Persona results by model and condition, continued.}\label{tab:persona-shares:1}
\footnotesize\setlength{\tabcolsep}{4pt}
\begin{NiceTabular}{@{}l;c;c@{}}
\toprule
Condition & Attribute share of the task (\%) & Persona share of the prompt (\%) \\
\midrule
\multicolumn{3}{@{}l}{\textit{Gemma 4 31B, floor $f_{\mathrm{res}}$ $9.5\times10^{-7}$}} \\
no persona & \begin{tabular}[t]{@{}c@{}}$37.7$\\$[28.0, 40.5]$\end{tabular} & $0.0$ \\
\dottedrule
the generic description & \begin{tabular}[t]{@{}c@{}}$41.9$\\$[35.1, 46.6]$\end{tabular} & \begin{tabular}[t]{@{}c@{}}$2.4$\\$[1.8, 3.6]$\end{tabular} \\
\dottedrule
the price-conscious description & \begin{tabular}[t]{@{}c@{}}$45.5$\\$[40.3, 50.2]$\end{tabular} & \begin{tabular}[t]{@{}c@{}}$5.8$\\$[2.9, 8.0]$\end{tabular} \\
\dottedrule
the detailed description & \begin{tabular}[t]{@{}c@{}}$49.2$\\$[42.4, 54.2]$\end{tabular} & \begin{tabular}[t]{@{}c@{}}$4.5$\\$[3.3, 6.0]$\end{tabular} \\
\addlinespace
\multicolumn{3}{@{}l}{\textit{Gemma 4 12B, floor $f_{\mathrm{res}}$ $7.2\times10^{-7}$}} \\
no persona & \begin{tabular}[t]{@{}c@{}}$26.3$\\$[19.6, 29.6]$\end{tabular} & $0.0$ \\
\dottedrule
the generic description & \begin{tabular}[t]{@{}c@{}}$33.7$\\$[28.1, 38.4]$\end{tabular} & \begin{tabular}[t]{@{}c@{}}$4.2$\\$[0.9, 4.8]$\end{tabular} \\
\dottedrule
the price-conscious description & \begin{tabular}[t]{@{}c@{}}$40.8$\\$[33.2, 45.3]$\end{tabular} & \begin{tabular}[t]{@{}c@{}}$7.0$\\$[6.3, 8.5]$\end{tabular} \\
\dottedrule
the detailed description & \begin{tabular}[t]{@{}c@{}}$40.6$\\$[35.2, 45.5]$\end{tabular} & \begin{tabular}[t]{@{}c@{}}$3.5$\\$[2.9, 4.8]$\end{tabular} \\
\bottomrule
\end{NiceTabular}
\end{table}
\endgroup{}
\floatnotes{persona_shares}

\begingroup\long\def\floatNotesBody{
Boxes summarize 16 flight tasks; points are tasks. Dashed lines mark the no-persona median. Top: attribute mass divided by task mass. Bottom: persona-clause mass divided by whole-prompt mass. The two shares have different denominators.
}%
\begin{figure}[tbp]
\centering
\includegraphics[width=\textwidth]{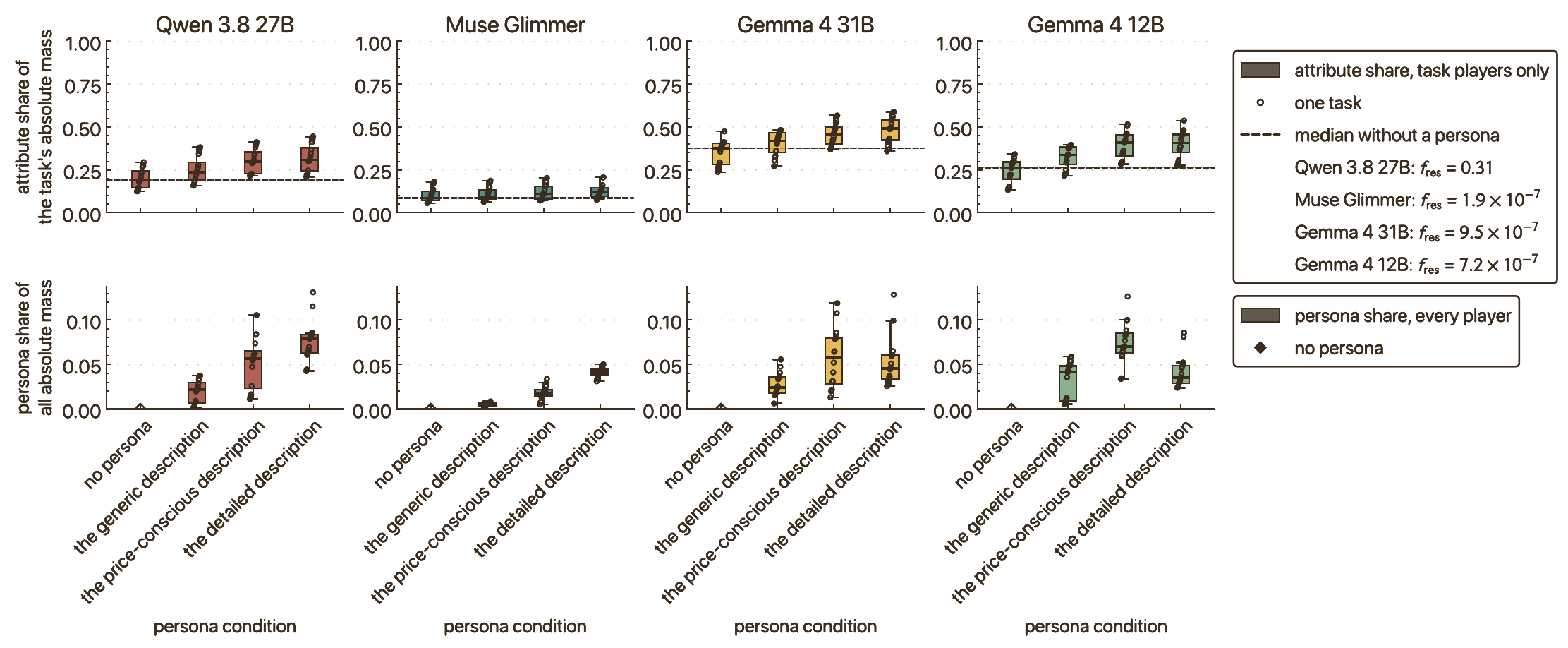}
\caption{The attribute share of the task's mass and the persona's share of the whole prompt by persona condition, by model.}\label{fig:e6}
\end{figure}
\endgroup{}
\figurenotes{e6_personas}

\begingroup\long\def\floatNotesBody{
Players follow prompt order. Shading and squares: attributes; circles: context; triangles: function words, punctuation, or line breaks; diamonds: persona clauses. Bars: 95\% Monte Carlo intervals in log-odds. Tan bands: plus or minus the measurement floor.
}%
\begin{figure}[tbp]
\centering
\includegraphics[width=0.97\textwidth]{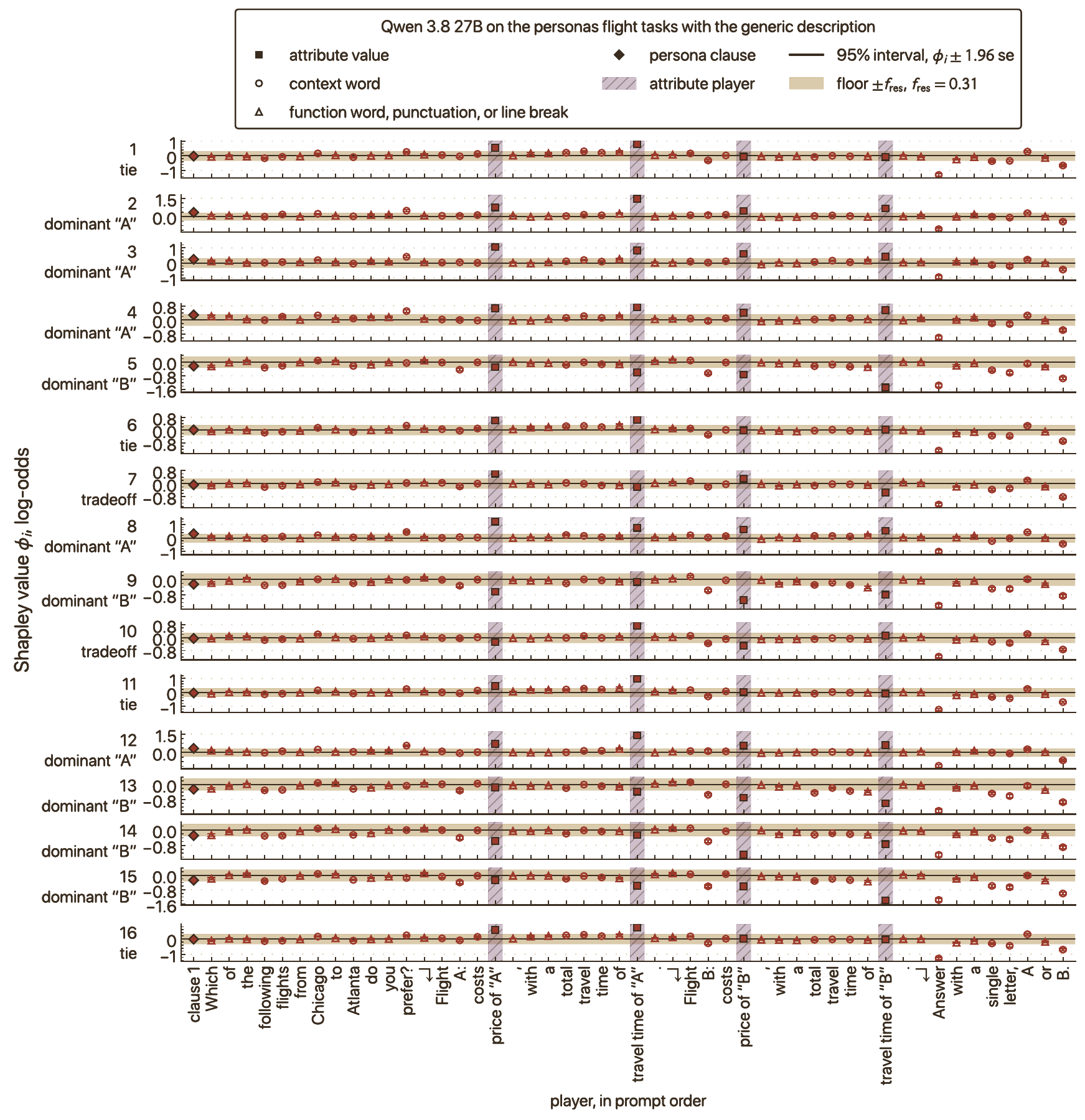}
\caption{Word-level Shapley values on every flight task for Qwen 3.8 27B with the generic description (personas), in the task order of Table~\appref{app:prompts:tasks:flight}.}\label{fig:shapley-e6-flight-persona-1-qwen3-8-27b}
\end{figure}
\endgroup{}
\figurenotes{shapley_e6_flight_persona-1_qwen3-8-27b}

\begingroup%
\begin{figure}[tbp]
\centering
\includegraphics[width=0.97\textwidth]{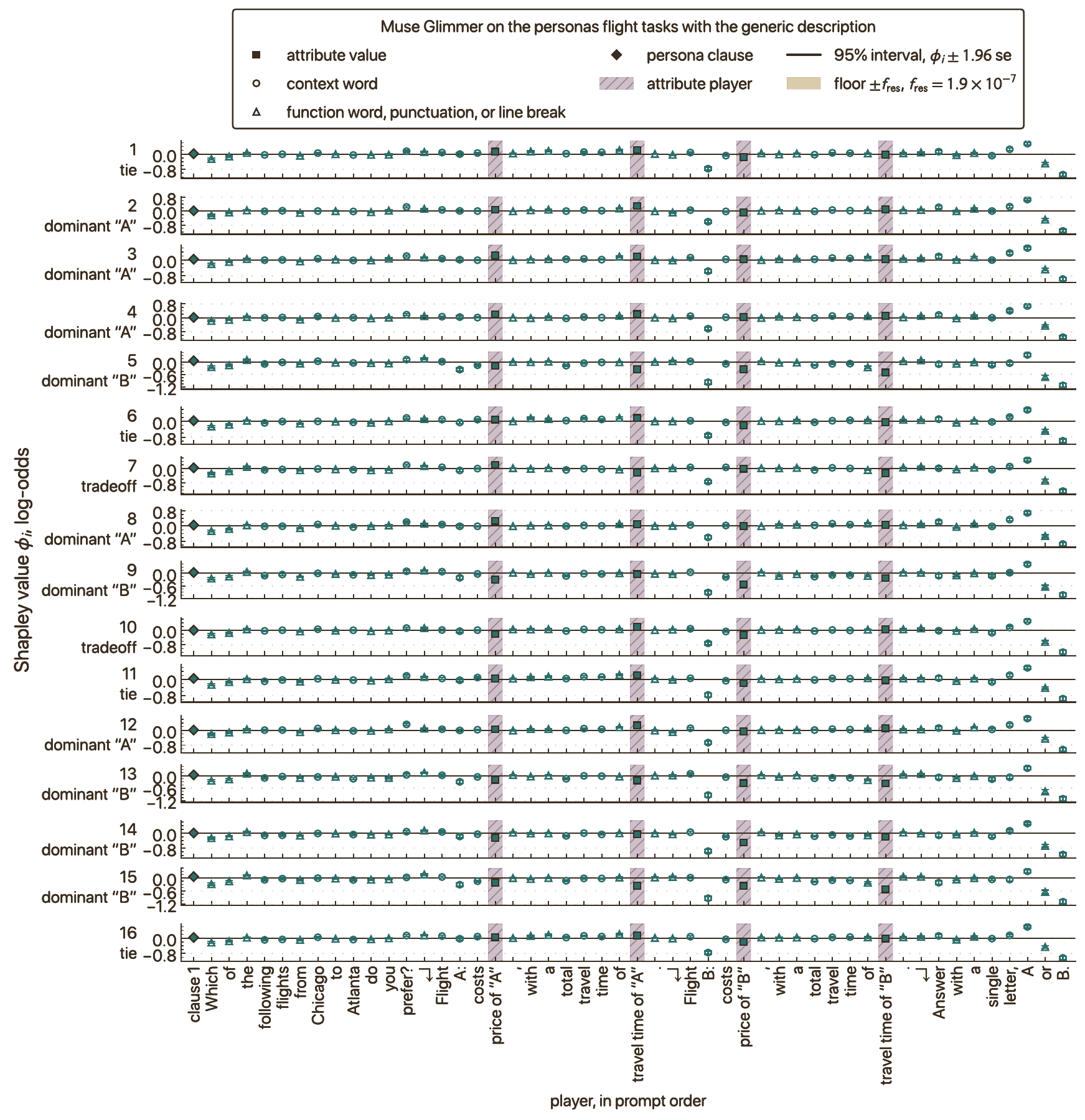}
\caption{Word-level Shapley values on every flight task for Muse Glimmer with the generic description (personas), drawn as in Figure~\appref{fig:shapley-e6-flight-persona-1-qwen3-8-27b}, with the persona's clauses first.}\label{fig:shapley-e6-flight-persona-1-muse-glimmer}
\end{figure}
\endgroup{}

\begingroup%
\begin{figure}[tbp]
\centering
\includegraphics[width=0.97\textwidth]{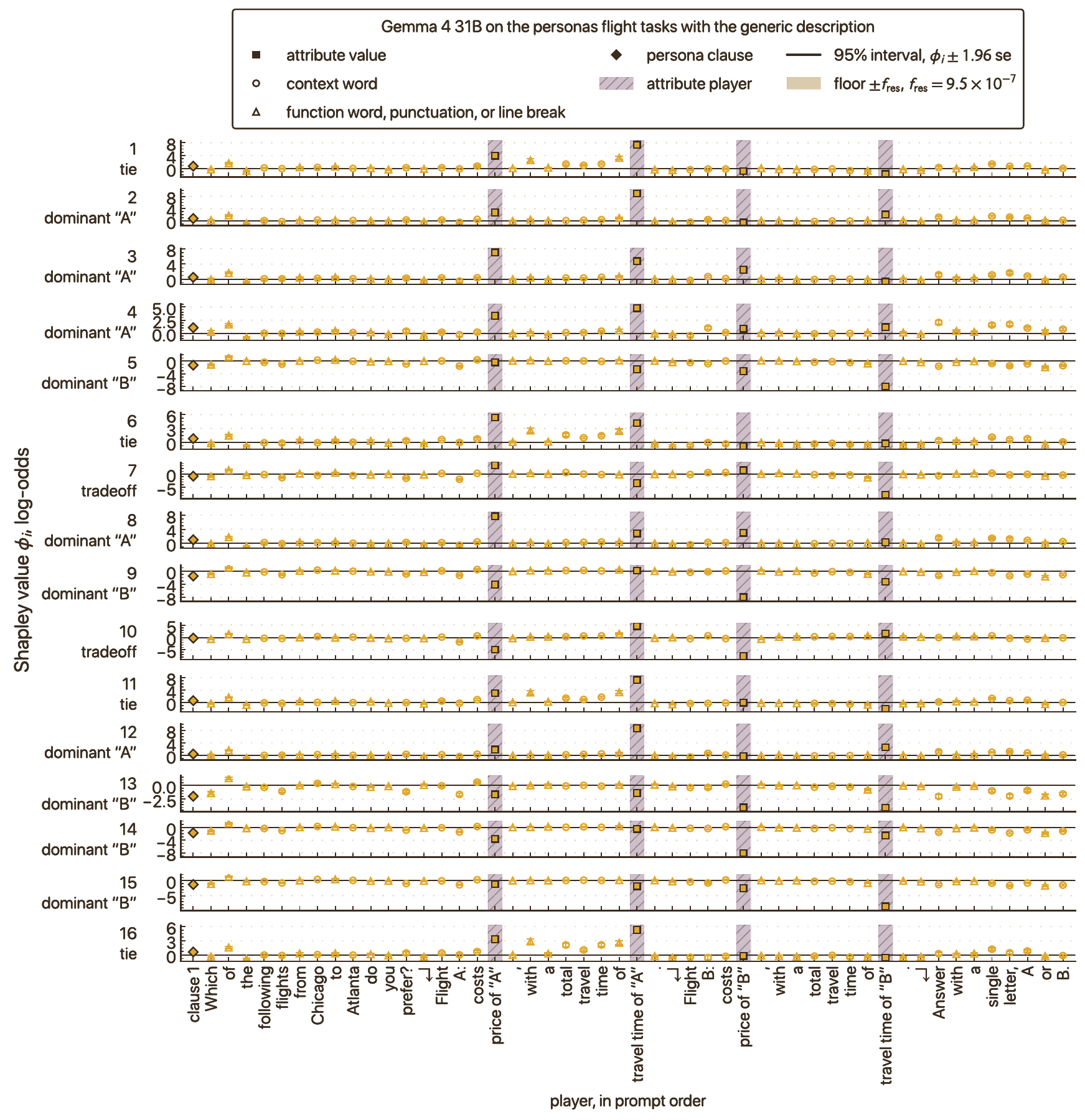}
\caption{Word-level Shapley values on every flight task for Gemma 4 31B with the generic description (personas), drawn as in Figure~\appref{fig:shapley-e6-flight-persona-1-qwen3-8-27b}, with the persona's clauses first.}\label{fig:shapley-e6-flight-persona-1-gemma4-31b}
\end{figure}
\endgroup{}

\begingroup%
\begin{figure}[tbp]
\centering
\includegraphics[width=0.97\textwidth]{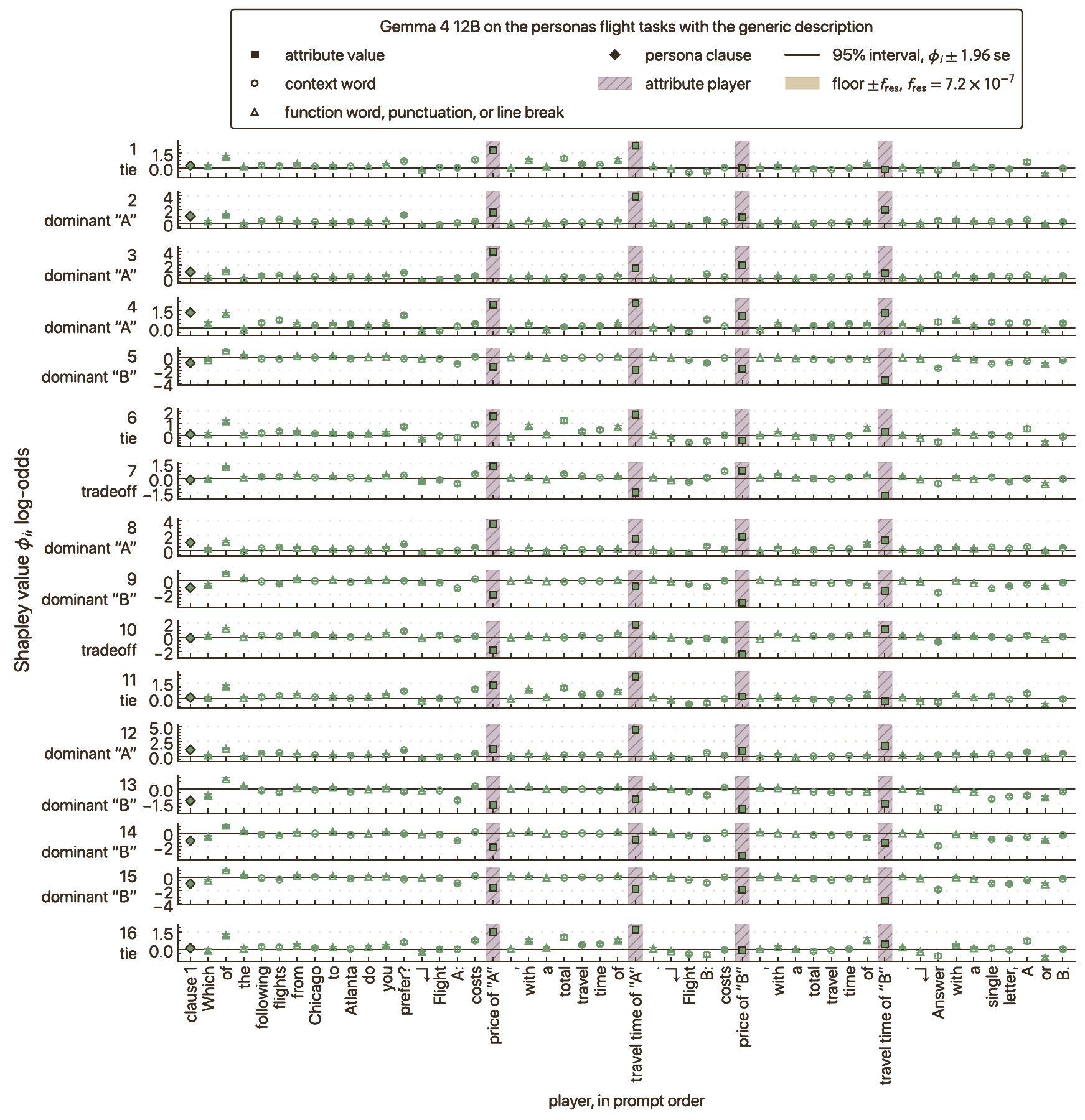}
\caption{Word-level Shapley values on every flight task for Gemma 4 12B with the generic description (personas), drawn as in Figure~\appref{fig:shapley-e6-flight-persona-1-qwen3-8-27b}, with the persona's clauses first.}\label{fig:shapley-e6-flight-persona-1-gemma4-12b}
\end{figure}
\endgroup{}

\begingroup%
\begin{figure}[tbp]
\centering
\includegraphics[width=0.97\textwidth]{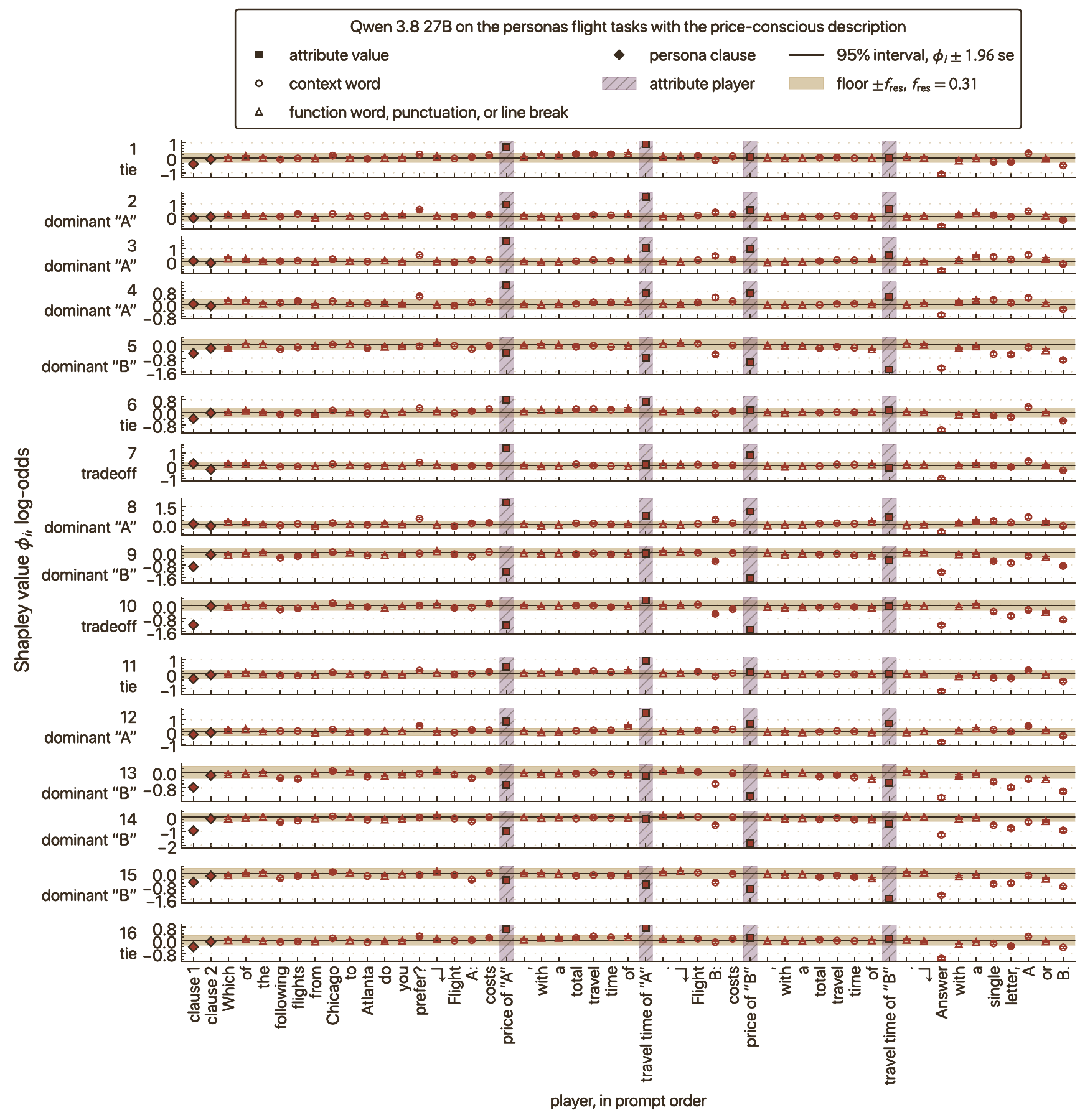}
\caption{Word-level Shapley values on every flight task for Qwen 3.8 27B with the price-conscious description (personas), drawn as in Figure~\appref{fig:shapley-e6-flight-persona-1-qwen3-8-27b}, with the persona's clauses first.}\label{fig:shapley-e6-flight-persona-2-qwen3-8-27b}
\end{figure}
\endgroup{}

\begingroup%
\begin{figure}[tbp]
\centering
\includegraphics[width=0.97\textwidth]{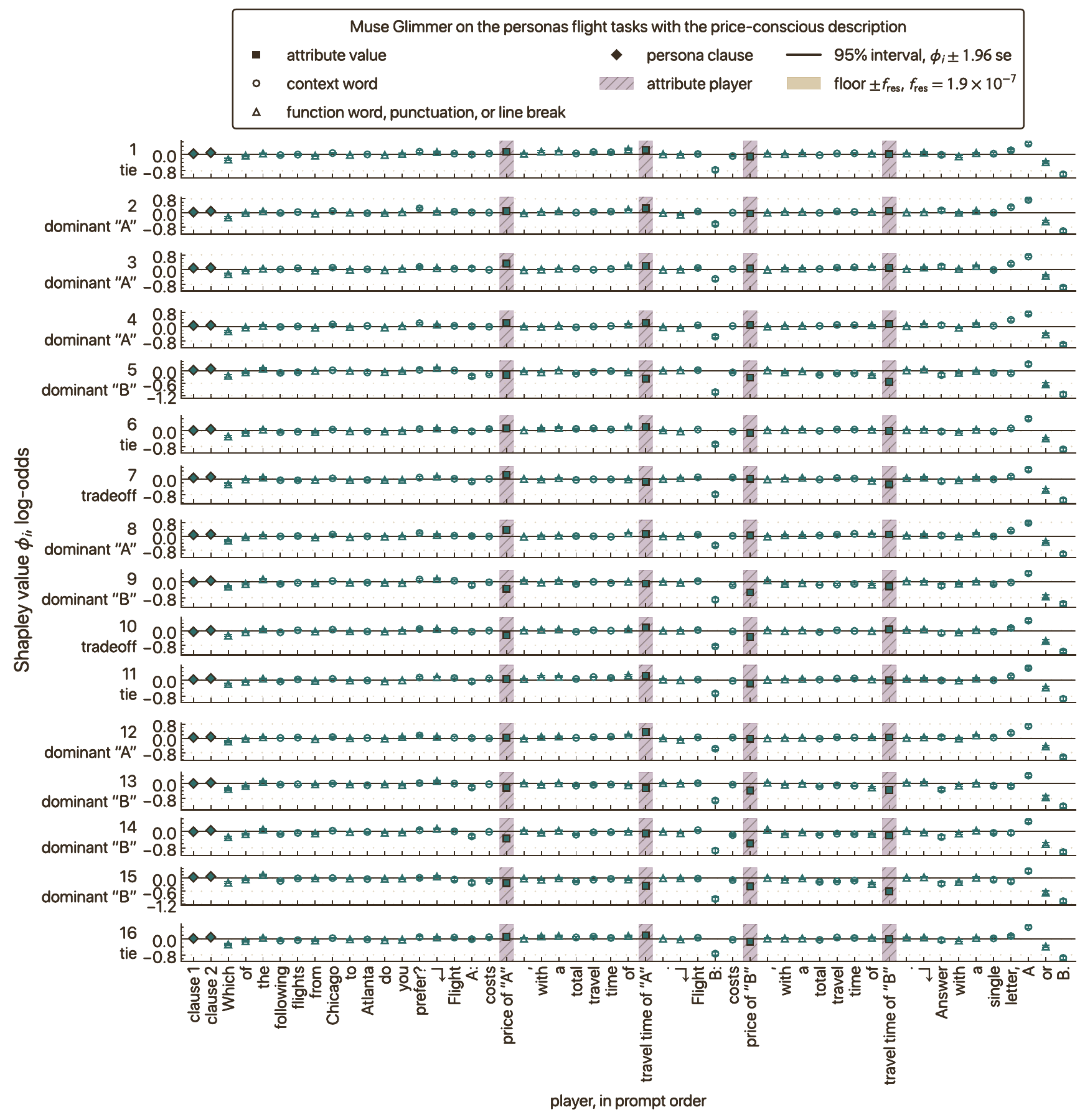}
\caption{Word-level Shapley values on every flight task for Muse Glimmer with the price-conscious description (personas), drawn as in Figure~\appref{fig:shapley-e6-flight-persona-1-qwen3-8-27b}, with the persona's clauses first.}\label{fig:shapley-e6-flight-persona-2-muse-glimmer}
\end{figure}
\endgroup{}

\begingroup%
\begin{figure}[tbp]
\centering
\includegraphics[width=0.97\textwidth]{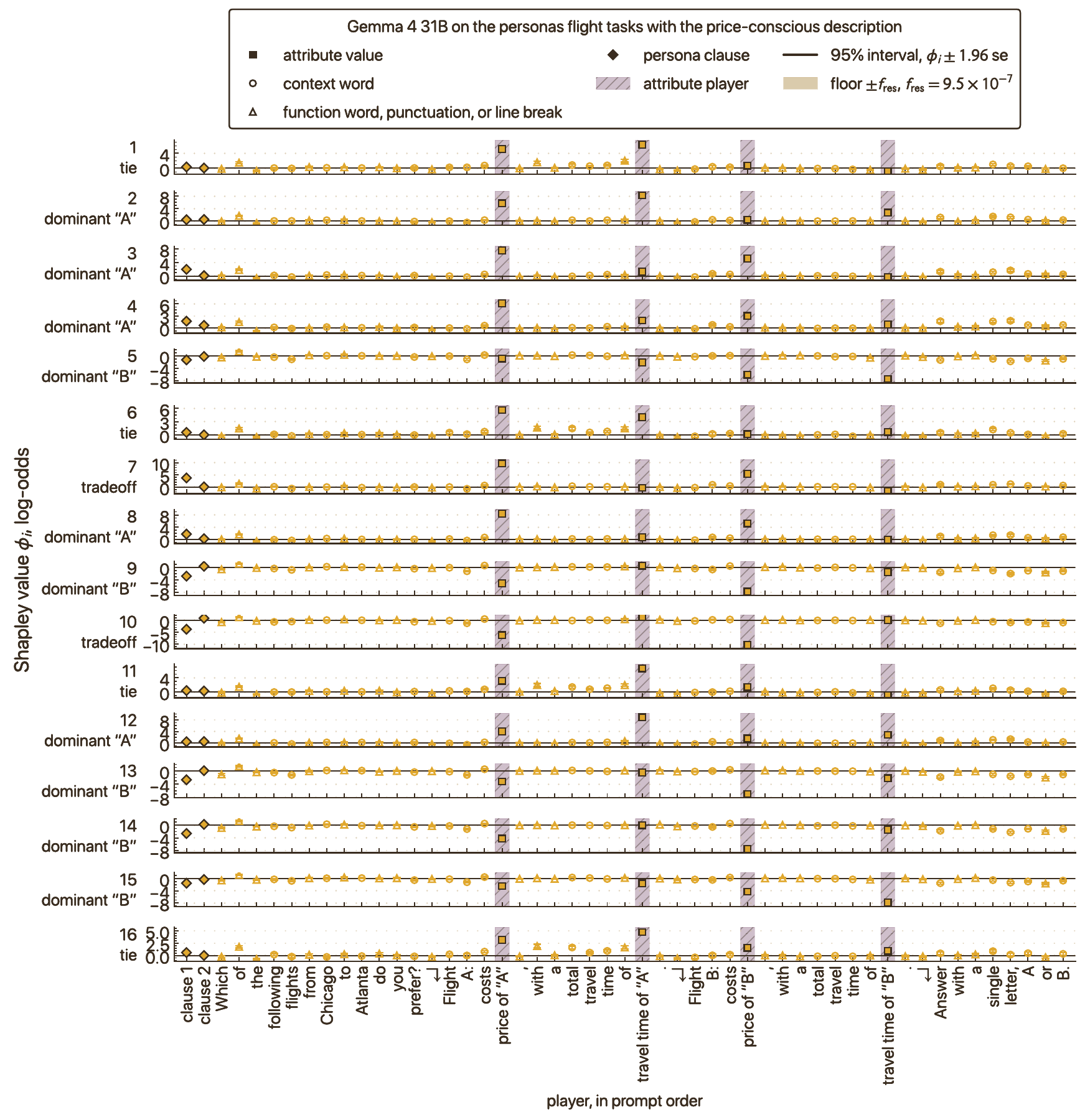}
\caption{Word-level Shapley values on every flight task for Gemma 4 31B with the price-conscious description (personas), drawn as in Figure~\appref{fig:shapley-e6-flight-persona-1-qwen3-8-27b}, with the persona's clauses first.}\label{fig:shapley-e6-flight-persona-2-gemma4-31b}
\end{figure}
\endgroup{}

\begingroup%
\begin{figure}[tbp]
\centering
\includegraphics[width=0.97\textwidth]{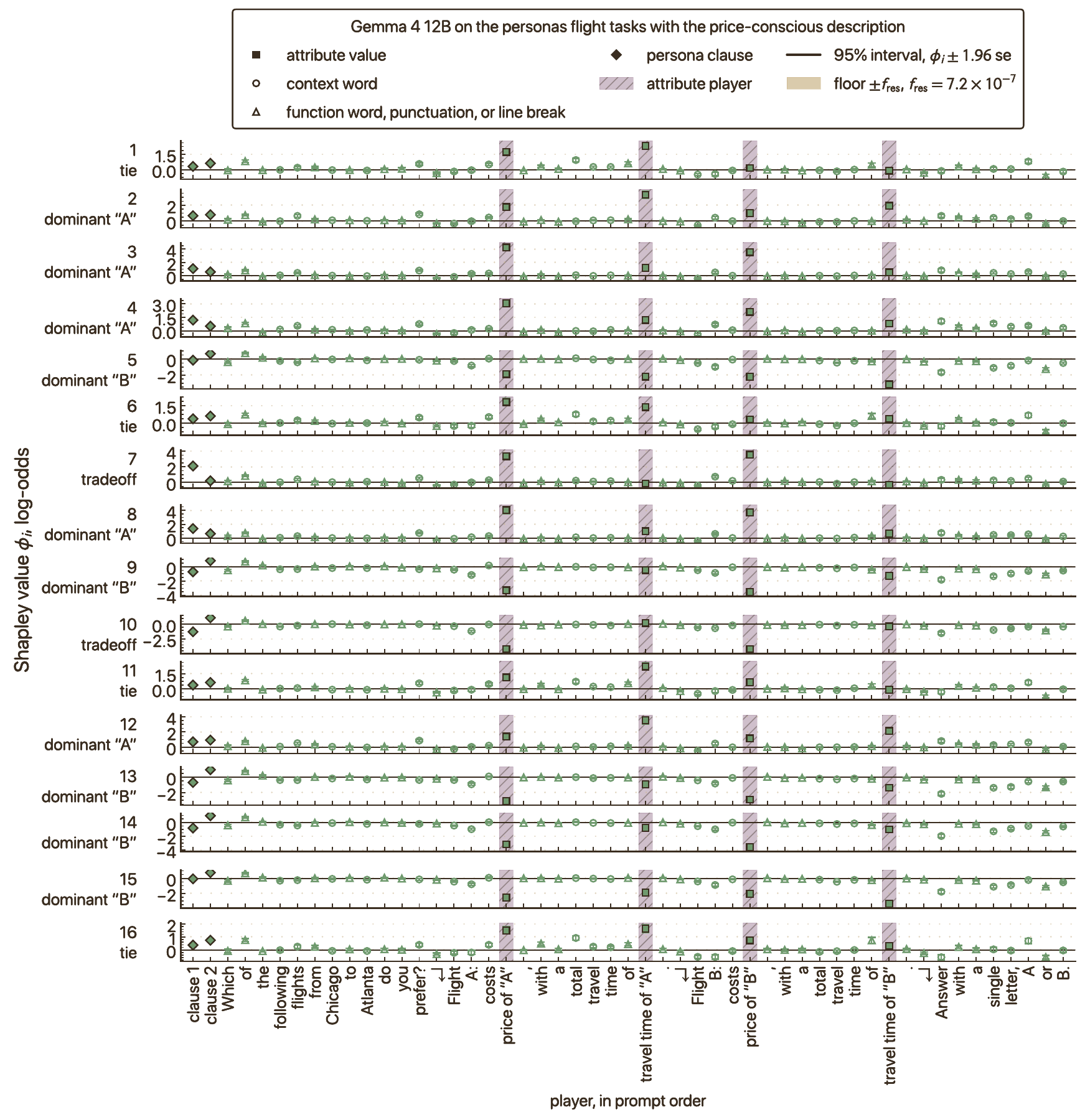}
\caption{Word-level Shapley values on every flight task for Gemma 4 12B with the price-conscious description (personas), drawn as in Figure~\appref{fig:shapley-e6-flight-persona-1-qwen3-8-27b}, with the persona's clauses first.}\label{fig:shapley-e6-flight-persona-2-gemma4-12b}
\end{figure}
\endgroup{}

\begingroup%
\begin{figure}[tbp]
\centering
\includegraphics[width=0.97\textwidth]{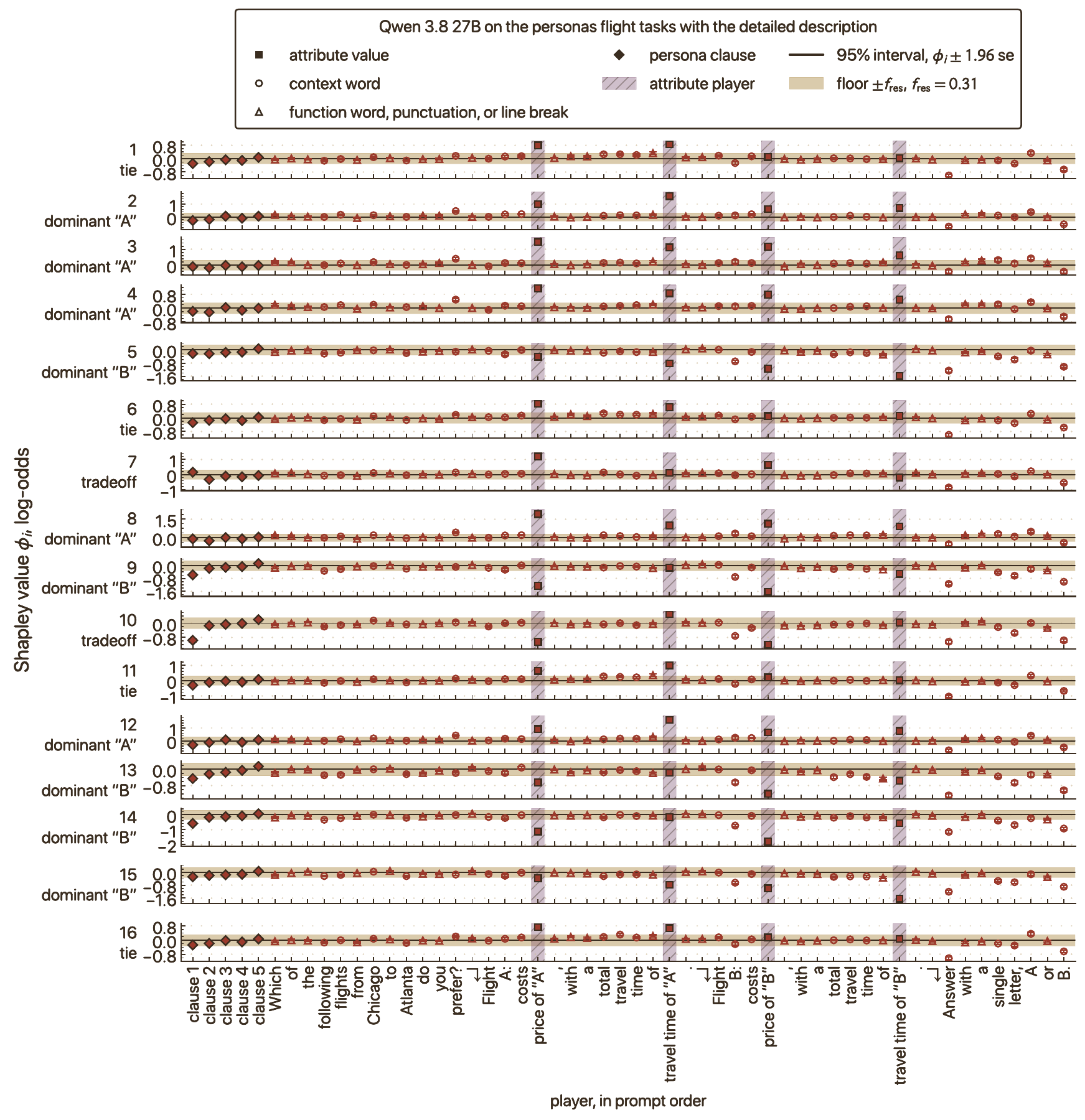}
\caption{Word-level Shapley values on every flight task for Qwen 3.8 27B with the detailed description (personas), drawn as in Figure~\appref{fig:shapley-e6-flight-persona-1-qwen3-8-27b}, with the persona's clauses first.}\label{fig:shapley-e6-flight-persona-3-qwen3-8-27b}
\end{figure}
\endgroup{}

\begingroup%
\begin{figure}[tbp]
\centering
\includegraphics[width=0.97\textwidth]{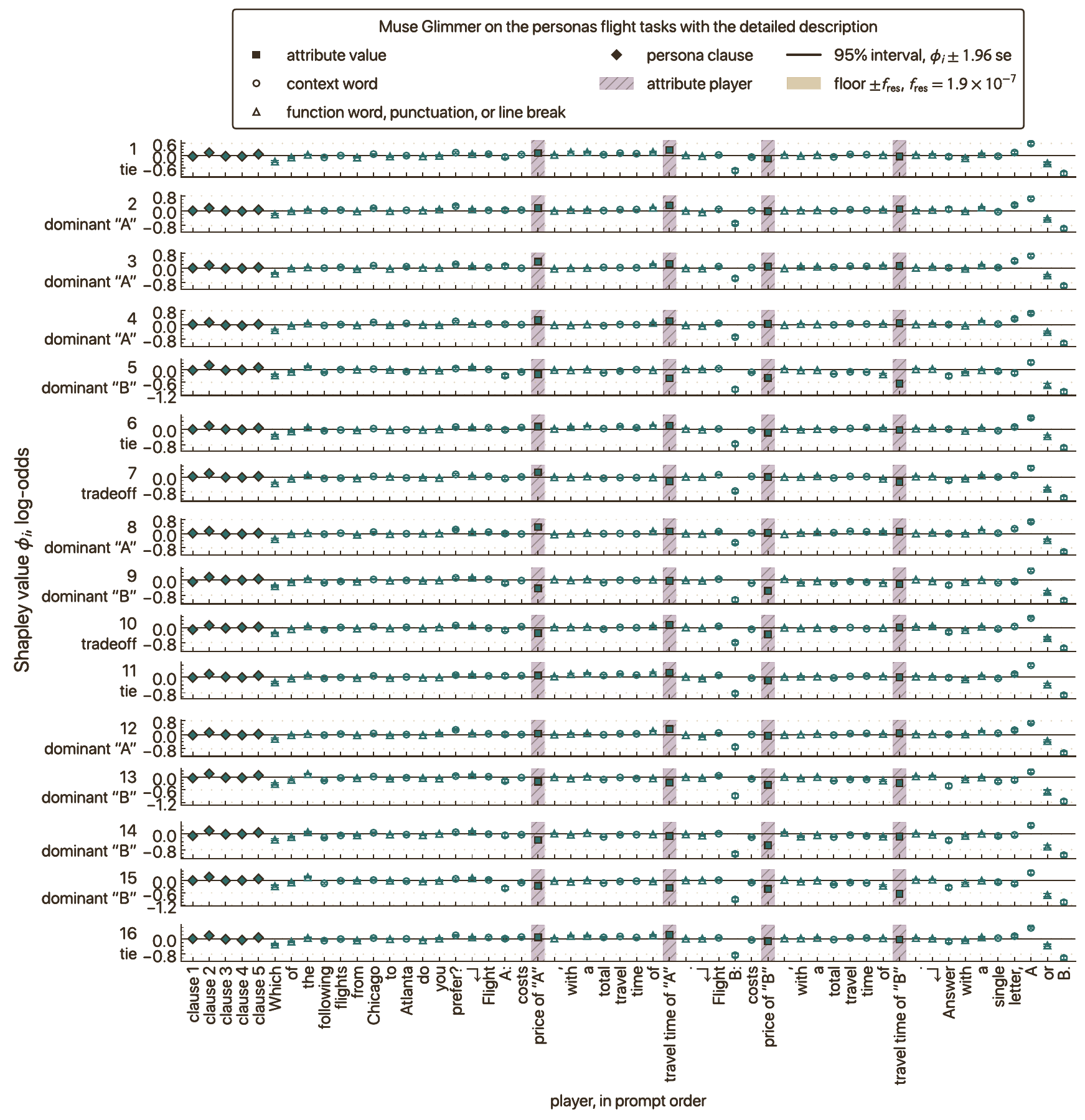}
\caption{Word-level Shapley values on every flight task for Muse Glimmer with the detailed description (personas), drawn as in Figure~\appref{fig:shapley-e6-flight-persona-1-qwen3-8-27b}, with the persona's clauses first.}\label{fig:shapley-e6-flight-persona-3-muse-glimmer}
\end{figure}
\endgroup{}

\begingroup%
\begin{figure}[tbp]
\centering
\includegraphics[width=0.97\textwidth]{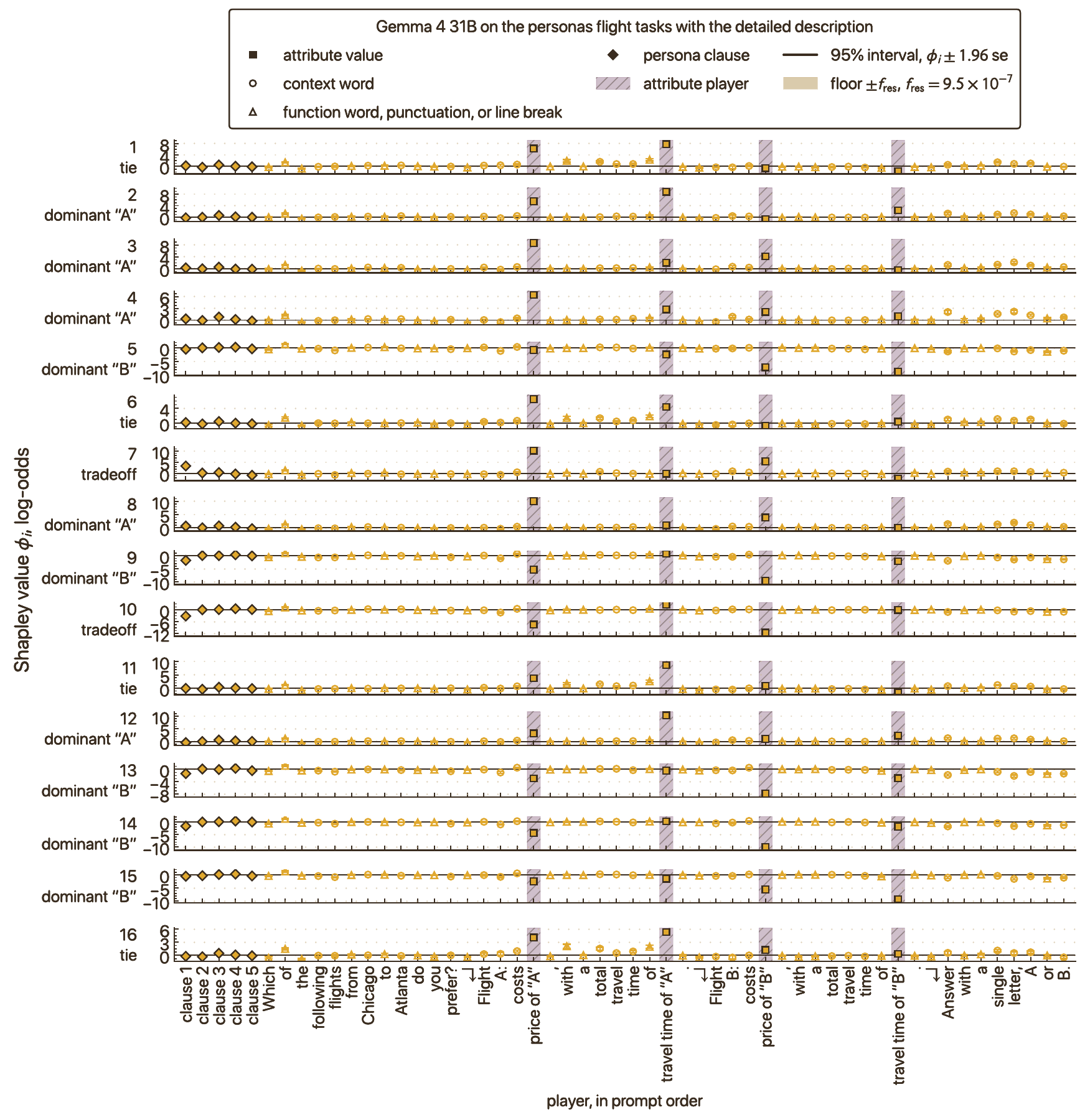}
\caption{Word-level Shapley values on every flight task for Gemma 4 31B with the detailed description (personas), drawn as in Figure~\appref{fig:shapley-e6-flight-persona-1-qwen3-8-27b}, with the persona's clauses first.}\label{fig:shapley-e6-flight-persona-3-gemma4-31b}
\end{figure}
\endgroup{}

\begingroup%
\begin{figure}[tbp]
\centering
\includegraphics[width=0.97\textwidth]{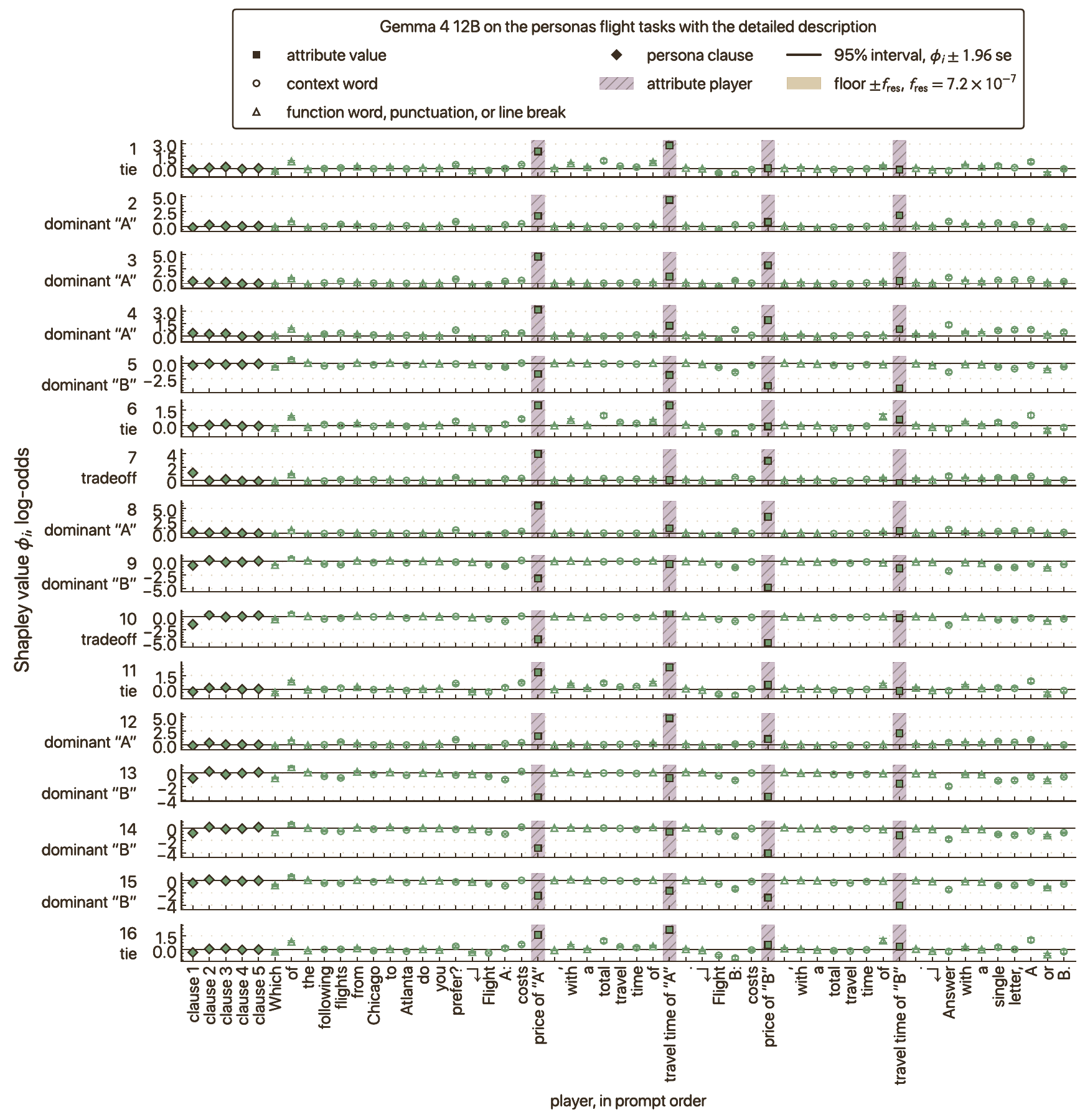}
\caption{Word-level Shapley values on every flight task for Gemma 4 12B with the detailed description (personas), drawn as in Figure~\appref{fig:shapley-e6-flight-persona-1-qwen3-8-27b}, with the persona's clauses first.}\label{fig:shapley-e6-flight-persona-3-gemma4-12b}
\end{figure}
\endgroup{}

\subsection{Willingness to pay}\label{part:economics-reproduction}

For each coalition $S$ of edits, I fit ordinary least squares over tasks $j$:
\begin{equation}\label{eq:online-wtp}
 f_{jS}=\alpha_S+\beta_S^{\mathrm{price}}(\mathrm{price}_{Aj}-\mathrm{price}_{Bj})
                    +\beta_S^{\mathrm{attr}}(\mathrm{attr}_{Aj}-\mathrm{attr}_{Bj})+\varepsilon_{jS}.
\end{equation}
Here $\mathrm{price}$ is an option's price, $\mathrm{attr}$ is its travel time, distance, or storage, and $\alpha_S$ is an intercept. Constant slopes summarize this two-level design. Willingness to pay (WTP) is $\beta_S^{\mathrm{attr}}/\beta_S^{\mathrm{price}}$ per hour or mile saved and $-\beta_S^{\mathrm{attr}}/\beta_S^{\mathrm{price}}$ per additional gigabyte. The calculation reports a ratio only if $\beta_S^{\mathrm{price}}<0$ and $|\beta_S^{\mathrm{price}}|\ge2\,\mathrm{se}(\widehat\beta_S^{\mathrm{price}})$; otherwise it is undefined. This is the implemented coefficient-ratio rule.

Within-phrasing standard errors use the delta method with covariance $\widehat\sigma^2(Z^\top Z)^{-1}$, where $Z$ contains the intercept and attribute differences and $\widehat\sigma^2$ is residual sum of squares divided by residual degrees of freedom. The conventional uncorrelated, constant-variance residual approximation describes fit to fixed scores, not human sampling. Two levels identify conditional slopes, not a valuation curve. The design fixes each attribute's level gap, while the coefficient ratio is estimated from the model's scores; a ratio near the level-gap ratio describes these responses and is not an algebraic consequence of two levels.

Table~\appref{tab:wtp} includes all three categories and the standard deviation (SD) across phrasings. Figure~\appref{fig:e3-wtp} displays the 64 combinations of the six flight formatting edits. Muse's retained flight range is \tnRetainedWtpMinFlightMuseGlimmer{} to \tnRetainedWtpMaxFlightMuseGlimmer{} dollars per hour. A spread below a wide fit-based standard error need not be negligible. Saturated dominance scores and tie-task asymmetries can leave substantial regression residuals.

\begingroup\long\def\floatNotesBody{
Units: dollars per hour for flights, dollars per mile for hotels, dollars per GB for smartphones. Base is unedited. SD is sample standard deviation across phrasings. Defined counts ratios meeting the price-coefficient rule (Equation~\appeqref{eq:online-wtp}). Columns distinguish formatting from retained wording.
}%
\begin{table}[htbp]
\centering
\caption{Willingness to pay by template and model across the formatting coalitions and the retained wording coalitions.}\label{tab:wtp}
\footnotesize\setlength{\tabcolsep}{3pt}
\begin{NiceTabular}{@{}l;r;r;r;r;r;r;r;r;r@{}}
\toprule
 & \multicolumn{5}{c}{Formatting coalitions} & \multicolumn{4}{c}{Retained wording coalitions} \\
\cmidrule(lr){2-6}\cmidrule(lr){7-10}
Model & Base & Median & Min & Max & SD & Defined & Median & Min & Max \\
\midrule
\multicolumn{10}{@{}l}{\textit{Flight}} \\
Qwen 3.8 27B & $54.59$ & $56.37$ & $52.33$ & $59.85$ & $1.51$ & $1024$ & $57.52$ & $49.78$ & $64.54$ \\
\dottedrule
Muse Glimmer & $60.85$ & $66.94$ & $54.88$ & $85.52$ & $10.17$ & $1024$ & $61.79$ & $46.95$ & $85.52$ \\
\dottedrule
Gemma 4 31B & $50.94$ & $51.12$ & $48.50$ & $54.80$ & $1.49$ & $1024$ & $50.01$ & $44.09$ & $56.51$ \\
\dottedrule
Gemma 4 12B & $50.83$ & $50.17$ & $45.97$ & $55.19$ & $2.08$ & $1024$ & $52.50$ & $45.24$ & $59.78$ \\
\addlinespace
\multicolumn{10}{@{}l}{\textit{Hotel}} \\
Qwen 3.8 27B & $16.27$ & $16.74$ & $15.64$ & $17.40$ & $0.42$ & $2048$ & $17.50$ & $15.04$ & $21.04$ \\
\dottedrule
Muse Glimmer & $12.05$ & $13.22$ & $11.45$ & $15.90$ & $0.93$ & $2039$ & $15.87$ & $11.45$ & $29.47$ \\
\dottedrule
Gemma 4 31B & $16.86$ & $17.11$ & $15.80$ & $18.90$ & $0.85$ & $2048$ & $17.99$ & $14.87$ & $20.70$ \\
\dottedrule
Gemma 4 12B & $15.00$ & $15.80$ & $13.79$ & $17.67$ & $0.81$ & $2048$ & $17.68$ & $12.80$ & $24.42$ \\
\addlinespace
\multicolumn{10}{@{}l}{\textit{Smartphone}} \\
Qwen 3.8 27B & $0.81$ & $0.79$ & $0.74$ & $0.84$ & $0.02$ & $2048$ & $0.74$ & $0.61$ & $0.92$ \\
\dottedrule
Muse Glimmer & $0.98$ & $1.03$ & $0.90$ & $1.20$ & $0.08$ & $1863$ & $1.04$ & $0.59$ & $1.64$ \\
\dottedrule
Gemma 4 31B & $0.68$ & $0.68$ & $0.64$ & $0.73$ & $0.02$ & $2048$ & $0.71$ & $0.52$ & $0.95$ \\
\dottedrule
Gemma 4 12B & $0.81$ & $0.82$ & $0.76$ & $0.88$ & $0.03$ & $2048$ & $0.76$ & $0.63$ & $1.02$ \\
\bottomrule
\end{NiceTabular}
\end{table}
\endgroup{}
\floatnotes{wtp}

\begingroup\long\def\floatNotesBody{
Boxes show medians and interquartile ranges; whiskers reach the furthest coalition within 1.5 interquartile ranges. Points are coalitions. Dashed lines mark unedited phrasing; bands show its WTP plus or minus median within-phrasing standard error. Legends give measurement floors.
}%
\begin{figure}[tbp]
\centering
\includegraphics[width=\textwidth]{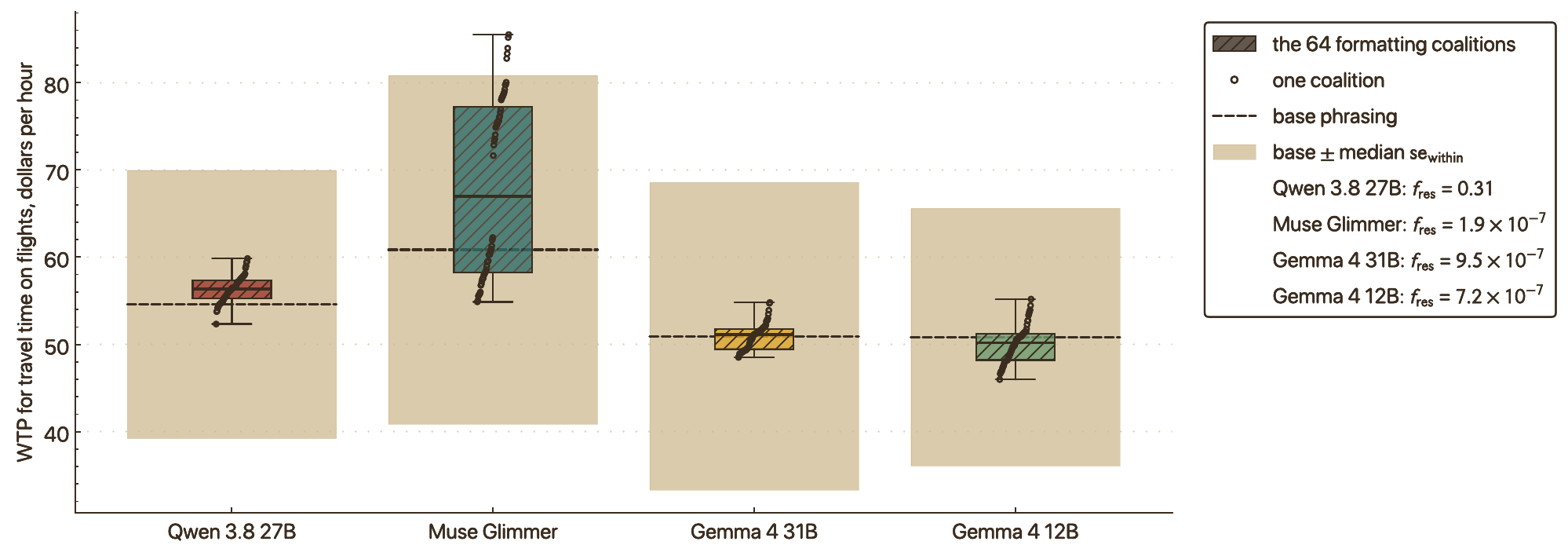}
\caption{Willingness to pay for travel time on flights across the 64 coalitions of the 6 formatting edits, by model, in dollars per hour.}\label{fig:e3-wtp}
\end{figure}
\endgroup{}
\figurenotes{e3_wtp}

For the example flight trade-off, the fitted score is $\widehat f=\alpha_S+3\beta_S^{\mathrm{price}}(W-50)$, with willingness to pay $W$ in dollars per hour. Crossing the physical level-gap ratio of fifty therefore need not reverse fitted choice when the intercept is nonzero; observed choice also includes the residual. Attribution rescaled by a price allocation, or a response shift divided by a fitted price slope, likewise does not establish an experimentally equivalent fare change.

Table~\appref{tab:wtp-personas} supplies the persona regressions. Their respondent descriptions change meaning; their coefficient differences should not be attributed to equivalent formatting alone.

\begingroup\long\def\floatNotesBody{
Willingness to pay (WTP) for travel time uses dollars per hour. The first group covers 64 formatting coalitions; the second also varies persona rephrasings. SD is sample standard deviation across coalitions. Summaries include only defined coefficient ratios; the descriptions change respondent meaning.
}%
\begin{table}[htbp]
\centering
\caption{Willingness to pay by model and persona condition, one panel per model.}\label{tab:wtp-personas}
\footnotesize\setlength{\tabcolsep}{4pt}
\begin{NiceTabular}{@{}l;r;r;r;r;r;r;r;r@{}}
\toprule
 & \multicolumn{5}{c}{Formatting coalitions} & \multicolumn{3}{c}{Persona and formatting coalitions} \\
\cmidrule(lr){2-6}\cmidrule(lr){7-9}
Condition & Base & Median & Min & Max & SD & Median & Min & Max \\
\midrule
\multicolumn{9}{@{}l}{\textit{Qwen 3.8 27B, floor $f_{\mathrm{res}}$ $0.31$}} \\
no persona & $54.59$ & $56.37$ & $52.33$ & $59.85$ & $1.51$ &  &  &  \\
\dottedrule
the generic description & $53.45$ & $52.71$ & $49.45$ & $56.08$ & $1.57$ & $52.71$ & $49.45$ & $56.08$ \\
\dottedrule
the price-conscious description & $30.84$ & $31.12$ & $28.88$ & $33.27$ & $1.19$ & $31.32$ & $28.87$ & $34.12$ \\
\dottedrule
the detailed description & $34.21$ & $35.20$ & $33.64$ & $38.34$ & $1.45$ & $35.76$ & $33.39$ & $39.81$ \\
\addlinespace
\multicolumn{9}{@{}l}{\textit{Muse Glimmer, floor $f_{\mathrm{res}}$ $1.9\times10^{-7}$}} \\
no persona & $60.85$ & $66.94$ & $54.88$ & $85.52$ & $10.17$ &  &  &  \\
\dottedrule
the generic description & $57.30$ & $64.75$ & $51.77$ & $81.98$ & $9.42$ & $64.75$ & $51.77$ & $81.98$ \\
\dottedrule
the price-conscious description & $46.45$ & $49.39$ & $42.29$ & $61.82$ & $5.73$ & $49.93$ & $42.29$ & $66.04$ \\
\dottedrule
the detailed description & $47.14$ & $51.79$ & $44.23$ & $63.65$ & $5.69$ & $52.55$ & $41.81$ & $70.63$ \\
\bottomrule
\end{NiceTabular}
\end{table}
\begin{table}[htbp]
\ContinuedFloat
\centering
\caption{Willingness to pay by model and persona condition, continued.}\label{tab:wtp-personas:1}
\footnotesize\setlength{\tabcolsep}{4pt}
\begin{NiceTabular}{@{}l;r;r;r;r;r;r;r;r@{}}
\toprule
 & \multicolumn{5}{c}{Formatting coalitions} & \multicolumn{3}{c}{Persona and formatting coalitions} \\
\cmidrule(lr){2-6}\cmidrule(lr){7-9}
Condition & Base & Median & Min & Max & SD & Median & Min & Max \\
\midrule
\multicolumn{9}{@{}l}{\textit{Gemma 4 31B, floor $f_{\mathrm{res}}$ $9.5\times10^{-7}$}} \\
no persona & $50.94$ & $51.12$ & $48.50$ & $54.80$ & $1.49$ &  &  &  \\
\dottedrule
the generic description & $59.40$ & $55.92$ & $53.28$ & $65.58$ & $2.22$ & $55.92$ & $53.28$ & $65.58$ \\
\dottedrule
the price-conscious description & $25.33$ & $25.37$ & $25.13$ & $25.63$ & $0.13$ & $25.54$ & $25.12$ & $26.04$ \\
\dottedrule
the detailed description & $26.56$ & $26.58$ & $26.11$ & $28.28$ & $0.54$ & $27.20$ & $25.88$ & $31.52$ \\
\addlinespace
\multicolumn{9}{@{}l}{\textit{Gemma 4 12B, floor $f_{\mathrm{res}}$ $7.2\times10^{-7}$}} \\
no persona & $50.83$ & $50.17$ & $45.97$ & $55.19$ & $2.08$ &  &  &  \\
\dottedrule
the generic description & $52.42$ & $47.19$ & $42.60$ & $55.88$ & $3.98$ & $47.19$ & $42.60$ & $55.88$ \\
\dottedrule
the price-conscious description & $24.83$ & $26.01$ & $24.25$ & $27.69$ & $0.76$ & $26.70$ & $24.12$ & $29.99$ \\
\dottedrule
the detailed description & $26.26$ & $26.15$ & $25.14$ & $28.17$ & $0.70$ & $26.82$ & $24.80$ & $29.95$ \\
\bottomrule
\end{NiceTabular}
\end{table}
\endgroup{}
\floatnotes{wtp_personas}
 \clearpage
\section{Robustness checks}\label{part:constructions}

I vary the reference, payoff, players, classification, and edit set to show which attribution question each choice defines. The tables of this appendix use the Qwen flight tasks, except Table~\appref{tab:grid-formulations}, which compares the two games across models. Read their distributions as descriptions of these fixed tasks; a larger attribute share does not establish a better simulated respondent.

\subsection{References and score scales}

Table~\appref{tab:grid-references} compares underscores with empty strings and five random-token references. A random reference draws replacement tokens from the model vocabulary with the recorded seed; it does not represent a population of natural questions. Qwen's chat template trims message-edge whitespace. Thus empty-reference coalitions with missing edge words can also lose edge spacing. The evaluated reference includes that template behavior.

The median attribute share changes from \tnGridShareMedianWord\% under underscores to \tnGridShareMedianEmpty\% under empty replacement. This difference concerns the reference prompt, even though the complete task stays fixed. Sign agreement and rank correlations locate additional changes that the scalar share can conceal.

Conditional Shapley values instead draw absent words from their distribution given the present words, keeping coalition prompts natural. That conditional distribution is intractable here. Conditional values can also credit a word because it predicts other words the model uses, so they answer a different question from replacing a word by a fixed reference \citep{app:janzing2020feature,app:sundararajan2020many}.

\begingroup\long\def\floatNotesBody{
Primary uses underscores; empty uses empty strings; random rows use vocabulary draws. Share divides classified attribute mass by total absolute Shapley mass: task median [interquartile range]. Against primary, Same sign counts matching player signs; $\rho$ gives median within-task Spearman correlation.
}%
\begin{table}[htbp]
\centering
\caption{Reference vectors of the robustness checks on the 16 flight tasks of Qwen 3.8 27B.}\label{tab:grid-references}
\footnotesize\setlength{\tabcolsep}{4pt}
\begin{NiceTabular}{@{}l;c;r;r@{}}
\toprule
Reference & Attribute share (\%) & Same sign & $\rho$ \\
\midrule
primary & \begin{tabular}[t]{@{}c@{}}$19.1$\\$[14.6, 24.5]$\end{tabular} &  &  \\
\dottedrule
empty & \begin{tabular}[t]{@{}c@{}}$24.6$\\$[21.6, 27.2]$\end{tabular} & $556$ of $784$ & $0.67$ \\
\dottedrule
random-1 & \begin{tabular}[t]{@{}c@{}}$21.8$\\$[20.8, 23.5]$\end{tabular} & $586$ of $784$ & $0.65$ \\
\dottedrule
random-2 & \begin{tabular}[t]{@{}c@{}}$21.8$\\$[19.4, 22.5]$\end{tabular} & $554$ of $784$ & $0.64$ \\
\dottedrule
random-3 & \begin{tabular}[t]{@{}c@{}}$23.3$\\$[22.4, 25.4]$\end{tabular} & $567$ of $784$ & $0.64$ \\
\dottedrule
random-4 & \begin{tabular}[t]{@{}c@{}}$22.4$\\$[20.9, 23.8]$\end{tabular} & $569$ of $784$ & $0.59$ \\
\dottedrule
random-5 & \begin{tabular}[t]{@{}c@{}}$19.4$\\$[18.6, 20.4]$\end{tabular} & $582$ of $784$ & $0.58$ \\
\bottomrule
\end{NiceTabular}
\end{table}
\endgroup{}
\floatnotes{grid_references}

Table~\appref{tab:grid-payoffs} recomputes the word game using log-odds, conditional probability, and log-probability from the same scores. The probability and log-probability median shares are \tnGridShareMedianProbability{} and \tnGridShareMedianLogProbability\%. A nonlinear score transformation changes marginal contributions and their allocation. Multiplying a payoff by a nonzero constant instead scales all values equally and leaves the absolute share unchanged. Reversing the mathematical log-odds orientation negates every value; rewriting answer labels in the prompt need not do so.

\begingroup\long\def\floatNotesBody{
Payoffs: log-odds $f$, the probability $p(\mathrm{A};x)=1/(1+e^{-f(x)})$ of ``A'', and its natural logarithm $\log p(\mathrm{A};x)$. Share divides classified attribute mass by total absolute Shapley mass: task median [interquartile range]. Against log-odds, Same sign counts matching player signs; $\rho$ gives median within-task Spearman correlation.
}%
\begin{table}[htbp]
\centering
\caption{Payoffs of the robustness checks on the 16 flight tasks of Qwen 3.8 27B: the word games under three payoffs computed from the same scores.}\label{tab:grid-payoffs}
\footnotesize\setlength{\tabcolsep}{4pt}
\begin{NiceTabular}{@{}l;c;r;r@{}}
\toprule
Payoff & Attribute share (\%) & Same sign & $\rho$ \\
\midrule
log-odds $f$ & \begin{tabular}[t]{@{}c@{}}$19.1$\\$[14.6, 24.5]$\end{tabular} &  &  \\
\dottedrule
$p(\mathrm{A};x)$ & \begin{tabular}[t]{@{}c@{}}$16.4$\\$[13.5, 22.3]$\end{tabular} & $695$ of $784$ & $0.94$ \\
\dottedrule
$\log p(\mathrm{A};x)$ & \begin{tabular}[t]{@{}c@{}}$24.2$\\$[19.8, 26.8]$\end{tabular} & $684$ of $784$ & $0.94$ \\
\bottomrule
\end{NiceTabular}
\end{table}
\endgroup{}
\floatnotes{grid_payoffs}

\subsection{Players and classification}

Table~\appref{tab:span-membership} lists the words that belong to each of the flight template's ten spans.

\begingroup\long\def\floatNotesBody{
Angle-bracket placeholders denote the displayed attribute values with their units. Semicolons separate fragments belonging to the same span. Each span restores all its member words together; absent members retain their individual replacements.
}%
\begin{table}[tbp]
\centering
\caption{Membership of the flight template's ten spans.}\label{tab:span-membership}
\footnotesize\setlength{\tabcolsep}{4pt}
\begin{NiceTabular}{@{}>{\raggedright\arraybackslash}p{2.8cm};>{\raggedright\arraybackslash}p{10.5cm}@{}}
\toprule
Span & Member text \\
\midrule
question & line 1: \texttt{\protect\frenchspacing Which of the following flights}; line 1: \texttt{\protect\frenchspacing do you prefer?} \\
\dottedrule
route & line 1: \texttt{\protect\frenchspacing from Chicago to Atlanta} \\
\dottedrule
label of “A” & line 2: \texttt{\protect\frenchspacing Flight A:} \\
\dottedrule
price of “A” & line 2: \texttt{\protect\frenchspacing costs <price of A>,} \\
\dottedrule
travel time of “A” & line 2: \texttt{\protect\frenchspacing with a total travel time of <travel time of A>.} \\
\dottedrule
label of “B” & line 3: \texttt{\protect\frenchspacing Flight B:} \\
\dottedrule
price of “B” & line 3: \texttt{\protect\frenchspacing costs <price of B>,} \\
\dottedrule
travel time of “B” & line 3: \texttt{\protect\frenchspacing with a total travel time of <travel time of B>.} \\
\dottedrule
instruction & line 4: \texttt{\protect\frenchspacing Answer with a single letter, A or B.} \\
\dottedrule
line breaks & All line-break players between the prompt lines. \\
\bottomrule
\end{NiceTabular}
\end{table}
\endgroup{}
\floatnotes{span_membership}

Table~\appref{tab:grid-resolutions} compares detailed words with ten spans. The span game restores all words in a span together and retains member-wise replacements. Its complete and reference prompts therefore match the word game's endpoints, but its coalition contexts differ. Each span takes the lowest-numbered tier among its members. Attribute spans consequently include surrounding language; their share is not restricted to numeric values.

Table~\appref{tab:grid-classifications} instead leaves the word game fixed and changes the numerator from Tier~1 to Tiers~1 and~2. Under the generous span classification in Table~\appref{tab:grid-resolutions}, nearly all text joins the numerator. A large resulting share largely reflects that definition. Neither relabeling contributions nor merging players supplies a consumer-validity test.

\begingroup\long\def\floatNotesBody{
Spans inherit their lowest-numbered member tier. Strict attributes: the attribute values (Tier~1); generous: the attribute values and the context words (Tiers~1 and~2). Shares divide classified attribute mass by total absolute Shapley mass: task medians [interquartile ranges]. Above word level counts tasks with increased strict shares.
}%
\begin{table}[htbp]
\centering
\caption{Resolutions of the robustness checks on the 16 flight tasks of Qwen 3.8 27B: the word games against exact games of span players.}\label{tab:grid-resolutions}
\footnotesize\setlength{\tabcolsep}{4pt}
\begin{NiceTabular}{@{}l;r;c;c;r@{}}
\toprule
Resolution & Players & Attribute share (\%) & Generous share (\%) & Above word level \\
\midrule
word & $49$ & \begin{tabular}[t]{@{}c@{}}$19.1$\\$[14.6, 24.5]$\end{tabular} & \begin{tabular}[t]{@{}c@{}}$75.7$\\$[74.8, 77.8]$\end{tabular} &  \\
\dottedrule
span & $10$ & \begin{tabular}[t]{@{}c@{}}$40.8$\\$[35.8, 44.1]$\end{tabular} & \begin{tabular}[t]{@{}c@{}}$99.1$\\$[98.2, 99.4]$\end{tabular} & $16$ of $16$ \\
\bottomrule
\end{NiceTabular}
\end{table}
\endgroup{}
\floatnotes{grid_resolutions}
\begingroup\long\def\floatNotesBody{
Each cell gives the task median and bracketed interquartile range. Share divides classified attribute mass by total absolute Shapley mass. The generous scheme includes context words. Only category membership changes; the underlying word values stay fixed.
}%
\begin{table}[htbp]
\centering
\caption{Classification schemes of the robustness checks on the 16 flight tasks of Qwen 3.8 27B, from the word games.}\label{tab:grid-classifications}
\footnotesize\setlength{\tabcolsep}{4pt}
\begin{NiceTabular}{@{}l;l;c@{}}
\toprule
Scheme & Attribute tiers & Share (\%) \\
\midrule
strict & 1 & \begin{tabular}[t]{@{}c@{}}$19.1$\\$[14.6, 24.5]$\end{tabular} \\
\dottedrule
generous & 1 and 2 & \begin{tabular}[t]{@{}c@{}}$75.7$\\$[74.8, 77.8]$\end{tabular} \\
\bottomrule
\end{NiceTabular}
\end{table}
\endgroup{}
\floatnotes{grid_classifications}

Table~\appref{tab:grid-adversarial} keeps the four values separate and pools the remaining text. The median share is \tnGridShareMedianAdversarial\%. This exact game leaves contributions inside the pooled player unexamined. Pooling changes interaction allocation and can cancel opposing signed values. Equation~\appeqref{eq:online-pooling} supplies a hypothetical decrease, so the observed increases are not a general bound.

\begingroup\long\def\floatNotesBody{
Pooling leaves attribute values separate and restores the remainder together. Share divides classified attribute mass by total absolute Shapley mass; each cell gives the task median and bracketed interquartile range. Above word level counts increased shares. Interactions prevent a universal direction of change.
}%
\begin{table}[htbp]
\centering
\caption{Pooled grouping of the robustness checks on the 16 flight tasks of Qwen 3.8 27B.}\label{tab:grid-adversarial}
\footnotesize\setlength{\tabcolsep}{4pt}
\begin{NiceTabular}{@{}l;r;c;r@{}}
\toprule
Grouping & Players & Attribute share (\%) & Above word level \\
\midrule
word & $49$ & \begin{tabular}[t]{@{}c@{}}$19.1$\\$[14.6, 24.5]$\end{tabular} &  \\
\dottedrule
pooled & $5$ & \begin{tabular}[t]{@{}c@{}}$64.7$\\$[53.9, 71.9]$\end{tabular} & $16$ of $16$ \\
\bottomrule
\end{NiceTabular}
\end{table}
\endgroup{}
\floatnotes{grid_adversarial}

\subsection{Alternative edits and agreement between the two games}

Table~\appref{tab:grid-oracle} compares hand-authored replacements with replacements generated by Gemma 4 31B for the declared spans. I apply the same exclusions to both sets and recompute their reduced games. Among the retained operations, only the instruction rewrite differs: ``Reply with one'' versus ``Respond with a single.'' High agreement therefore concerns a narrowly changed set. It is not independent confirmation across broadly different edits.

\begingroup\long\def\floatNotesBody{
Gemma 4 31B's only retained difference: ``Reply with one'' versus ``Respond with a single''. Task medians [interquartile ranges] summarize ratios and correlations. Price over edits: price magnitude/total edit magnitude. Above one counts ratios exceeding one. $\rho$: Spearman correlation of absolute values across retained players.
}%
\begin{table}[htbp]
\centering
\caption{Edit sets of the robustness checks on the 16 flight tasks of Qwen 3.8 27B: the hand-authored edits against the set another model wrote.}\label{tab:grid-oracle}
\footnotesize\setlength{\tabcolsep}{4pt}
\begin{NiceTabular}{@{}l;c;r;c@{}}
\toprule
Edit set & Price over edits & Above one & $\rho$ with hand-authored \\
\midrule
hand-authored & \begin{tabular}[t]{@{}c@{}}$1.4$\\$[0.41, 3.8]$\end{tabular} & $8$ of $16$ &  \\
\dottedrule
alternative & \begin{tabular}[t]{@{}c@{}}$1.8$\\$[0.69, 3.5]$\end{tabular} & $10$ of $16$ & \begin{tabular}[t]{@{}c@{}}$0.76$\\$[0.63, 0.93]$\end{tabular} \\
\bottomrule
\end{NiceTabular}
\end{table}
\endgroup{}
\floatnotes{grid_oracle}

For each retained edit, Table~\appref{tab:grid-formulations} compares its absolute edit value with the sum of absolute word values it touches. A replacement touches overlapping word spans; an edit within an intervening space touches the adjacent words. Spearman rank correlation compares those quantities across edits within each task. The intact-edit and replacement-reference games evaluate different intervention sets. Positive correlation is descriptive and is not required for either calculation to be correct.

\begingroup\long\def\floatNotesBody{
$\rho$ ranks edit magnitudes against touched-word absolute mass across 10 retained edits. Each cell gives the task median and bracketed interquartile range. Positive counts tasks with positive correlation. The games evaluate different interventions and references; agreement is descriptive.
}%
\begin{table}[htbp]
\centering
\caption{Agreement of the reduced edit games with the word games on the 16 flight tasks, by model.}\label{tab:grid-formulations}
\footnotesize\setlength{\tabcolsep}{4pt}
\begin{NiceTabular}{@{}l;r;c;r@{}}
\toprule
Model & Tasks & $\rho$ & Positive \\
\midrule
Qwen 3.8 27B & $16$ & \begin{tabular}[t]{@{}c@{}}$0.17$\\$[-0.14, 0.47]$\end{tabular} & $10$ of $16$ \\
\dottedrule
Muse Glimmer & $16$ & \begin{tabular}[t]{@{}c@{}}$0.13$\\$[0.02, 0.24]$\end{tabular} & $12$ of $16$ \\
\dottedrule
Gemma 4 31B & $16$ & \begin{tabular}[t]{@{}c@{}}$0.27$\\$[-0.03, 0.43]$\end{tabular} & $11$ of $16$ \\
\dottedrule
Gemma 4 12B & $16$ & \begin{tabular}[t]{@{}c@{}}$0.07$\\$[-0.14, 0.18]$\end{tabular} & $10$ of $16$ \\
\bottomrule
\end{NiceTabular}
\end{table}
\endgroup{}
\floatnotes{grid_formulations}
 \clearpage
\section{Reproducibility}\label{part:reproducibility}

I document the model inputs, the saved records, and the analysis software and sequence needed to recompute the retained results from the saved scores. Table~\appref{tab:repro:models} identifies the checkpoints and answer tokens; Table~\appref{tab:repro:runs} records scoring work. Distinguish recomputing attribution from saved scores, regenerating exhibits from aggregates, and rerunning inference. These operations require different inputs and costs.

\subsection{Models and software}

Weights use 16-bit brain floating point, without quantization. Scoring requests raw label log-probabilities and generates one token per prompt. Longer generation is used only to write the alternative edit text.

Each task is a user message rendered through the model's chat template, with the persona supplied as a system message where present. Qwen and Gemma use their reasoning-disable switches. Muse uses its lowest reasoning strength and an assistant prefix addressing the user; its template date is fixed at 2026-09-29. Switching reasoning off makes the first generated token the answer, so that each score is read directly from the model for a fixed prompt. The games extend to a model that reasons before it answers, whose score is the log-odds of its final answer averaged over the reasoning it writes and estimated by sampling, with the pieces of the prompt as its players. Qwen trims message-edge whitespace. Table~\appref{tab:repro:models} provides the exact model repositories, checkpoint identifiers, and label-token identifiers.

\begingroup\long\def\floatNotesBody{
Revision is the pinned Hugging Face checkpoint hash, split across two lines. Label identifiers are the answer tokens for ``A'' and ``B''.
}%
\begin{table}[htbp]
\centering
\caption{Checkpoints and answer labels.}\label{tab:repro:models}
\footnotesize
\begin{NiceTabular}{@{}l;>{\raggedright\arraybackslash}p{2.7cm};>{\raggedright\arraybackslash}p{3.0cm};>{\raggedright\arraybackslash}p{1.9cm}@{}}
\toprule
Model & Repository & Revision & Label tokens \\
\midrule
Qwen 3.8 27B & \texttt{\scriptsize Qwen}/\allowbreak \texttt{\scriptsize Qwen3.8-27B} & \texttt{\scriptsize 1d4bf0f2ff6012fd8203}\newline \texttt{\scriptsize 9f2fa52739d0dd7c60c0} & ``A'' $32$\newline ``B'' $33$ \\
\dottedrule
Muse Glimmer & \texttt{\scriptsize meta-models}/\allowbreak \texttt{\scriptsize Muse-Glimmer-30B} & \texttt{\scriptsize a4e59da52a7bc87ae725}\newline \texttt{\scriptsize 1dd5545c0dd437c44b68} & ``A'' $45$\newline ``B'' $46$ \\
\dottedrule
Gemma 4 31B & \texttt{\scriptsize google}/\allowbreak \texttt{\scriptsize gemma-4-31B-it} & \texttt{\scriptsize 842da3794eaa0b77d5f0}\newline \texttt{\scriptsize 8bae87a17459d91ff475} & ``A'' $236776$\newline ``B'' $236799$ \\
\dottedrule
Gemma 4 12B & \texttt{\scriptsize google}/\allowbreak \texttt{\scriptsize gemma-4-12B-it} & \texttt{\scriptsize 707f0a3b8a3c7ad586ed}\newline \texttt{\scriptsize 01e27eafbad8a27dd0f7} & ``A'' $236776$\newline ``B'' $236799$ \\
\bottomrule
\end{NiceTabular}
\end{table}
\endgroup{}
\floatnotes{repro_models}

The analysis environment is Python 3.12.13 with NumPy 2.5.3, pandas 3.0.6, PyArrow 25.0.1, SciPy 1.18.1, Matplotlib 3.11.2, PyYAML 6.0.3, and Pydantic 2.13.5, each pinned in \texttt{uv.lock}. The analysis and inference environments are distinct. The configuration files specify chat handling, precision settings, and checks; completion records preserve the settings actually used.

\subsection{Seeds and data definitions}

The permutation seed uses the first eight bytes of a 256-bit Secure Hash Algorithm (SHA) digest of \texttt{template|task\_id|variant|model\_key}, interpreted as a big-endian integer and masked to 63 bits. NumPy's \texttt{default\_rng(seed)} draws orders sequentially, so increasing the order count extends the stream. Bootstrap draws use \texttt{default\_rng([seed, 1])}. The engine and raw-label sampler use seed zero. The random-token reference seeds are one through five; their realized replacements are saved. Alternative-edit generation retains its accepted replacement strings and sampling record.

Prompt identifiers hash the encoded system/user-message pair, permitting exact-text reuse across studies. Under \texttt{data/raw/manifests/}, prompt records contain \texttt{prompt\_id, system, user}. Chain records contain \texttt{game\_id, t, k, player, mask, prompt\_id}: the order index, restoration step, added player, coalition mask, and scored prompt. Exact-coalition records contain \texttt{game\_id, mask, prompt\_id}. Metadata supplies game identifiers, seeds, and player and order counts. Component definitions come from \texttt{code/config/design.yaml} and the game builders.

Score records contain \texttt{prompt\_id, l\_a, l\_b, run\_id}. Aggregation searches all study folders for a model because reused prompts may have been scored elsewhere. Completion records report checkpoint and engine settings, extraction checks, counts, and elapsed time. Table~\appref{tab:repro:runs} distinguishes required prompts from newly scored prompts.

\begingroup\long\def\floatNotesBody{
Needed counts distinct study prompts; Scored excludes previously scored prompts. Edit-game and alternative edit runs scored original sets; retained games reuse subsets without new inference.
}%
\begin{table}[htbp]
\centering
\caption{Scoring runs, one row per model and study, summed over a study's shards.}\label{tab:repro:runs}
\footnotesize
\begin{NiceTabular}{@{}l;l;r;r@{}}
\toprule
Model & Study & Needed & Scored \\
\midrule
Qwen 3.8 27B & presentation sweeps & $346$ & $346$ \\
\dottedrule
Qwen 3.8 27B & word games, flights & $710{,}524$ & $710{,}508$ \\
\dottedrule
Qwen 3.8 27B & edit games & $196{,}608$ & $196{,}575$ \\
\dottedrule
Qwen 3.8 27B & framing & $829{,}613$ & $796{,}845$ \\
\dottedrule
Qwen 3.8 27B & word games, hotels and smartphones & $1{,}204{,}649$ & $1{,}204{,}617$ \\
\dottedrule
Qwen 3.8 27B & personas & $2{,}301{,}544$ & $2{,}301{,}544$ \\
\dottedrule
Qwen 3.8 27B & robustness checks & $4{,}704{,}036$ & $4{,}653{,}634$ \\
\dottedrule
Qwen 3.8 27B & alternative edit set & $65{,}536$ & $57{,}344$ \\
\addlinespace
Muse Glimmer & presentation sweeps & $346$ & $330$ \\
\dottedrule
Muse Glimmer & word games, flights & $710{,}580$ & $710{,}580$ \\
\dottedrule
Muse Glimmer & edit games & $196{,}608$ & $196{,}575$ \\
\dottedrule
Muse Glimmer & framing & $829{,}528$ & $796{,}760$ \\
\dottedrule
Muse Glimmer & word games, hotels and smartphones & $1{,}204{,}538$ & $1{,}204{,}506$ \\
\dottedrule
Muse Glimmer & personas & $2{,}301{,}481$ & $2{,}301{,}481$ \\
\bottomrule
\end{NiceTabular}
\end{table}
\begin{table}[htbp]
\ContinuedFloat
\centering
\caption{Scoring runs, continued.}\label{tab:repro:runs:1}
\footnotesize
\begin{NiceTabular}{@{}l;l;r;r@{}}
\toprule
Model & Study & Needed & Scored \\
\midrule
Gemma 4 31B & presentation sweeps & $346$ & $330$ \\
\dottedrule
Gemma 4 31B & word games, flights & $710{,}589$ & $710{,}589$ \\
\dottedrule
Gemma 4 31B & edit games & $196{,}608$ & $196{,}575$ \\
\dottedrule
Gemma 4 31B & framing & $829{,}596$ & $796{,}828$ \\
\dottedrule
Gemma 4 31B & word games, hotels and smartphones & $1{,}204{,}490$ & $1{,}204{,}458$ \\
\dottedrule
Gemma 4 31B & personas & $2{,}301{,}584$ & $2{,}301{,}584$ \\
\addlinespace
Gemma 4 12B & presentation sweeps & $346$ & $327$ \\
\dottedrule
Gemma 4 12B & word games, flights & $710{,}536$ & $705{,}835$ \\
\dottedrule
Gemma 4 12B & edit games & $196{,}608$ & $188{,}388$ \\
\dottedrule
Gemma 4 12B & framing & $829{,}529$ & $796{,}761$ \\
\dottedrule
Gemma 4 12B & word games, hotels and smartphones & $1{,}204{,}563$ & $1{,}204{,}531$ \\
\dottedrule
Gemma 4 12B & personas & $2{,}301{,}644$ & $2{,}301{,}644$ \\
\midrule
Total &  & $25{,}742{,}376$ & $25{,}539{,}495$ \\
\bottomrule
\end{NiceTabular}
\end{table}
\endgroup{}
\floatnotes{repro_runs}

\subsection{Working analysis procedure}

The retained view lives in \texttt{data/retained/}. It holds excluded edits off and recomputes the reduced games from stored coalition scores. Word-game uncertainty recomputes complete marginal-contribution rows from chains and scores. The pipeline compares recomputed means and standard errors with saved attribution files before producing contrasts. This is a record-integrity check using the same estimator.

With complete raw manifests, scores, and completion records available, the aggregate procedure is
\begin{quote}\small\ttfamily
PYTHONPATH=code uv run python -m tokennoise.regenerate --strict
\end{quote}
It rebuilds the underlying aggregates and retained view without new inference. Missing raw inputs prevent that rebuild. They do not prevent exhibit generation from the supplied aggregates:
\begin{quote}\small\ttfamily
PYTHONPATH=code uv run python -m analysis.make\_all\\
paper/build.sh\\
PYTHONPATH=code uv run python -m analysis.manuscript\_qa
\end{quote}
The factory writes tables, figures, notes, and the shared number macros. The build compiles the online document before the main paper so external exhibit references resolve. Data-column definitions reside in \texttt{data/README.md} and \texttt{data/docs/retained.md}. Individual modules use \texttt{python -m tokennoise.retained TARGET [MODEL ...]}, with targets \texttt{e3}, \texttt{agreement}, \texttt{e4}, \texttt{word-e2}, \texttt{word-e4}, \texttt{word-e5}, \texttt{word-e6}, \texttt{example}, \texttt{oracle}, and \texttt{wtp}. Recomputing from the raw data requires all shared score folders and the saved random-reference and alternative-edit records, not just the selected study's scores.
 \clearpage
\section{Extended literature}\label{part:literature}

I organize the additional literature by the question it informs: representing respondents, designing an elicitation task, and allocating a response. These connections locate my proposed audit without making attribution a test of human fidelity. The reference list belongs to this online document and includes the methodological sources cited in its other parts.

\subsection{Simulated respondents and research targets}

\citet{app:goli2024can} examined the use of language models to elicit preferences, including intertemporal choices. Their task-specific discrepancies make the target of a simulation consequential. In the current version of their market-research paper, \citet{app:brand2026usingllms} reported that willingness-to-pay estimates derived from language-model responses are sometimes comparable with estimates from human studies, but often inaccurate and in some cases wrong-signed. They positioned generated data as a supplement to human studies, not a substitute. Inspecting prompt dependence complements that external validation; it cannot replace it.

Different research applications also demand different evidence. \citet{app:li2024determining} studied automated perceptual analysis. Their application concerns perceptions rather than establishing the validity of every simulated choice. \citet{app:wang2026large} developed a data-augmentation approach combining generated and observed choice data. That integration differs from treating generated respondents as direct replacements for people. Neither application supplies a justified cutoff for the attribute share defined here.

Personas create a further distinction between describing a respondent and demonstrating that a simulation represents that respondent. \citet{app:toubia2025twin} provided data for building and evaluating individual digital twins. The supplied profiles in my controlled tasks instead expand the input being attributed. Their effects can be meaningful even when they differ from equivalent rephrasings of an unchanged profile. This distinction explains why persona credit is reported separately from task credit.

\subsection{Prompt design and interpretation}

\citet{app:brucks2025prompt} showed that experimental prompt architecture can induce methodological artifacts. Here architecture refers to presentation choices, such as arrangement and framing, rather than an identified neural mechanism. \citet{app:gui2023challenge} gave a causal-inference account of another difficulty: a changed prompt can alter the model's inference about variables the researcher left unstated. Holding the visible attribute values fixed therefore does not by itself prove that every textual intervention preserves the intended problem.

Evidence from simulated surveys and evaluation tasks supports examining the protocol explicitly. \citet{app:bisbee2024synthetic} documented sensitivity of synthetic survey responses to wording and model changes. \citet{app:sclar2024quantifying} examined prompt-format sensitivity, while \citet{app:mizrahi2024state} showed that instruction paraphrases affect measured performance. In few-shot tasks, demonstration order is another source of variation \citep{app:lu2022fantastically}. My tasks contain no demonstrations; that result concerns a related input dimension, not one measured here.

More recent work separates several levels of diagnosis. \citet{app:leng2026prism} examined local sensitivity in preference measurement by relating paraphrase changes to a recovered preference direction. \citet{app:netzer2026synthetic} proposed diagnostics for evaluating digital-twin outputs without requiring ground truth for every target question. \citet{app:qin2026interactions} decomposed output scores into input interactions and examined their changes under prompt variation. \citet{app:cummins2026threat} studied analytic flexibility across configurations of simulated-data studies. These works motivate clear definitions of the object being compared. My games fix the scoring protocol and allocate response contrasts over declared component or edit coalitions; they do not infer the validity of an entire study from a single profile.

\subsection{Allocation and direct sensitivity}

The Shapley rule averages marginal contributions over inclusion orders \citep{app:shapley1953value}. Its marketing applications include online advertising attribution \citep{app:berman2018beyond}; its prediction-explanation applications include feature contributions \citep{app:strumbelj2014explaining,app:lundberg2017unified}. The reference matters because it defines the contrast \citep{app:merrick2020explanation}. Different choices of game can therefore produce different allocations even for the same complete input \citep{app:sundararajan2020many}.

\citet{app:hackmann2024word} masked words individually and measured effects on generated outputs. Their approach is a direct word-importance procedure, not a Shapley estimator. A matched removal contrast answers what a component contributes when the rest of the prompt is intact. Averaging over coalition contexts answers a different allocation question. Full-factorial response tables can expose interactions directly; Shapley divides their contributions under a stated rule. Neither method creates information beyond its evaluated inputs.

Global sensitivity analysis asks another question: how input variation under a distribution contributes to output variance. Shapley effects apply the allocation rule to such variance-based games \citep{app:owen2014sobol,app:song2016shapley}. My fixed-reference games instead allocate a particular endpoint contrast. They need no population distribution over natural prompts, but still require chosen replacements, players, and coalition weights. Input removal can produce unfamiliar configurations, a general limitation of unrestricted feature removal \citep{app:hooker2021unrestricted}. I therefore report each game's Shapley values together with the alone and left-out effects, preserving the distinction between explaining a response and validating its use.
 \bibliographystyle{plainnat}
\end{document}